%% file: merged.tex
\documentclass[preprint,5p,times,twocolumn]{elsarticle}

\input{math_commands.tex}

\usepackage{hyperref}
\usepackage{url}
\usepackage{amssymb}       
\usepackage{algpseudocode}
\usepackage{algorithm}
\usepackage{amsfonts}       
\usepackage{graphicx}
\usepackage{subfigure}
\usepackage{natbib}

\journal{Pattern Recognition}

\begin{document}

\begin{frontmatter}



\title{ECLAD: Extracting Concepts with Local Aggregated Descriptors}


\author[rwth]{Andrés Felipe Posada-Moreno\fnref{thanks}}
\ead{andres.posada@dsme.rwth-aachen.de}
\author[perfios]{Nikita Surya\fnref{thanks}}
\ead{nikita.surya@rwth-aachen.de}
\author[rwth]{Sebastian Trimpe}
\ead{trimpe@dsme.rwth-aachen.de}

\fntext[thanks]{First two authors have equal contribution to this work.}

\affiliation[rwth]{organization={Institute for Data Science in Mechanical Engineering, RWTH Aachen University},
            addressline={Dennewartstr. 27}, 
            city={Aachen},
            postcode={52068},
            state={North Rhine-Westphalia},
            country={Germany}}
\affiliation[perfios]{organization={Perfios Software Solutions Private Ltd},
            addressline={HM Vibha Towers, 5th Floor, No. 66/5-25, Adugodi},
            city={Bengaluru},
            postcode={560030},
            state={Karnataka},
            country={India}}

\begin{abstract}
Convolutional neural networks (CNNs) are increasingly being used in critical systems, where robustness and alignment are crucial. In this context, the field of explainable artificial intelligence has proposed the generation of high-level explanations of the prediction process of CNNs through concept extraction. While these methods can detect whether or not a concept is present in an image, they are unable to determine its location. What is more, a fair comparison of such approaches is difficult due to a lack of proper validation procedures. To address these issues, we propose a novel method for automatic concept extraction and localization based on representations obtained through pixel-wise aggregations of CNN activation maps. Further, we introduce a process for the quantitative comparison and validation of concept-extraction techniques based on synthetic datasets with pixel-wise annotations of their main components, mitigating possible confirmation biases induced by human visual inspection. Extensive experimentation on both synthetic and real-world datasets demonstrates that our method outperforms state-of-the-art alternatives.
\end{abstract}



\begin{keyword}
concept extraction \sep explainable artificial intelligence \sep convolutional neural networks

\MSC{68T30} \sep \MSC{68T45}
\end{keyword}

\end{frontmatter}


\section{Introduction}
\label{sec:Introduction}

As convolutional neural networks (CNN) become increasingly used in critical real-world applications (e.g., quality control \citep{Wang2018-xh} or medical diagnosis \citep{Benjamens2020}), there is an urgent need to understand their inner workings. This has led to a growing adoption of explainability methods during the lifecycle of models \citep{DBLP:journals/jair/BurkartH21,DBLP:conf/ACMdis/DhanorkarWQXP021,DBLP:journals/corr/abs-2104-08952} in an effort to increase transparency and trust, convey a sense of causality, ensure alignment, and make adjustments when necessary \citep{DBLP:conf/fat/BhattXSWTJGPME20,DBLP:journals/inffus/ArrietaRSBTBGGM20}.

In particular, post-hoc visual explanations of CNNs have proven to be useful for detecting undesired biases or unexpected behaviors in models \citep{DBLP:journals/jimaging/SinghSL20,DBLP:journals/tnn/TjoaG21}. In recent years, post-hoc visual explanations have been tackled by either (i) adopting feature attribution methods \citep{DBLP:journals/ijcv/SelvarajuCDVPB20,DBLP:conf/icml/ShrikumarGK17,DBLP:conf/aaai/QiKL20}, or (ii) mining higher level features through concept extraction (CE) techniques \citep{DBLP:conf/icml/KimWGCWVS18,DBLP:conf/nips/GhorbaniWZK19,DBLP:conf/nips/YehKALPR20}. Explanations provided by the first approach are termed \emph{local explanations} as they focus on analyzing single data instances, while those of the second approach are \emph{global explanations} as they focus on obtaining features pertaining to the understanding of the model as a whole. Although these two approaches are widely used, both have significant limitations.

\begin{figure}[h]
\centering
\subfigure[Class OK.]{
\label{fig:metal casting dataset intro class OK}
\includegraphics[width=0.13\textwidth]{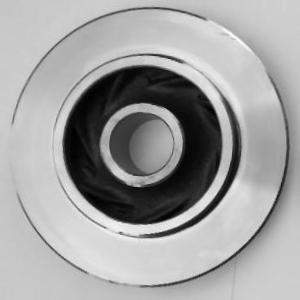}
\includegraphics[width=0.13\textwidth]{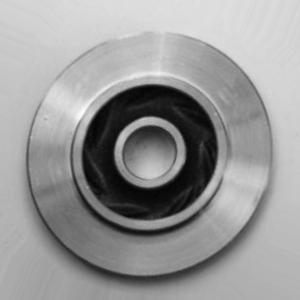}
\includegraphics[width=0.13\textwidth]{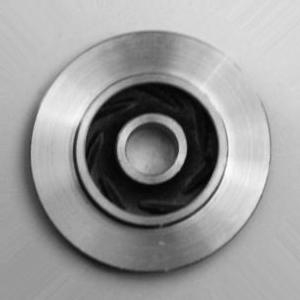}
}

\subfigure[Class defective.]{
\label{fig:metal casting dataset intro class defective}
\includegraphics[width=0.13\textwidth]{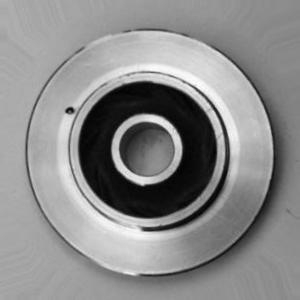}
\includegraphics[width=0.13\textwidth]{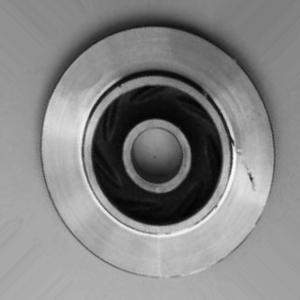}
\includegraphics[width=0.13\textwidth]{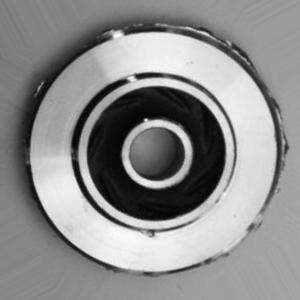}
}
\caption{Dataset metal casting, composed of two classes, \emph{OK} with a well cast metal part (a), and \emph{defective} with a metal part containing pinholes, scratches, or deformed edges (b).}
\label{fig:metal casting dataset intro}
\end{figure}

As a practical example, let us consider a CNN model for the classification of metal casting parts in a quality control process (good or defective) \citep{dabhi2020casting}, as seen in Figure \ref{fig:metal casting dataset intro}. 
During the lifecycle of the CNN, explainable artificial intelligence (XAI) may be used to detect undesired behaviors and to better understand which features are present in the acquired data. 
That is, we can use XAI techniques to understand which regions of the image are important for a single prediction (feature attribution), or which visual cues (e.g. pinholes, scratches, and deformed edges) the model differentiates and uses in its prediction process (Concept extraction). 
We want a model which makes predictions based on the pinholes, scratches, and deformed edges, or the lack of them. Understanding if this is the case, increases trust in the model, mitigates the risk of undesired biases, and ensures a high-level alignment with expert knowledge.

Feature attribution (local explanation) methods can be used to determine (for \emph{single images}) whether the pixels in the scratched or deformed regions are important for image classification, yet they do not tell us which groups of pixels are contextually related (composing a scratch), or whether the model distinguishes pixels in a scratch from pixels in an edge. This means, feature attribution methods indicate how much a pixel contributes to a prediction, but it does not explain which visual cues were learned by a model and consistently used in the prediction process. Moreover, recent studies have shown that feature attribution methods can be noisy and misleading \citep{DBLP:conf/nips/AdebayoGMGHK18}.

\begin{figure*}[h]
\includegraphics[width=0.98\textwidth]{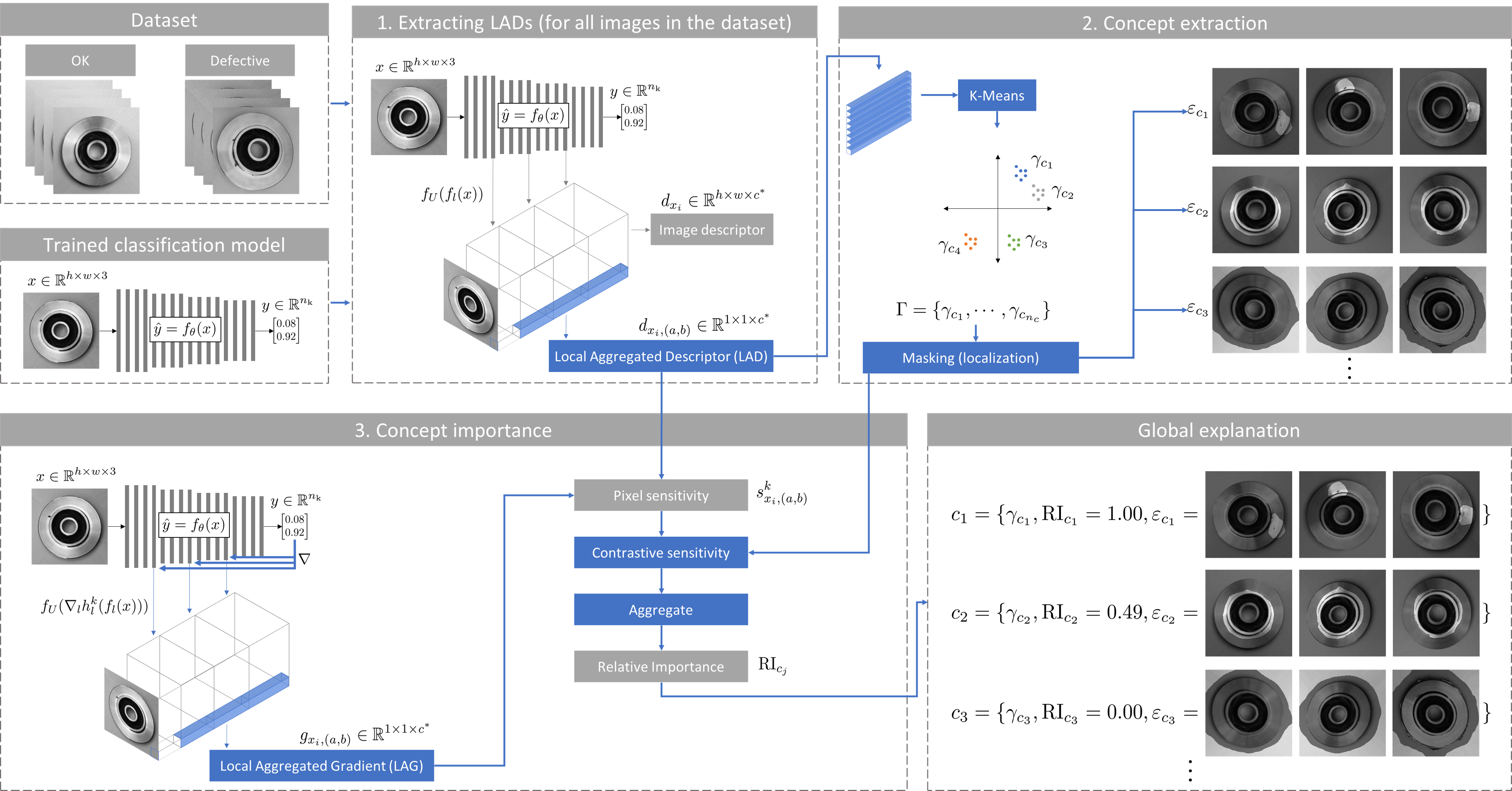}

\caption{
\textit{Extracting concepts with local aggregated descriptors} (ECLAD). The proposed concept extraction technique analyzes a trained classification model and a dataset. The technique is based on pixel-wise representations named Local Aggregated Descriptors (LADs) as seen in step 1. All descriptors from a dataset are clustered to find patterns of how a CNN learned to encode images (as seen in step 2). For new images, the extracted patterns can be localized through a masking process. Localization of concepts on single images allows for the aggregation of pixel-wise sensitivities to compute an importance score, which bridges local explanations (local sensitivities) and global patterns (as seen in step 3). Finally, the resulting set of concepts are described by a centroid $\gamma$, a relative importance score, and a set of example images.}
\label{fig:ECLAD main}
\end{figure*}

Concept extraction (global explanation) methods can analyze a model in the context of a dataset and return \emph{different sets of images} representing concepts -- in the case of the example mentioned above, samples of scratches or edges. These sets represent the concepts learned by a model during training and are accompanied by a score denoting their importance in the model's prediction process \citep{DBLP:conf/icml/KimWGCWVS18,DBLP:conf/nips/GhorbaniWZK19}. When explaining new instances, these methods can determine whether a concept is present, but not where it is, i.e., current methods do not localize the pixels containing each concept (where is the scratch, or the well-formed edges). This is a severe limitation in many applications, as posterior to the CE process, the results are not being used to explain abnormal behaviors in detail, increasing the possibility of biased interpretations. For example, a problematic instance of a piece with darker edges and a shiny patch elsewhere may be erroneously detected as defective by a CNN. A human may then erroneously interpret the edge as the problematic region, whereas the unusual shiny region was the confounding factor. The same issue makes the objective comparison and validation of CE techniques difficult, as interpreting which cues relate to a concept as well as its relation to the ground truth requires human intervention (though visual inspection).

Our work focuses on concept-based explanations, i.e., global explanations, and specifically on their inability to provide a straightforward concept localization. We propose \emph{Extracting Concepts with Local Aggregated Descriptors} (ECLAD) as a method for CE that -- posterior to its global execution -- is able to localize concepts and quantify their importance for a single image prediction, as shown in Figure \ref{fig:ECLAD main}. In contrast to previous CE methods, we do not encode an image as a single flattened activation map, but rather as a set of pixel-wise aggregations of the activation maps of multiple layers. We call these pixel-wise representations \emph{local aggregated descriptors} (LADs). 

\begin{figure*}[h]
\subfigure[ECLAD]{
\label{fig:ECLAD}
\includegraphics[width=0.98\textwidth]{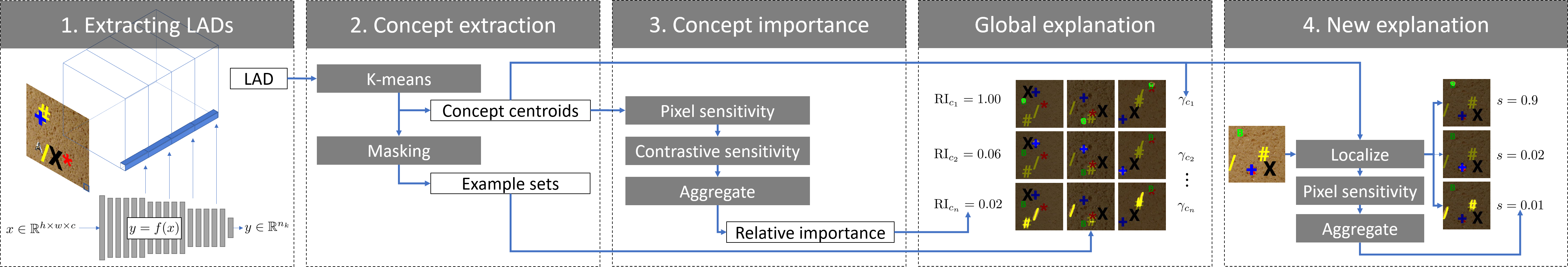}
}

\subfigure[ACE]{
\label{fig:ACE}
\includegraphics[width=0.98\textwidth]{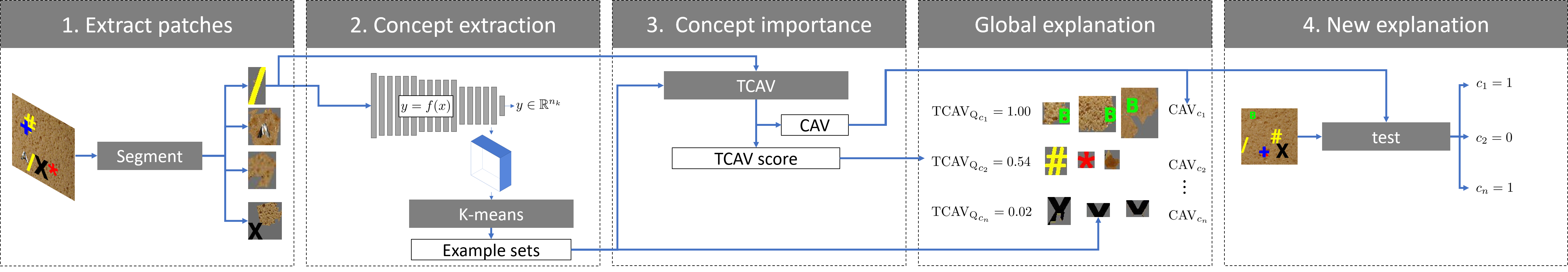}
}
\caption{Proposed concept extraction technique ECLAD \ref{fig:ECLAD} in comparison to a state-of-the-art alternative ACE \ref{fig:ACE}. ECLAD extracts a representation per pixel (LAD) before clustering and extracting concepts. ACE segments each image and uses a single representation to describe each patch before clustering. For new explanations, ECLAD provides the localization of each concept in the image, whereas ACE only tests whether each concept exists in the image.}
\label{fig:CE}
\end{figure*}

As an additional contribution, we address the challenge of quantitatively comparing and validating CE techniques. Current alternatives require human intervention to associate important concepts with the ground truth, making difficult a consistent and scalable scoring of CE methods. We propose an alternative process for the comparison and validation of CE methods that requires no human visual inspection to assess the correctness of extracted concepts. We achieve this by spatially associating labelled components on images (primitives) with regions related to the extracted concepts. As other CE methods do not provide a straightforward localization of the concepts, we segment the images and test each patch in search of every extracted concept. This process can be used to compare and validate any other CE technique and provide an objective and consistent performance measure, accounting for factors such as different CNN architectures and randomness. In this paper, we use this process to validate the correct performance of ECLAD, comparing it with other CE algorithms through a set of new synthetic datasets.

In summary, our main contributions are:
\begin{itemize}
    \item We propose a concept extraction method based on local aggregated descriptors that can extract global concepts and localize them in single images.
    \item We propose a process for validating concept extraction techniques by using pixel-level ground truth to relate extracted concepts with primitives of a synthetic dataset.
    \item We compare and validate our methods with multiple synthetic datasets (making them public), as well as real-world use cases. The experimental results of ECLAD outperform the state of the art with respect to how concepts are scored (importance correctness) while maintaining a comparable representation correctness.
\end{itemize}
\section{Related work on concept extraction}
\label{sec:Related Work}

The goal of concept-based explanation methods is to extract high-level features that relate to the decision-making process of a CNN. To achieve this, ante-hoc approaches have proposed custom CNN architectures, constraining the representations learned in their latent spaces \citep{DBLP:conf/nips/ChenLTBRS19,DBLP:conf/icml/KohNTMPKL20,DBLP:journals/natmi/ChenBR20,DBLP:conf/fruct/UtkinDKK21,DBLP:journals/corr/abs-1907-07165,DBLP:journals/corr/abs-2109-04518}.
In contrast, post-hoc approaches mine for sets of images which have distinctive representations within the latent space of CNNs. This means that representations from images containing a concept are similar, and differ from those of images without said concept. This has been achieved by extracting patches from images, and clustering them based on a defined representation within the latent space of the analyzed CNN \citep{DBLP:conf/nips/GhorbaniWZK19,DBLP:conf/cvpr/GeXXZKCIW21}.

Our work lies in the category of post-hoc concept extraction (CE), where the standard approach for extracting concepts is the algorithm \emph{Automatic Concept-based Explanations} (ACE) \citep{DBLP:conf/nips/GhorbaniWZK19}, depicted in Figure \ref{fig:ACE}. ACE uses the segmentation technique \textit{Simple Linear Iterative Clustering} (SLIC) \citep{DBLP:journals/pami/AchantaSSLFS12} to extract patches at multiple scales. Then, the extracted patches are resized and encoded through the CNN, by computing the flattened activation map of a layer close to the top of the CNN.
Afterwards, ACE clusters these representations and scores the importance of each cluster using the concept-testing algorithm TCAV \citep{DBLP:conf/icml/KimWGCWVS18}. Other studies have built on the CE capabilities of ACE, by focusing on concept completeness \citep{DBLP:conf/nips/YehKALPR20} (ConceptShap), or structural relations between concepts \citep{DBLP:conf/cvpr/GeXXZKCIW21,DBLP:journals/corr/abs-2008-06457,DBLP:journals/corr/abs-2007-07477}. In essence, these methods assess whether an image contains a concept, and to what extent that concept influences the prediction of said image, but not where the concept is located, omitting relevant spatial information. 
Our approach uses LADs to represent each pixel, rather than employing a single flattened activation map describing a whole image (as seen in Figure \ref{fig:ECLAD}).  
In contrast to the above-mentioned works, this approach allows for a straightforward localization of pixels considered part of a concept. 
In addition, said localization reflects the internal representation of the models, instead of being biased by a segmentation technique used over the raw data.

The second main contribution of this paper is a method for comparing and validating CE techniques. Such a process has proven challenging for most studies to date, and three principal approaches have been followed. 
The first consists of using image classification datasets such as ImageNet \citep{DBLP:conf/cvpr/DengDSLL009}, performing CE over a trained model, and visually (qualitatively) inspecting the results for specific classes \citep{DBLP:conf/nips/GhorbaniWZK19,DBLP:journals/corr/abs-2106-08641,DBLP:conf/cvpr/GeXXZKCIW21,DBLP:journals/corr/abs-2008-06457,DBLP:journals/corr/abs-2007-07477}. 
The second approach builds on the first, performing a user study to either measure the meaningfulness of extracted concepts, or linking them to the main unannotated attributes of each class \citep{DBLP:conf/nips/GhorbaniWZK19,DBLP:conf/nips/YehKALPR20}.
The third approach utilizes datasets (synthetic or natural) with labels denoting the presence of an attribute in each image \citep{DBLP:journals/corr/abs-1907-07165,DBLP:conf/nips/YehKALPR20}. This approach allows for a quantitative evaluation of the correlation between the labels and the extracted concepts, yet it does not ensure that the same visual cue is responsible for both.
Our proposed comparison and validation approach uses tailored synthetic datasets with pixel-level annotations for each primitive (visual cues or concepts present on each image) to objectively relate them to extracted concepts through a distance metric. 
This measure enables an automatic assessment of whether the important extracted concepts coincide with the primitives used to compose the images of each class. 
In addition, the completely automated process allows for a comparison at a scale, taking into account the stochastic effects of training models and the differences of CNN architectures.
To our knowledge, this is the first CE validation technique that uses pixel-wise annotations to verify whether visual cues from an important extracted concept are related to dataset primitives without human intervention.
\section{ECLAD}
\label{sec:ECLAD}

We present \emph{Extracting Concepts with Local Aggregated Descriptors} (ECLAD) as an explanation method for CNNs that extracts concepts (meaningful representations that a model has learned) using a pixel-wise aggregation of activation maps. Its main premise is that the activation maps of the multiple low, mid, and high level layers can be re-scaled and composed at a pixel level to obtain a comprehensive description of how a neural network encodes a location of an image (including its surrounding context). Consequently, this encoding can be used to mine for concepts.

We introduce ECLAD in six steps. 
First, we introduce our notion of a concept in CNNs, and how to extract them.
Second, we propose and specify what we mean by local aggregated descriptors (LADs), as the core component of our approach. 
Third, we describe the process of CE by clustering LADs.
Fourth, we explain the main idea behind concept localization.
Fifth, we propose a metrics of the relative importance of each extracted concept. 
Finally, we provide the pseudocode of ECLAD, and summarize its usage.

\subsection{Concept Extraction}

We consider the task of CE as the process of analyzing a trained model $\hat{y} = f(x)$ and a dataset $E = \{ (x_1 , y_1), \cdots, (x_{n_E} , y_{n_E}) \}$, to automatically extract a set $C = \{ c_1, \cdots, c_{n_c}\}$ of patterns which the model has learned to differentiate, and to score the importance of said patterns towards the prediction process of a model.
These patterns are referred to as \textit{concepts}, representing an idea or abstraction. In the image domain, these concepts relate to specific visual cues present in multiple images which share similar representations within the latent space of a CNN.

On practical terms, each extracted concept $c_j$ is described by a vector representation $v_{c_j}$ within the CNN, an importance score $I_{c_j}$, and an example set $\varepsilon_{c_j}$. 
The vector representation $v_{c_j}$ is an entity which allows the classification of whether or not an image or region contains the concept $c_j$.
$v_{c_j}$ can be the normal of a plane classifying images with or without the concept's visual cues \citep{DBLP:conf/icml/KimWGCWVS18,DBLP:conf/nips/GhorbaniWZK19}, a specific dimension on a lower dimensional projection of an activation map \citep{DBLP:conf/nips/YehKALPR20}, or a centroid of a cluster (of representations) in our case.
The score $I_{c_j}$ denotes importance of the concept in the prediction process of the model. It has been computed as the fraction of images in a class containing a concept \citep{DBLP:conf/icml/KimWGCWVS18,DBLP:conf/nips/GhorbaniWZK19}, as the average expected marginal contribution of the concept \citep{DBLP:conf/nips/YehKALPR20}, or as an aggregation of the local importance of the related visual cues in our case.
Finally, the example set $\varepsilon_{c_j}$ is composed of images or patches from the input domain $X$ which share and highlight the visual cues related to the concept $c_j$. It is used as a human understandable proxy for the pattern learned by the CNN while training.

On an abstract level, CE methods are composed of three subtasks. 
First, the selection of a relevant representation of the latent space of a CNN. This representation defines the type of patterns which will be mined, and reflects the granularity of the used information. 
Second, the concept extraction itself, it is the mining of patterns within the selected representations of the analyzed dataset. This mining process directly affects the choice of $v_{c_j}$, the boundaries for the detection of each concept, and the example sets $\varepsilon_{c_j}$ visually describing each mined pattern.
Third, the scoring of each extracted concept $c_j$ to estimate the importance $I_{c_j}$ of said pattern in the prediction process of the model. On the subsections below, we introduce our approach.

\subsection{Local aggregated descriptors}

\begin{figure}[tb!]
\includegraphics[width=0.48\textwidth]{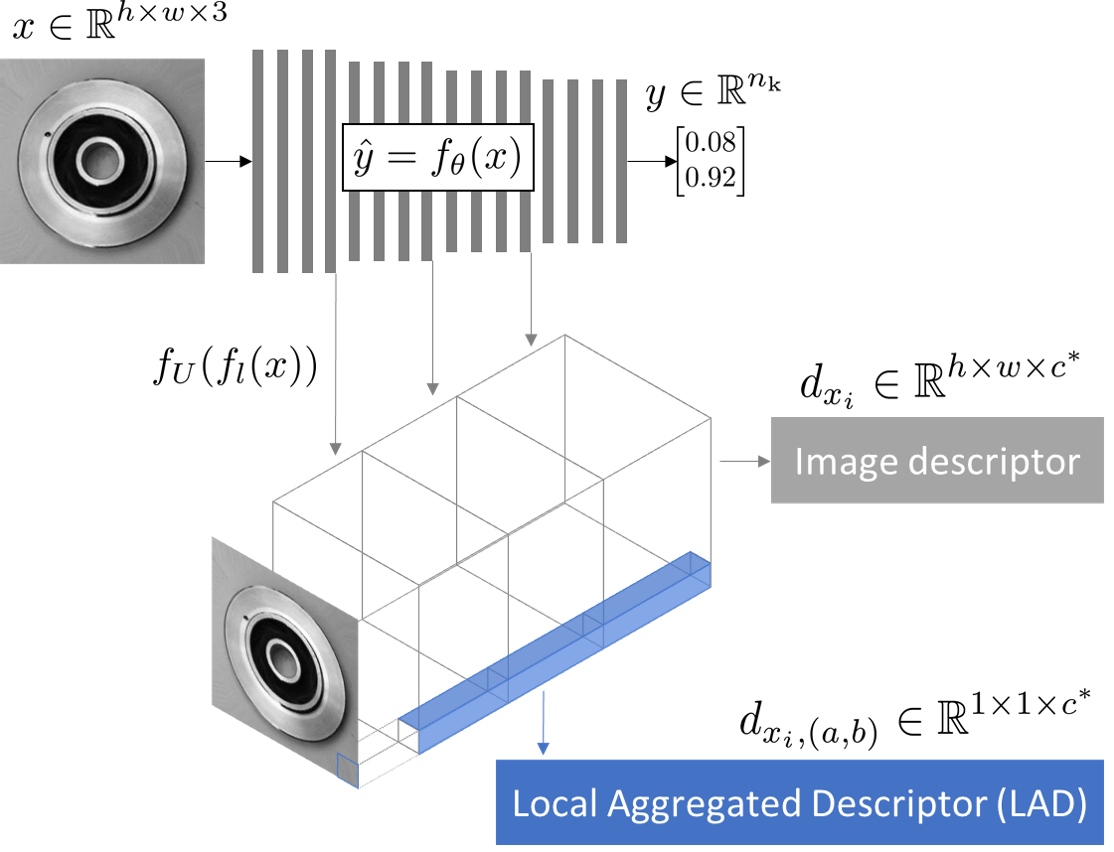}

\caption{Visual description of Local Aggregated Descriptors (LADs). 
Image descriptors are obtained by extracting, upscaling, and concatenating activation maps from a defined set of layers of a CNN. LADs are the pixel-wise components of said descriptor.
This local vector representation contains information of how a CNN encodes a pixel and its surrounding context at different abstraction levels.}
\label{fig:LAD}
\end{figure}

The first challenge of concept extraction techniques is finding meaningful representations related to the latent space of a model. Specifically, this representation should reflect how a model encodes information, allow for the mining of patterns, and serve as a basis for estimating the importance of said patterns for the prediction process of the model. 
This is where we propose the notion of \textit{Local Aggregated Descriptors} (LADs). The key idea of LADs is to use a pixel-wise descriptor of how models encode a region at different levels of abstraction. We achieve this by upscaling and aggregating the activation maps of multiple layers along a CNN, as seen in Figure \ref{fig:LAD}.

We base the selection of the representation used in ECLAD on how CNNs work. Typical CNN classification models approximate the mapping of images $x$ of dimensions $(h,w,3)$ (for RGB images), to a vector $y$ corresponding to the probability of the input belonging to $n_\mathrm{k}$ classes with a function $f: x \in \mathbb{R}^{h \times w \times 3} \to y \in \mathbb{R}^{n_\mathrm{k}}$. 
This mapping is performed by composing multiple layers of convolutions, activation functions and other pooling mechanisms. Through the composition of these mechanisms, CNNs encode the information of the input data into different latent spaces after each layer. 
In CNNs, a partial evaluation until a layer $l$, yields an activation map $a_{l} = f_{l}(x)$, belonging to a latent space $a_l \in \mathbb{R}^{h_l \times w_l \times c_l}$, where the dimensions depend on the input as well as the type and quantity of layers evaluated.


To exploit the progressive information encoding of the CNNs, we propose the aggregation of the activation maps of a predefined set $L=\{ l_1, \cdots , l_{n_\mathrm{l}}\}$ of $n_\mathrm{l}$ layers. In addition, to take into account the base translation invariance of convolutional layers, we consider how a pixel and its surroundings are encoded within these layers.
We obtain the aggregated descriptor of an image $x_i$, by first computing $a_{l}$ for each layer $l \in L$. Then, we upscale each $a_{l}$ to the spatial dimensions of $x_i$ using bilinear interpolation ($f_U$), as illustrated in Figure \ref{fig:LAD}. Finally, we concatenate the resulting maps alongside their third dimension (depth)
\begin{equation}
d_{x_i} = \left [ f_U(f_{l_1}(x_i)) \, \cdots \, f_U(f_{l_{n_\mathrm{l}}}(x_i)) \right ].
\label{eq:aggregation function}
\end{equation}
We obtain the descriptor $d_{x_i} \in \mathbb{R}^{h \times w \times c^*}$, where $c^*$ is the sum of the number of units for all layers in $L$. \emph{Local aggregated descriptors} (LADs) refer to each pixel $d_{x_i,(a,b)} \in \mathbb{R}^{1 \times 1 \times c^*}$ of the tensor $d_{x_i}$, where $(a,b)$ denotes the position along the width and height of $d_{x_i}$. LADs contain information about how the CNN encodes a pixel and its surrounding context in different abstraction levels, providing a practical latent representation for understanding how a model interprets an input image.

LADs offer two main benefits in comparison with the representations used in ACE \citep{DBLP:conf/nips/GhorbaniWZK19} and ConceptShap \citep{DBLP:conf/nips/YehKALPR20}.
First, LADs provide a representation for each pixel and its surrounding. This ensures any feature invariance or equivariance learned by a model will be reflected in our representation. In addition, this allows mining for patterns without having to segment images into patches.
Second, we take into account multiple levels of abstraction, which results in more granular concepts, reflecting not only the encodings of the top layers, but also mid-level representations which also contain important information for explaining CNN models.

\subsection{Mining patterns}

The subtask of mining patterns is directly related to the chosen latent representation.
For example, ACE proposes extracting concepts classwise, this is done by first segmenting all images using SLIC, then resizing and encoding each patch on the model, before performing clustering over the selected representations of all patches. In contrast, ConceptShap learns an intermediate lower dimensional projection, while maintaining the model's performance. Each dimension of this projection is then treated as a concept.
On one side, patch encoding induces significant cost, by requiring hundreds of evaluations of the model per image. On another side, projections, can lose information, specially in the cases of correlated visual features.

We avoid both pitfalls with the introduction of LADs, and a direct mining of concepts without segmentation or re-training.
The main idea behind our concept extraction approach relies on directly mining patterns from the LADs of all images in the dataset $E$,
\begin{equation*}
D = \{ d_{x_i,(a,b)} \mid a \in \{1, \dots, h\} , b \in \{1, \dots, w\} , x_i \in E \}.
\label{eq:LADs set}
\end{equation*}
To mitigate the computational cost, we use batches $E_b$ of $n_i$ images, compute the set of their LADs $D_b$, and apply the minibatch \textit{k}-means \citep{DBLP:conf/www/Sculley10} algorithm. This allows for an iterative evaluation of the complete dataset, providing global concepts which can be important for the prediction of one or more classes. Subsequently, we obtain the set $\Gamma=\{ \gamma_{c_1}, \cdots, \gamma_{c_{n_\mathrm{c}}}\}$ of centroids $\gamma_{c_j} \in \mathbb{R}^{1 \times 1 \times c^*}$ defining the concepts. Each centroid $\gamma_{c_j}$ is the vector representation of the concept $c_j$ (akin to $v_{c_j}$ from ACE).



In comparison to other methods, directly mining concepts from LADs results in multiple benefits. 
First, the number of evaluations of the analyzed model is reduced, as each image is evaluated only once. Second, in contrast with ACE, LADs centroids do not assume a mean centered latent space \citep{DBLP:journals/natmi/ChenBR20} (assumed in concept activation vectors). 
Third, in contrast with ConceptShap, correlated visual features will only be merged in a single concept if they are encoded similarly within the latent space of the model.
Fourth, the obtained concepts are defined by LAD centroids, which can be directly used for localization. 

\subsection{Concept localization}

The task of concept localization consists in assessing which visual cues are related to a concept. This can be a non-trivial process, depending on the choice of latent representation used for concept extraction. For example, approaches which use a single flattened representation per image, allow for the detection of whether the concept is present or not in an image, but not where. In our approach, the chosen representation (LADs) allow for a straightforward localization, by comparing the distance of each LAD in an image to the set of centroids $\Gamma$ describing the concepts.

To locate a concept $c_j$ in an image $x_i$, we create a mask $m_{x_i}^{c_j} \in \mathbb{R}^{h \times w \times 1}$ by analyzing each $d_{x_i,(a,b)}$ and assessing whether it belongs to the cluster defined by $\gamma_{c_j}$, 
\begin{equation}
m_{x_i,(a,b)}^{c_j} = 
    \begin{cases}
    1       & \quad \underset{c_q}{\mathrm{argmin}}\,(\| d_{x_i,(a,b)} - \gamma_{c_q} \|)=c_j\,,\\
    0       & \quad  \text{otherwise.}
  \end{cases}
\label{eq:mask pixels}
\end{equation}
This allows for a direct evaluation of an image, identifying not only whether it contains a concept, but also where it is located (localization). We use the masks $m_{x_i}^{c_j}$ to attenuate unrelated pixels by a factor $\lambda \in (0,1]$ and obtain the human-understandable sets of examples
\begin{equation}
\varepsilon_{c_j} = \{(1-\lambda) \, m_{x_i}^{c_j} \odot x_i + \lambda x_i \mid x_i \in E \}.
\label{eq:example sets}
\end{equation}

\subsection{Concept importance}

The computation of a concept importance score, aims to provide a single metric conveying how much influence a concept has in the prediction process of a model. For example, other approaches measure either the proportion of images where the concept has a positive influence in the prediction (ACE), or the expected average contribution of a concept (ConceptShap) \citep{DBLP:journals/corr/abs-1907-07165,DBLP:conf/nips/GhorbaniWZK19}. Nonetheless, these can be problematic, as the scores do not take into account the specific visual cues of a concept, but the complete image. Thus, having limitations when dealing with correlated concepts. These approaches also disregard any link between local (image-wise) explanations and the global (dataset-wise) importance score of a concept.
In contrast, our proposed approach centers on computing the sensitivity of a prediction with respect to the localized visual cues of a concept. These sensitivities are then aggregated for the complete dataset and contrasted for different classes and concepts. Our approach ensures that only visual cues related to a concept are being used for the importance computation, and provides a direct link between image-wise sensitivities, and the obtained concept importance scores.

\textbf{First, we propose the computation of the sensitivity of a prediction on a pixel level using LADs}. For this, we use the same approach as before and also aggregate gradients to obtain $g_{x_i} \in \mathbb{R}^{h \times w \times c^*}$, 
\begin{equation*}
\begin{split}
g_{x_i} = 
\big[
f_U(\nabla_{l_1} h_{l_1}^{k}(f_{l_1}(x_i))) \, \cdots \,
f_U(\nabla_{l_{n_\mathrm{l}}} h_{l_{n_\mathrm{l}}}^{k}(f_{l_{n_\mathrm{l}}}(x_i))) 
\big],
\end{split}
\label{eq:gradient descriptor}
\end{equation*}
where $\nabla_{l} h_{l}^{k}(f_{l}(x))$ is the gradient of the prediction for the class $k$ with respect to an activation map $a_l = f_{l}(x)$. Using the aggregate gradient $g_{x_i}$ as a basis, we define the local aggregated gradient as $g_{x_i,(a,b)}$ similar to the LADs. Then, the sensitivity of a pixel, denoted as $s_{x_i,(a,b)}^{k}$, becomes the dot product between its local aggregated gradient and its LAD,
\begin{equation}
s_{x_i,(a,b)}^{k} = (g_{x_i,(a,b)})^{T} \cdot d_{x_i,(a,b)}.
\label{eq:pixel sensitivity}
\end{equation}

\textbf{Second, we introduce the aggregation of sensitivity scores for the visual cues of each concept}. For this, we introduce the set of sensitivities (towards class $k$) for all pixels in all images of $E_k$ (images of class $K$) belonging to the concept $c_j$, as
\begin{equation}
S_{c_j}^{k, E_{k}} = \{ s_{x_i,(a,b)}^{k} \mid x_i \in E_k , \; m_{x_i,(a,b)}^{c_j}=1 \}.
\label{eq:aggregated sensitivity}
\end{equation}
On it own, the average $\overline{S_{c_j}^{k, E_{k}}}$ can be used as a measure of how sensitive the prediction of class $k$ is with respect to the concept $c_j$. Nonetheless, if the concept $c_j$ is present in images of all classes, with a positive sensitivity towards the class $k$, it is then considered a bias, and not a differentiating concept.

\textbf{Taking possible biases into account, we propose the usage of a contrastive sensitivity} $\mathrm{CS}_{c_j}^{k}$, measuring the difference between the average sensitivity of a concept $c_j$ towards the class $k$ for all images of a class $k$ minus the average for the rest of the dataset:
\begin{equation}
\mathrm{CS}_{c_j}^{k} = 
\overline{S_{c_j}^{k, E_{k}}}  - 
\overline{S_{c_j}^{k, E \setminus E_{k}}} \,\,.
\label{eq:contrastive sensitivity}
\end{equation}
This measure quantifies how sensitive are the predictions of a class $k$ with respect to a concept $c_j$, while taking into account possible biases. It must be highlighted that in cases where a concept only appears in images of class $k$, $\mathrm{CS}_{c_j}^{k}$ will be equal to $\overline{S_{c_j}^{k, E_{k}}}$. Yet, this measure can result impractical when dealing with a higher number of classes.

\textbf{Finally, we propose the usage of a relative importance score}, which measures the importance of a concept, taking into account all classes, and presented in a relative scale between concepts.
We propose the relative importance measure $\mathrm{RI}_{c}$,
\begin{equation}
\mathrm{RI}_{c_j} = 
\mathrm{CS}_{c_j}^{k_{c_j}} /
\underset{c_q,k}{\mathrm{max}}(\mid \mathrm{CS}_{c_q}^{k} \mid) \; ; \; k_{c_j} = \underset{k}{\mathrm{argmax}}(\mid \mathrm{CS}_{c_j}^{k} \mid).
\label{eq:relative imporance}
\end{equation}
The relative importance ($\mathrm{RI}_{c_j}$) is a scaled value denoting the highest contrastive sensitivity of a concept, normalized across all concepts. Moreover, $\mathrm{RI}_{c_j}$ allows for the extraction of attributive and counterfactual concepts. 

As a general intuition, the most important concept used by a model will be scored with a magnitude of 1.0. 
This score would mean  that the model learned to differentiate its visual cues, and the concept strongly influences the prediction of a class.
Similarly, non-important concepts in the prediction process of a model will be scored with a magnitude of 0.0,
meaning that the visual cues were learned but are functionally useless when making a prediction.

Our importance score is directly tied to (A) the spatial regions containing a concept, and (B) the magnitude of the sensitivity of units in the selected layers for different class images. The objective of this metric is to better represent the inner workings of a CNN. By doing so, we avoid three known limitations of TCAV and concept Shapely values. First, by relying on (A) we avoid issues computing the importance score of co-occurring concepts, a known limitation of Shapely values. Second, by relying on (B) we avoid issues computing the importance score of concepts which are in a similar general direction when the latent space of a CNN is not zero centered. Third, (A) and (B) allow us to give a relative importance to co-occurring concepts by comparing the magnitude of their sensitivities. Our importance metric aims to directly reflect these dynamics of CNNs (and their internal activations), which are not captured through either TCAV scores or Shapely values.

\subsection{Pseudocode}

ECLAD is designed to be first executed over a complete dataset to generate a global explanation. The resulting set of centroids $\Gamma$ can then be used to localize each concept for new input images. The global execution of ECLAD is described in Algorithm \ref{alg:ECLAD (global)}.

\begin{algorithm}[H]
   \caption{ECLAD (global)}
   \label{alg:ECLAD (global)}
\begin{algorithmic}[1]
    \Require model $f$, dataset $E$, number of output classes $n_\mathrm{k}$, layers $L$, number of concepts $n_\mathrm{c}$, mini batch size $n_\mathrm{i}$, mask attenuation $\lambda$
    \For{$E_b \in E$} \Comment{ $\mid E_b \mid = n_\mathrm{i}$ \qquad \qquad \qquad \qquad \qquad}
        \State $D_b \leftarrow \mathrm{GetLADs}(f, E_b, L)$ 
        \State $\Gamma \leftarrow \mathrm{MiniBatchKmeans}(D_b, n_\mathrm{c})$
    \EndFor
    \For{$\gamma_{c_j} \in \Gamma$}
        \State $\varepsilon_{c_j} \leftarrow \{(1-\lambda) ( \mathrm{Mask}(x_i, \gamma_{c_j}) \odot x_i ) + \lambda x_i \mid x_i \in E \}$
        \For{$k \in \{1, 2, \dots, n_\mathrm{k}\}$}
            \State create $S_{c_j}^{k, E_{k}}, S_{c_j}^{k, E_{k^\prime}}$
            \State compute $\mathrm{CS}_{c_j}^{k}$
        \EndFor
        \State compute $\mathrm{RI}_{c_j}$
    \EndFor
    \State \textbf{return} $\{ ( \gamma_{c_j}, \varepsilon_{c_j}, \mathrm{RI}_{c_j} ) \mid \gamma_{c_j} \in \Gamma\}$
\end{algorithmic}
\end{algorithm}
As a result of executing ECLAD, we obtain for each concept $c_j$: a centroid $\gamma_{c_j}$ which serves as an anchor; an example set $\varepsilon_{c_j}$ of human-understandable visualizations; and a relative importance score $\mathrm{RI}_{c_j}$, which describes how important each concept is for the overall predictions of the model.

We perform the localization of each concept $c_j$ in a new image by extracting the mask $m_{x_i}^{c_j}$, for each concept defined by $\gamma_{c_j} \in \Gamma$. The resulting masks serve as an explanation of where the different concepts are located. In addition, the average sensitivities of all pixels in an image belonging to a concept can be used as local measures of importance.

\section{Validation of concept extraction techniques}
\label{sec:Validation}

We propose a method for the quantitative comparison and validation of CE techniques based on pixel-level annotations of synthetic datasets. This method is not meant to replace usability studies with humans, which seek to understand explanations in human-AI systems \citep{DBLP:journals/corr/abs-1902-01876}. Rather, it is an approach to score CE techniques purely based on quantitative metrics on synthetic datasets in a consistent and scalable way. 

The validation of CE methods relies on the assumption that the model learned the intended features of a dataset. Specifically because the task of CE aims to extract concepts related to what a model learns, and not necessarily what the structure of a dataset is.
Yet, guaranteeing that a model learns the intended features is often non-trivial, as many factors intervene on what is learned, including randomness, a model's architecture, and possible spurious correlations or biases present in the high dimensional datasets \citep{DBLP:journals/natmi/GeirhosJMZBBW20}. For this reason, we propose the comparison and validation of CE methods using a set of controlled and relatively simple synthetic datasets. In this controlled scenario, and through multiple runs with different random seeds and model architectures, we obtain a quantitative evaluation of the performance of CE techniques.

For the design of our validation process, we consider a case, where we have an unbiased classification dataset of images and their labels. Within all high level features contained in the data, we have a subset of \emph{important features} which are the differentiating factors between the labels, and a subset of \emph{unimportant features}. We build upon the assumption that after training, a model learns to predict the labels by detecting a subset of the important features (possibly disregarding correlated features \citep{DBLP:journals/natmi/GeirhosJMZBBW20}).
Then, a CE algorithm analyzes the model and dataset, extracting a set of concepts and scoring their importance. In this case, the results of both, extracted concepts and their importance should be aligned with the intended features of the dataset.

We denote as \textit{aligned} concepts those spatially related to the important features of the dataset (which were learned by the model).
Similarly, we denote as \textit{unaligned} concepts those representing unimportant or unannotated features of the dataset that are irrelevant for performing the desired task.
In an ideal case, where the features of a dataset were perfectly learned by a trained model, we propose scoring the performance of a CE method by measuring how aligned the resulting concepts are with the intended features of the dataset. We compare the correctness of the concepts in terms of their spatial localization and importance scores, by using two proxy metrics named \textbf{representation correctness} and \textbf{importance correctness}. More details on these metrics are provided below.


With the ideal case in mind, we propose a comparison and validation procedure that aims to evaluate representation and importance correctness of a CE technique.
First, we create a set of synthetic datasets, including masks for the base components of the images. Second, we train a set of models for each dataset. Third, we execute the CE method. Fourth, we compute localization masks for each extracted concept for each dataset image. Fifth, we associate each concept to the ground truth masks using a spatial distance metric. This allows us to classify the concepts as aligned or not. Finally, we quantify the \textbf{importance correctness} and \textbf{representation correctness} of the extracted concepts.
This process can be repeated for multiple CE techniques, comparing the introduced metrics.

\subsection{Synthetic datasets}
\label{sec:synth datasets}

The main challenge in testing CE methods is the lack of ground truth regarding which features are learned by a model and how relevant they are for its prediction process. This leads to the assumption that the model has learned the intended features of a dataset. 
Yet, it is also known that models are susceptible to learning shortcuts, spurious correlations, or biases which unintentionally are present in the training data \citep{DBLP:journals/natmi/GeirhosJMZBBW20}. 

\begin{figure}[tb]
\centering
\subfigure[Class A.]{
\label{fig:example dataset AB A}
\includegraphics[width=0.09\textwidth]{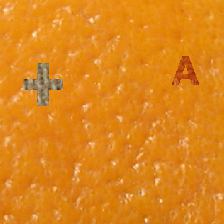}
\includegraphics[width=0.09\textwidth]{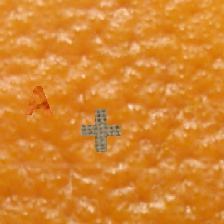}
}
\subfigure[Class B.]{
\label{fig:example dataset AB B}
\includegraphics[width=0.09\textwidth]{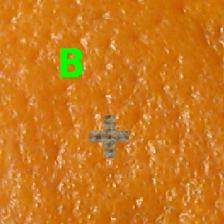}
\includegraphics[width=0.09\textwidth]{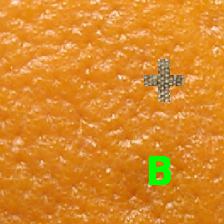}
}
\subfigure[Primitives $p_1$, $p_2$, $p_3$, and $p_4$.]{
\label{fig:example dataset AB primitives}
\includegraphics[width=0.09\textwidth]{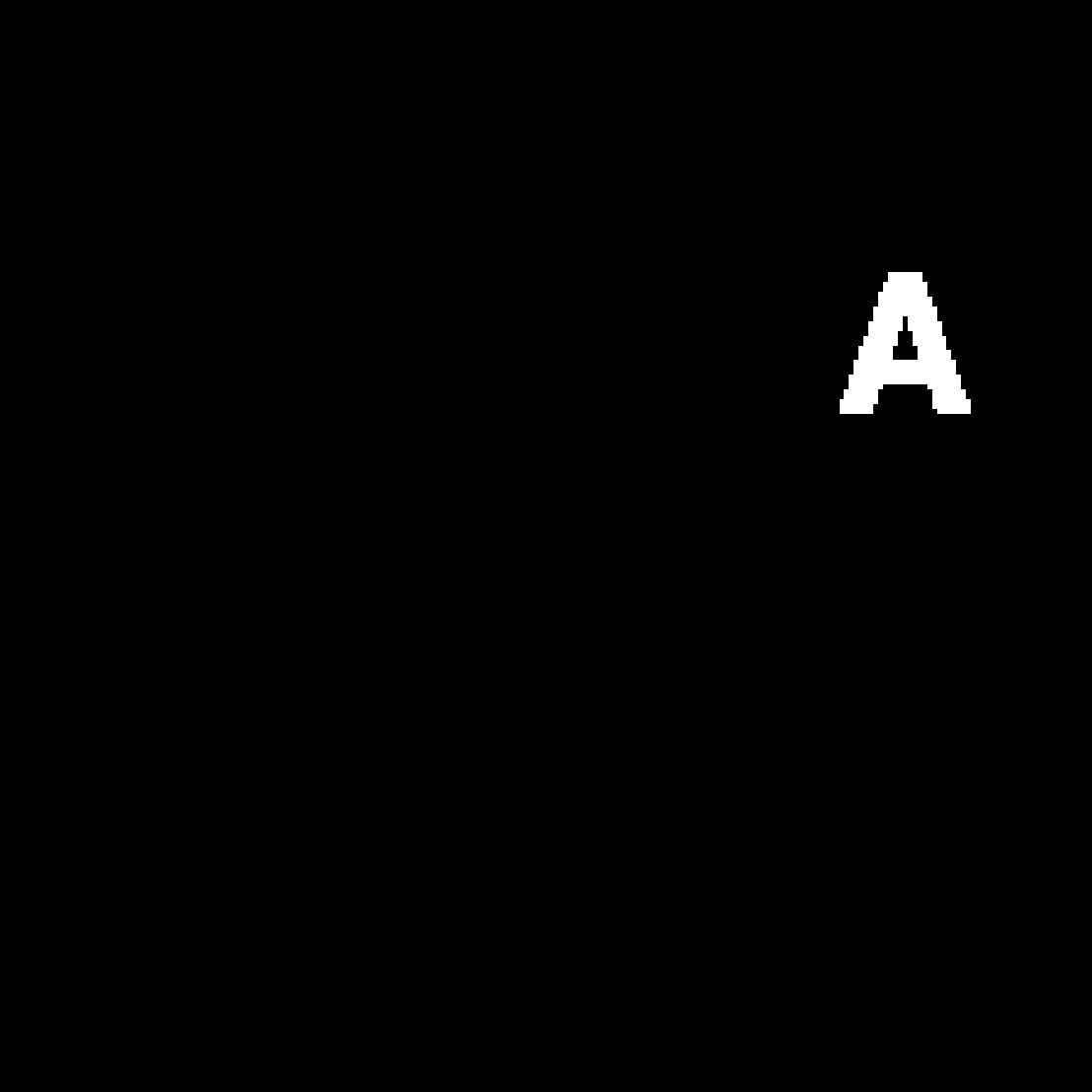}
\includegraphics[width=0.09\textwidth]{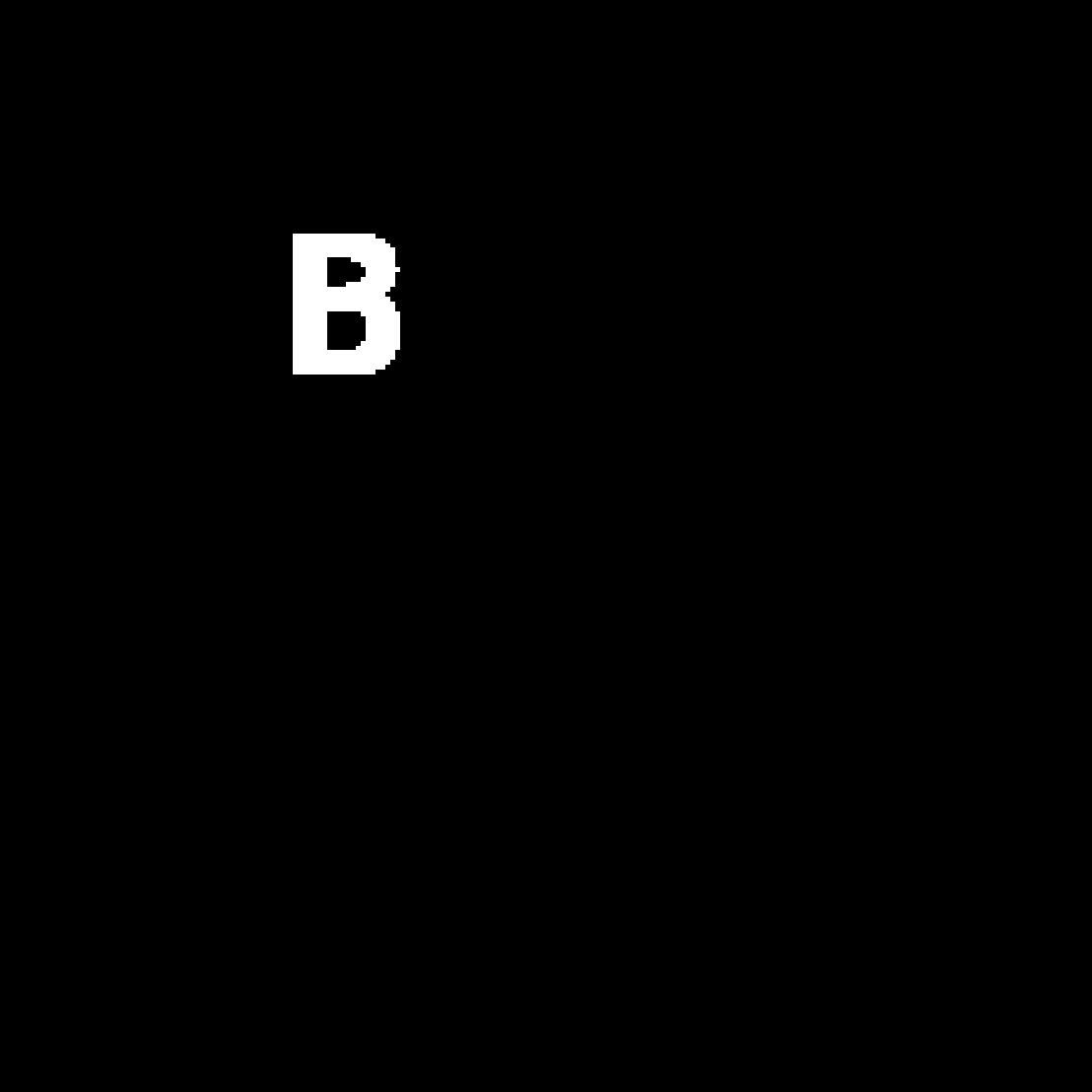}
\includegraphics[width=0.09\textwidth]{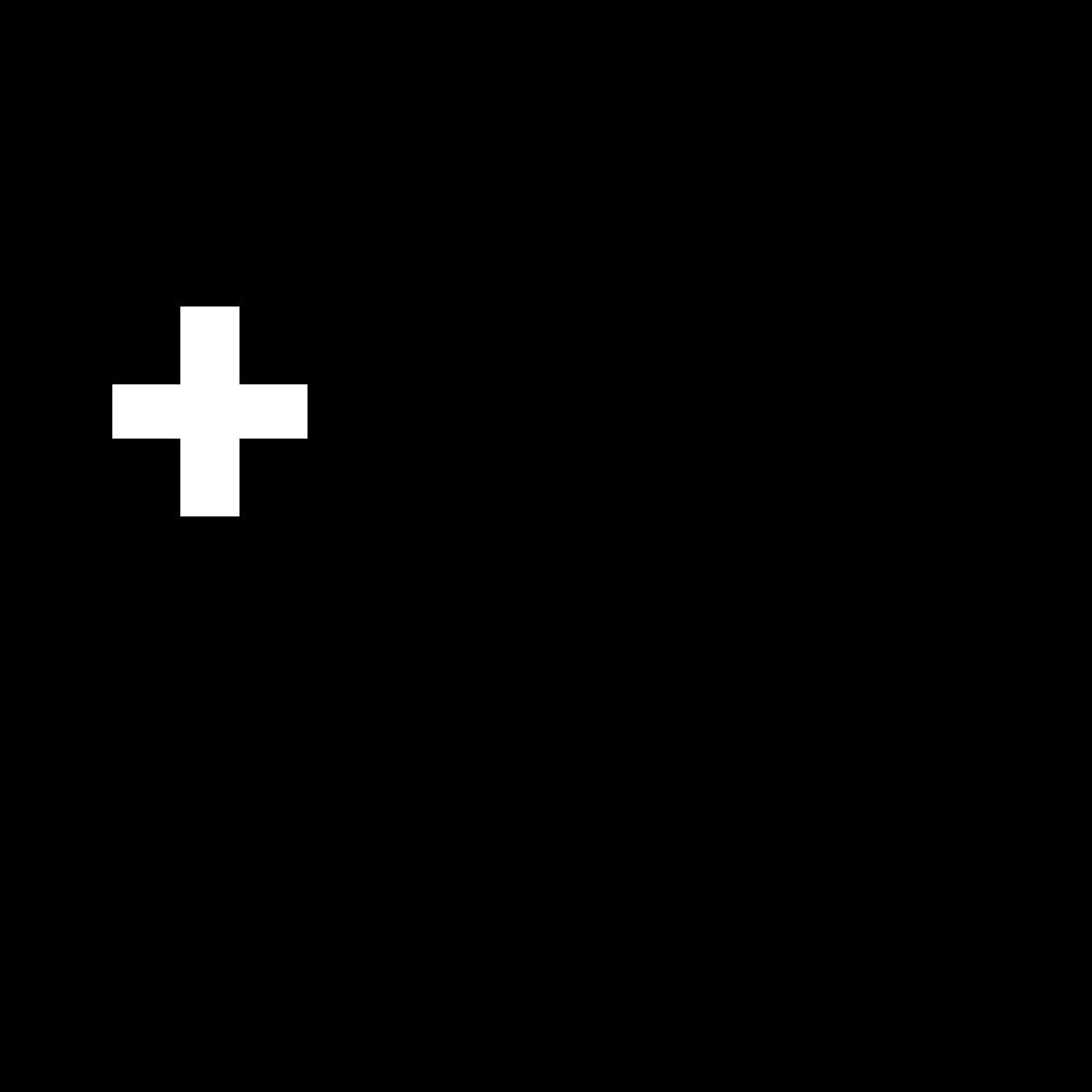}
\includegraphics[width=0.09\textwidth]{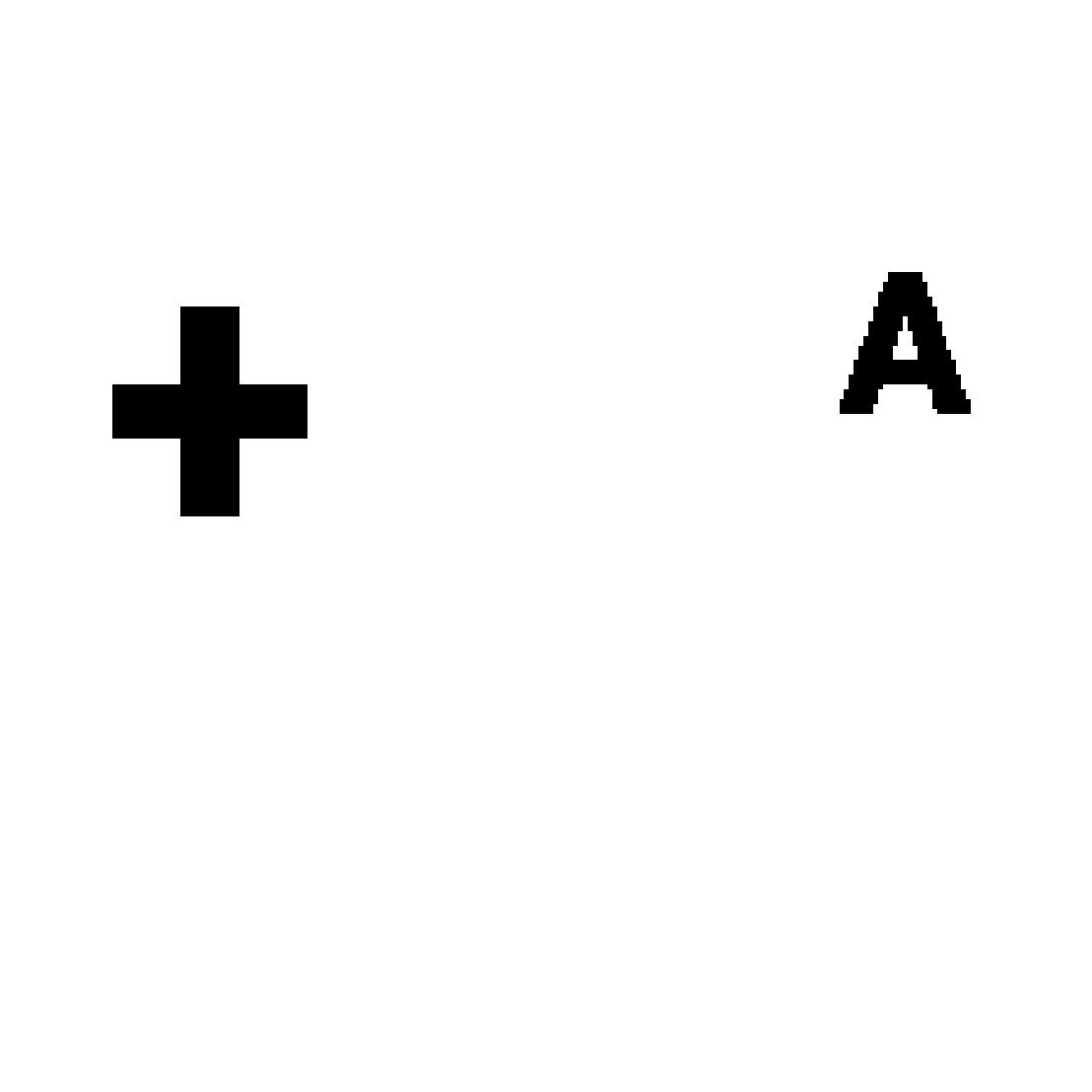}
}
\caption{Synthetic dataset AB. \ref{fig:example dataset AB A} and \ref{fig:example dataset AB B} show examples of classes \emph{A} and \emph{B}, respectively. \ref{fig:example dataset AB primitives} shows example masks $m_{x_i}^{p_o}$ for the primitives \emph{A}, \emph{plus}, \emph{background} and \emph{B} respectively.}
\label{fig:AB dataset examples}
\end{figure}

To mitigate the abovementioned phenomenon, we propose the usage of synthetic datasets, carefully balancing all features present in the data. Similarly, the synthetic datasets allow for the creation of ground truth annotations locating each element of the images. 
We focus on the low-complexity task of character classification, creating six synthetic datasets for the validation of CE techniques such as ECLAD (e.g., dataset AB in Figure \ref{fig:AB dataset examples}). The proposed synthetic datasets are described in detail in Appendix \ref{apd:synthetic datasets}.

We generate the images of each classification task by overlapping multiple elements. Each element (e.g., red A, a gray background) is the combination of multiple features (e.g., ``is red'', ``has the form of an A''), and is generated by a mask (denoted \emph{primitive}) filled with a specific texture. In each dataset, we select a subset of features and their primitives to define each class, marking them as important, and creating the labels. Unimportant features are balanced between classes. Thus, each datapoint is composed of an image $x_i$, their label $y_i$, and a mask $m_{x_i}^{p_o}$ for each primitive $p_o$.

\subsection{Model training}
\label{sec:val model training}
Once a dataset is generated, the next step in our validation process is to train a set of models. As the main assumption of this process is that each model learns perfectly the created datasets, we train each model until convergence, using a reduce-on-plateau learning rate scheduler. 

To ensure a broad testing of each concept extraction technique, we train multiple models with different architectures and random seeds. 
This ensures that the final aggregated results will explore the variations induced by the stochastic nature of the training, as well as the different flows of information in the latent space of different architectures.

In contrast with other validation procedures, our quantitative approach allows for a consistent comparison of CE algorithms. This leads to insights regarding the stability of the CE algorithms, and their generalization capabilities with respect to models architectures, minimizing confirmation biases when interpreting the obtained results.


\subsection{Concept extraction}

After obtaining each trained model, the next step is to execute each concept extraction algorithm that will be compared. This means, that each CE technique is used to analyze the trained model and dataset. Depending on the nature of the CE technique, it has to be executed for the complete dataset, or for each class separately. In the case of ACE-based methods, the execution is class specific, yielding a set of concepts for each class. Nonetheless, the concepts of all classes are aggregated and analyzed together. In contrast, ConceptShap and ECLAD are executed once for the complete dataset, generating a single set of concepts related to the general prediction process of the model.

As a result, a set of concepts is generated, with their respective vector representation, importance score, and example sets. In the case of ACE-based methods, this vector representation is the \emph{concept activation vector} (CAV) \citep{DBLP:conf/icml/KimWGCWVS18}. In the case of ConceptShap, each concept related to an axis in a lower dimensional projection between layers \citep{DBLP:conf/nips/YehKALPR20}. This axis can then be used as a vector representation of the concept. In our case, ECLAD provides a centroid $\gamma_{c_j}$ associated with every concept.

The representation vectors are used for concept localization, and the importance scores are used for correctness quantification. As long as these two are provided (importance score and vector representation), a CE technique can be validated using the proposed aproach.

\subsection{Concept localization}

The key idea behind the quantitative validation of CE techniques, is to be able to associate concepts and the features of a dataset. We perform this association based on a distance metric, comparing the visual cues related to both for each image in the dataset. To do so, we rely on the ground truth created for each image $x_i$ in the synthetic dataset, which contains a mask $m_{x_i}^{p_o}$ for each primitive $p_o$. Similarly, for each image $x_i$, we compute the mask $m_{x_i}^{c_j}$, locating the visual cues related to each concept $c_j$.

We use the results of the previous step to localize each extracted concept $c_j$ in each image $x_i$ of the synthetic dataset. For ACE, we perform concept localization by segmenting each image and testing whether each patch contains a concept. For ConceptShap, we upscale the projected lower dimensional space to the original shape of the input image $x_i$, and obtain a mask for each dimension (which related to the concepts). In the case of ECLAD, we use the descriptor $d_{x_i}$ of every image, and compute a mask $m_{x_i}^{c_j}$ for every concept, as described in Equation \ref{eq:mask pixels}. As a result of this step, every tested CE approach generates a binary mask $m_{x_i}^{c_j}$ for each concept $c_j$ of each model, for every image $x_i$ in a dataset $E$.

\subsection{Concept association}

The process of associating concepts and the important features of a dataset has previously been performed through human inspection \citep{DBLP:conf/nips/GhorbaniWZK19,DBLP:conf/cvpr/GeXXZKCIW21}. This association allows the comparison of extracted concepts and the dataset's intended features. It allows for a subjective judgement of the correctness of the CE methods.
To perform this association automatically, we introduce the distance $\mathrm{DST}_{p_o,c_j}$, measuring how close a concept is to the features of a dataset. Intuitively, if a concept $c_j$ and a primitive $p_o$ are located on the same regions, consistently through a dataset (e.g. overlapping), the spatial distance $\mathrm{DST}_{p_o,c_j}$ will be small. In this case, we consider that the concept $c_j$ and primitive $p_o$ are spatially associated.

To measure the \textit{spatial association} of a concept and a feature, we consider partial overlapping as well as spatial closeness. We compute this distance through the comparison of the concept masks $m_{x_i}^{c_j}$ and the primitives $m_{x_i}^{p_o}$ of the features. In cases where a concept detects the surrounding of a primitive (when activation maps become off-centered through a CNN), existing metrics (e.g. Jaccard index, adjusted rand score) perform poorly, this limitations are discussed in the Appendix \ref{apd:Distance metric}.
We propose an expressive metric, computed by adding the Euclidean distance between each pixel on a mask to the nearest element of another mask. This metric results in a zero value if the mask of the primitive and the concept are overlapping, and increases as the masks separate. 
We compute a one-way distance between the mask $m_{x_i}^{p_o}$ of a primitive $p_o$ and the mask $m_{x_i}^{c_j}$ of a concept $c_j$ as
\begin{equation}
\mathrm{dst}_{p_o,c_j}(x_i) = 
\mathrm{sum}\left ( 
m_{x_i}^{p_o} \odot \mathrm{EDT}(m_{x_i}^{c_j\prime})
\right ),
\label{eq:distance}
\end{equation}
where $\mathrm{EDT}()$ refers to the Euclidean distance transform; $m_{x_i}^{c_j\prime}$ is the negated mask $m_{x_i}^{c_j}$; and $\odot$ denotes the element-wise multiplication of matrices. We estimate the association distance between $c_j$ and $p_o$ by computing the average two-way distance for $E$,  
\begin{equation}
\mathrm{DST}_{p_o,c_j} = 
\frac{1}{|E|}\sum_{x_i \in E} \mathrm{dst}_{p_o,c_j}(x_i) + \mathrm{dst}_{c_j,p_o}(x_i).
\label{eq:distance}
\end{equation}
Using this distance, we associate each concept to its closest primitive, $
p_{c_j} = 
\underset{p_o}{\mathrm{argmin}}(\mathrm{DST}_{p_o,c_j}).
$
Finally, we use this association to classify each concept as \textit{aligned} if $p_{c_j}$ is an important primitive and $\mathrm{DST}_{p_{c_j},c_j}$ is below a defined threshold $t_{\mathrm{DST}}$, or otherwise as \textit{unaligned}. The usage of $t_{\mathrm{DST}}$ reduces the number of aligned concepts lacking semantic meaning. 

\subsection{Correctness quantification}
\label{sec:val correctness quantification}
Based on an ideal case, we assume that the models learn a subset of the important features of the dataset when trained. Thus, the resulting extracted concepts must be aligned with both, the visual cues and the intended importance of the features (primitives) of the dataset. To quantify both properties, we introduce the metrics of \textit{representation correctness} and \textit{importance correctness}.

\textbf{Representation correctness} measures how spatially close are the visual cues of extracted \textit{aligned} concepts in comparison to the important features of the dataset. This measure is meant for the comparison of how specific are the (localized) explanations of different CE methods.
We compute the \textit{representation correctness}, as the negative average association distance of all \textit{aligned} concepts extracted from all models:
\begin{equation}
\mathrm{RC_{\mathrm{CE}}} = 
\frac{1}{|C_a|}\sum_{c_j \in C_a}
-\mathrm{DST}_{p_{c_j},c_j} \, ,
\label{eq:representation correctness}
\end{equation}
where $C_a$ denotes the set of all aligned concepts extracted from a model in the validation process. 
In an ideal case, the value of $\mathrm{RC_{\mathrm{CE}}}$ would be zero, meaning that there is a subset of extracted concepts which perfectly represents the important features of the datasets learned by the model. Nonetheless, models aggregate information through their layers, and activations associated to a feature can become off-centered or dilated in space.
This translates to lower representation correctness for different architectures, which still reflect the internal representations learned by the models.

\textbf{Importance correctness} measures how aligned are the importance scores of the extracted concepts in comparison to the intended importance of their related features (primitives). 
This means, extracted concepts which are spatially related to meaningful features of the dataset (\textit{aligned} concepts) should be score as important (e.g. 1.0).
In contrast, extracted concepts spatially related to meaningless features of the dataset (\textit{unaligned} concepts) should be scored as unimportant (e.g. 0.0).
To quantify the \textit{importance correctness}, we compute the average absolute importance of all aligned concepts $C_a$ minus the average absolute importance for the unaligned concepts $C_u$. We then normalize by the maximum importance of all concepts:
\begin{equation}
\mathrm{IC_{\mathrm{CE}}} = 
\frac{1}{\underset{c_q \in C_a \cup C_u}{\mathrm{max}}\left (\left | \mathrm{I}_{c_q} \right | \right)}
\left (
\frac{1}{|C_{a}|} \sum_{c_j \in C_a} \left | \mathrm{I}_{c_j} \right | -
\frac{1}{|C_{u}|} \sum_{c_j \in C_u} \left | \mathrm{I}_{c_j} \right |
\right ) \, ,
\label{eq:importance correctness}
\end{equation}
where $\left | \mathrm{I}_{c_j} \right |$ refers to the absolute value of the importance of $c_j$. In the case of ECLAD, we use the relative importance score $\mathrm{I}_{c_j}=\mathrm{RI}_{c_j}$. For ConceptShap, we use the Shapley values associated with each concept. Finally, for ACE, we scale the sensitivity score, $\mathrm{I}_{c_j}=2 \times \mathrm{TCAV}_Q - 1$, so that unimportant concepts have a value of 0, and important concepts have a value of 1 or -1.

In an ideal case, the value of $\mathrm{IC_{\mathrm{CE}}}$ would be close to 1.0, meaning that \textit{aligned} concepts were scored as important (close to 1.0), and \textit{unaligned} concepts were scored as unimportant (close to 0.0). Nonetheless, correlated important features can lead to shortcut learning, which can directly impact the representation of the model and their functional importance during the prediction process. This phenomenon can lead to obtaining $\mathrm{IC_{\mathrm{CE}}}$ scores lesser than 1.0, when evaluating the subsequent concept extraction.

The resulting evaluation metrics provide a consistent and quantitative score for the comparison and validation of CE techniques, which was previously lacking in related works. The current process requires synthetic datasets with pixel-wise annotations to mitigate common pitfalls of models learning. Nonetheless, this process can be extended with carefully obtained real-world datasets also containing pixel-wise annotations of the different components on the images.
\section{Results}
\label{sec:Results}

In this section, we present experimental results for our method ECLAD in comparison with ACE and ConceptShap.
For the validation and comparison of our method, we performed a series of experiments, considering six synthetic datasets, two industrial datasets with ground truth annotations (subsets of the MVTec-AD dataset \citep{DBLP:journals/ijcv/BergmannBFSS21}), five CNN architectures (ResNet-18, ResNet-34, DenseNet-121, EfficientNet-B0, and VGG16 \citep{DBLP:journals/corr/HeZRS15, DBLP:journals/corr/HuangLW16a, DBLP:conf/icml/TanL19, DBLP:journals/corr/SimonyanZ14a}), 20 random seeds, and the three CE methods mentioned above. Each experimental run was performed as described in Sections \ref{sec:val model training} to \ref{sec:val correctness quantification}. 

Experimental runs were executed for the combinations of datasets, models, CE methods and random seeds. Then, the results were aggregated for all random seeds, and are presented per dataset, model, and CE method. We provide a comparison of the performance of the different CE methods based on the introduced metrics $\mathrm{RC_{\mathrm{CE}}}$ and $\mathrm{IC_{\mathrm{CE}}}$. More details on the experimental setup are provided in Appendix \ref{apd:experimental setup}.

The main findings of the experiments can be summarized as follows:
\begin{itemize}
    \item Concepts extracted with ECLAD are closely related to the relevant features that the models learn. This was measured through a high representation correctness across all experiments.
    \item Importance scores provided by ECLAD outperform those of ACE and ConceptShap in all experiments. The importance of concepts related to relevant features of the dataset are scored high, and irrelevant concepts are scored low.
    \item ECLAD explanations provide reasonable and meaningful insights in real world scenarios where understanding models is critical.
    \item Our validation procedure allows for a quantitative comparison of CE methods, providing consistent metrics which reflect the performance of the CE techniques.
\end{itemize}

We report the key findings of our experiments in the sections below. To do so, we first analyze a run of each CE method on an example case, using the \textit{AB} synthetic dataset (see Figure \ref{fig:scatter results}). Then, we provide the aggregated results of the performance metrics for all random seeds of representative datasets. Next, we provide an example of the execution of ECLAD in a real world use case (without annotations). Finally, we discuss the method's performance and limitations.

\subsection{Example case}

This subsection presents the results of executing ECLAD, ACE, and ConceptShap, over a single ResNet-34 trained on the \textit{AB} synthetic dataset. This example provides representative insights which generalize to the rest of the experiments. 

The AB dataset was described in Section \ref{sec:synth datasets}, and it's a procedurally generated dataset of two classes. Class \textit{A} is composed of images including the character ``A'', and class \textit{B} is composed of images containing the character ``B''. 
An intrusive element was added to all images in the form of a character ``+''. 
In an ideal case, a trained model will learn to differentiate both primitives (characters ``A'' and ``B'') and use them in its prediction process. 
Consequently, the extracted concepts should also contain the relevant characters ``A'' and ``B'', scoring a subset of them as important. 
In addition, if a concept is extracted related to the character ``+'', it must be scored with an importance close to 0.0.

\begin{figure}[h!]
\centering
\subfigure[Concepts extracted using ECLAD]{
\label{fig:scatter ECLAD}
\includegraphics[width=0.41\textwidth]{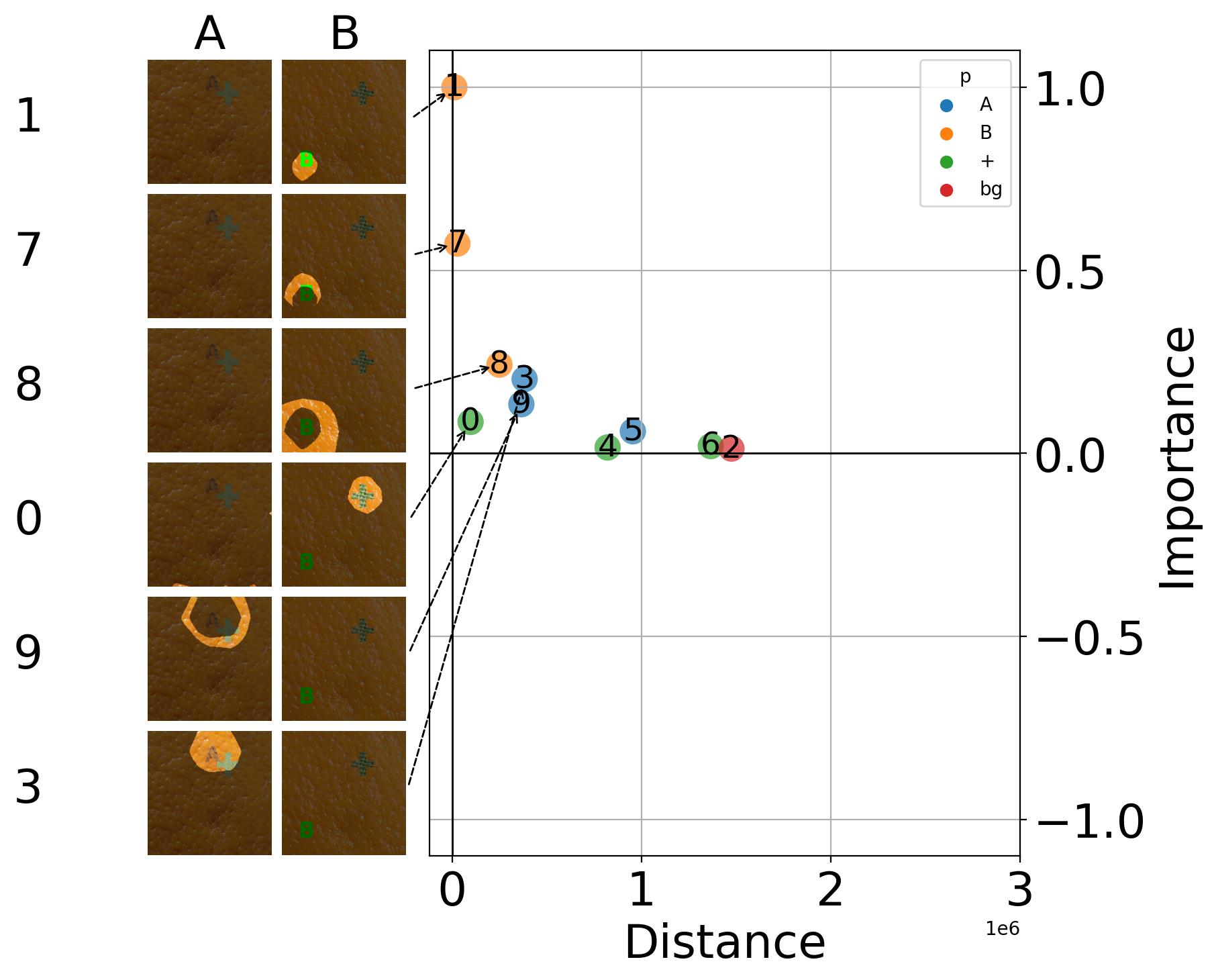}
}

\subfigure[Concepts extracted using ACE.]{
\label{fig:scatter ACE}
\includegraphics[width=0.41\textwidth]{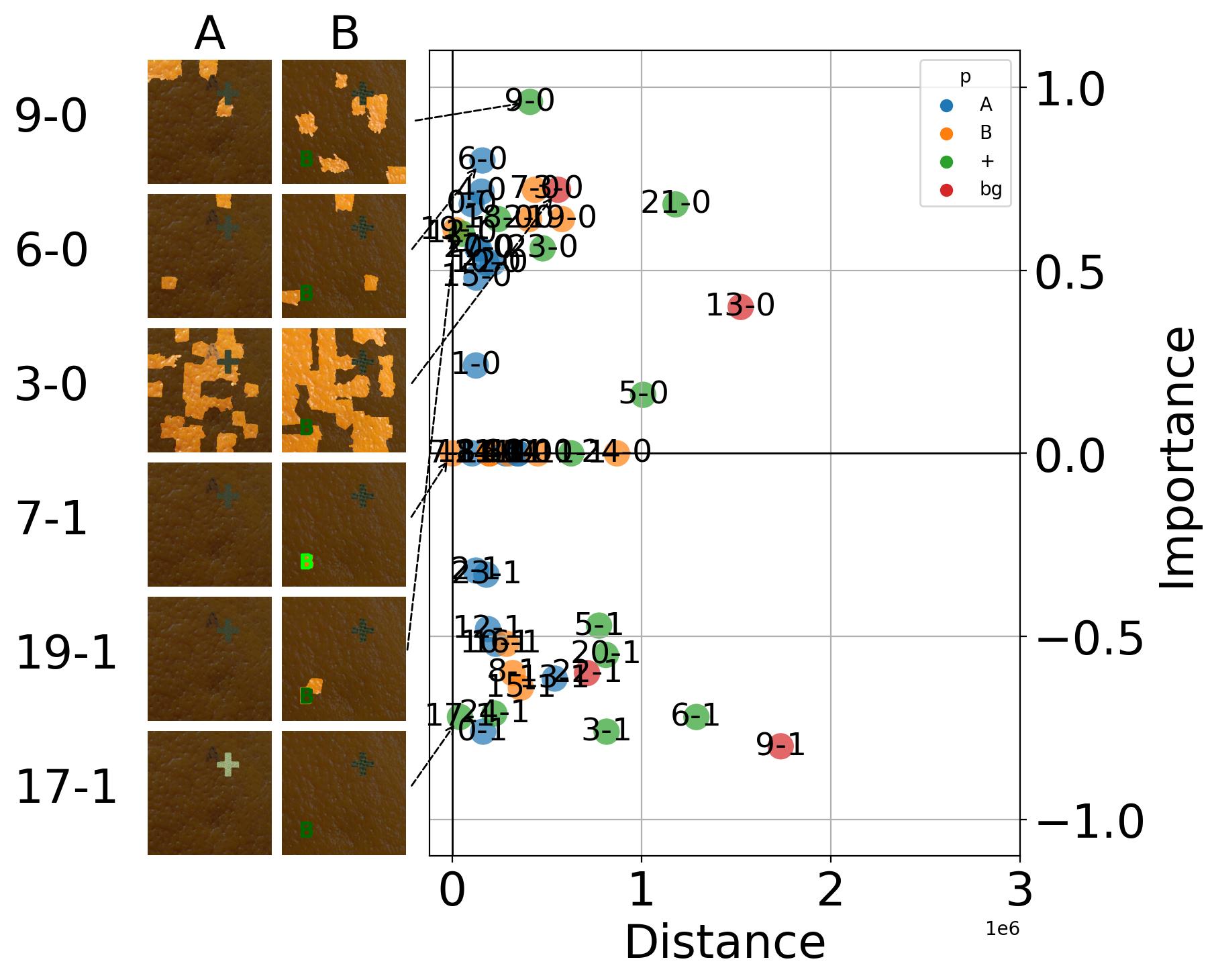}
}

\subfigure[Concepts extracted using ConceptShap]{
\label{fig:scatter ConceptShap}
\includegraphics[width=0.41\textwidth]{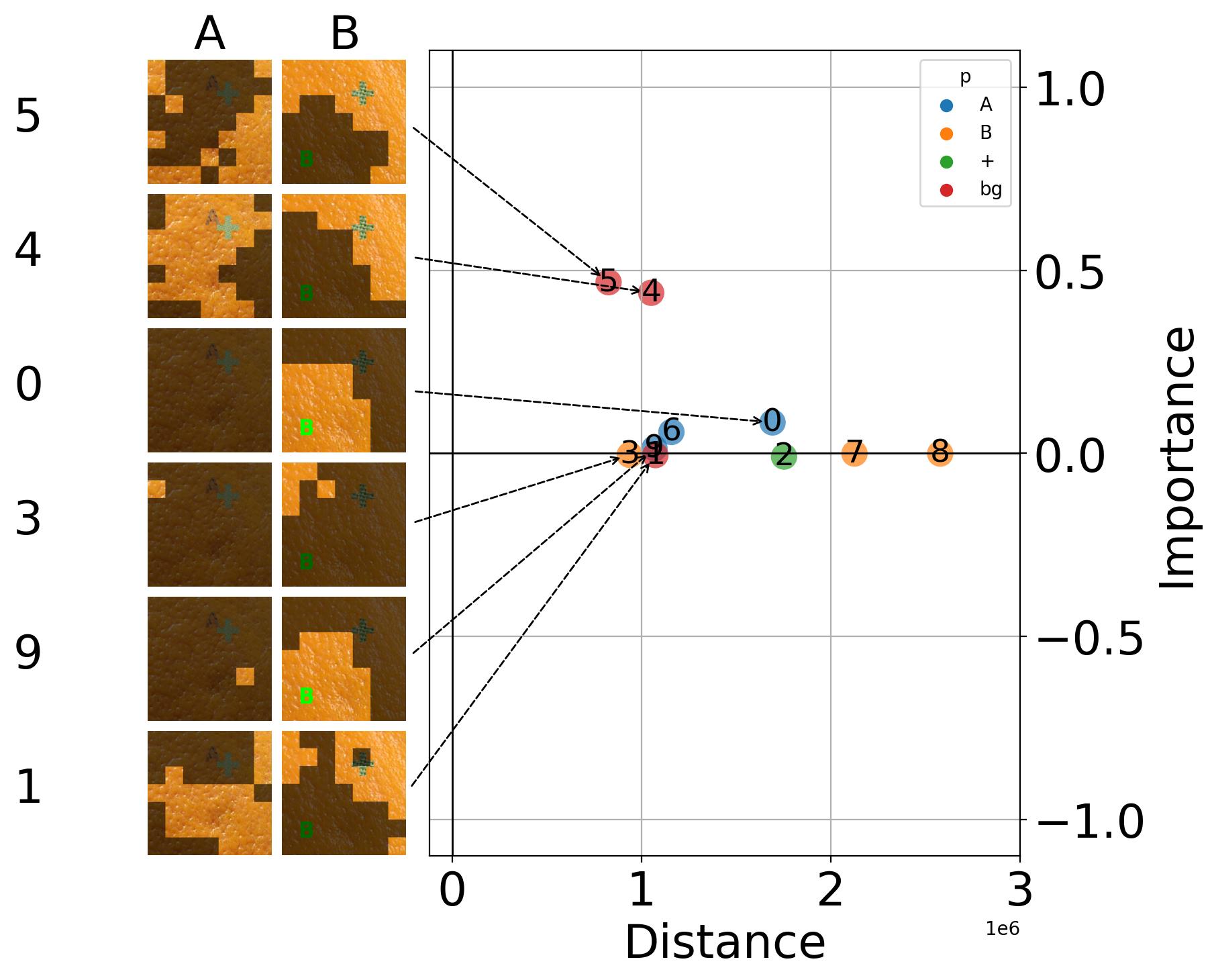}
}
\caption{
Concepts extracted from a ResNet-34 trained on the \textit{AB} dataset. The extracted concepts from each CE method are plotted in relation to their importance (y-axis), and the distance (x-axis) towards their closest primitive (hue). In an ideal case, important concepts will be closely related to the important primitives (e.g., $c_1$ in \ref{fig:scatter ECLAD}), while concepts unrelated to important primitives will be scored as unimportant (e.g., $c_0$ \ref{fig:scatter ECLAD}).
}
\label{fig:scatter results}
\end{figure}

In Figure \ref{fig:scatter results}, we provide a visual representation of the results of executing ECLAD, ACE, and ConceptShap on the trained ResNet-34. Each subfigure contains a scatter plot on the right and a set of examples for selected concepts on the left. Each datapoint on the scatter plots refers to an extracted concept, located based on its importance score (y-axis), and their distances (x-axis) towards the closest primitive (hue). On the left, each row represents a concept, starting with the concept identifier on the first column, and providing examples of the concepts for each class in the second and third columns.

On a perfect case, extracted concepts should be close to either the x or y axis. Extracted concepts which relate to relevant features (\textit{aligned} concepts), should be located close to the y-axis, and a subset of them must be scored as highly important (close to the coordinate $(0,1)$).
Concepts unrelated to relevant features (\textit{unaligned} concepts), must be located close to the x-axis, scored as unimportant by the CE methods.
In comparison with ACE and ConceptShap, the results of ECLAD (seen in Figure \ref{fig:scatter ECLAD}) closer resemble the ideal CE scenario.

\textbf{Concepts extracted with ECLAD closely relate to the relevant features of the dataset, while being directly related to the latent representations of the analyzed model}. 
In Figure \ref{fig:scatter ECLAD}, concepts $c_1$, $c_3$, and $c_0$ correspond to the main features of the dataset (characters ``B'', ``A'', and ``+'' respectively). In addition, these concepts were extracted directly from the activations of the model (though LADs). Meaning that the model learned to differentiate these visual cues through the training process.
Figure \ref{fig:scatter ACE} shows the concepts extracted by ACE, where $c_{7-1}$ relates to the character ``B'' and $c_{17-1}$ relates to ``'+'. Nonetheless, there was no associated concept related to the character ``A''. Moreover, multiple concepts (e.g. $c_{9-0}$ and $c_{6-0}$) are related to random patches of the same background. This behavior can be caused by biases induce by the segmentation and encoding process, or when obtaining a non-centered latent representation of concepts. As a consequence, ACE concepts do not directly relate to how a model encodes or learns to differentiate regions on the images. 
Figure \ref{fig:scatter ConceptShap} shows the concepts extracted by ConceptShap, where $c_0$ and $c_9$ relate to ``B'', and there are no concepts related to either ``A'' or ``+''. These concepts correspond to the axis of a lower dimensional projection, which maintain the performance of the model. Thus, this projection can also suffer from shortcut learning, disregarding important insights about the model. If the projection contains ``B'', it is enough to maintain the performance of the model, disregarding the existance of concepts related to ``A'', even if they exist within the latent space of the CNN. Similarly, as ConceptShap assumes each axis of its projection contains a disentangled representation, which is not necessarily the case, as seen in $c_0$ and $c_9$.
As a highlight, ECLAD extracts concepts directly from local representations (LADs), which provides better insights regarding how a model encodes an image, and what it learns to differentiate to perform a prediction. This significantly mitigates issues generated by the segmentation of input images or learning projection, which may not directly reflect the behavior of a model.

\textbf{Importance scores provided by ECLAD are more aligned with the intended features of the datasets, while being directly related to the local sensitivity of the model}.
In Figure \ref{fig:scatter ECLAD}, \textit{unaligned} concepts ($c_0$, $c_4$, $c_6$, and $c_2$) are scored with low importance (close to 0.0). Whereas a subset of the \textit{aligned} concepts ($c_1$, and $c_7$) are scored as important. Moreover, these importance scores are a direct aggregation of the sensitivity of the regions containing the concepts, directly reflecting the decision process of the model.
Figure \ref{fig:scatter ACE} shows how both \textit{aligned} (e.g. $c_{7-1}$) and \textit{unaligned} (e.g. $c_{9-0}$) concepts can be wrongly scored as either important or not. This problem can arise, as the TCAV scores do not take into account the magnitude of the sensitivity of the prediction with respect to the concept vectors, but only the proportion of images (for a class) where these sensitivities are positive. Thus, marginal sensitivities (or biases) of the model with respect to concepts can translate into significant changes on the scoring of their importance.
In Figure \ref{fig:scatter ConceptShap}, \textit{aligned} concepts $c_0$ and $c_1$ are scored as unimportant and \textit{unaligned} concepts $c_4$ and $c_5$ are scored as important. This issue can be the result of either a learned artifact during the concept extraction process which is prone to shortcut learning, or the known limitation of shapley values (used by ConceptShap) when scoring co-ocurring or complimentary elements.
As a highlight, ECLAD importance scoring relies on the aggregated sensitivity of the visual cues of each concept. This process takes into account not only the sign of the sensitivity, but also the magnitude and location, which mitigates limitations of previously proposed importance scoring processes.

\subsection{Aggregation across multiple runs}

\begin{figure*}[h!]
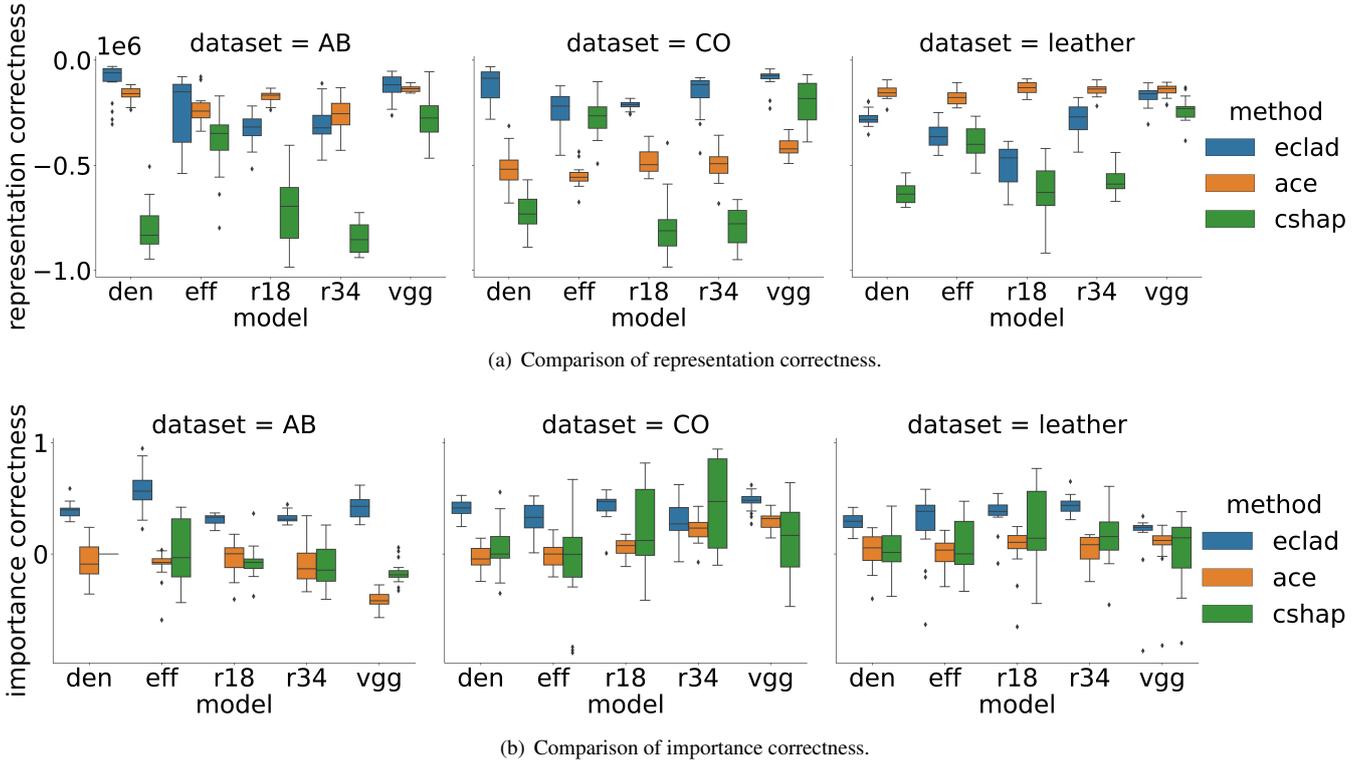

\centering
\subfigure[Comparison of representation correctness.]{
\label{fig:boxplot RC}
\includegraphics[width=0.98\textwidth]{boxplot_representation_correctness_t100E06_full}
}

\subfigure[Comparison of importance correctness.]{
\label{fig:boxplot IC}
\includegraphics[width=0.98\textwidth]{boxplot_importance_correctness_t100E06_full}
}

\caption{
Comparison of representation correctness and importance correctness for five models (DenseNet-121, EfficientNet-B0, ResNet-18, ResNet-34, and VGG16) trained on the AB and CO synthetic datasets. An ideal CE method will have a representation correctness (negative distance between aligned concepts and important primitives) close to zero, and an importance correctness (relative difference between the importance of aligned and unaligned concepts) close to one. In all plots, higher is better.
}
\label{fig:boxplots}
\end{figure*}

Experimental runs were executed for the combinations of datasets, models, and random seeds.
We selected a subset of the datasets (synthetic datasetAB, synthetic dataset CO, and leather subset from MVTec), which provide representative results and insights generalizable to our study.
The aggregated results for these datasets are shown in Figure \ref{fig:boxplots}.

The aggregated representation correctness results are shown in Figure \ref{fig:boxplot RC}. The runs of each dataset and model architecture were aggregated for all 20 random seeds.
As a general insight, \textbf{ECLAD representation correctness is comparable with ACE and consistently better than ConceptShap, while directly relating concepts to their local representations within the latent space of models.}.
ConceptShap representation lack spatial information (e.g. $c_0$ and $c_9$ in Figure \ref{fig:scatter ConceptShap}), and are consistently worse than ECLAD and ACE. 
This is caused by the loss of information during the projection process of the concept extraction.
In contrast, ACE representations are obtained using SLIC, thus, it performs better in simple cases, where relevant features have a clear boundary (e.g. leather dataset, or character ``B'' in Figure \ref{fig:scatter ACE}). 
Finally, ECLAD representations directly reflect the local encodings of CNNs, which provides a consistent extraction of the relevant features across all cases.
In particular, ECLAD performs better in cases where features are not as clearly defined (e.g. character ``A'' in Figure \ref{fig:scatter ECLAD}, or the right trace differentiating a ``O'' from an ``C''), reflecting the visual cues which the models learn to differentiate.

The aggregated importance correctness results are shown in Figure \ref{fig:boxplot IC}. As a general insight, \textbf{ECLADs importance scoring allows for a better differentiation between aligned and unaligned concepts, providing a consistently better importance correctness.}
In the case of ConceptShap, the Shapely value of inversely correlated concepts (e.g., characters ``A'' and ``B''), can be truncated, independent of the extent of the activations and actual contribution of a region to a prediction (e.g., $c_4$ and $c_9$ in Figure. \ref{fig:scatter ConceptShap}).
In contrast, ACE relies on TCAV scores, assuming that each concept is encoded in a distinctive direction in the latent space of a network. This assumption can entangle spurious patterns with a concept (e.g., random patches and the character ``A'' in $c_{3-0}$ in Figure \ref{fig:scatter ACE}). 
These limitations of both ACE and ConceptShap are severe, and translate in a consistent erroneous scoring of \textit{unaligned} or \textit{aligned} concepts, quantified in Figure \ref{fig:boxplot IC}.
ECLAD bootstraps the importance scores using the sensitivity of the regions containing the visual cues of the concepts.
This allows for a consistent, more stable, and reliable scoring of concept importance.
This can be seen in Figure \ref{fig:scatter results}, where the most important concepts extracted by ECLAD ($c_1$ and $c_7$) are closely associated with the features that should be important for the model.
Globally, this is reflected in ECLAD having a larger importance correctness score for most datasets and model combinations, as seen in Figure \ref{fig:boxplot IC}.

\subsection{Real world use cases}

\begin{figure*}[h!]
\centering
\subfigure[Class 0 - No DR.]{
\label{fig:APTOS dataset class 0}
\includegraphics[height=0.09\textheight]{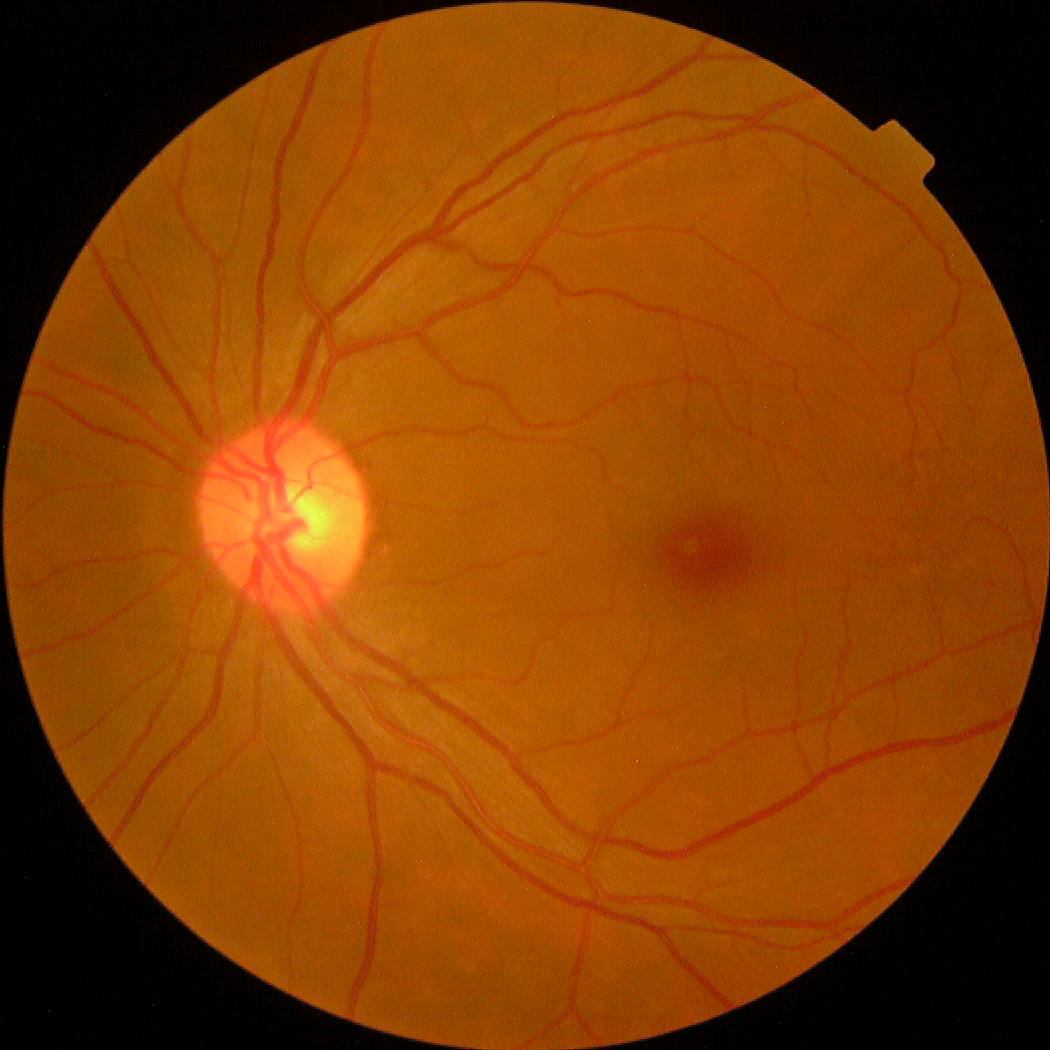}
\includegraphics[height=0.09\textheight]{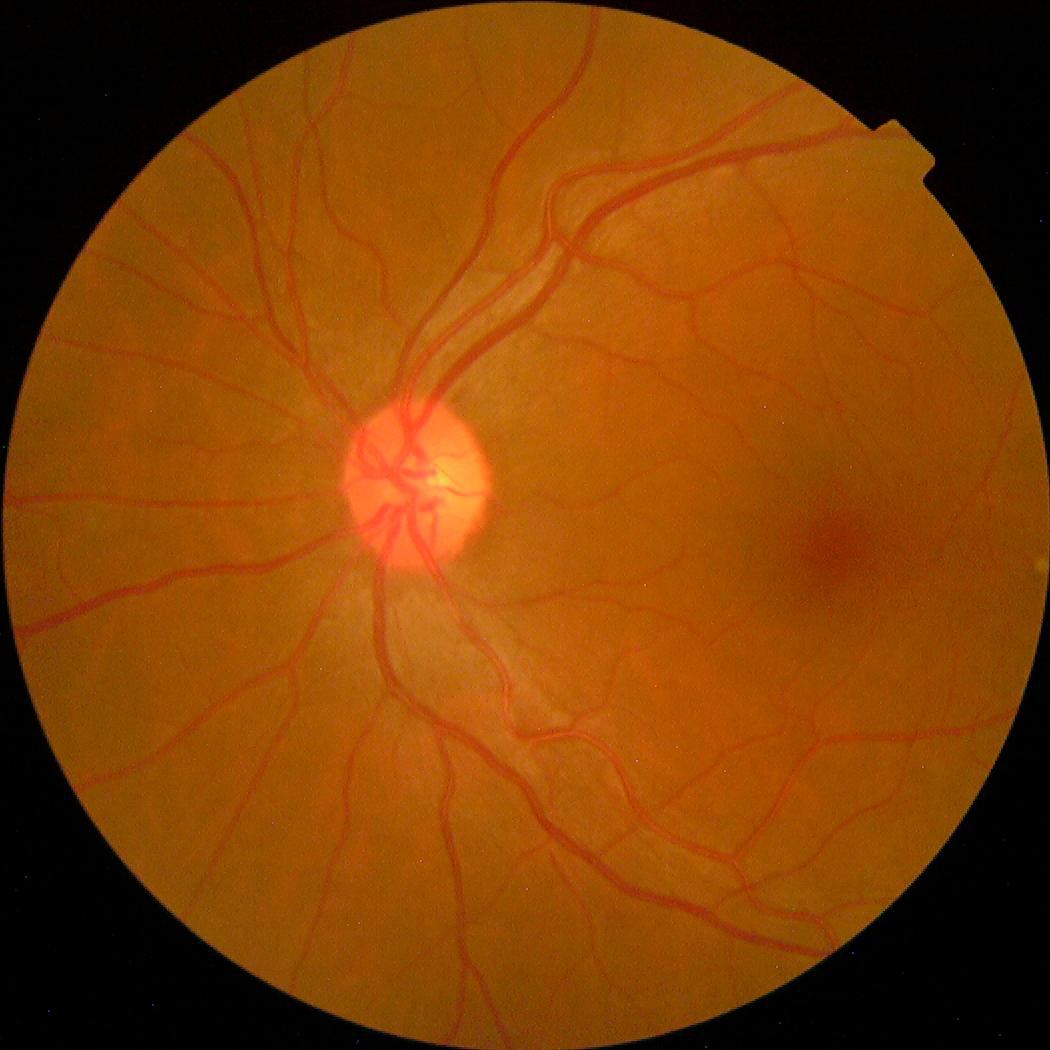}
}
\subfigure[Class 1 - Mild.]{
\label{fig:APTOS dataset class 1}
\includegraphics[height=0.09\textheight]{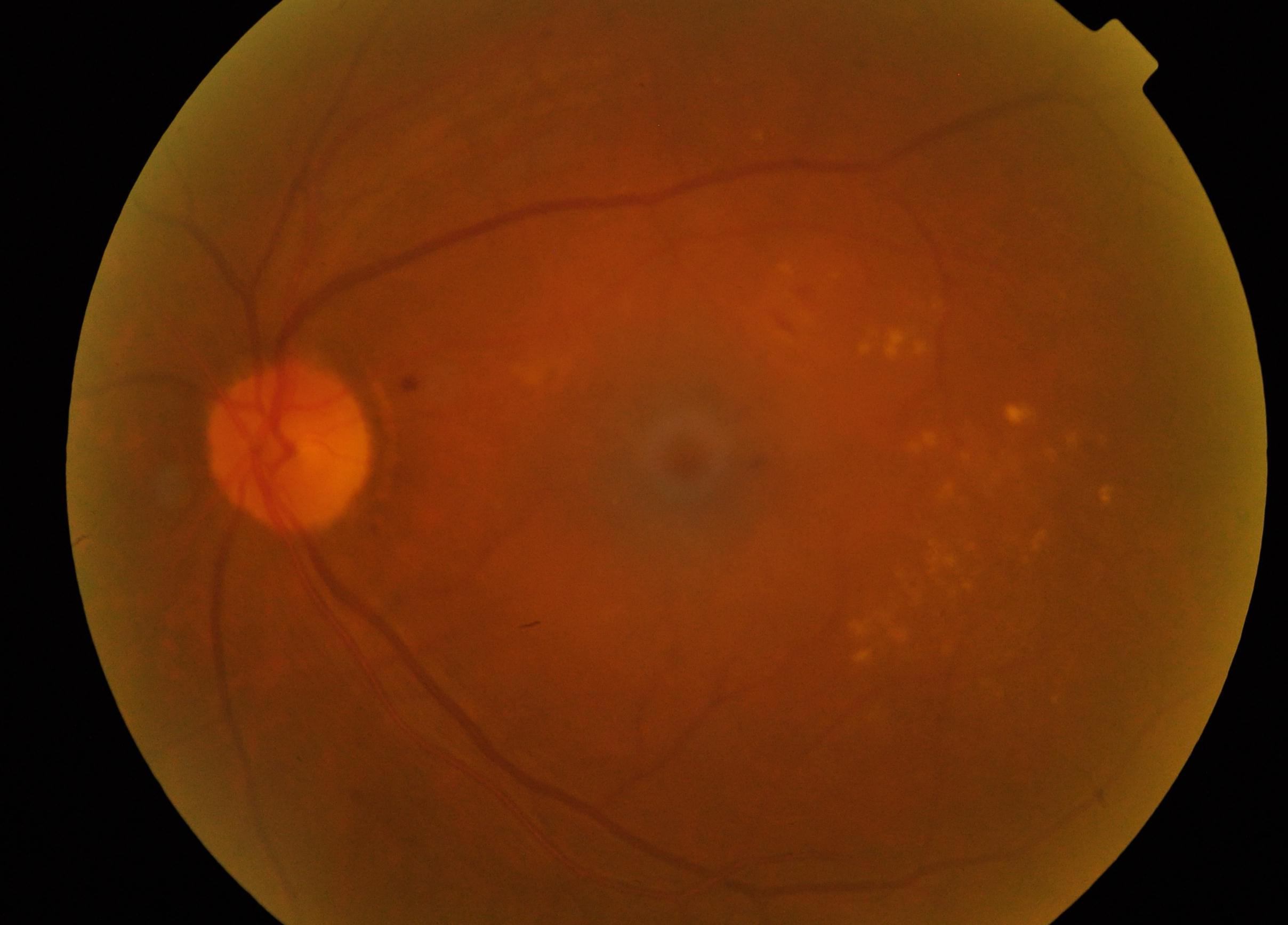}
\includegraphics[height=0.09\textheight]{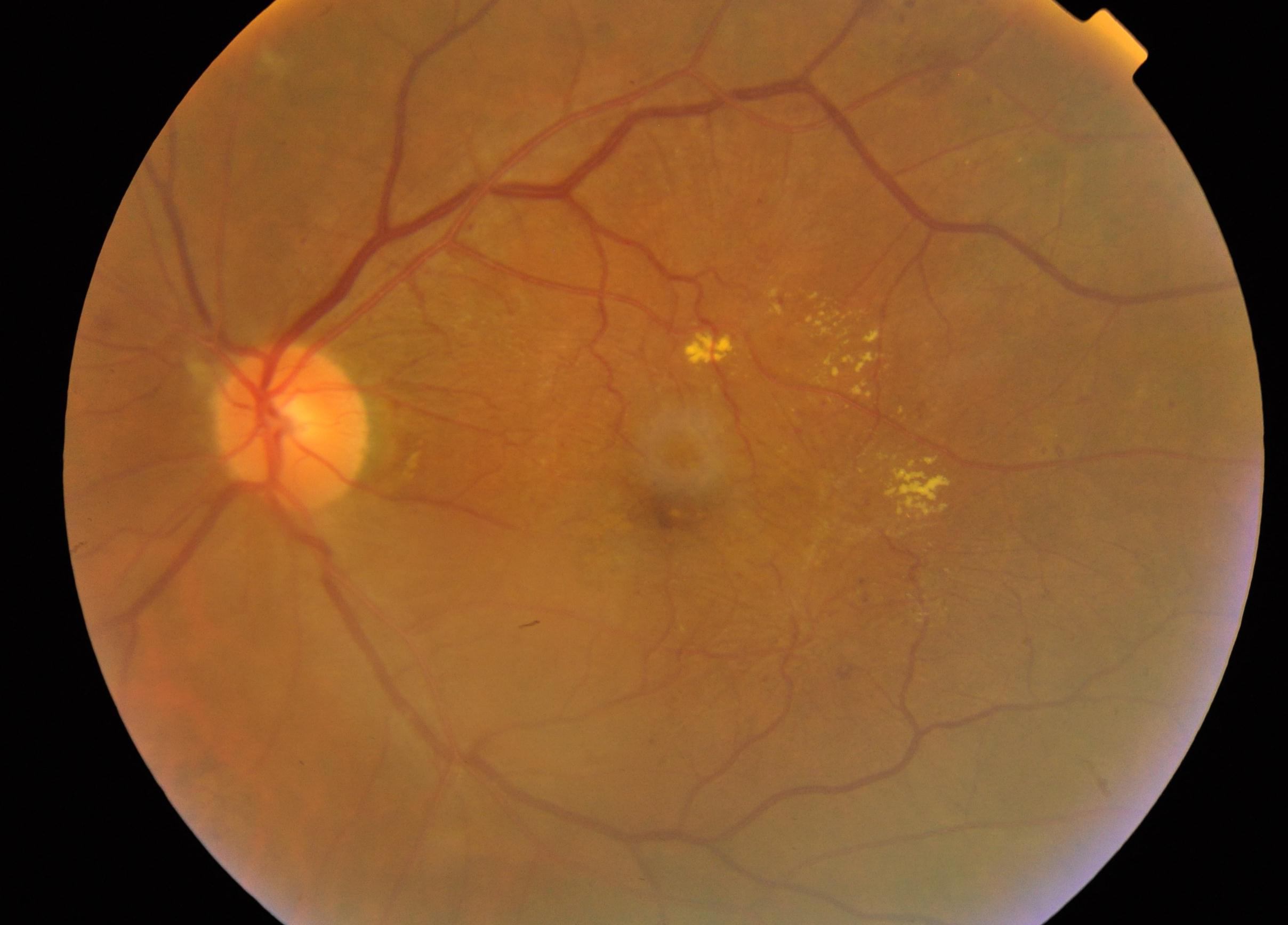}
}
\subfigure[Class 2 - Moderate.]{
\label{fig:APTOS dataset class 2}
\includegraphics[height=0.09\textheight]{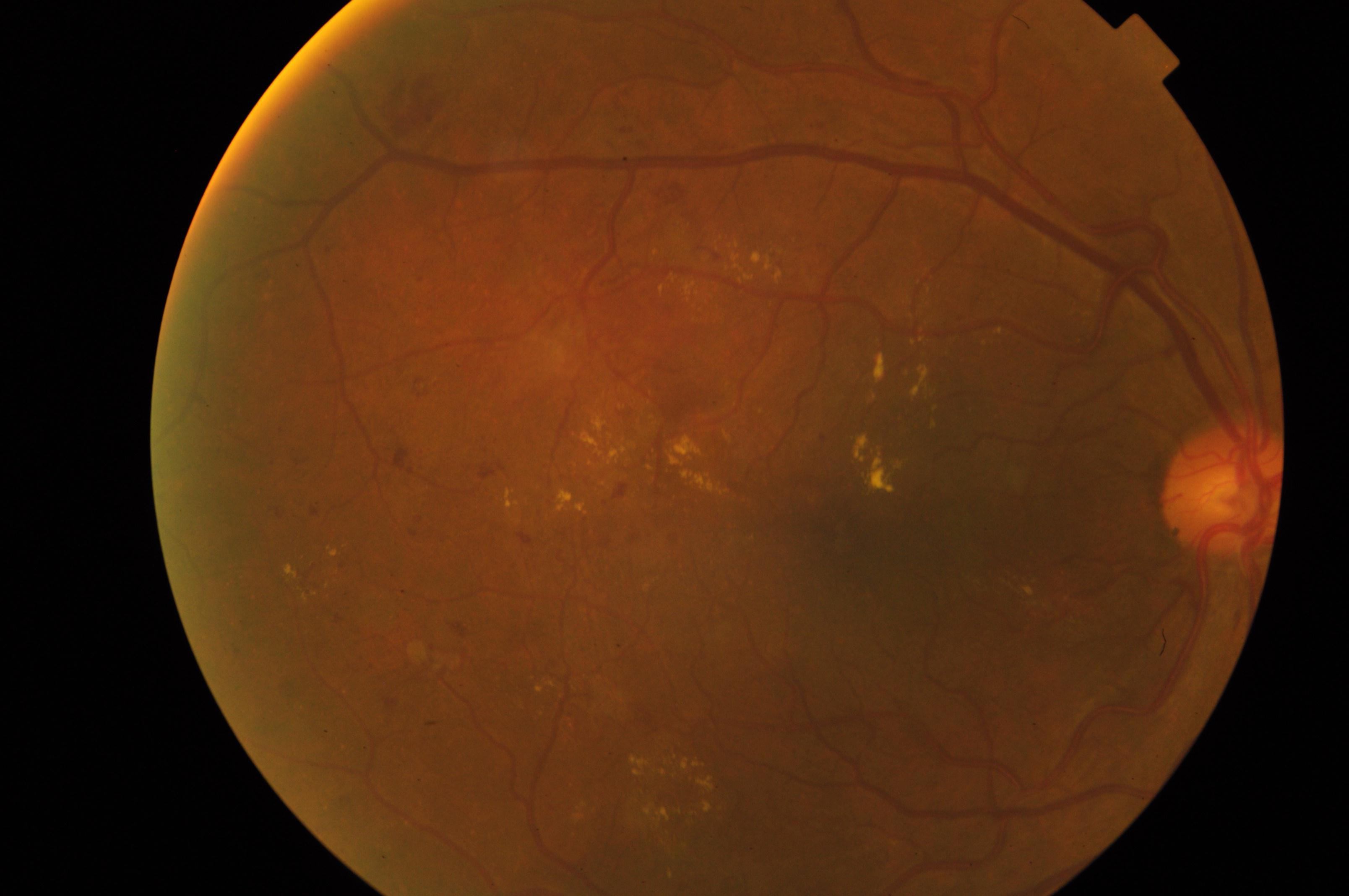}
\includegraphics[height=0.09\textheight]{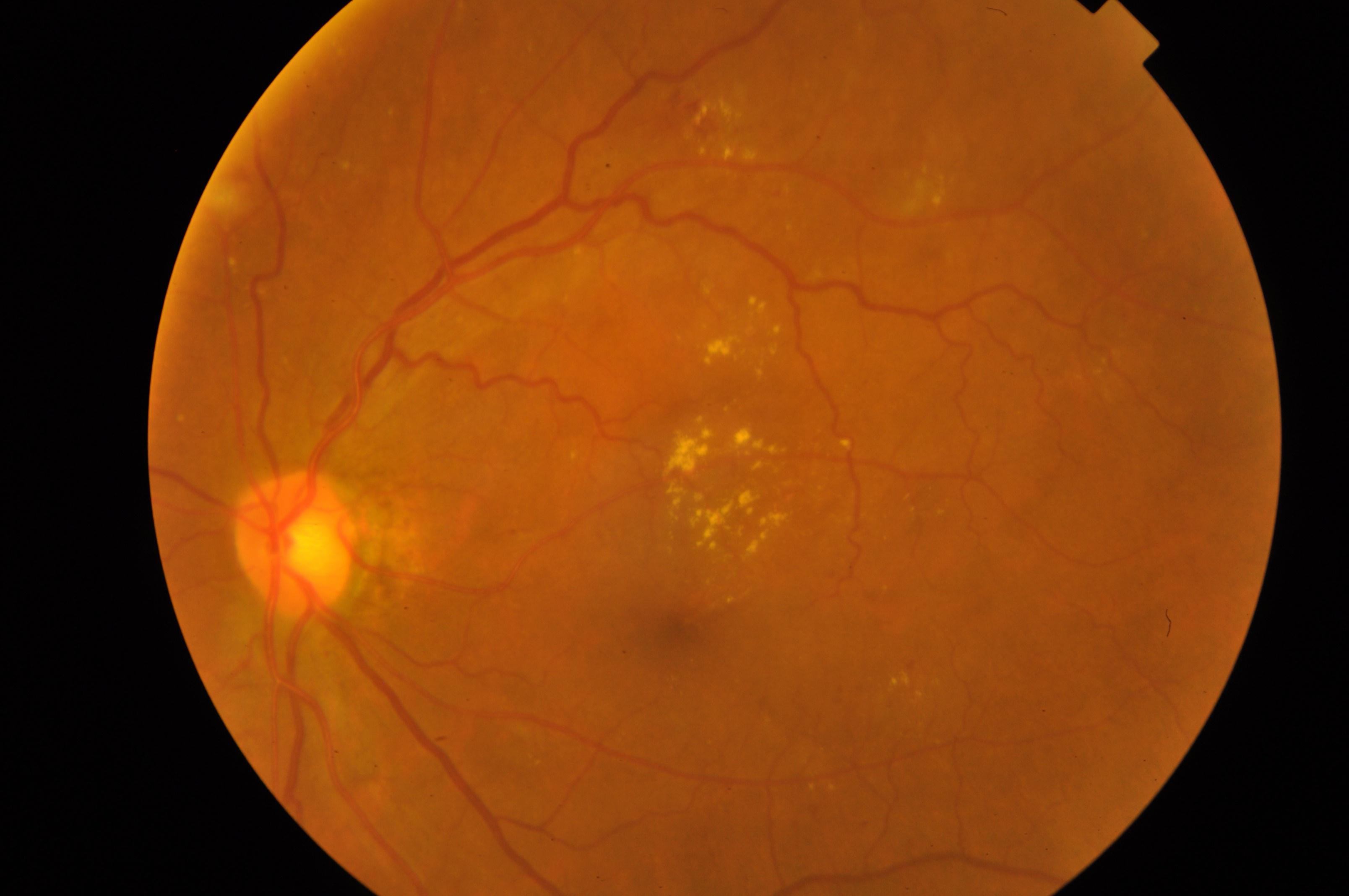}
}
\subfigure[Class 3 - Severe.]{
\label{fig:APTOS dataset class 3}
\includegraphics[height=0.09\textheight]{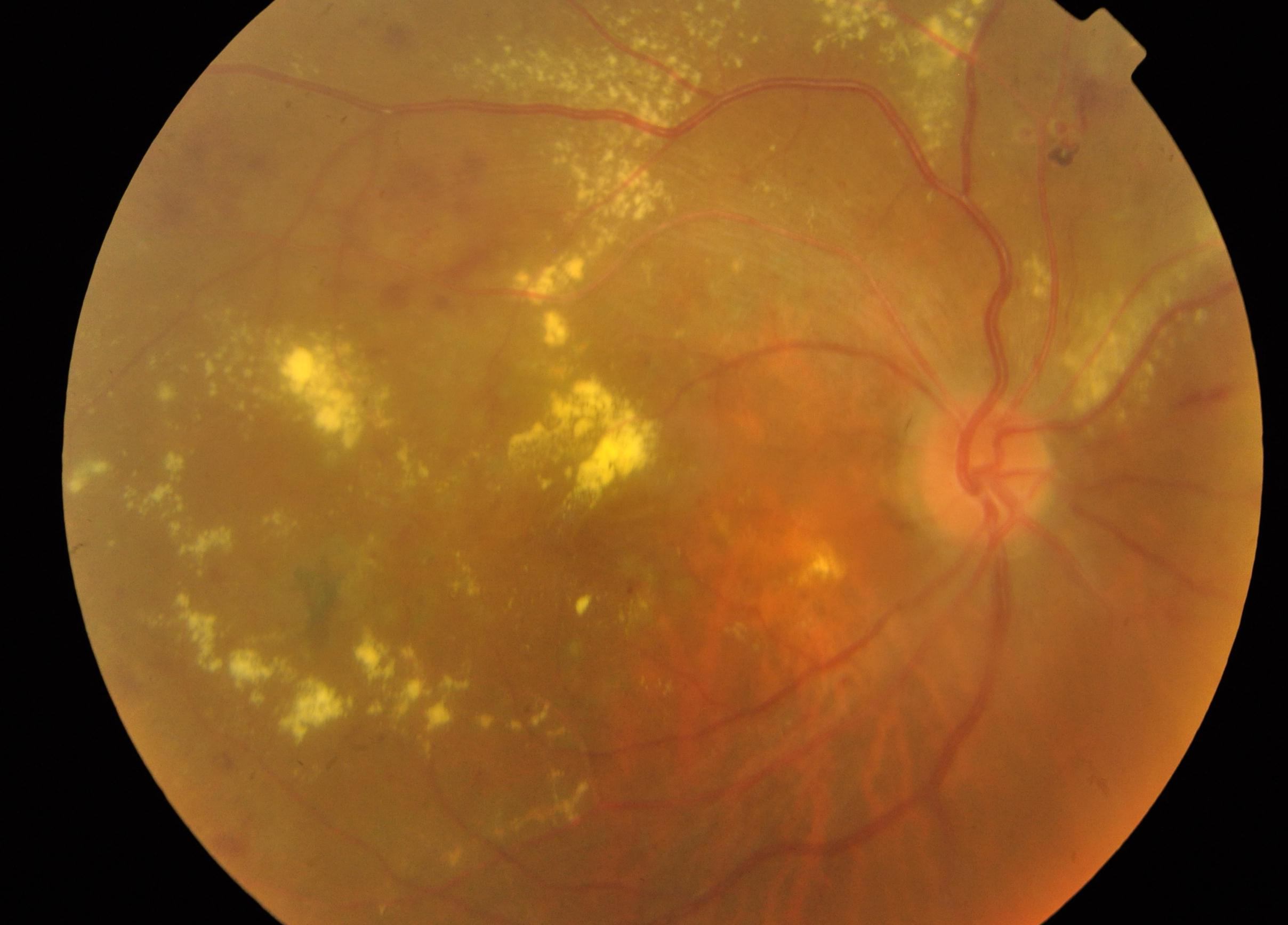}
\includegraphics[height=0.09\textheight]{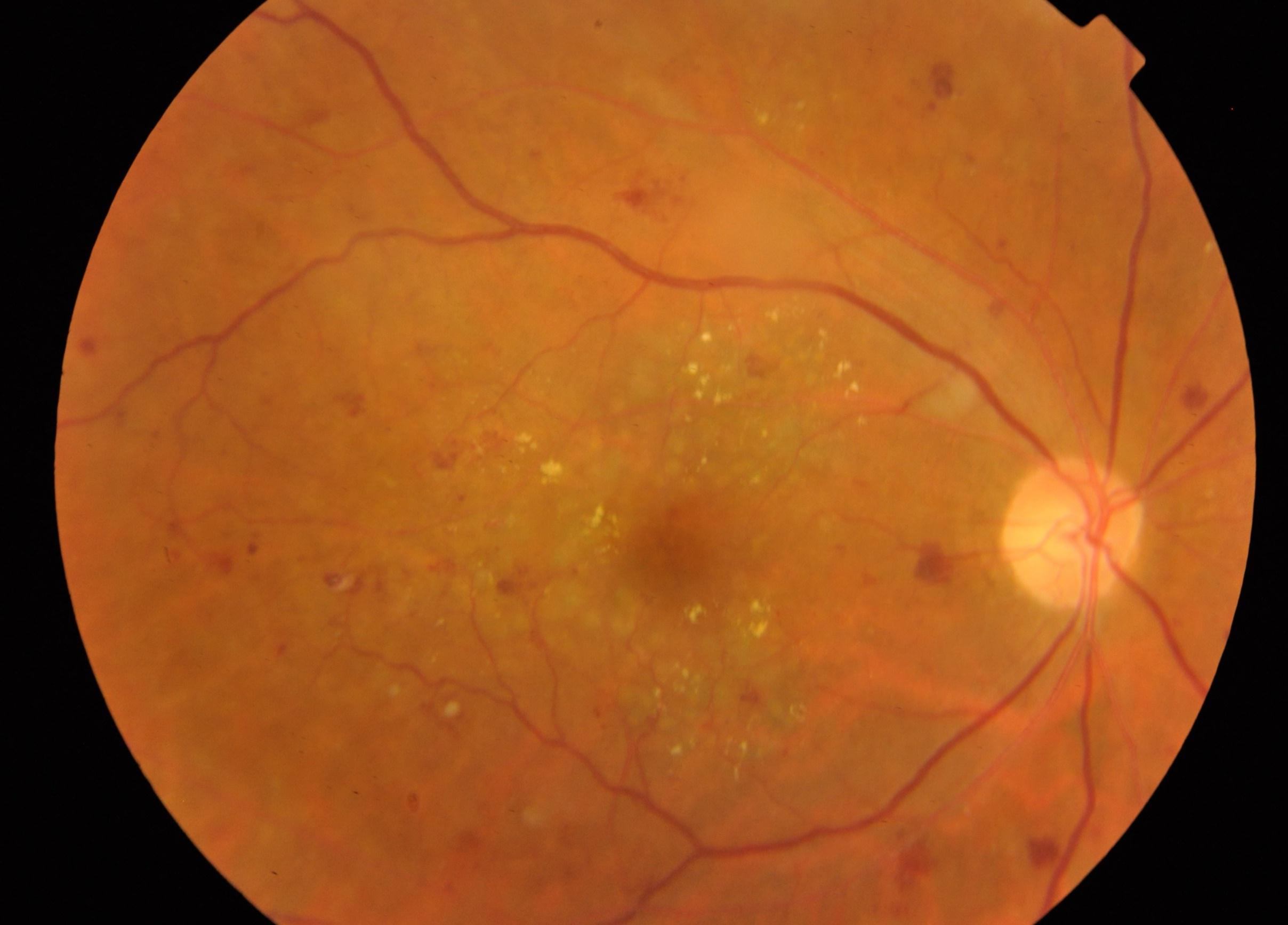}
}
\subfigure[Class 4 - Proliferative DR.]{
\label{fig:APTOS dataset class 4}
\includegraphics[height=0.09\textheight]{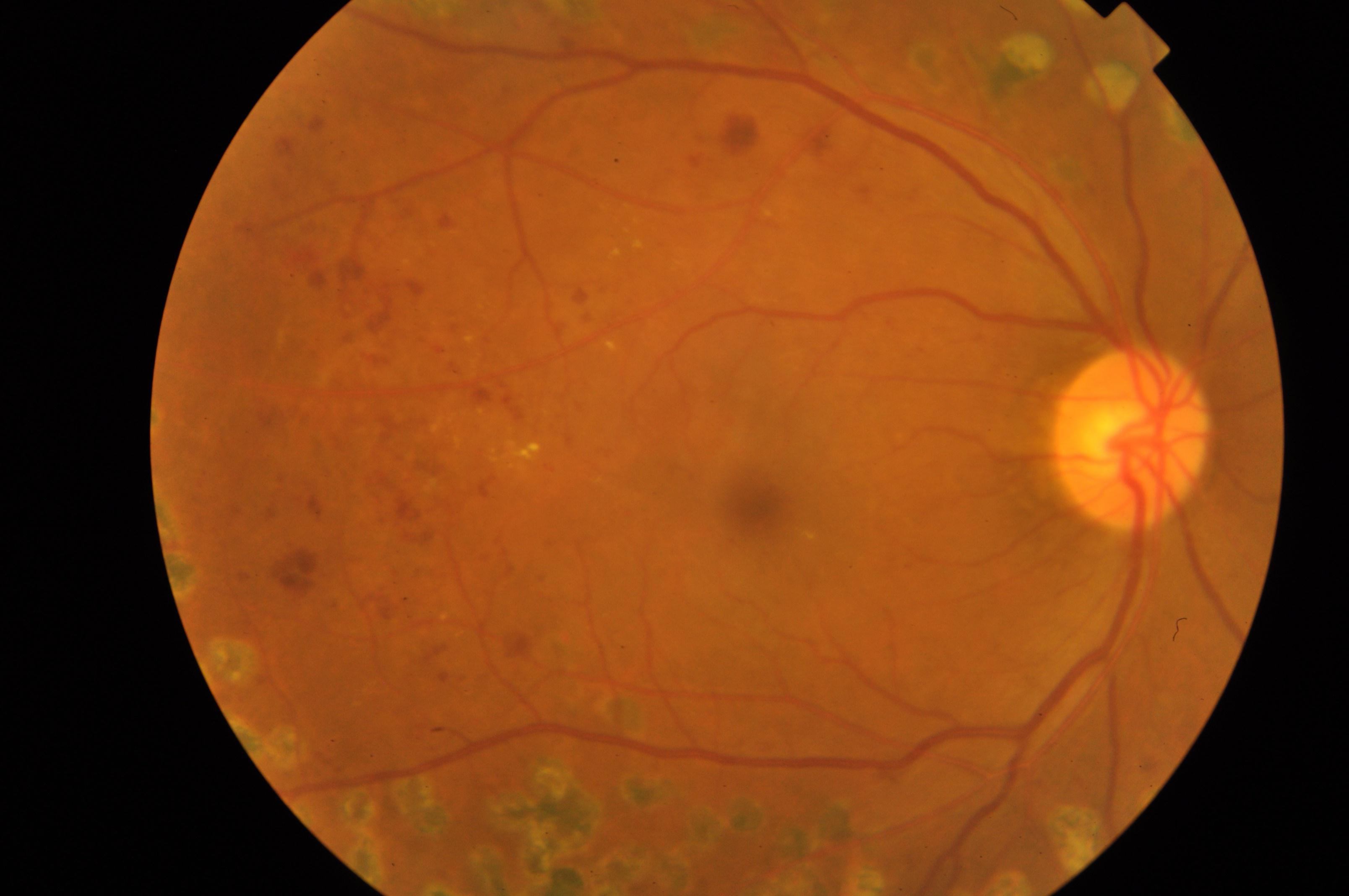}
\includegraphics[height=0.09\textheight]{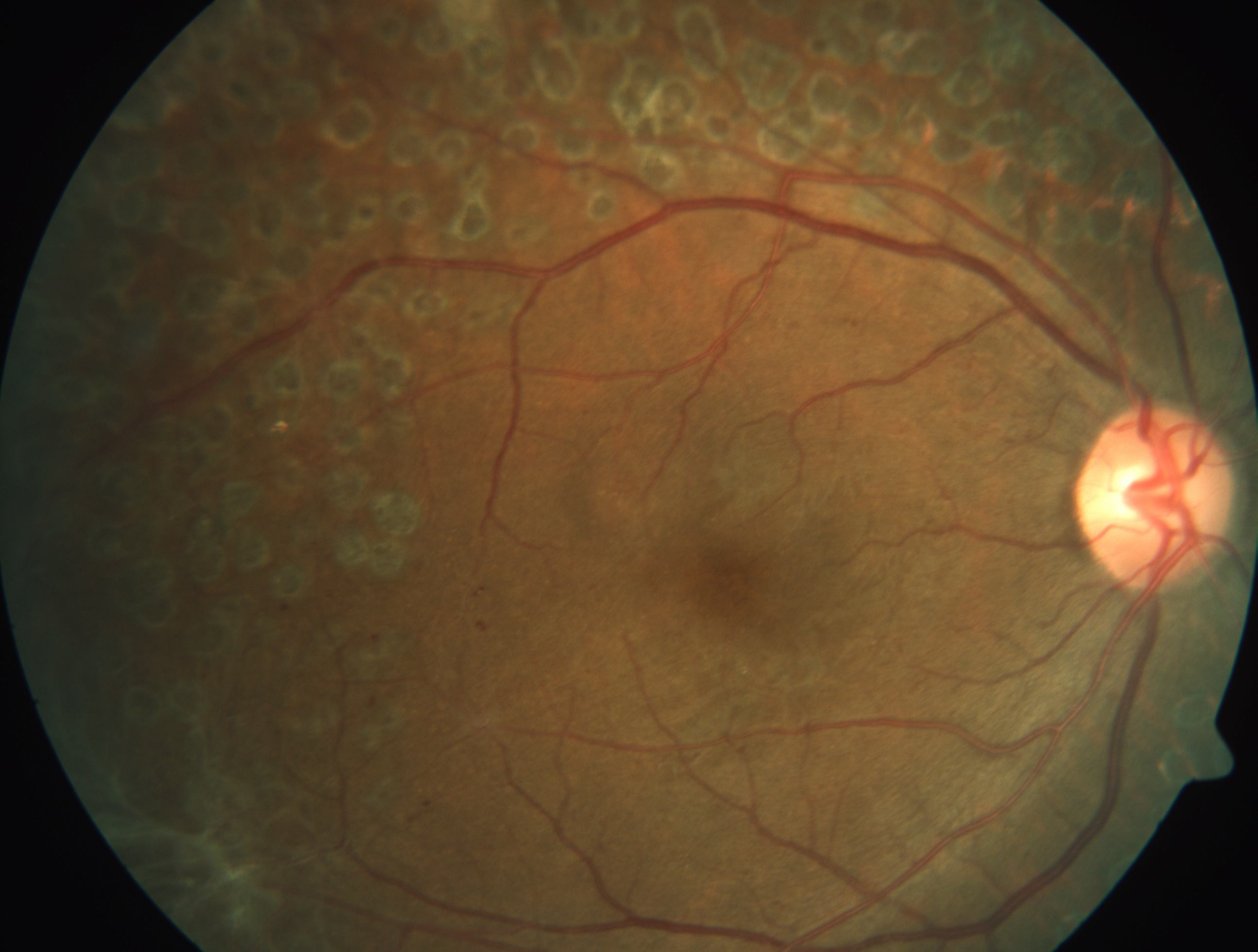}
}

\subfigure[Concept $c_1$ (micro-aneurysm).]{
\label{fig:aptos aneurysm}
\includegraphics[height=0.18\textheight]{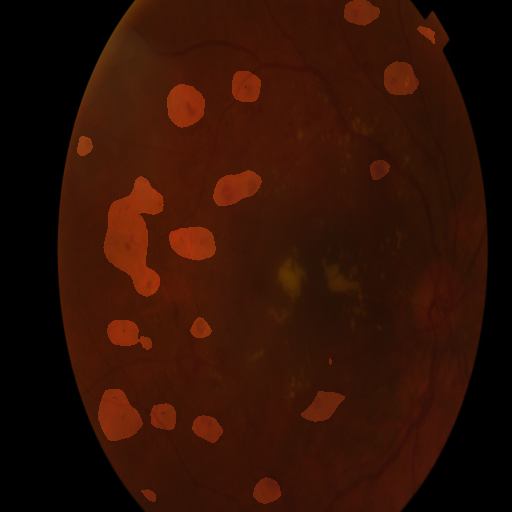}
}
\subfigure[Concept $c_2$ (exudates).]{
\label{fig:aptos exudates}
\includegraphics[height=0.18\textheight]{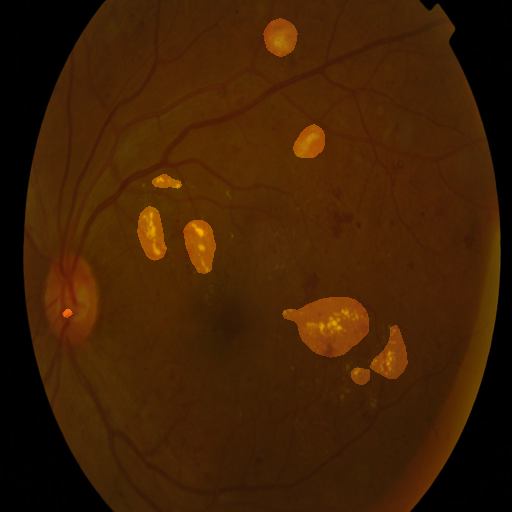}
}
\subfigure[Concept $c_3$ (blood vessels).]{
\label{fig:aptos blood vessels}
\includegraphics[height=0.18\textheight]{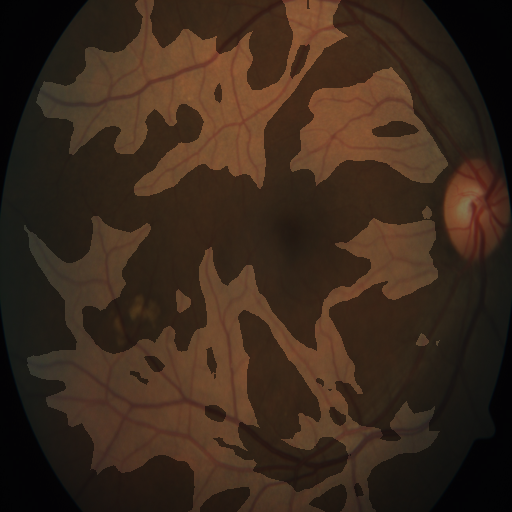}
}

\caption{Dataset diabetic retinopathy classification \cite{aptos2019}, composed of five classes, \emph{0 - No DR} with a healthy retina image (Figure \ref{fig:APTOS dataset class 0}), 
\emph{0 - No DR} with a healthy retina image (Figure \ref{fig:APTOS dataset class 0}), 
\emph{1 - Mild} with a retina image with mild retinopathy  (Figure \ref{fig:APTOS dataset class 1}), 
\emph{2 - Moderate} with a retina image with moderate retinopathy  (Figure \ref{fig:APTOS dataset class 2}), 
\emph{3 - Severe} with a retina image with severe retinopathy  (Figure \ref{fig:APTOS dataset class 3}), and
\emph{4 - Proliferative DR} with a retina image with proliferative retinopathy  (Figure \ref{fig:APTOS dataset class 4}).
Most important concepts ($\mathrm{RI}_{c_j}\geq0.7$) extracted with ECLAD from a DenseNet-121 trained on the APTOS dataset \citep{aptos2019}. Concepts $c_1$ with an importance score of 1.0, relates to the visual cues of micro-aneuryms (Figure \ref{fig:aptos aneurysm}). Concept $c_2$ with an importance score of 0.8, relates to the visual cues of exudates (Figure \ref{fig:aptos exudates}). Concept $c_3$ with an importance score of 0.7, relates to thinner blood vessels (Figure \ref{fig:aptos blood vessels}).
ECLAD is able to extract and localize important/meaningful concepts related to the visual cues used by domain experts to perform a similar task.
}
\label{fig:APTOS dataset}
\end{figure*}

In addition to a quantitative validation of our method ECLAD, we also explore its application in real world use cases. 
Specifically, we explore the usage of ECLAD for understanding models in one medical diagnosis and one quality control use cases. In both scenarios, the understanding of how the models work has significant implications.
Thus, XAI methods are often used for to ensure alignment with domain experts, and mitigate risks of undesired behavior.

In these experiments, we first trained a DenseNet-121 using each dataset. Then, we analyzed the obtained model using ECLAD, extracting ten concepts. Finally, though visual inspection, we compared the obtained concepts with common visual cues used by experts of each domain.

\textbf{The diabetic retinopathy classification \cite{aptos2019} (APTOS)}, is a medical imaging dataset containing retina images taken using fundus photography. The images are classified in one of five classes, depending on the severity of their diabetic retinopathy. Examples of said classes are shown in Figure \ref{fig:APTOS dataset}. Domain experts diagnose the diabetic retinopathy severity based on visual cues such as micro-aneurysms, hard exudates, hemorrhages and abnormal blood vessel growth. During the development of these models, extracting concepts allows for a better explanation, of which visual cues are used by models, and how important they are in their prediction process.

After the execution of ECLAD, three concepts with importance scores above 0.6 were extracted. In addition, these concepts were aligned with the visual cues used by domain experts, as seen in Figure \ref{fig:APTOS dataset}. 

The concepts $c_1$ and $c_2$, with importance scores of 1.0 and 0.8 respectively, relate to micro-aneuryms and hard-exudates. As secondary visual cues, the concepts $c_3$ and $c_4$ with importance scores of 0.7 and 0.57, relate to different formations of blood vessels. As a tangential insight, none of the concepts were specifically related to large hemorrhages. Paradoxically, the analyzed model does use a minimal set of visual cues important to human experts. Yet, not all cues that human experts would consider relevant are being used by the CNN in the prediction process.

\begin{figure*}[h!]
\centering
\subfigure[Class OK.]{
\label{fig:metal casting dataset class OK}
\includegraphics[height=0.10\textheight]{images_casting_data_cast_ok_0_88}
\includegraphics[height=0.10\textheight]{images_casting_data_cast_ok_0_271}
\includegraphics[height=0.10\textheight]{images_casting_data_cast_ok_0_233}
}
\subfigure[Class defective.]{
\label{fig:metal casting dataset class defective}
\includegraphics[height=0.10\textheight]{images_casting_data_cast_def_0_31}
\includegraphics[height=0.10\textheight]{images_casting_data_cast_def_0_3426}
\includegraphics[height=0.10\textheight]{images_casting_data_cast_def_0_2960}
}

\subfigure[Concept $c_4$ (defects).]{
\label{fig:metal casting C4}
\includegraphics[height=0.10\textheight]{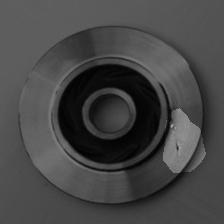}
\includegraphics[height=0.10\textheight]{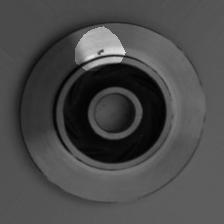}
}
\subfigure[Concept $c_0$ (good edges).]{
\label{fig:metal casting C0}
\includegraphics[height=0.10\textheight]{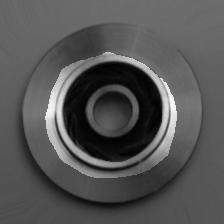}
\includegraphics[height=0.10\textheight]{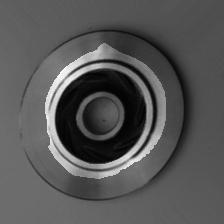}
}
\subfigure[Concept $c_1$ (background).]{
\label{fig:metal casting C1}
\includegraphics[height=0.10\textheight]{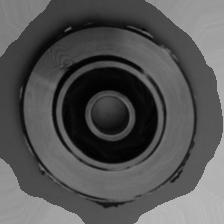}
\includegraphics[height=0.10\textheight]{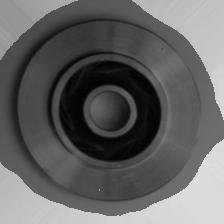}
}

\caption{
Metal casting dataset, composed of two classes, ``OK'' with a well cast metal part (Figure \ref{fig:metal casting dataset class OK}), and ``defective'' with a metal part containing pinholes, scratches, or deformed edges (Figure \ref{fig:metal casting dataset class defective}).
The most important concept ($\mathrm{RI}=1.0$) relates to all defects present in images, including pinholes, scratches and deformed edges, as seen in Figure \ref{fig:metal casting C4}.
Second most important concept ($\mathrm{RI}=0.49$) relates to well-formed inner edges \ref{fig:metal casting C0}.
Least important concept ($\mathrm{RI}=0.0$) relates to the background of the image, revealing that there are no biases in said region.
ECLAD provides insights, which in this case are aligned with expert knowledge, including intuitive visual cues (defects), non-intuitive complimentary cues (well-formed edges), and ensuring that unimportant (background) regions have no biases.
}
\label{fig:metal casting dataset}
\end{figure*}

\textbf{The metal casting classification dataset \cite{dabhi2020casting}}, is a visual quality control dataset containing images of metal cast parts.
The images are classified in two classes, ``defective'' for those with pinholes, scratches or deformed edges, and ``OK'' for the rest. 
Examples of said classes are shown in Figure \ref{fig:metal casting dataset}. 
During the development of these models, extracting concepts allows experts to ensure that the visual cues used by the model are correct, and that there are no undesired biases (e.g. background color). Ensuring alignment with experts often leads to increasing the trust in the models.

We extracted ten concepts using ECLAD, from which we present the two most important and the least important one.
The most important concept $c_4$ ($\mathrm{RI}=1.0$), related to all defects present in the ``defective'' class, including pinholes, scratches, and deformed edges (seen in Figure \ref{fig:metal casting C4}).
The second most important concept $c_0$ ($\mathrm{RI}=0.49$), related to the inner edges of non-defective parts, clearly excluding deformed edges (seen in Figure \ref{fig:metal casting C0}).
Finally, the least important concept $c_1$ ($\mathrm{RI}=0.00$), relates to the background, which was consistently differentiated in all images and deemed unimportant by the model.
These explanations allow a better understanding of how the model work, validating the alignment of domain knowledge (e.g. regions with defects are important), ensuring that no unexpected biases are present (e.g. background is not important), and providing non-intuitive yet insightful remarks (e.g. detecting correct edges can be as useful as detecting the deformed ones).

The two cases studied above prove difficult for both ACE and ConceptShap, yielding uninformative concepts and associated importance. Nonetheless, concepts extracted with ECLAD exploit local representations within the models, allowing for a robust performance and the extraction of insightful explanations.
\textbf{In both cases, the resulting explanations of ECLAD can be used for a better communication with domain experts, allowing for human understandable visualizations of how models understand images, and how important those visual cues are.}

\subsection{Performance and limitations}

\textbf{Ablation studies} were performed to explore the impact of key components of ECLAD, obtaining the following key insights.
First, the results of using a single element in the set $L$ for concept extraction highly depends on the depth of the chosen layer. These results range from a small subset of high level concepts (disregarding mid level ones) to low level features such as edges, disregarding high level concepts. In comparison, by combining layers from multiple depths, ECLAD allows the extraction of mid and high level concepts without the complexity of fine-tuning the selected layer. 
Second, more than three layers help compensate halo effects on extracted concepts (representations dilate through the network), as well as mid level concepts which are not present in higher layers.
Third, higher number of concepts will cause a progressive slicing of important concepts (without affecting their $\mathrm{RI}_{c_j}$).
Finally, using coarse interpolation methods ($f_U$) will impact the boundaries of the extracted concepts, but not the concepts themselves.
The chosen parameters for the presented analysis are a balance between performance and computational cost.
The complete details on the ablation study are presented on the appendix \ref{apd:Ablation study}.

\textbf{A computational cost analysis} was performed comparing ECLAD, ACE, and ConceptShap.
As a general insight, ECLAD provides more granular explanations, scaling well to large datasets and number of concepts. It scales linearly for the number of classes $N_k$, which makes it preferable when dealing with a small number of classes (e.g. $N_k<20$).
ConceptShap scales well for a large number of classes, yet it scales poorly for the number of extracted concepts. As a caution, ConceptShap and Shapely values in general have issues when dealing with correlated concepts, which can be problematic when detecting spurious correlations and their importance for a CNN.
Finally, ACE scales better than ECLAD with respect to the number of classes, yet, it has a significant computational cost of executing SLIC and SDG linear classification. In this regard, ACE can be parallelized and executed per class, being a better fit for large datasets with a large number of classes (e.g. $N_k~1000$).
The complete details on the computational cost analysis are presented on the appendix \ref{apd:Computational cost}.

\section{Conclusions}
\label{sec:Conclusions}

We propose ECLAD as a concept extraction (CE) technique, based on local aggregate descriptors (LADs). Our algorithm focuses on how CNNs represent pixels internally, allowing a more reliable CE and importance scoring. In addition, it provides the novel ability to localize, in new instances, which regions of an image contain the visual cues related to each concept. 

As an orthogonal contribution, we propose an automatic comparison and validation process for CE techniques, which provides consistent and scalable metrics denoting the performance of CE methods. Our validation process is based on two novel metrics, measuring the importance and visual cues of concepts with respect to the ground truth of synthetic datasets. We provide six new synthetic datasets that can be used for testing CE methods. The proposed datasets and validation method proved effective in comparing ECLAD, ACE, and ConceptShap. As validation procedures that forego (possibly subjective) human judgement are largely missing in the area, we hope our contribution becomes helpful in providing a quantitative approach for evaluating CE (to compliment human studies).

Through our validation process, ECLAD proved a reliable alternative for analyzing models through concept-based explanations. The extracted concepts were consistently related to the main features of the analyzed datasets, which was reflected on a high representation correctness across all experiments. In addition, the importance scores provided by ECLAD better reflect the intended importance of their associated features, outperforming other methods across all datasets. The importance of concepts related to relevant features of the dataset are scored high, and irrelevant concepts are scored low.

While ECLAD performed reliably in the studied cases, the results also raise relevant questions for future research. First, during the initial CE, ECLAD can be more computationally expensive than CAV-based methods, as the base clustering is performed over the representations of pixels and not images or patches. Second, the localization of concepts in new images strongly depends on the CNN architecture being studied, as not all CNNs represent local information with the same fidelity. Finally, for more complex tasks with a higher number of features, the number of extracted concepts will have to be adjusted accordingly to avoid relevant features being clustered together.
\section*{Acknowledgements}
\label{sec:Acknowledgements}

Funded by the Deutsche Forschungsgemeinschaft (DFG, German Research Foundation) under Germany’s Excellence Strategy–EXC-2023 Internet of Production–390621612.

The authors would also like to thank 
Dr. Dominik Baumann, and Dr. Friedrich Solowjow
for the helpful feedback they provided on an early version of this paper.

\bibliographystyle{elsarticle-num-names} 
\bibliography{references} 


\newpage
\appendix
\newpage

\section{Appendix}
\label{apd:appendix}

The appendices of this work contain extended information pertaining to three principal topics. 
First, we discuss in detail the experimental setup required to perform the experiments in Appendix \ref{apd:experimental setup}. 
Second, we describe the new synthetic datasets in Appendix \ref{apd:synthetic datasets}. 
Third, we discuss the computational cost of ECLAD, ACE, and ConceptShap in Appendix \ref{apd:Computational cost}.
Fourth, We describe the ablation study performed over key components of ECLAD in Appendix \ref{apd:Ablation study}.
Finally, we compare the proposed association distance with other existing alternatives in Appendix \ref{apd:Distance metric}.

\section{Experimental setup}
\label{apd:experimental setup}

We performed all experiments in servers with Intel\textsuperscript{\textregistered} Xeon\textsuperscript{\textregistered} Gold 6330 CPU and a NVIDIA A100 GPU. We implemented ECLAD, ACE, and ConceptShap using Pytorch 1.11, and the different model architectures using the \emph{PyTorch Image Models} (TIMM) library. As part of the supplementary material, we make available the code of the experiments, as well as the created datasets under an MIT license. Both items available on the link: \href{https://drive.google.com/drive/folders/16CjAvk8H1VAD2-rNiy0HV3OmzDlrwXo5?usp=sharing}{https://drive.google.com/drive/folders/16CjAvk8H1VAD2-rNiy0HV3OmzDlrwXo5?usp=sharing}

\textbf{Analyzed model architectures}. During the experiments, we trained and analyzed five different CNN architectures (ResNet-18 \citep{DBLP:journals/corr/HeZRS15}, ResNet-34 \citep{DBLP:journals/corr/HeZRS15}, DenseNet-121 \citep{DBLP:journals/corr/HuangLW16a}, EfficientNet-B0 \citep{DBLP:conf/icml/TanL19}, and VGG16 \citep{DBLP:journals/corr/SimonyanZ14a}). For each model, we selected four layers for executing ECLAD, and the last one ($l_4$) was used for ConceptShap. For ACE, we used the output of the average pooling before the fully connected layers of each model, as advised in TCAV \citep{DBLP:conf/icml/KimWGCWVS18}. The list of layers $L={l_1, l_2, l_3, l_4}$, and $l_{tcav}$ for each model are provided in Table \ref{tbl:layers}.

\begin{table*}[h!]
\caption{Synthetic datasets.}
\label{tbl:layers}
\vskip 0.15in
\begin{center}
\begin{small}
\begin{sc}
\begin{tabular}{lll}  
Model           & layers            & Layer\\ \hline
ResNet-18       & $l_1$             & Layer 1, Block 1, ReLU 2\\
ResNet-18       & $l_2$             & Layer 2, Block 1, ReLU 2\\
ResNet-18       & $l_3$             & Layer 3, Block 1, ReLU 2\\
ResNet-18       & $l_4$             & Layer 4, Block 1, ReLU 2\\
ResNet-18       & $l_{tcav}$          & Average pooling before fc layers\\
ResNet-34       & $l_1$             & Layer 1, Block 2, ReLU 2\\
ResNet-34       & $l_2$             & Layer 2, Block 3, ReLU 2\\
ResNet-34       & $l_3$             & Layer 3, Block 5, ReLU 2\\
ResNet-34       & $l_4$             & Layer 4, Block 2, ReLU 2\\
ResNet-34       & $l_{tcav}$          & Average pooling before fc layers\\
DenseNet-121    & $l_1$             & Transition layer 1, conv\\
DenseNet-121    & $l_2$             & Transition layer 2, conv\\
DenseNet-121    & $l_3$             & Transition layer 3, conv\\
DenseNet-121    & $l_4$             & Dense block 4, Dense Layer 16, conv 2\\
DenseNet-121    & $l_{tcav}$          & Average pooling before fc layers\\
EfficientNet-B0 & $l_1$             & Block 3, Inverted residual 2, conv\_pwl\\
EfficientNet-B0 & $l_2$             & Block 4, Inverted residual 2, conv\_pwl\\
EfficientNet-B0 & $l_3$             & Block 5, Inverted residual 3, conv\_pwl\\
EfficientNet-B0 & $l_4$             & Block 6, Inverted residual 0, conv\_pwl\\
EfficientNet-B0 & $l_{tcav}$          & Average pooling before fc layers\\
VGG16           & $l_1$             & MaxPooling after Conv 2-2\\
VGG16           & $l_2$             & MaxPooling after Conv 3-3\\
VGG16           & $l_3$             & MaxPooling after Conv 4-3\\
VGG16           & $l_4$             & MaxPooling after Conv 5-3\\
VGG16           & $l_{tcav}$          & Average pooling before fc layers\\
\end{tabular}
\end{sc}
\end{small}
\end{center}
\vskip -0.1in
\end{table*}

\textbf{Model training}. We performed the model training using an SGD optimizer with a learning rate of 0.1 and momentum of 0.9. We used a \emph{reduce lr on plateau} scheduler with a factor of 0.1 based on the \emph{negative log likelihood loss} of the models. The data was split into 0.85 for training and 0.15 for testing, with mini-batches of 24 images sampled and balanced between the classes. In addition, we used random color jitter, and affine transforms for data augmentation. 

\textbf{ECLAD}. We perform all ECLAD analysis with the same sets of parameters. Each execution was performed using a maximum of 200 images from each class, extracting 10 concepts, and using 2 images per clustering minibatch (100352 LADs). The layers used for each model are shown in Table \ref{tbl:layers}.

\textbf{ACE}. We perform all ACE analysis with the same parameters used by \citet{DBLP:conf/nips/GhorbaniWZK19}. We performed a SLIC segmentation over 20 images of each class, with sigma of 1.0 and compactness of 20.0 for 15, 50, and 80 segments. Subsequently, we resized and padded each patch before evaluating it in a model to extract the activation map of the selected layer. We then extracted 25 clusters using k-means over the flattened activation maps. Finally, we used each group of clustered patches to perform 50 repetitions of TCAV and obtain the $\mathrm{TCAV}_Q$ score of each concept. 

\textbf{ConceptShap}. We perform all ConceptShap analysis with the same sets of parameters. 10 concepts were extracted at each run with $\beta = 1.0 \times {10}^{-7}$, ${\lambda}_1 = 1 \times 10^{-7}$, ${\lambda} _ 2 = 2 \times 10^{-7}$. These values were obtained empirically after exploring values in orders of magnitude from $1 \times 10^{-1}$ to $1 \times 10^{-10}$. The Shapely values for the concepts were approximated with Monte-Carlo sampling with $100 \times n_c$ samples. As a cutting threshold for localizing each concept, we used the mean values of the projection of the activation map over the concept vectors, which worked well in comparison to other fixed thresholds.


\section{Synthetic datasets}
\label{apd:synthetic datasets}

The six synthetic datasets created for the validation of ECLAD are summarized in the Table \ref{tab:datasets}. All synthetic datasets were created using alphabetical characters and filling them with either solid colors or textures from the KTH-TIPS dataset \citep{Fritz04thekth-tips}. Each dataset is composed of 200 RGB images of 224$\times$224 pixels per class. 

\begin{table*}[h]
\caption{Synthetic datasets}
\centering
\resizebox{\textwidth}{!}{%
\begin{tabular}{|l|l|l|l|}
\hline
Name      & Class 0                       & Class 1                         & Primitives                                                                         \\ \hline
AB        & A                             & B                               & A, B, +, background                                                                \\ \hline
ABplus    & A                             & B                               & A, B, *, /, \#, X, background                                                      \\ \hline
Big-Small & Big \textbackslash{}emph\{B\} & Small \textbackslash{}emph\{B\} & Big \textbackslash{}emph\{B\}, Small \textbackslash{}emph\{B\}, +, background      \\ \hline
CO        & C                             & O                               & C, O, +, background                                                                \\ \hline
colorGB   & B                             & G                               & representative character (green or blue), intrusive green character, +, background \\ \hline
isA       & isA                           & notA                            & A, other characters (B-H), background                                              \\ \hline
\end{tabular}%
}
\label{tab:datasets}
\end{table*}

Each subsection contains a description of how the synthetic dataset was created, including example images of each class and the primitives. Similarly, for each dataset, we provide a sample result for each analysis (ECLAD, ACE, ConceptShap) and model trained in said dataset. 

\newpage
\subsection{AB dataset}

\begin{figure}[h]
\centering
\subfigure[Examples class A]{
\label{fig:AB dataset class A}
\includegraphics[width=0.11\textwidth]{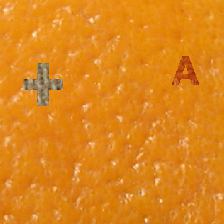}
\includegraphics[width=0.11\textwidth]{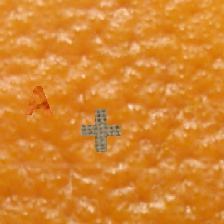}
}\subfigure[Examples class B]{
\label{fig:AB dataset class B}
\includegraphics[width=0.11\textwidth]{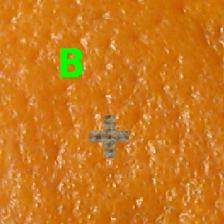}
\includegraphics[width=0.11\textwidth]{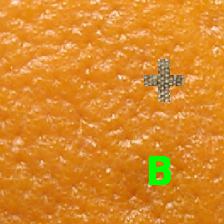}
}

\subfigure[Primitives $p_1$, $p_3$, and $p_4$ form class A.]{
\label{fig:AB dataset primitives class A}
\includegraphics[width=0.12\textwidth]{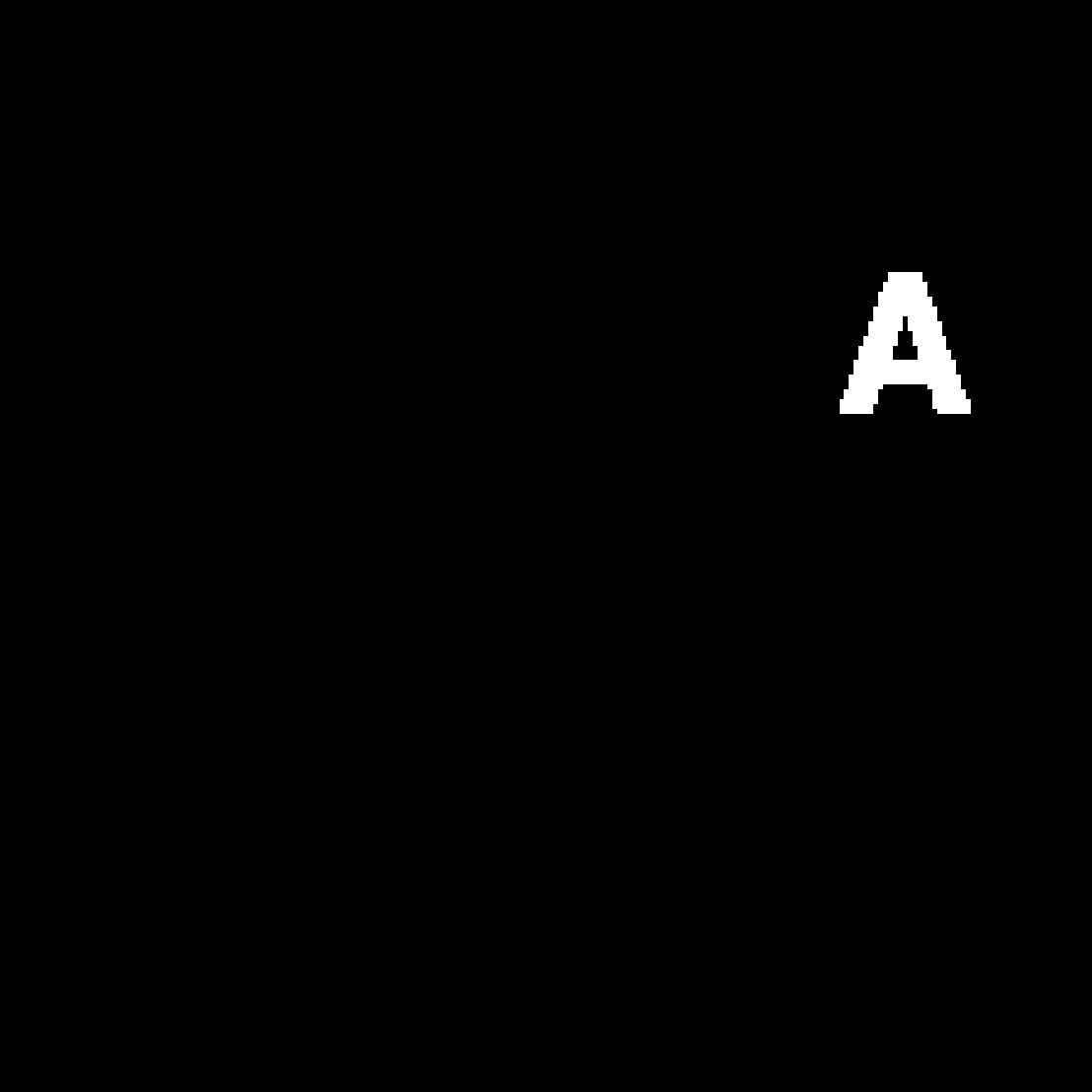}
\includegraphics[width=0.12\textwidth]{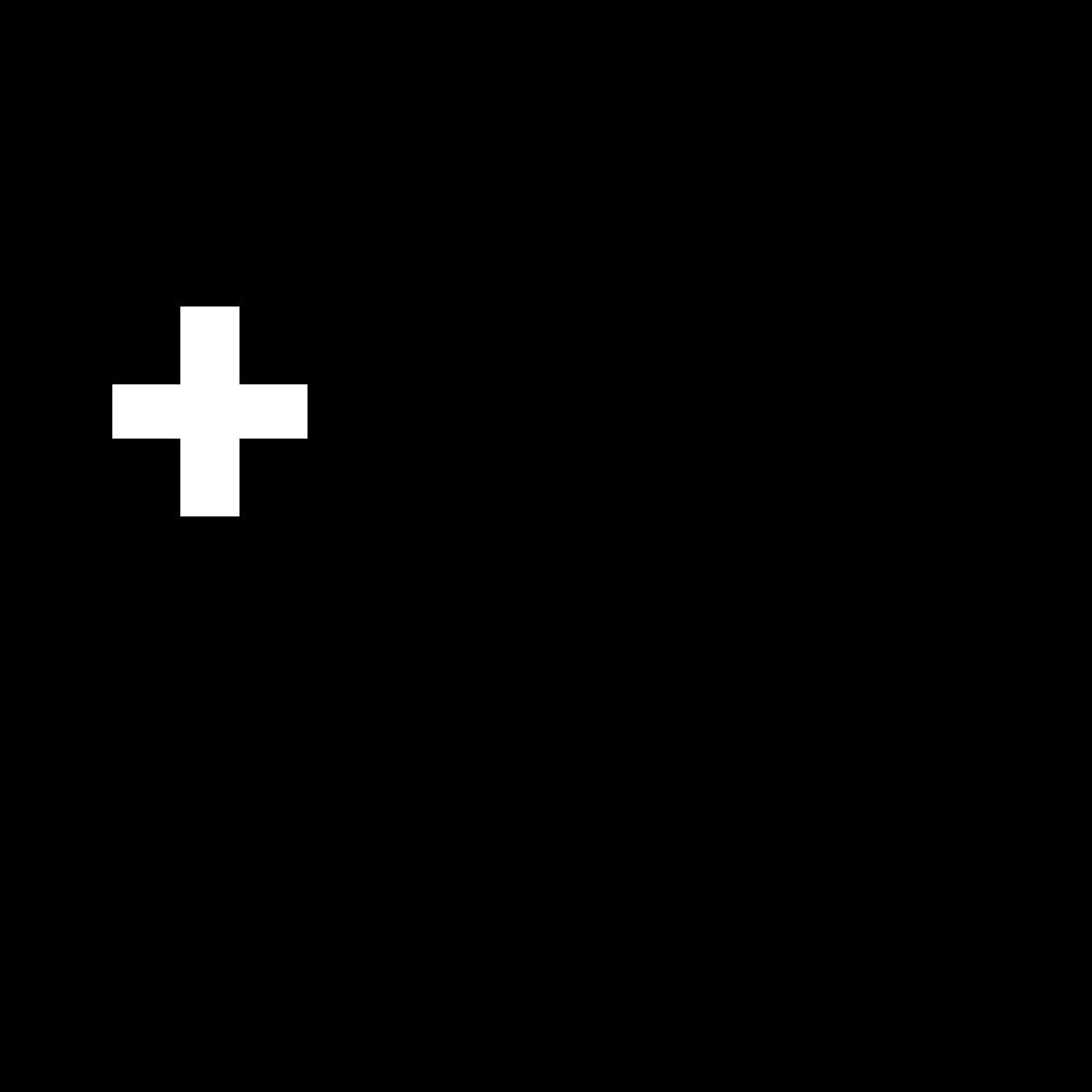}
\includegraphics[width=0.12\textwidth]{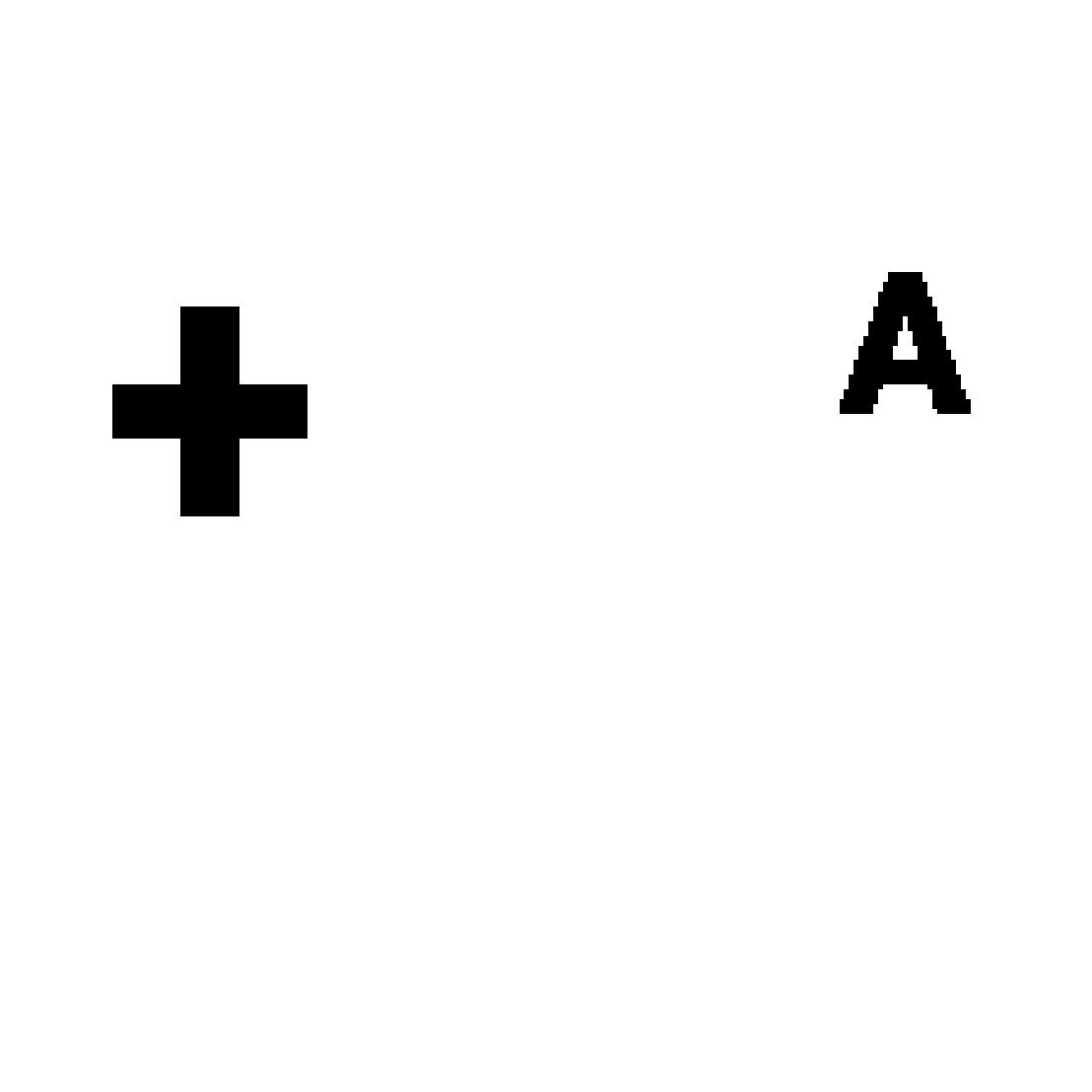}
}
\subfigure[Primitives $p_2$, $p_3$, and $p_4$ from class B.]{
\label{fig:AB dataset primitives class B}
\includegraphics[width=0.12\textwidth]{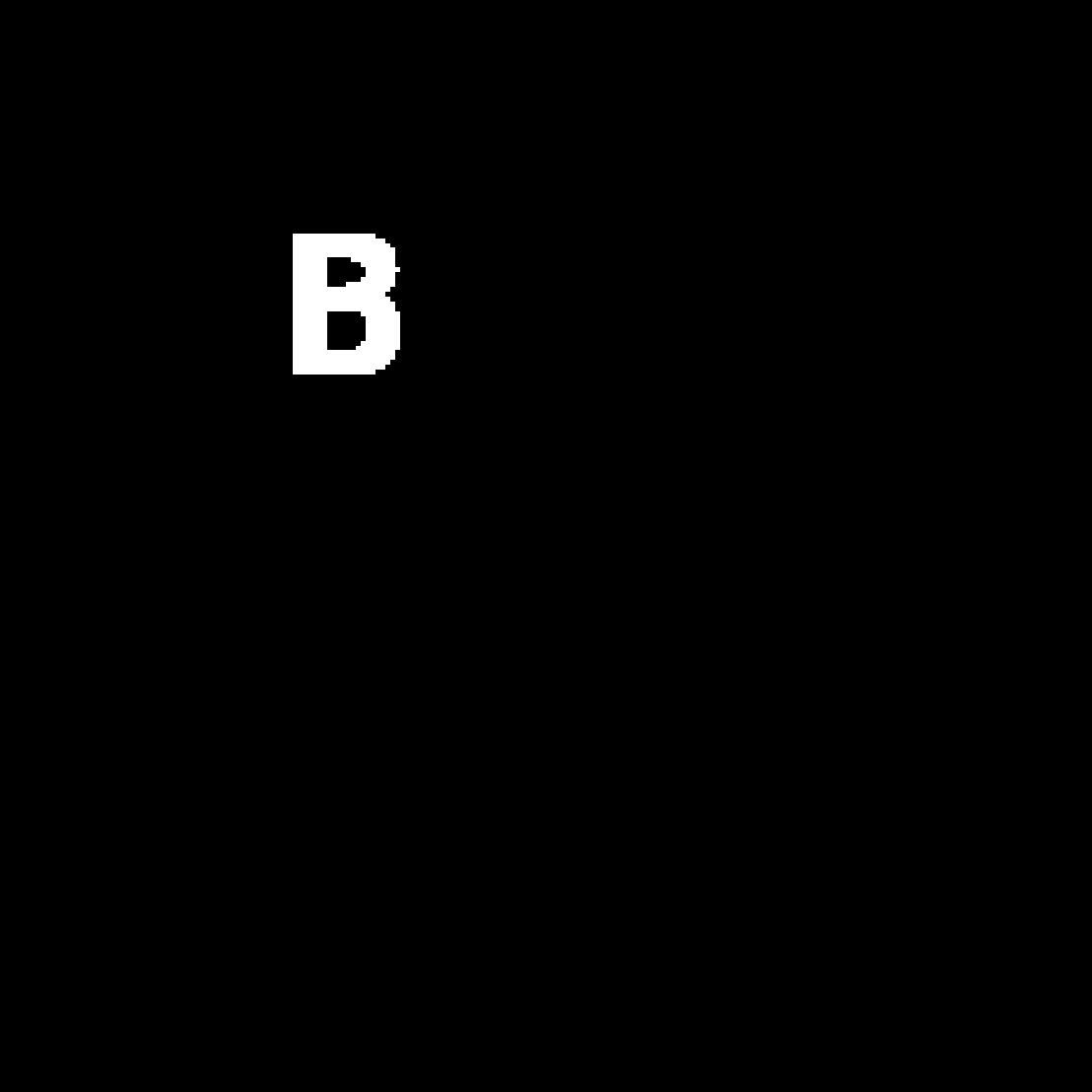}
\includegraphics[width=0.12\textwidth]{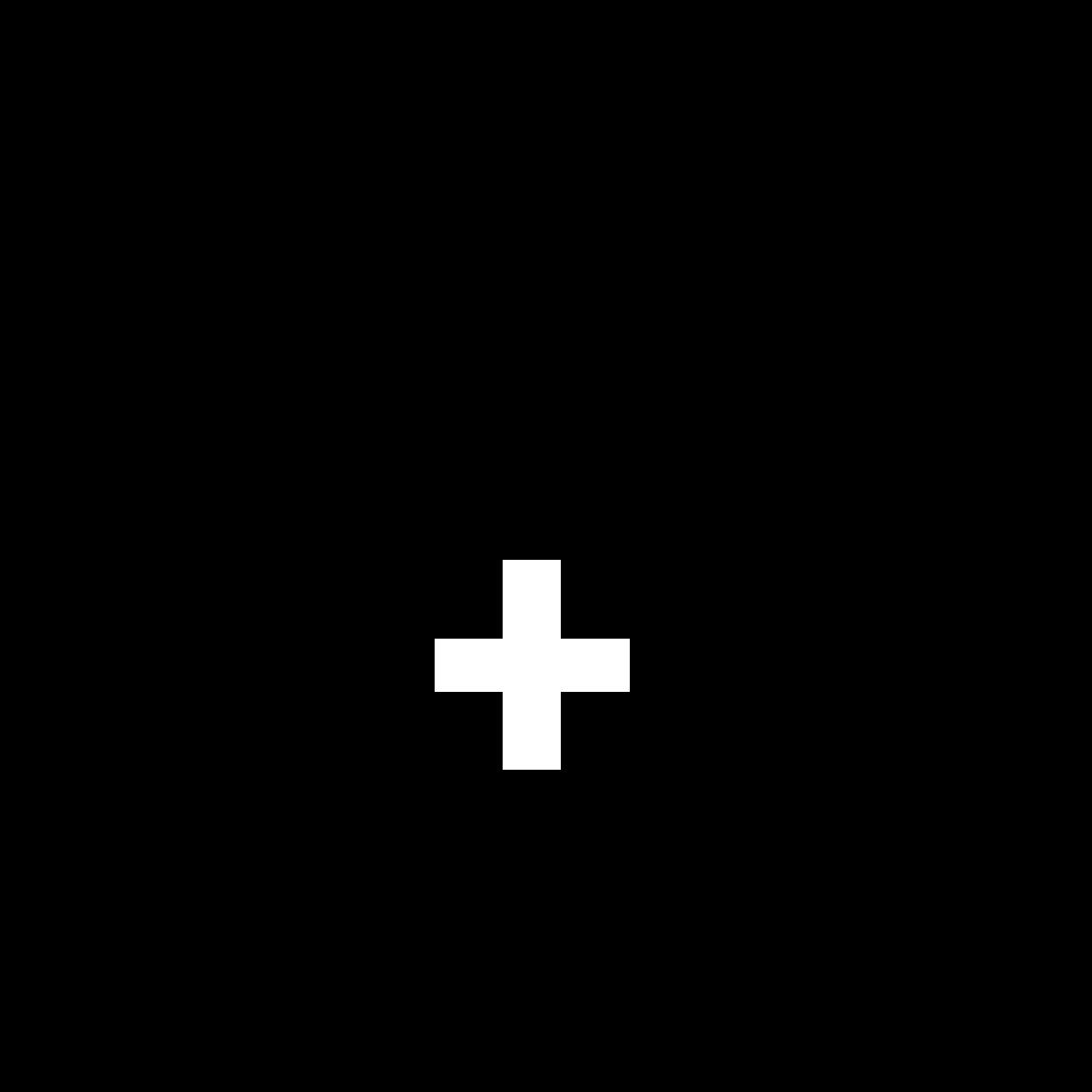}
\includegraphics[width=0.12\textwidth]{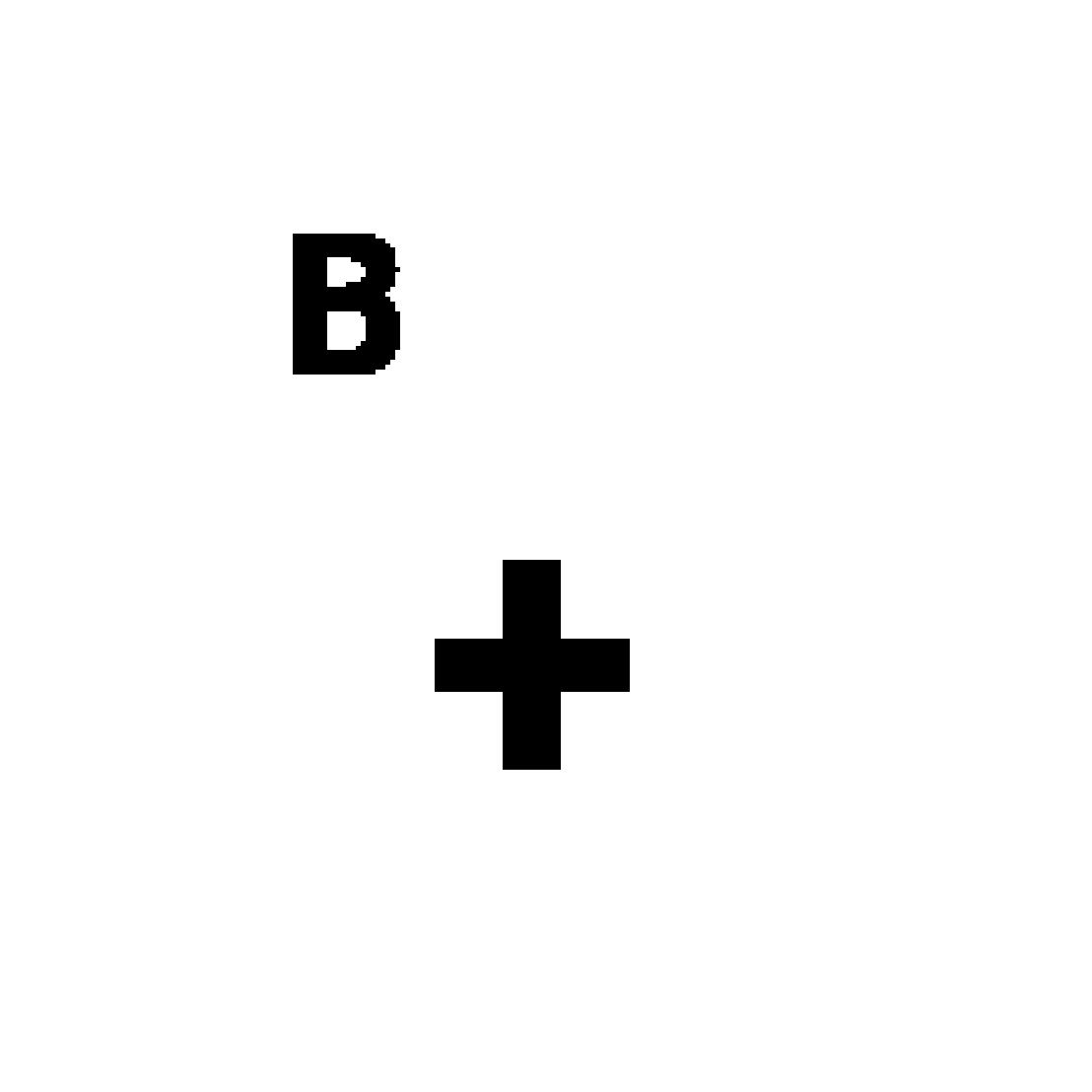}
}\caption{Dataset AB, composed of class A (\ref{fig:AB dataset class A}) and B (\ref{fig:AB dataset class B}). Primitives $p_1$, $p_3$, and $p_4$ from class A  are shown in Figure \ref{fig:AB dataset primitives class A}. Primitives $p_2$, $p_3$, and $p_4$ appearing in class B are shown in Figure \ref{fig:AB dataset class B}.}
\label{fig:AB dataset}
\end{figure}

The \emph{AB} dataset corresponds to a simple classification between images containing a character \emph{A} or a character \emph{B}. The character \emph{A}, denoted as the primitive $p_1$, is filled with a cork texture and only appears in class A. The character \emph{B}, denoted as the primitive $p_2$, is filled green, and only appears in class B. The primitive $p_3$ contains a \emph{$+$} filled with a cotton texture, which is an irrelevant feature appearing in all the images. Finally, $p_4$ refers to the background, filled with an orange peel texture. Examples of the class images and the primitives are presented in \ref{fig:AB dataset}.

This simple dataset can be used as a sanity check for CE methods. The characters A and B are different in both form and texture, which facilitates classification. Similarly, regardless of the principle for the classification (form or color), the primitives will still be the same for both base concepts.

\newpage
\subsubsection{ABplus dataset}

The \emph{ABplus} dataset also consists in the classification of images containing a character \emph{A} or a character \emph{B}, yet, it contains a higher number of intrusive elements in comparison with the dataset \emph{AB}. The primitive for the \emph{A} character is $p_1$, filled with an aluminum foil texture, appearing only in images from the class A. The primitive for the character \emph{B} is $p_2$, filled in green, appearing only in images from the class B. The rest of the primitives are balanced between the two classes, and only serve as irrelevant information. Primitives $p_3$, $p_4$, $p_5$, $p_6$, and $p_7$ refers to the symbols \textbf{*}, \textbf{/}, \textbf{\#}, and \textbf{X}, respectively, all filled in solid colors. Finally, $p_8$ refers to the background, filled with a sponge texture.

\begin{figure}[h]
\centering
\subfigure[Examples class A]{
\label{fig:ABplus dataset class A}
\includegraphics[width=0.11\textwidth]{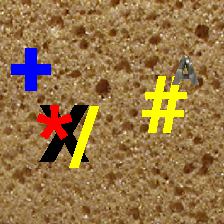}
\includegraphics[width=0.11\textwidth]{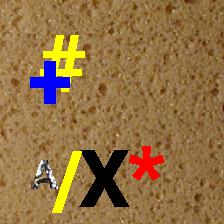}
}\subfigure[Examples class B]{
\label{fig:ABplus dataset class B}
\includegraphics[width=0.11\textwidth]{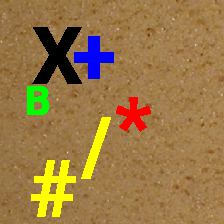}
\includegraphics[width=0.11\textwidth]{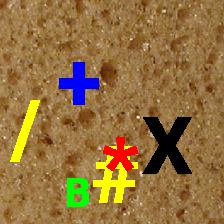}
}

\subfigure[Primitives $p_1$, $p_3$, $p_4$, and $p_5$ from class A.]{
\label{fig:ABplus dataset primitives class A}
\includegraphics[width=0.09\textwidth]{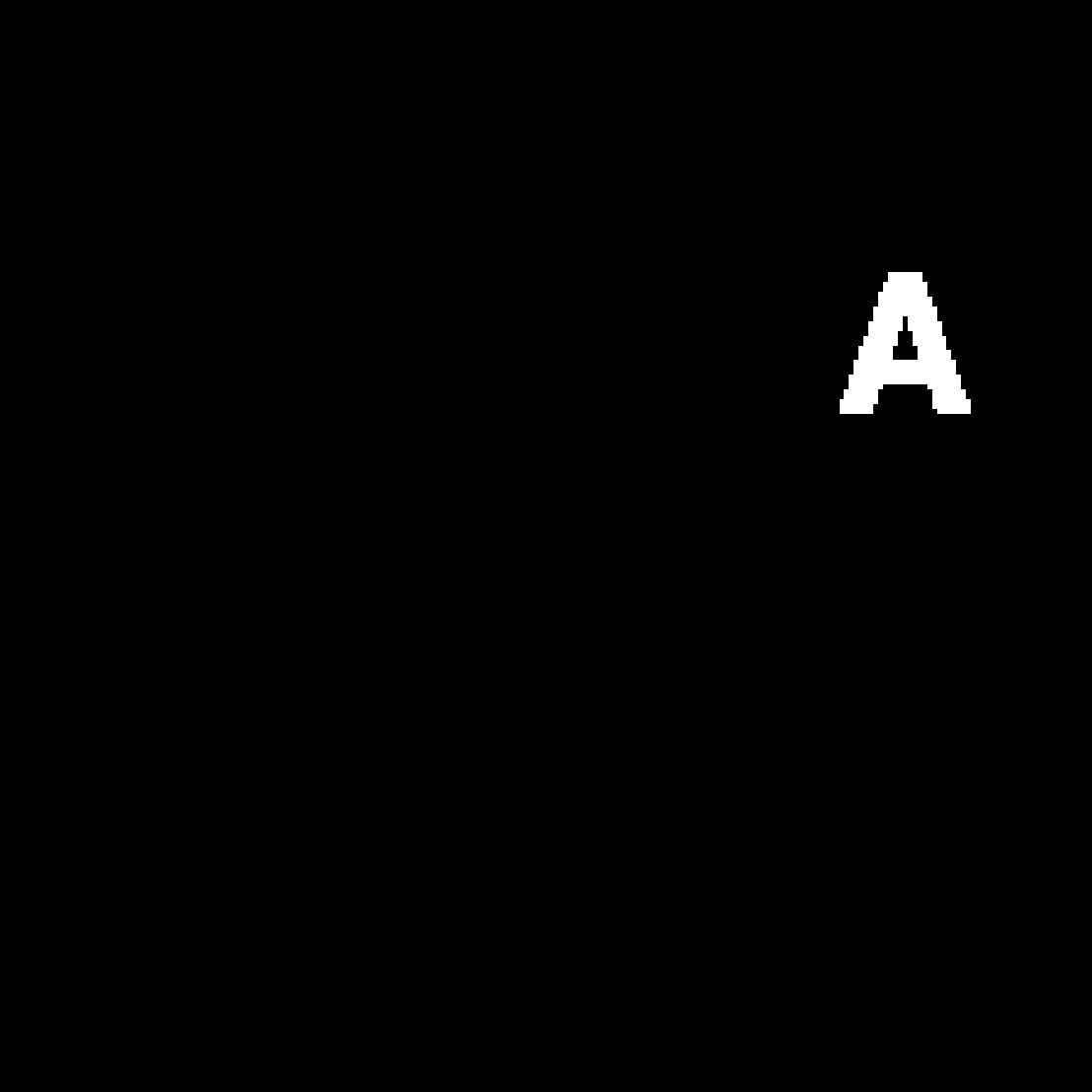}
\includegraphics[width=0.09\textwidth]{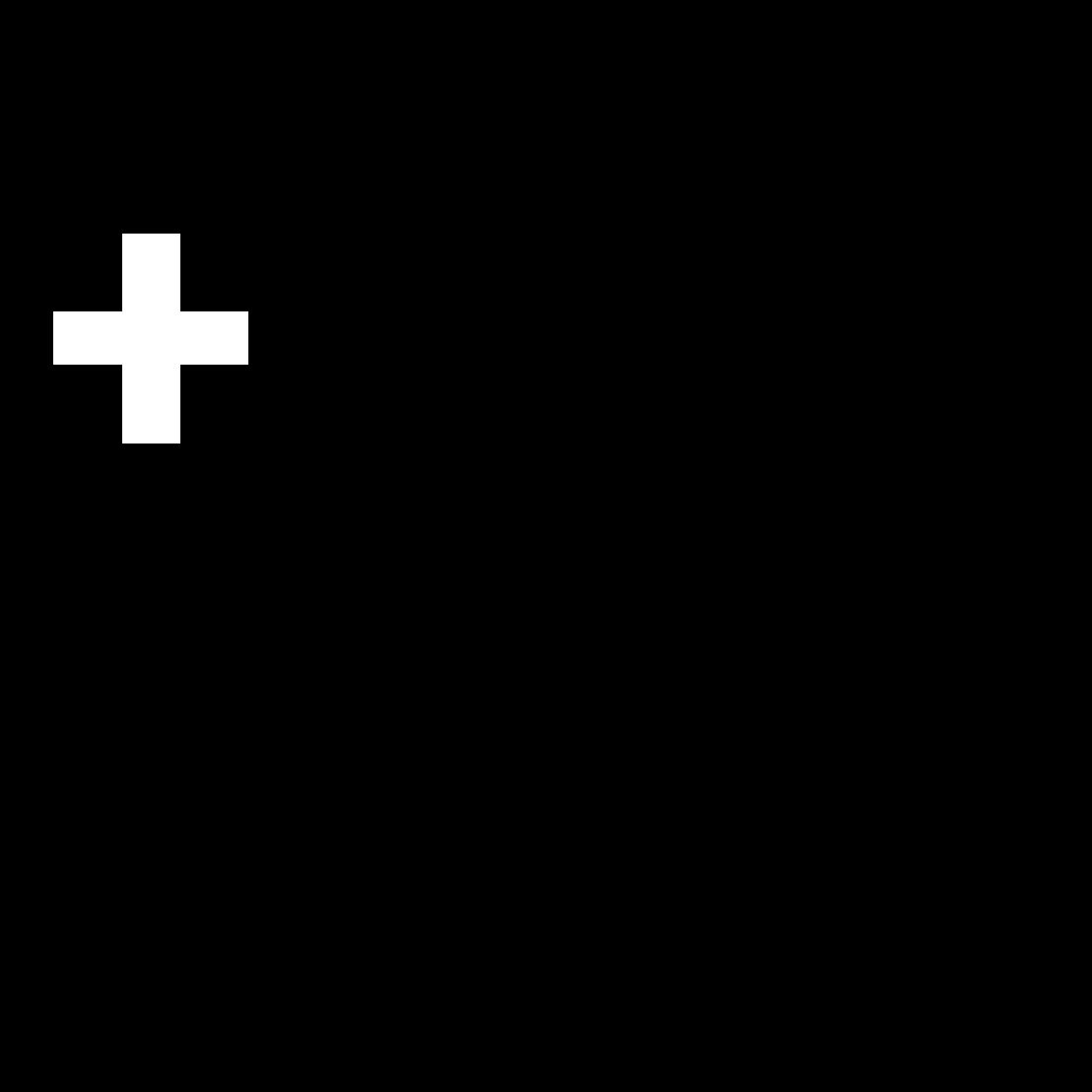}
\includegraphics[width=0.09\textwidth]{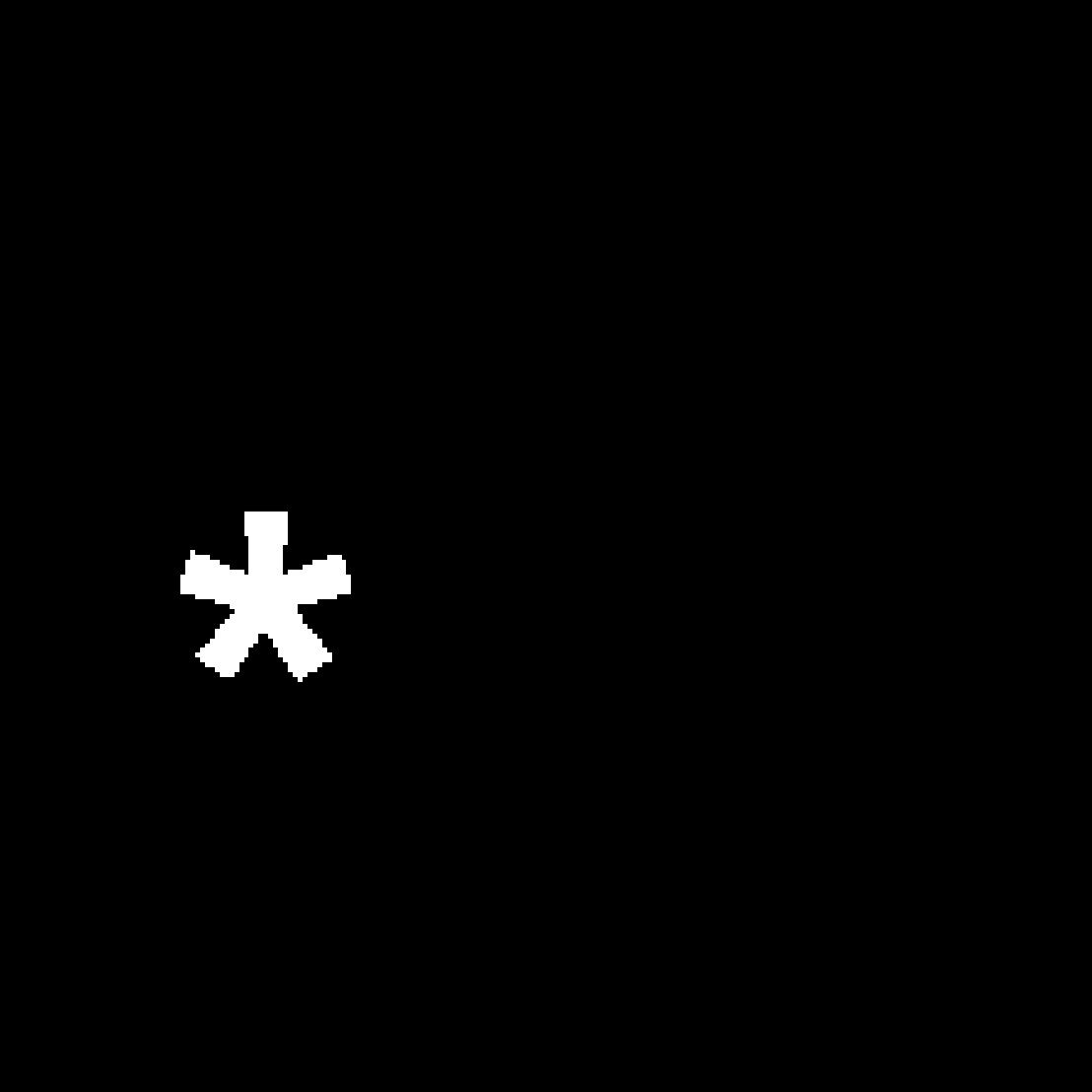}
\includegraphics[width=0.09\textwidth]{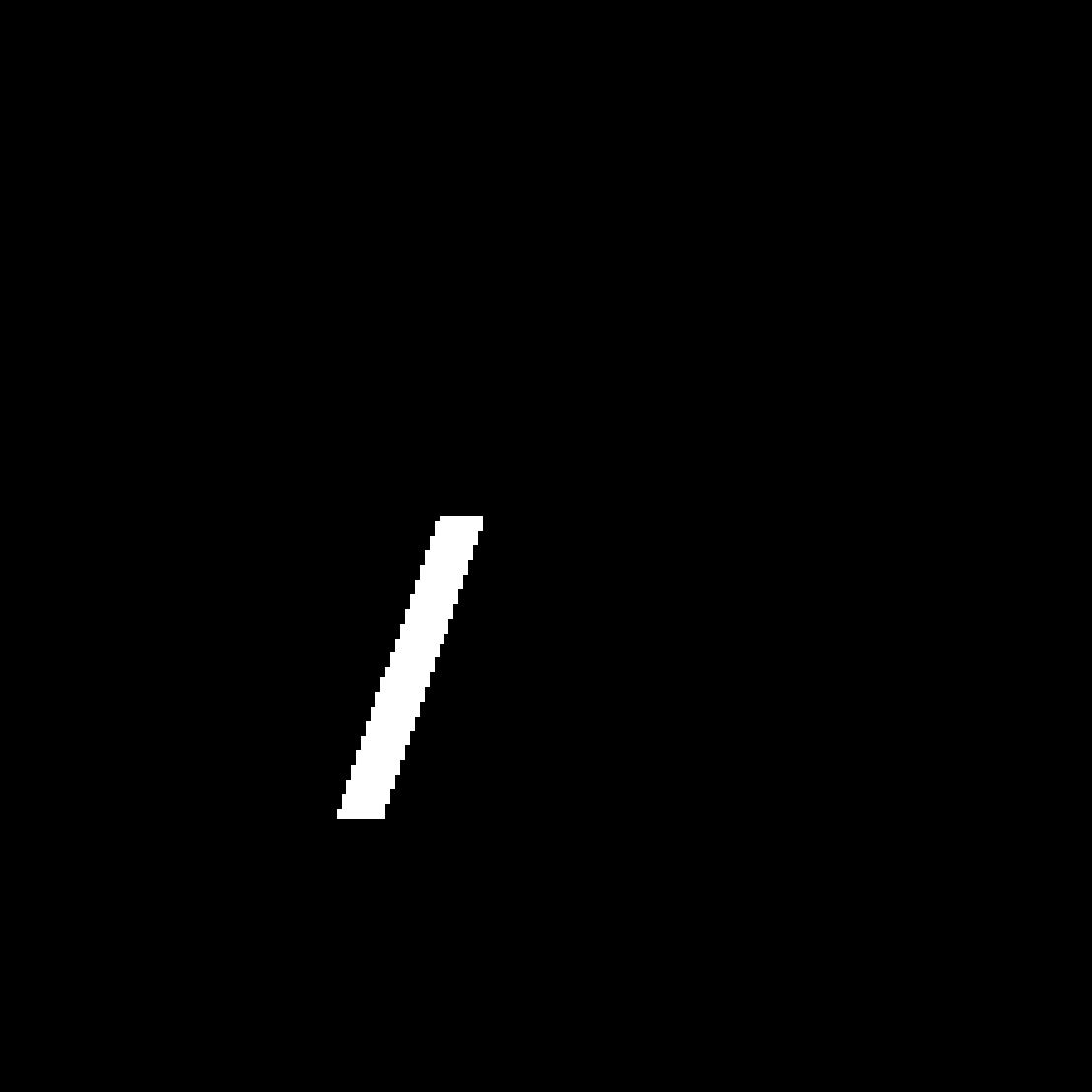}
}

\subfigure[Primitives $p_6$, $p_7$, and $p_8$ from class A.]{
\label{fig:ABplus dataset primitives class A}
\includegraphics[width=0.09\textwidth]{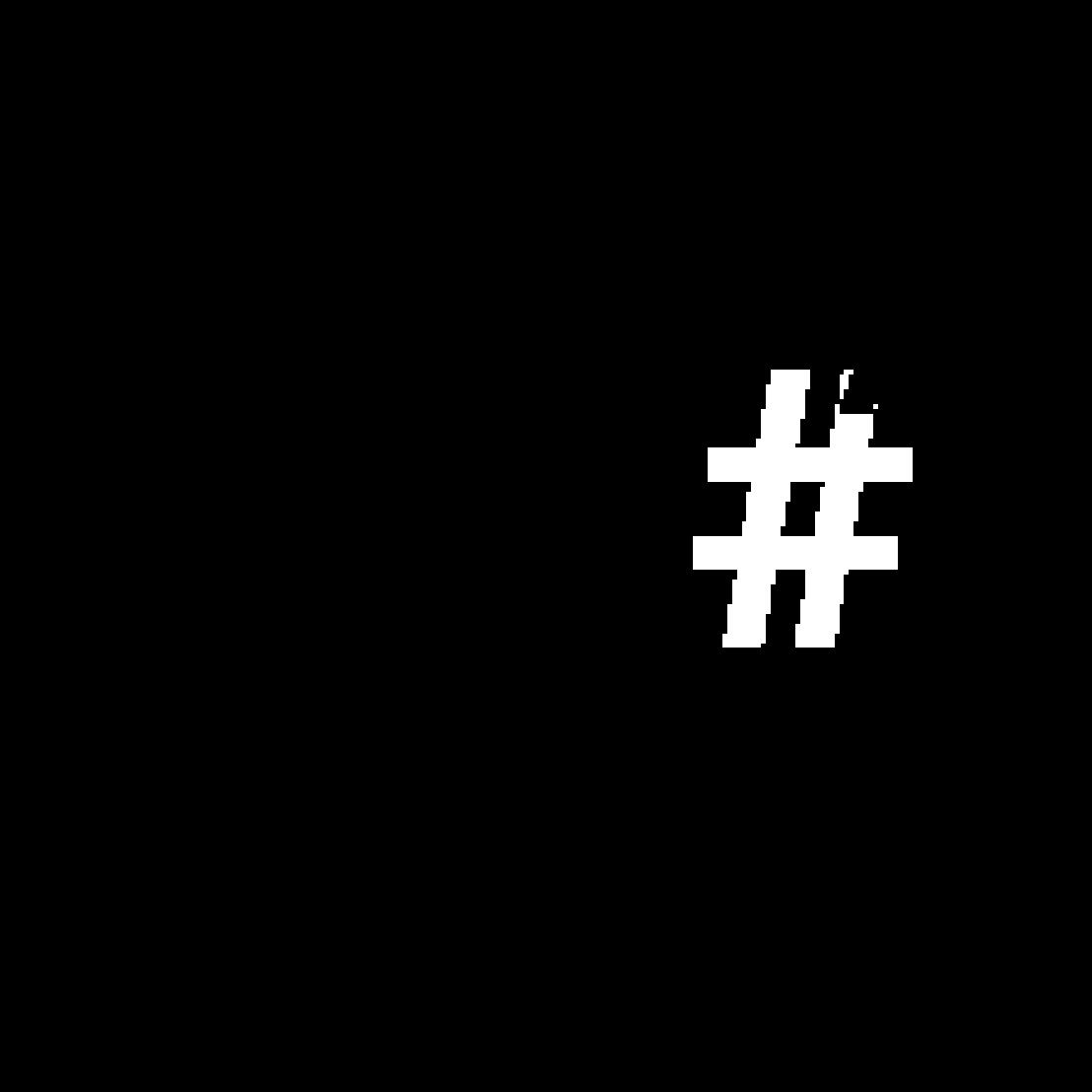}
\includegraphics[width=0.09\textwidth]{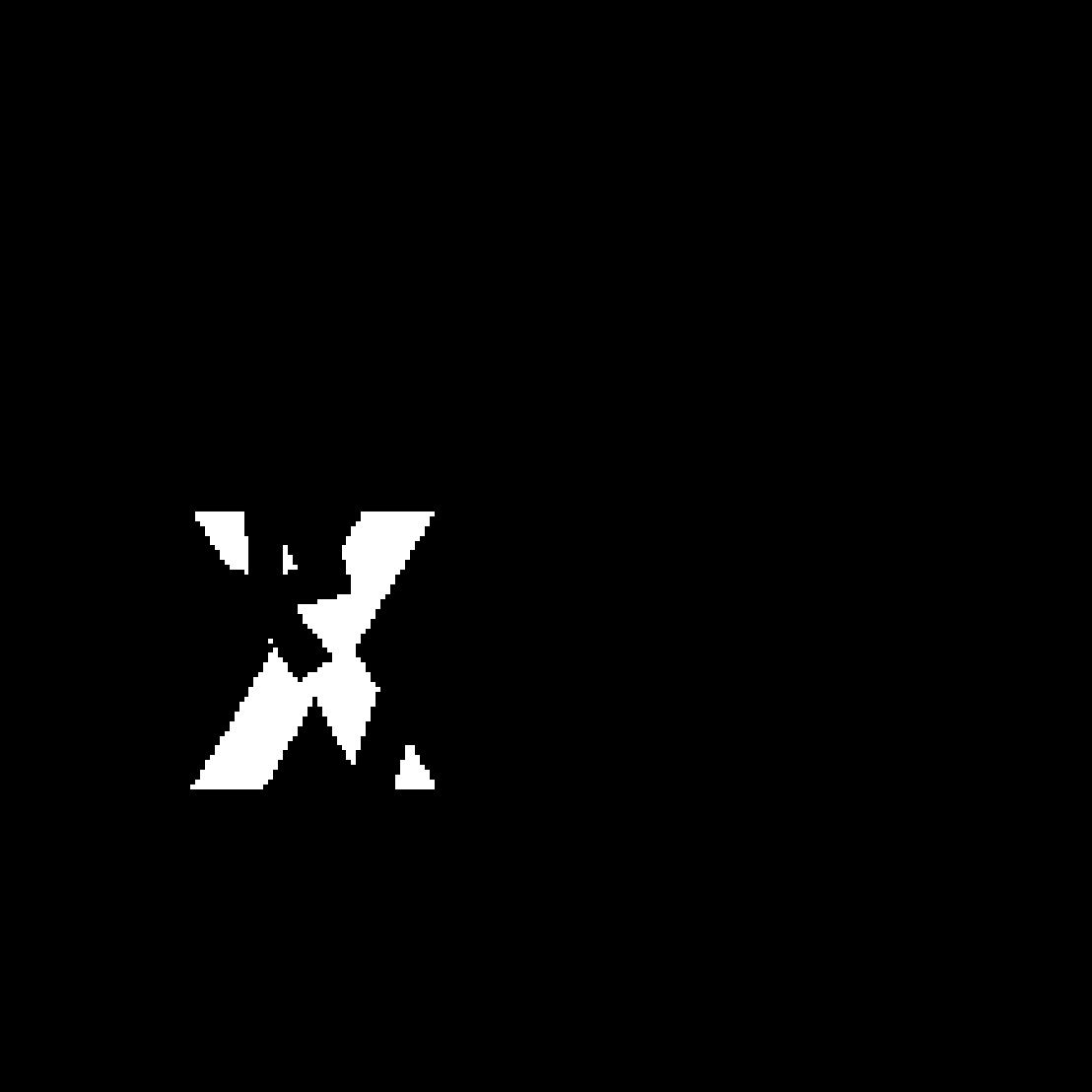}
\includegraphics[width=0.09\textwidth]{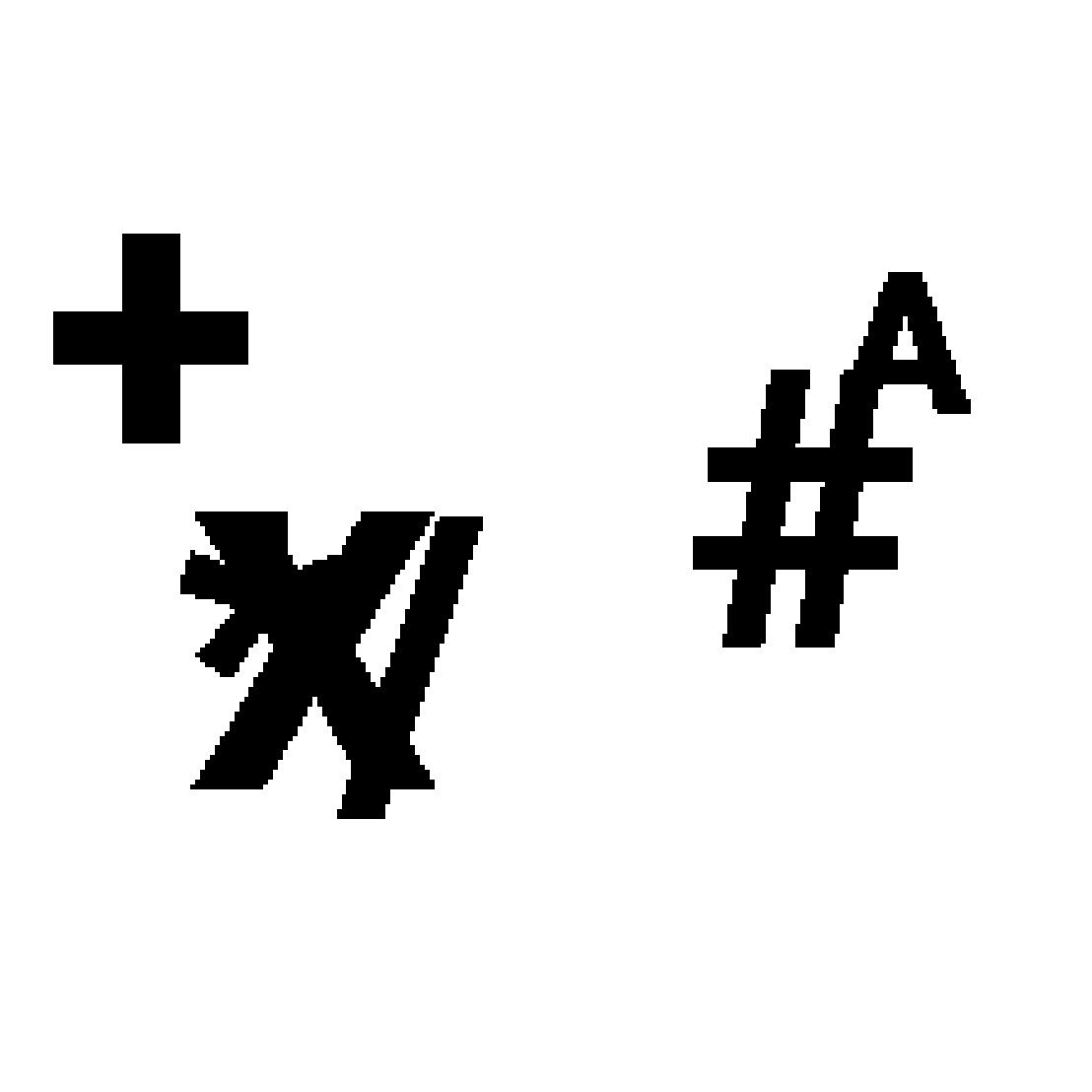}
}

\subfigure[Primitives $p_2$, $p_3$, $p_4$, and $p_5$ from class B.]{
\label{fig:ABplus dataset primitives class B}
\includegraphics[width=0.09\textwidth]{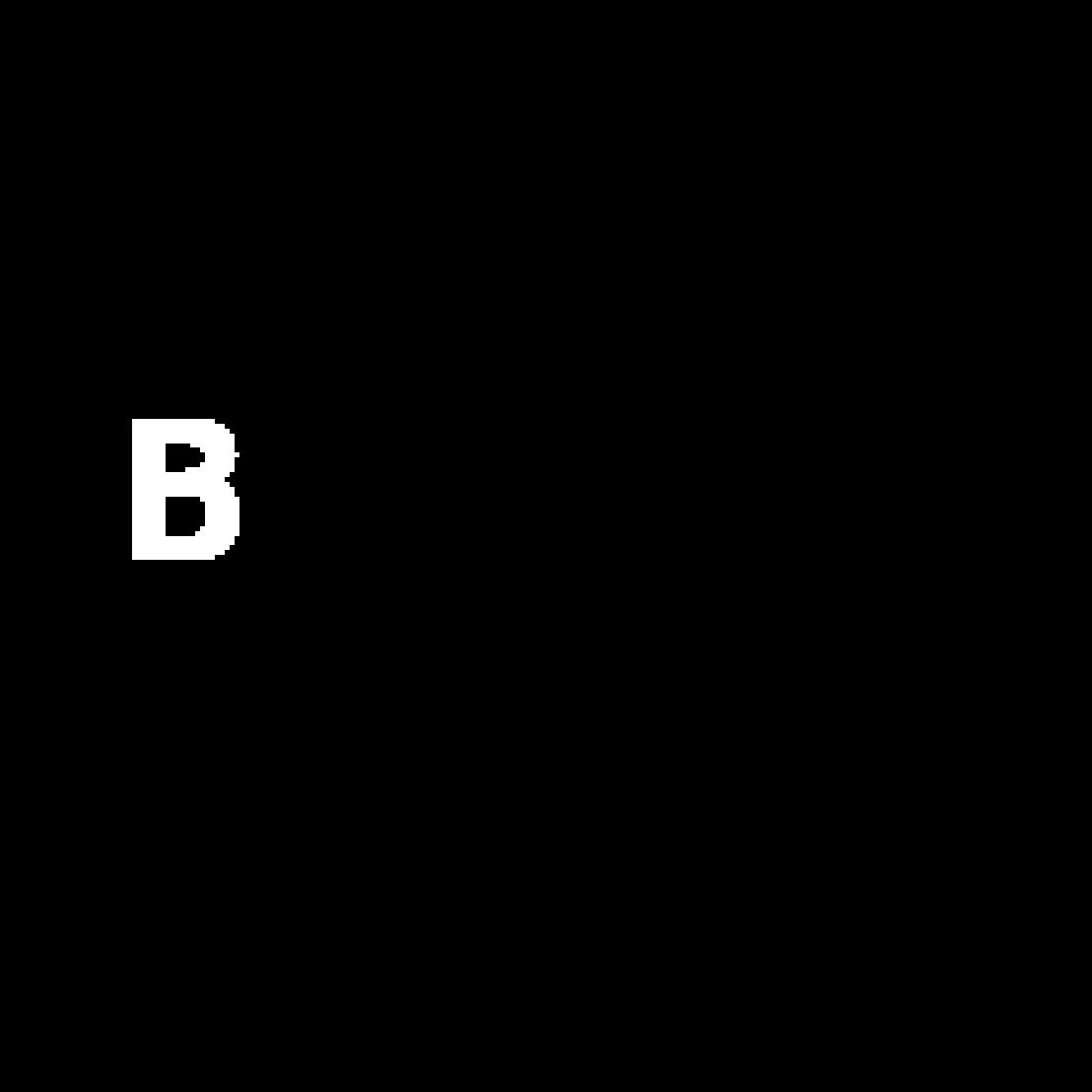}
\includegraphics[width=0.09\textwidth]{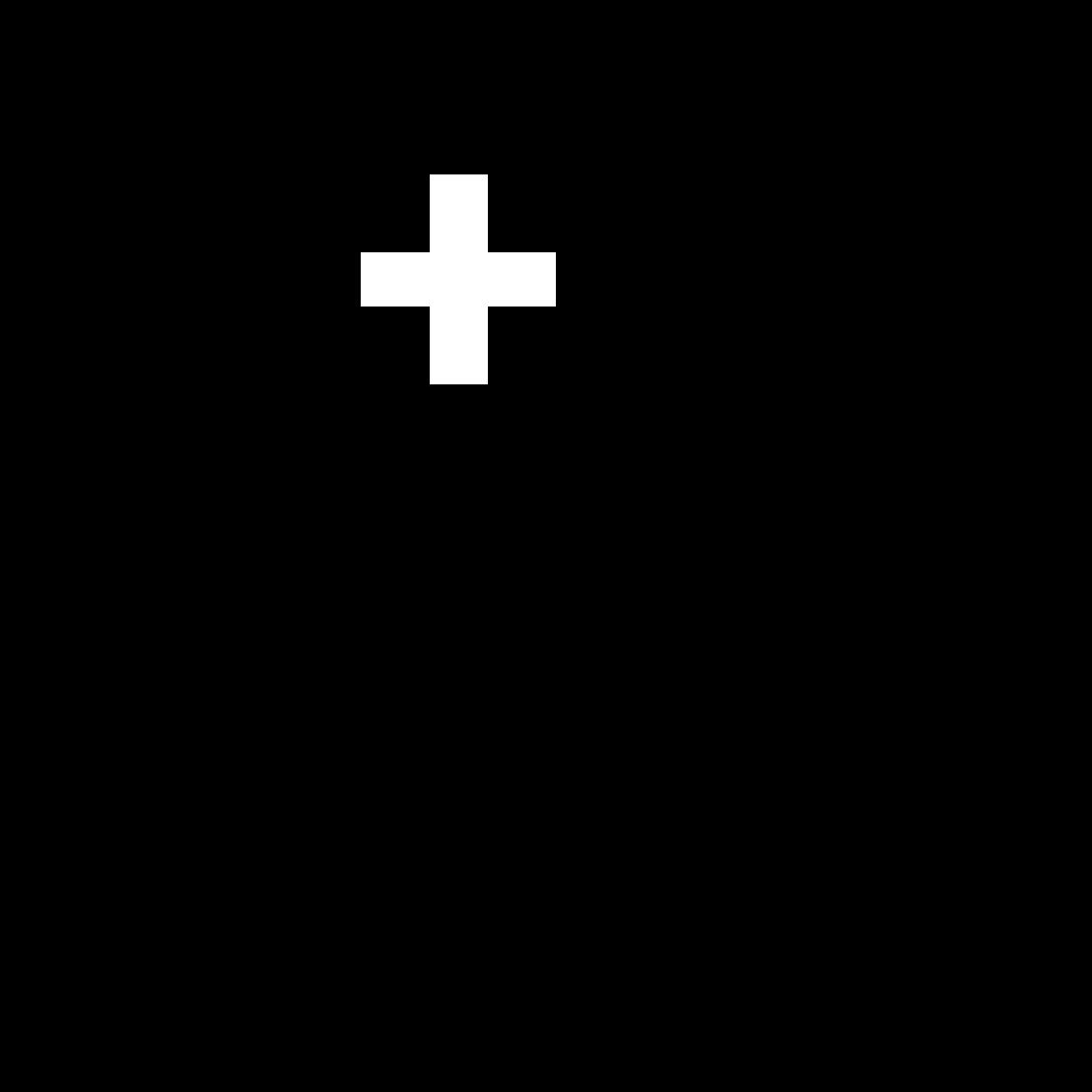}
\includegraphics[width=0.09\textwidth]{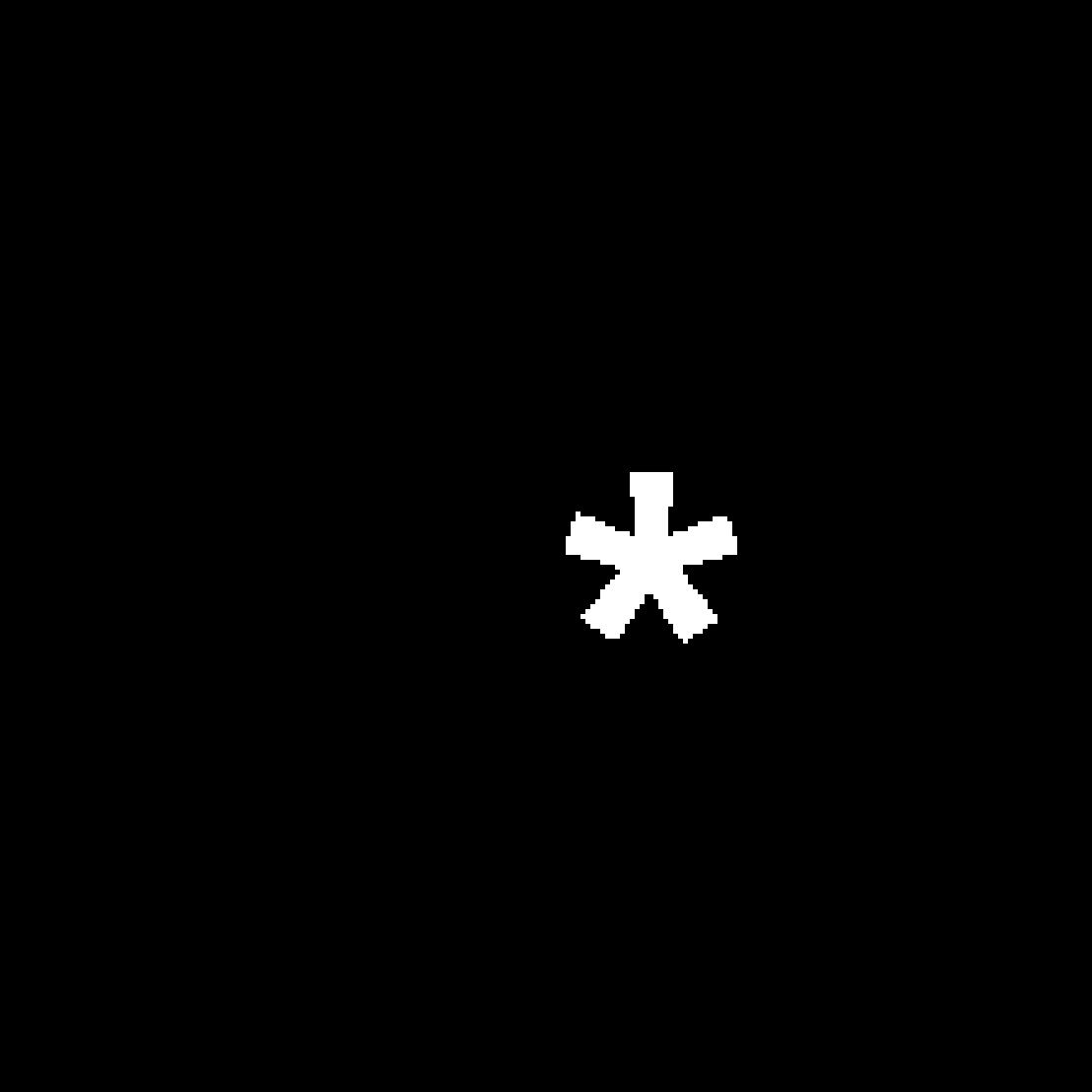}
\includegraphics[width=0.09\textwidth]{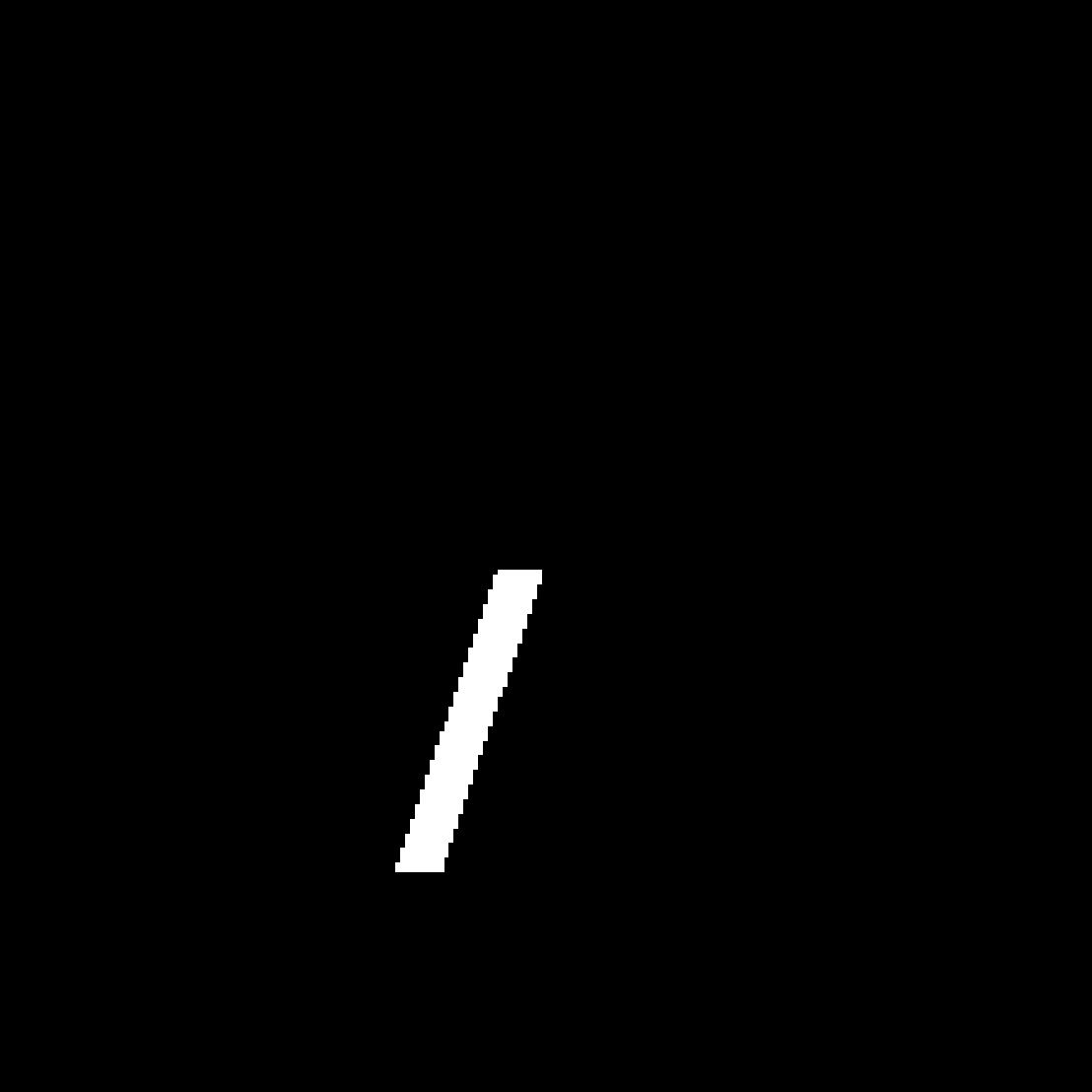}
}

\subfigure[Primitives $p_6$, $p_7$, and $p_8$ from class B.]{
\label{fig:ABplus dataset primitives class B}
\includegraphics[width=0.09\textwidth]{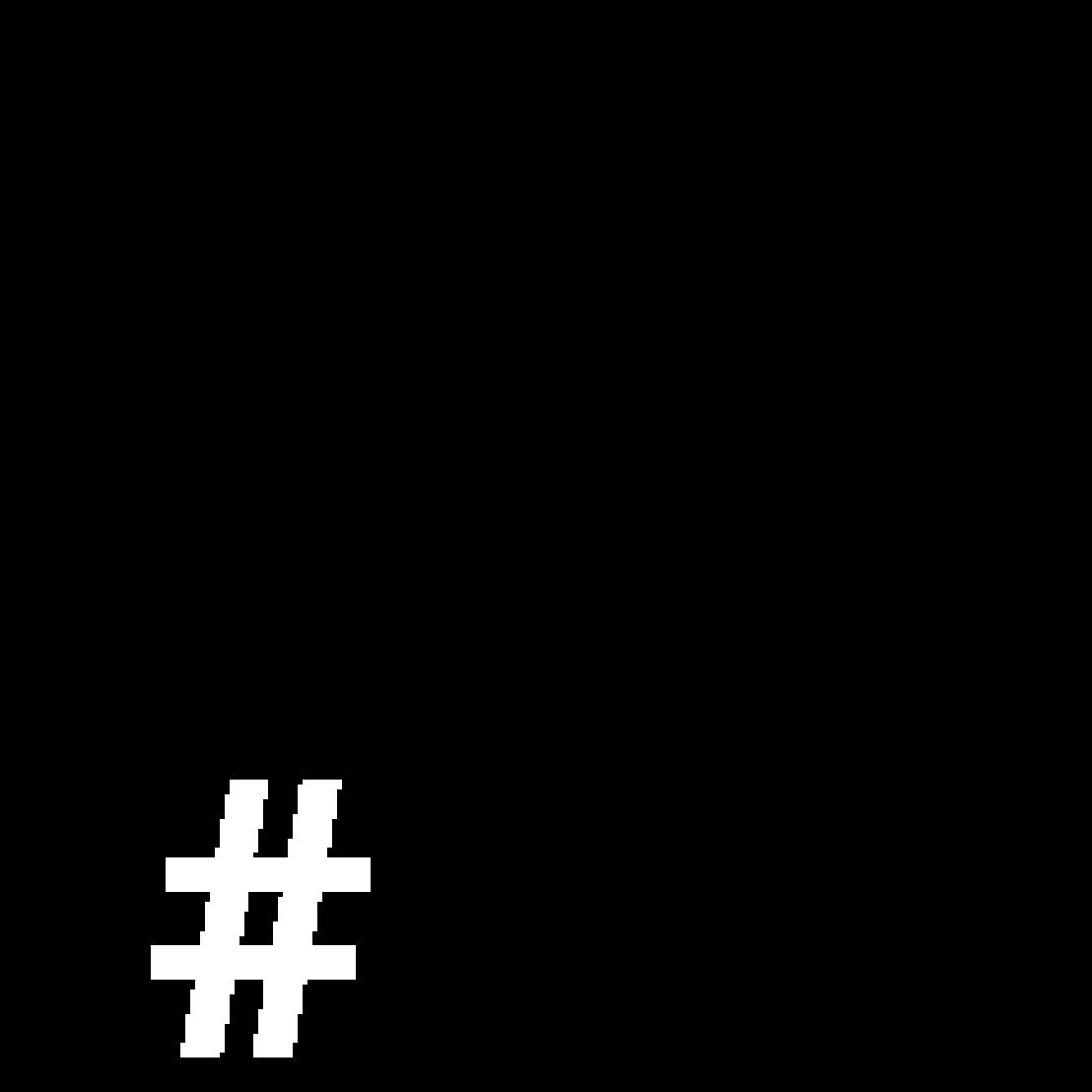}
\includegraphics[width=0.09\textwidth]{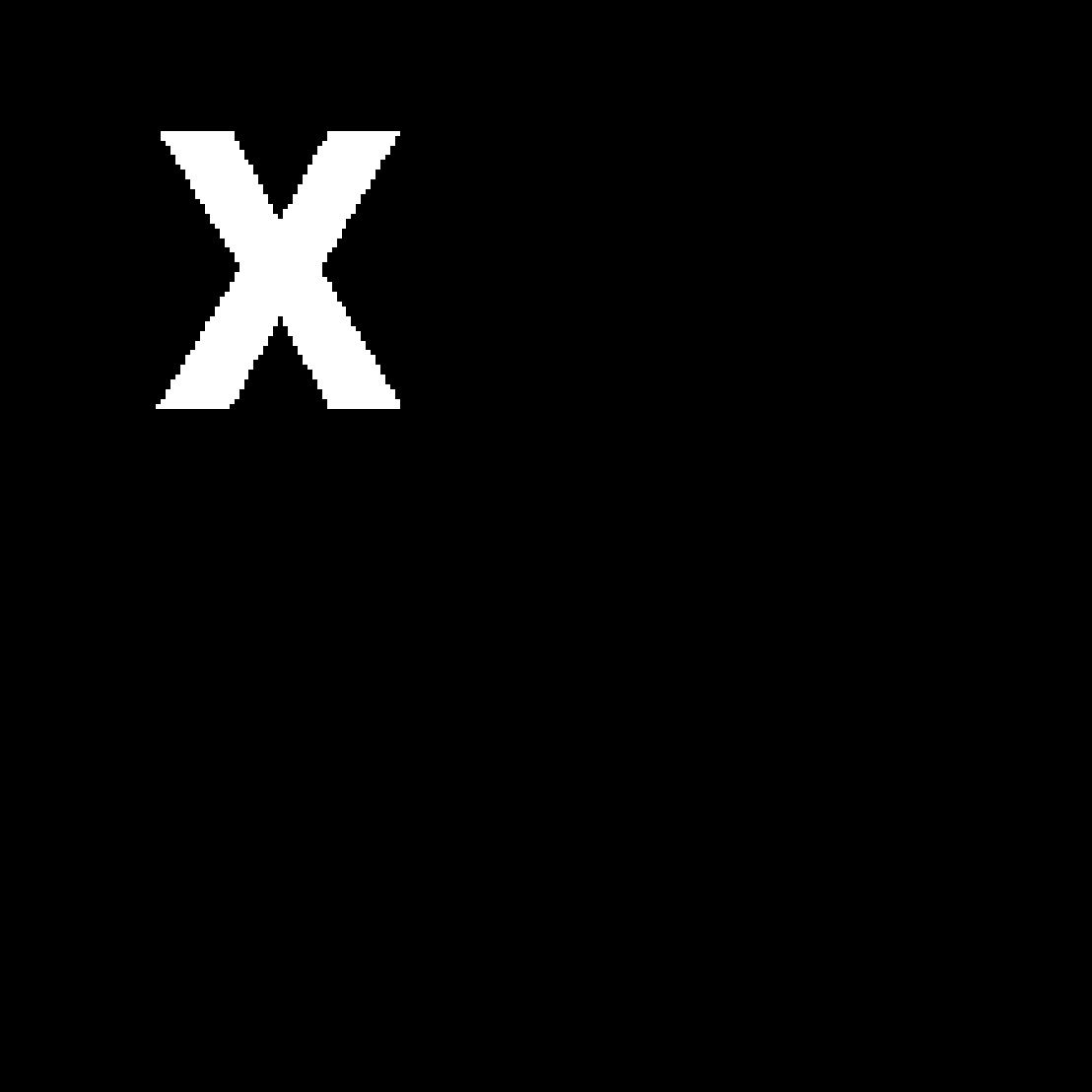}
\includegraphics[width=0.09\textwidth]{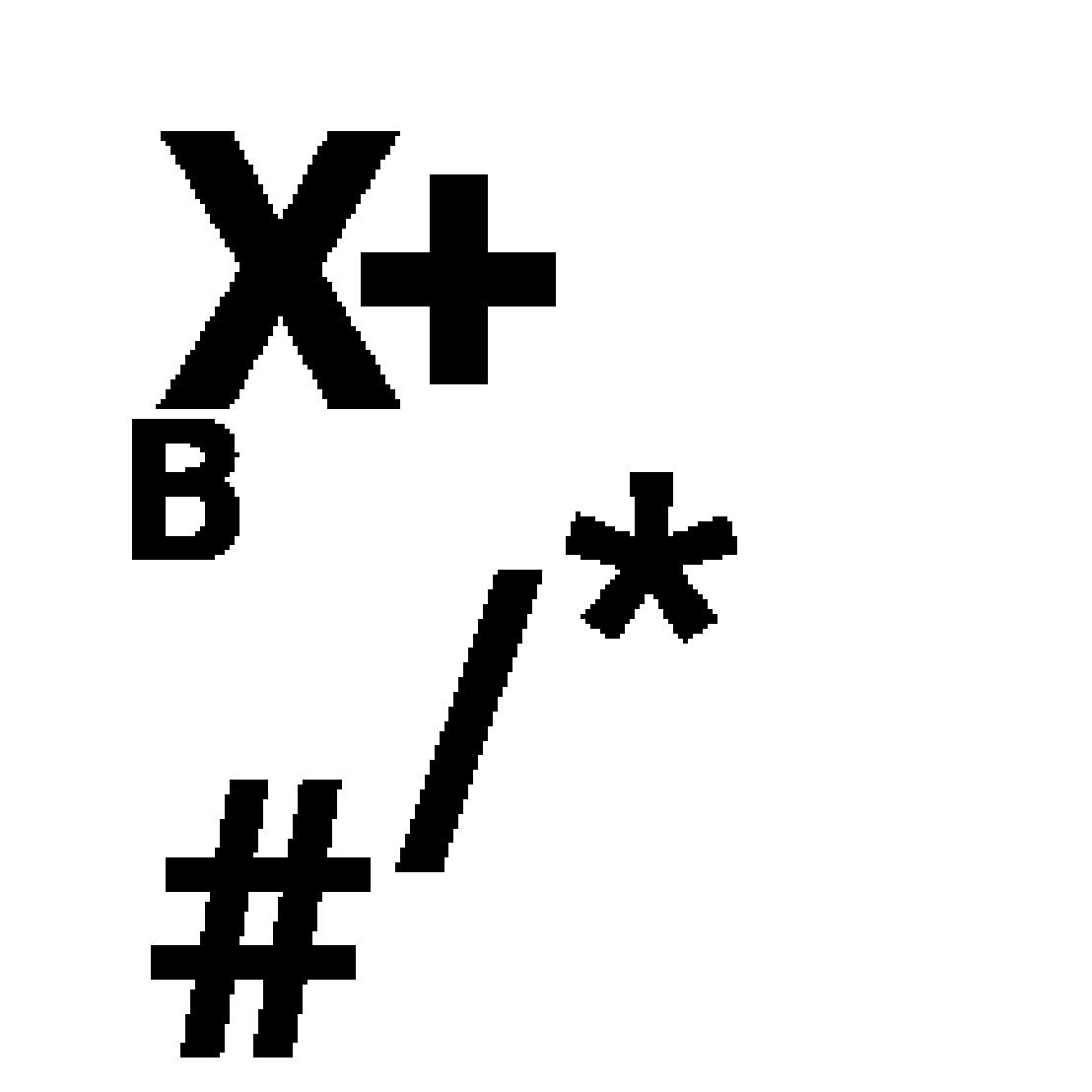}
}
\caption{Examples for both classes of the dataset ABplus are shown in Figures \ref{fig:ABplus dataset class A} (class A) and Figure \ref{fig:ABplus dataset class B} (class B). The primitives of class A are shown in Figure \ref{fig:ABplus dataset primitives class A} and the primitives of class B are shown in Figure (\ref{fig:ABplus dataset class B}).}
\label{fig:ABplus dataset}
\end{figure}

This dataset contains multiple irrelevant primitives with high contrast in random positions. This allows to test CE methods to observe and quantify their performance in cases with more feature variety. Thus, although a model may learn representations for some primitives, the irrelevant ones should be scored with low importance.

\newpage
\subsubsection{Big-Small dataset}

The \emph{Big-Small} dataset, contains images of two classes. The first one, class big has 100 pixels high letters \emph{B} filled in blue (primitive $p_1$). The second one, class small contains 40 pixels high letters \emph{B} also filled in blue (primitive $p_2$). The only difference between the two classes is the size of the letter \emph{B}. Primitive $p_3$, refers to an intrusive character \emph{$+$}, filled with a cotton textile and balanced between the two classes. Finally, the background of all images is annotated as the primitive $p_4$, and is filled with a cork texture.

\begin{figure}[h]
\centering
\subfigure[Examples class big]{
\label{fig:BigSmall dataset class Big}
\includegraphics[width=0.11\textwidth]{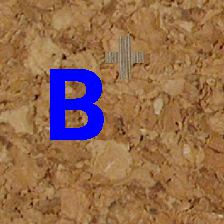}
\includegraphics[width=0.11\textwidth]{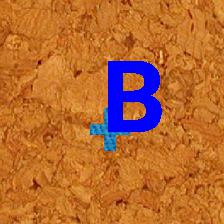}
}\subfigure[Examples class small]{
\label{fig:BigSmall dataset class Small}
\includegraphics[width=0.11\textwidth]{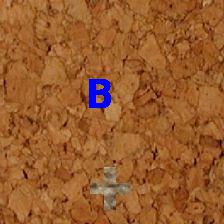}
\includegraphics[width=0.11\textwidth]{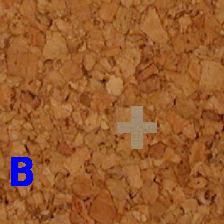}
}

\subfigure[Primitives $p_1$, $p_3$, and $p_4$ from class big.]{
\label{fig:BigSmall dataset primitives class Big}
\includegraphics[width=0.12\textwidth]{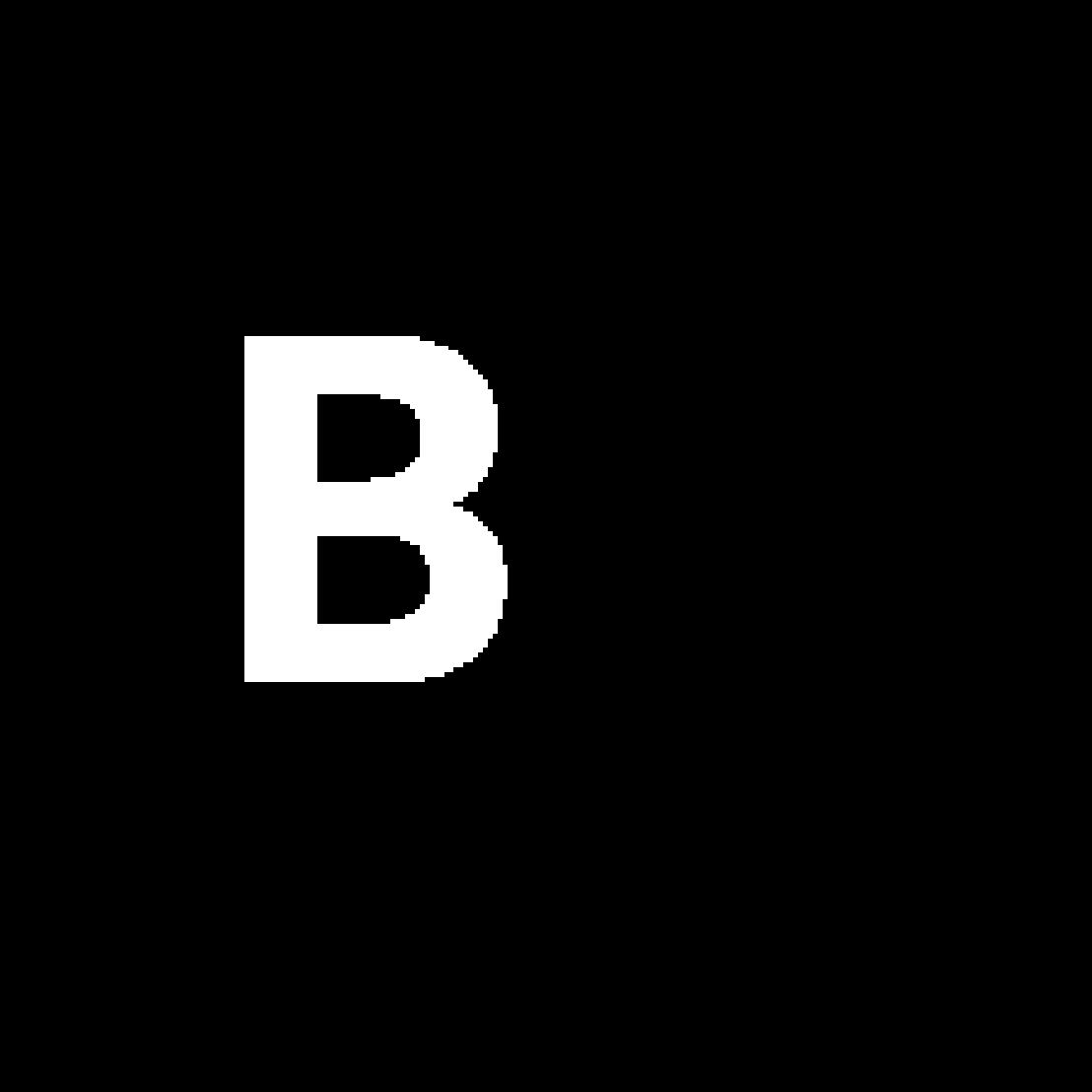}
\includegraphics[width=0.12\textwidth]{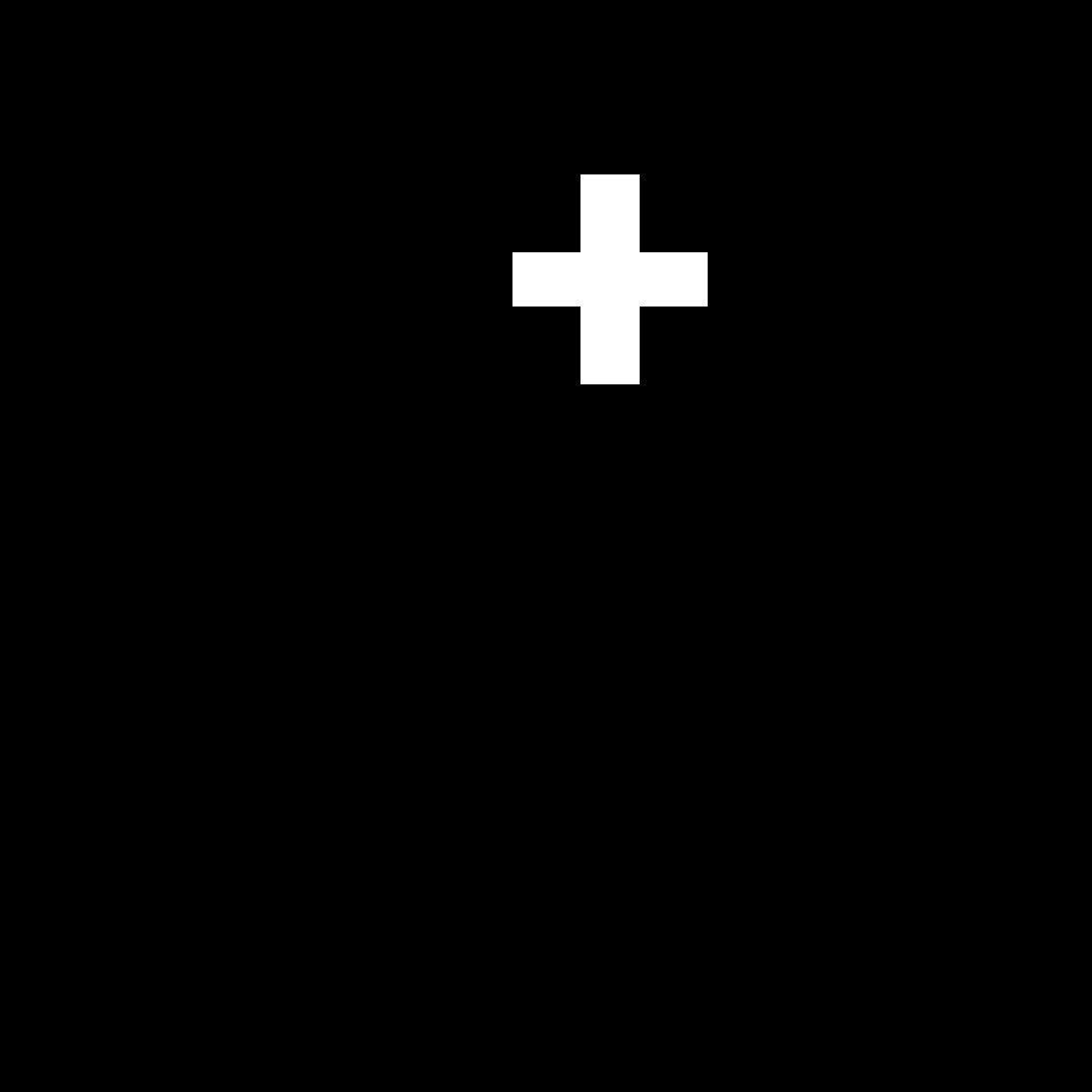}
\includegraphics[width=0.12\textwidth]{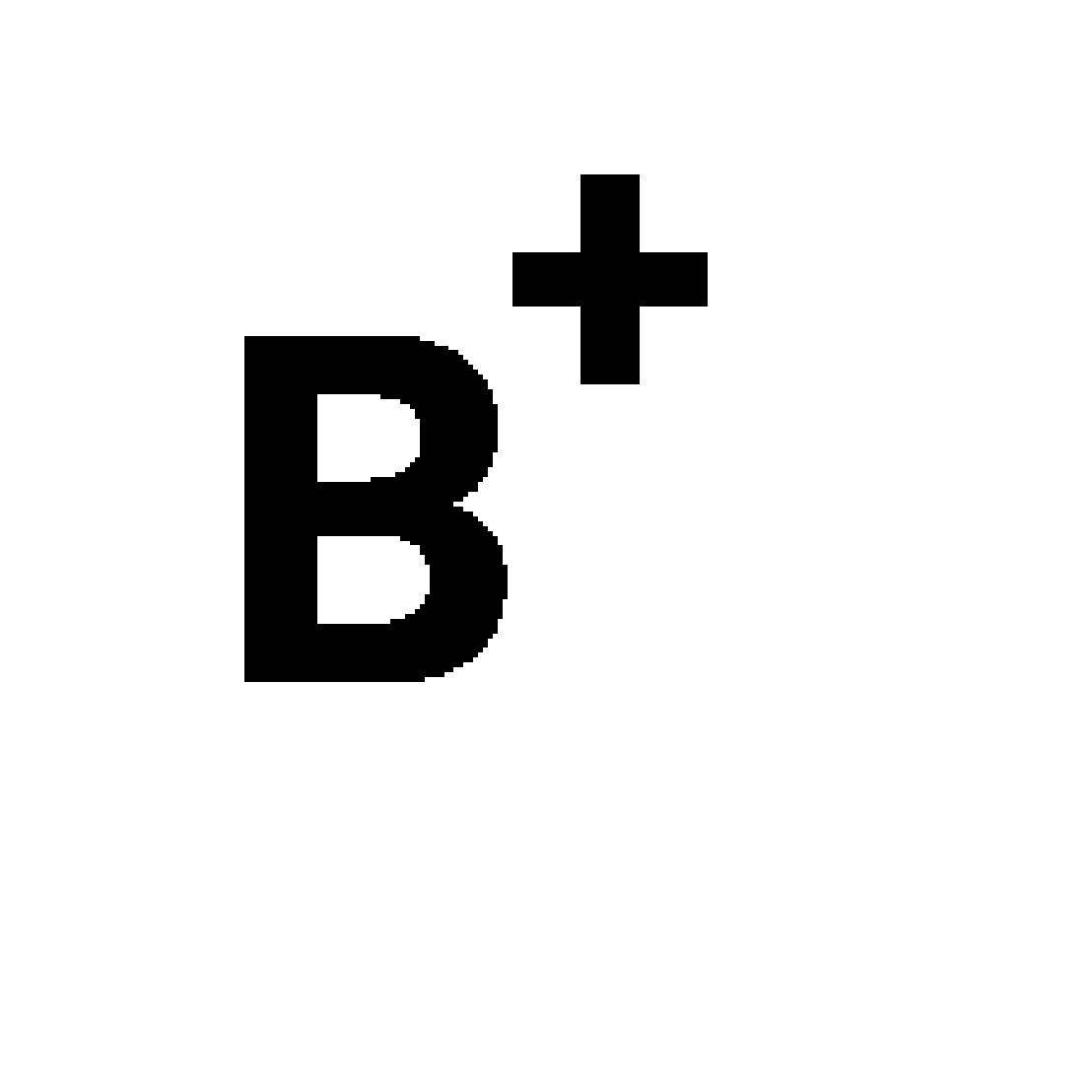}
}
\subfigure[Primitives $p_2$, $p_3$, and $p_4$ from class small.]{
\label{fig:BigSmall dataset primitives class Small}
\includegraphics[width=0.12\textwidth]{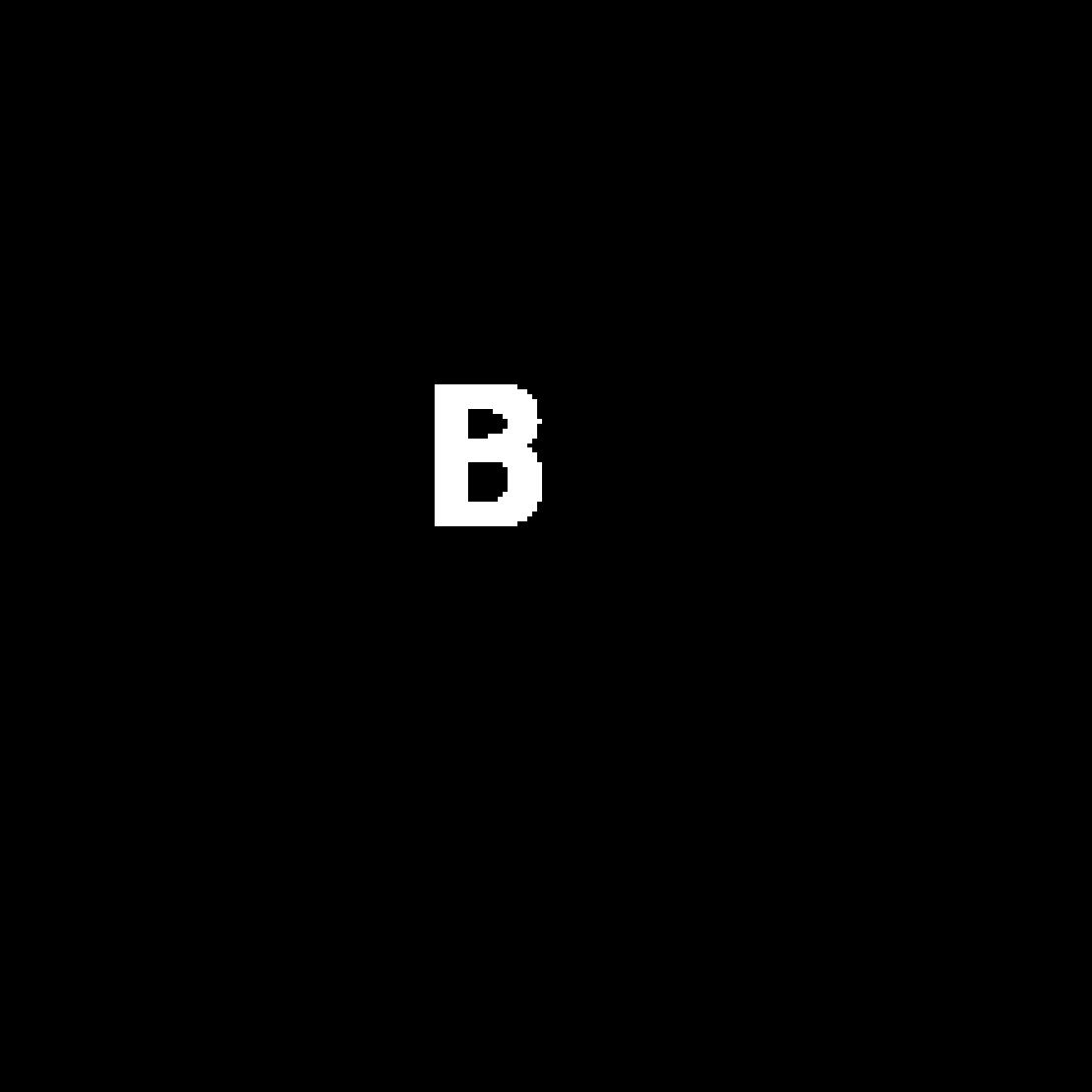}
\includegraphics[width=0.12\textwidth]{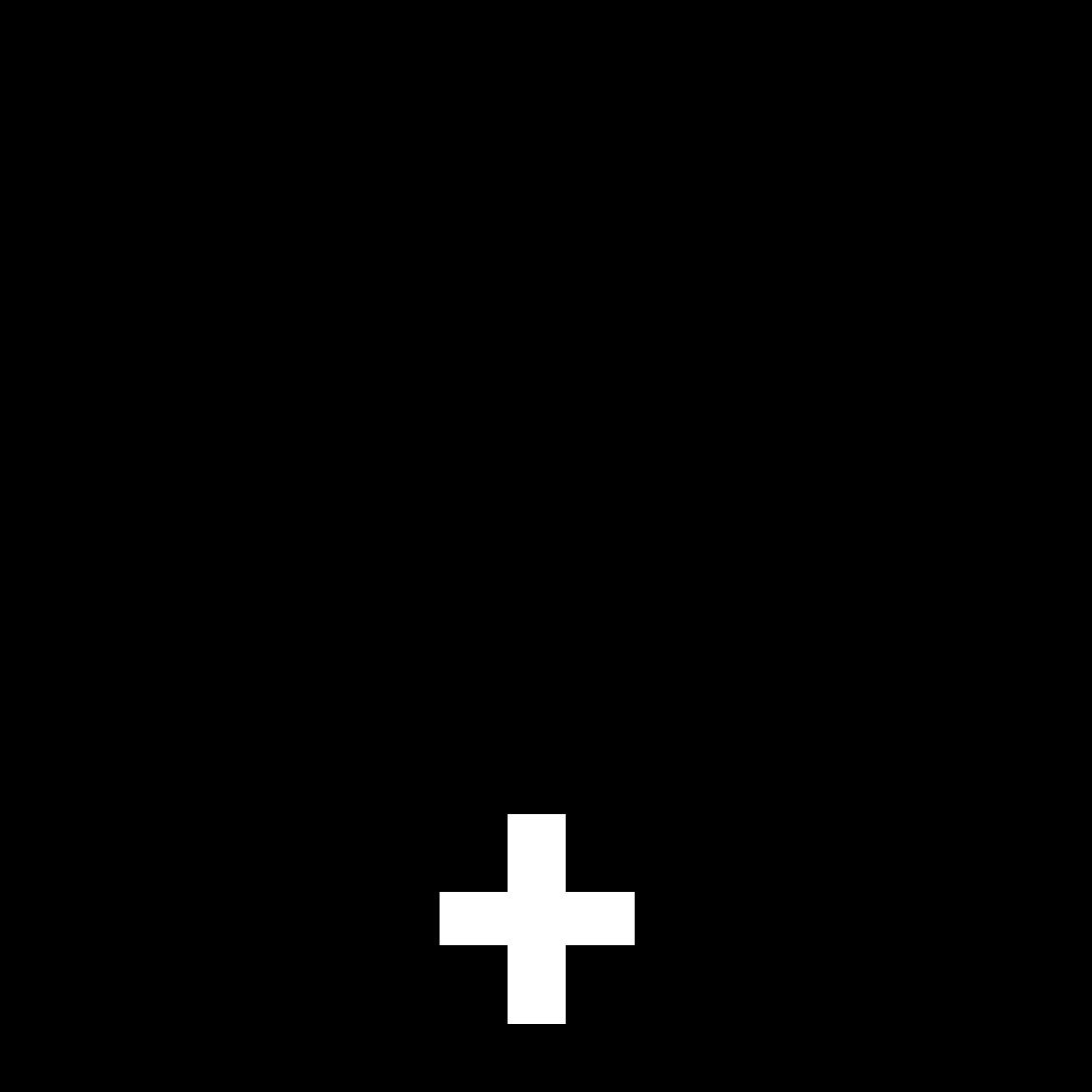}
\includegraphics[width=0.12\textwidth]{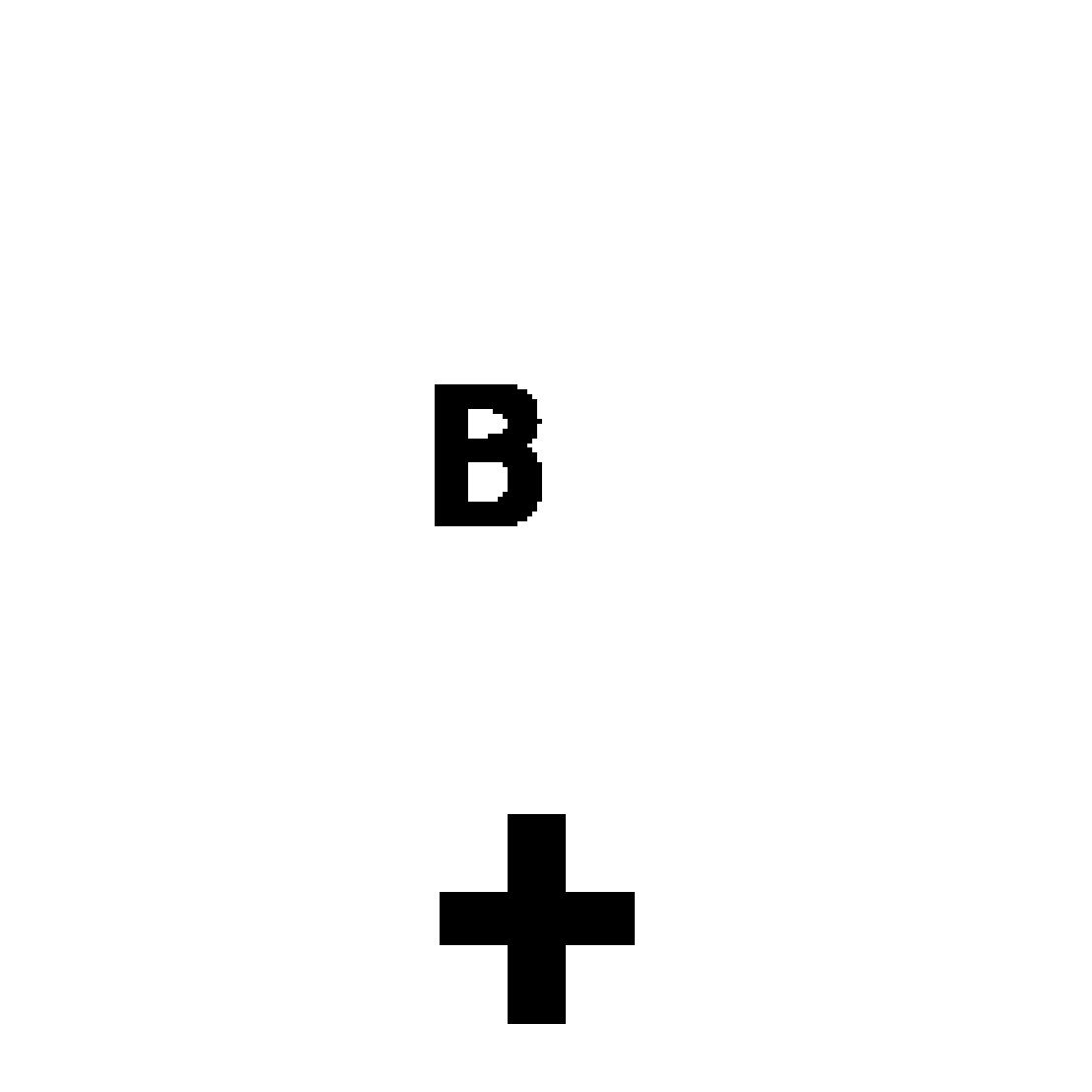}
}\caption{Big-Small dataset. Examples of class big are shown in Figure \ref{fig:BigSmall dataset class Big}, containing big characters \emph{B}, and examples of the class small are shown in Figure \ref{fig:BigSmall dataset class Small}, containing small \emph{B}s. Class Big contains primitives $p_1$, $p_2$, and $p_3$ (Figure \ref{fig:BigSmall dataset primitives class Big}); and class small contains primitives $p_1$, $p_2$, and $p_3$  shown in Figure (\ref{fig:BigSmall dataset primitives class Small}).}
\label{fig:BigSmall dataset}
\end{figure}

The Big-Small dataset aims to test how different CE techniques respond to instances where scale is the differentiating factor of classes. This is a common case in real-world applications such as quality control, or medical diagnosis.

\newpage
\subsubsection{CO dataset}

The task for the dataset \emph{CO} consists in differentiating images with a character \emph{C} (primitive $p_1$) in class C, or a character \emph{O} (primitive $p_2$) in class O. Both $p_1$ and $p_2$ are filled with an aluminum foil texture, and their only difference is that the \emph{O} is closed on the right. Primitive $p_3$ is the character \emph{$+$} filled with a cotton texture, which appears in both classes. Similarly $p_4$ refers to the background, filled with a cork texture.

\begin{figure}[h]
\centering
\subfigure[Examples class C]{
\label{fig:CO dataset class C}
\includegraphics[width=0.11\textwidth]{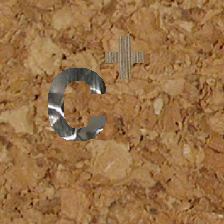}
\includegraphics[width=0.11\textwidth]{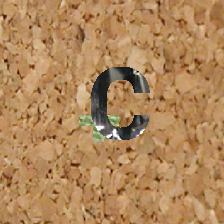}
}\subfigure[Examples class O]{
\label{fig:CO dataset class O}
\includegraphics[width=0.11\textwidth]{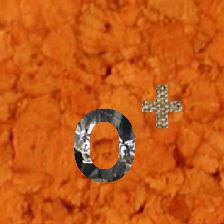}
\includegraphics[width=0.11\textwidth]{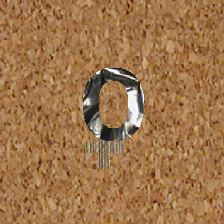}
}

\subfigure[Primitives $p_1$, $p_3$, and $p_4$ from Class C.]{
\label{fig:CO dataset primitives class C}
\includegraphics[width=0.12\textwidth]{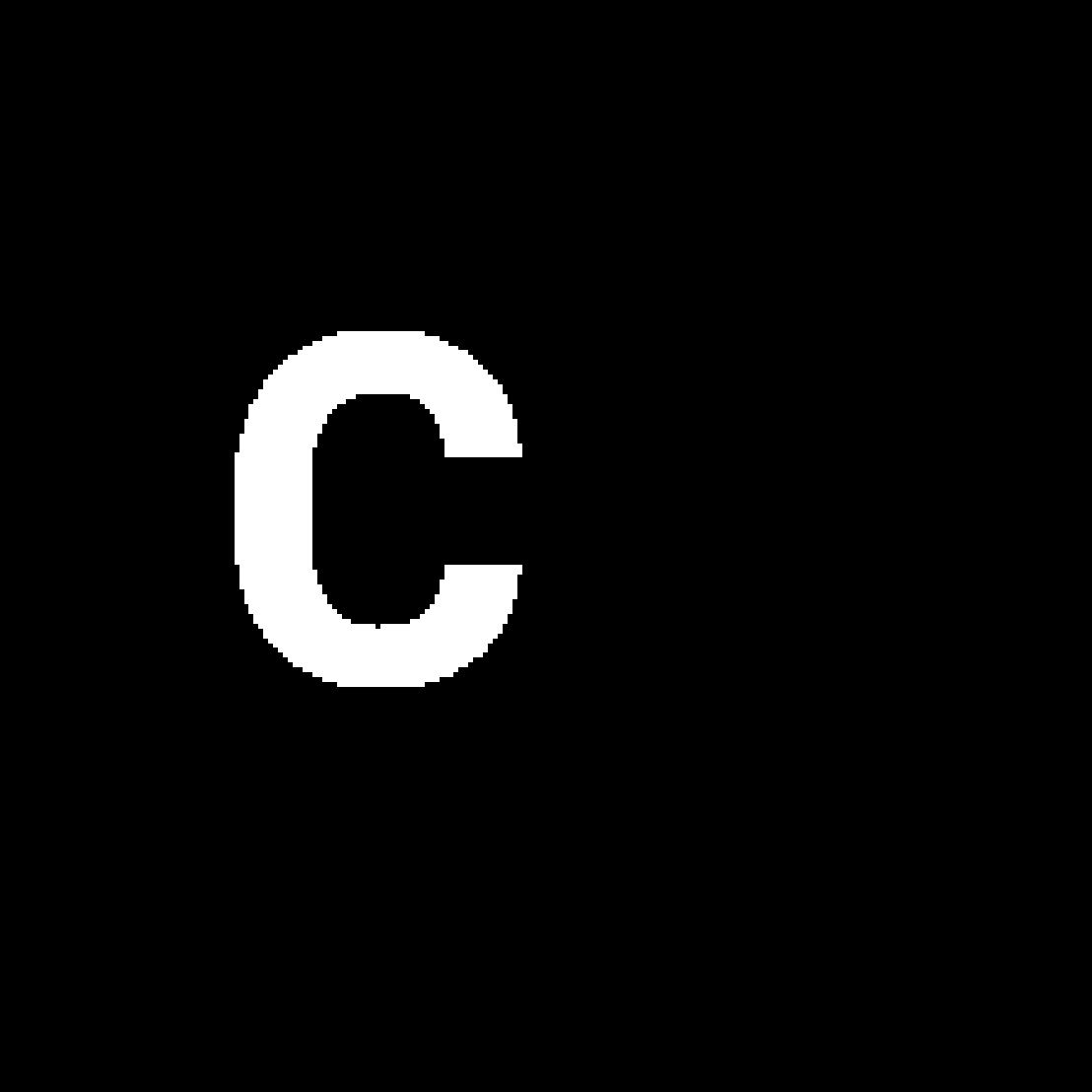}
\includegraphics[width=0.12\textwidth]{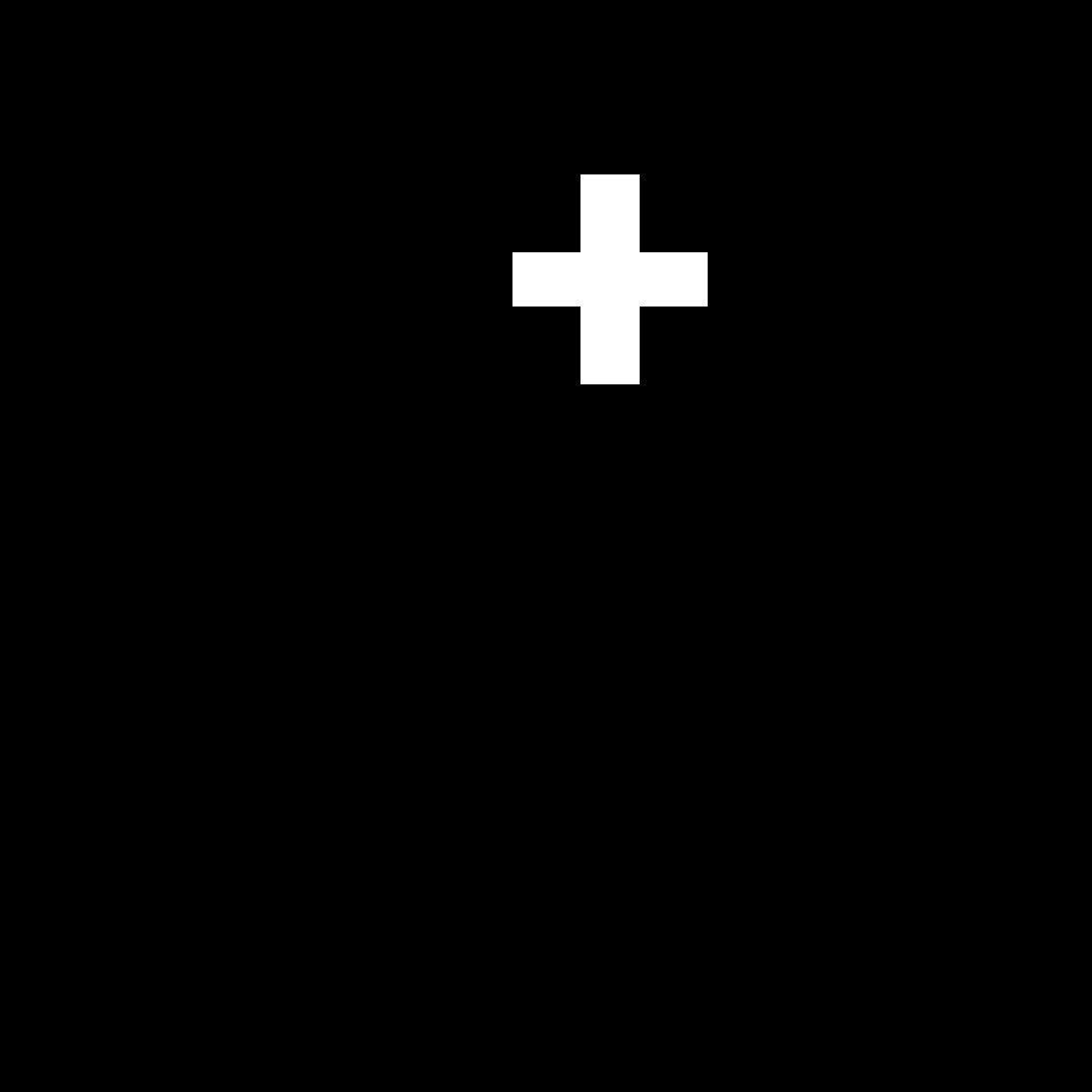}
\includegraphics[width=0.12\textwidth]{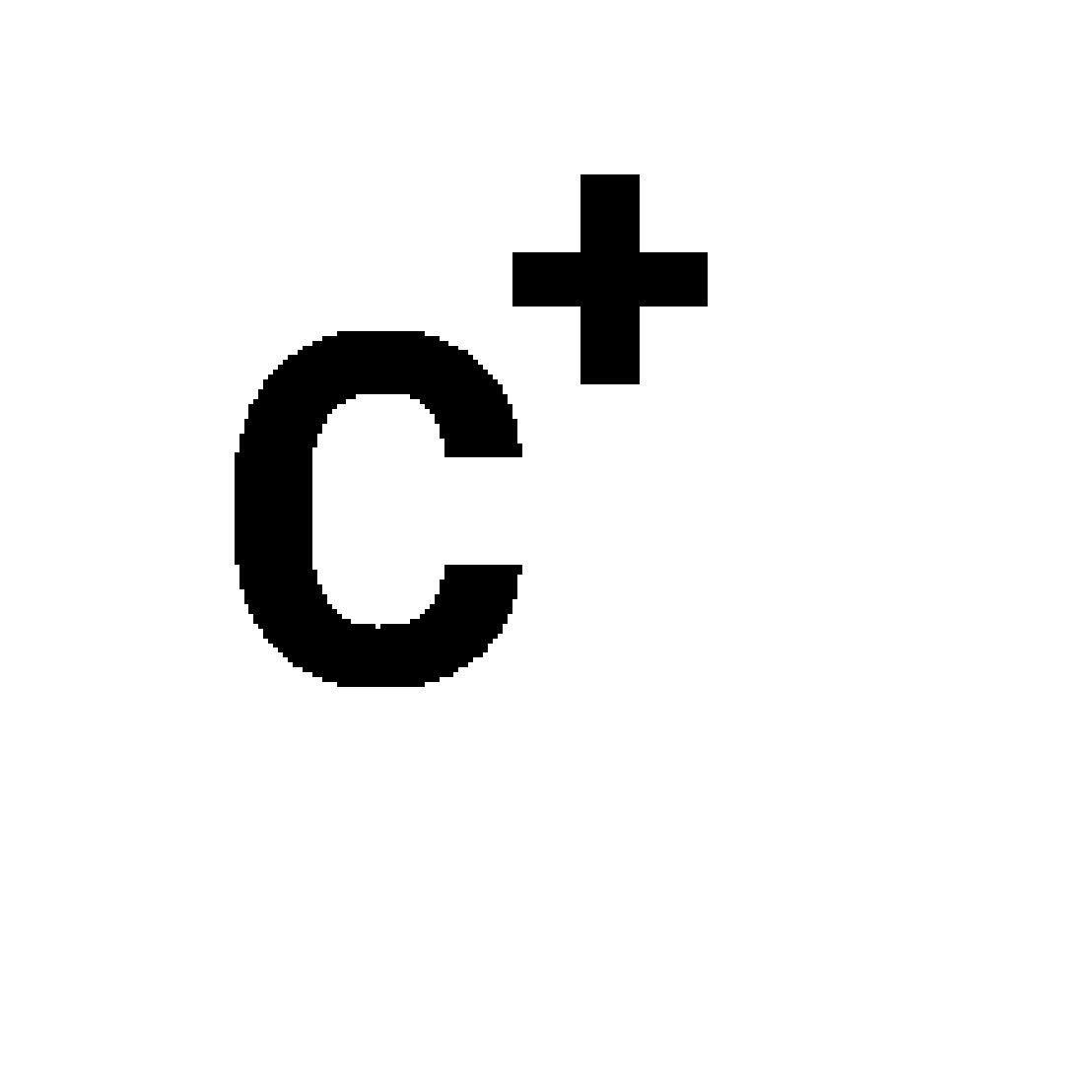}
}
\subfigure[Primitives $p_2$, $p_3$, and $p_4$ from class O.]{
\label{fig:CO dataset primitives class O}
\includegraphics[width=0.12\textwidth]{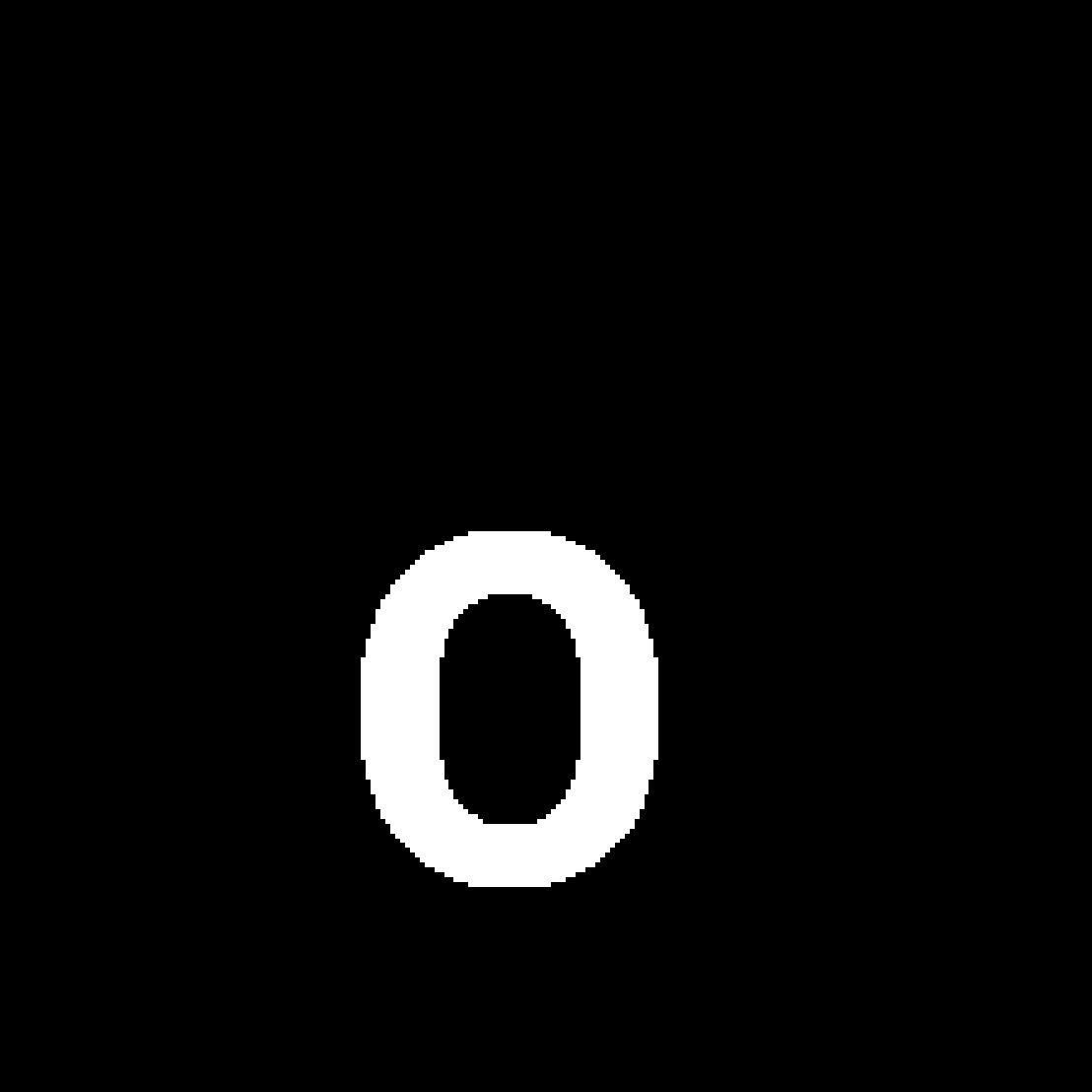}
\includegraphics[width=0.12\textwidth]{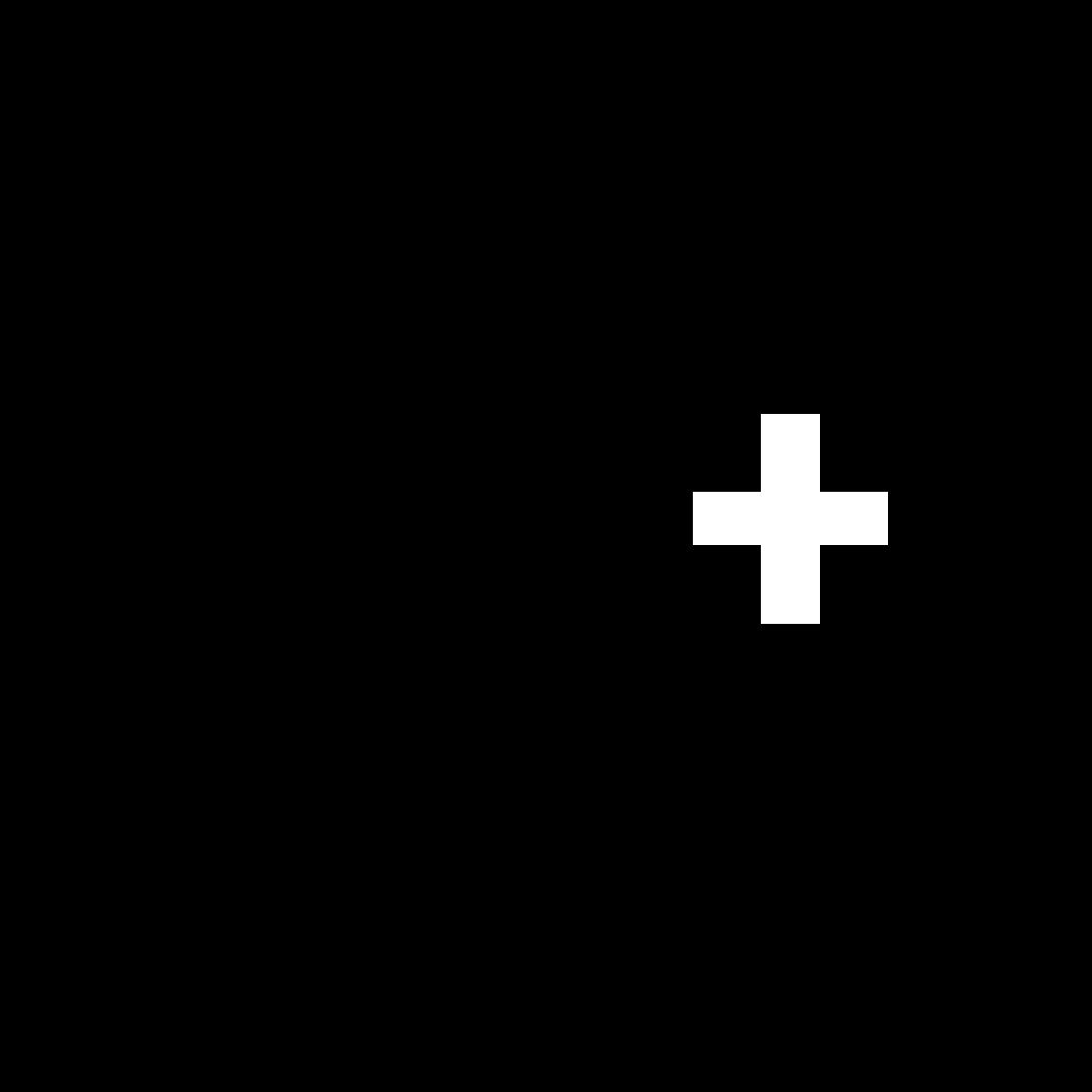}
\includegraphics[width=0.12\textwidth]{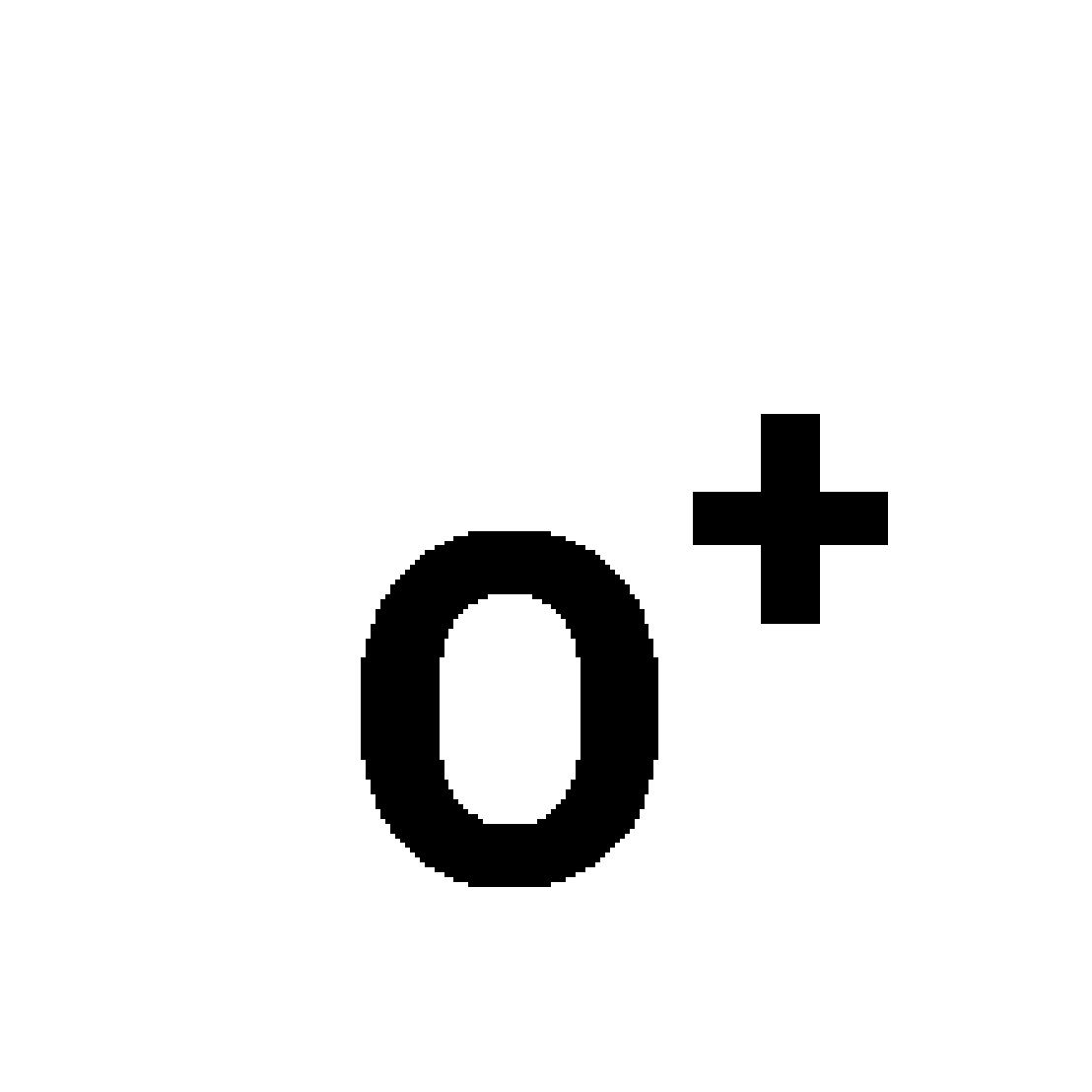}
}\caption{Examples of the dataset CO, where one class are images with the letter \emph{C} (Figure \ref{fig:CO dataset class C}), and the other has images containing the letter \emph{O} (Figure \ref{fig:CO dataset class O}. Primitives for the class C are shown in Figure \ref{fig:CO dataset primitives class C}, whereas primitives of class O are shown in Figure \ref{fig:CO dataset primitives class O}.}
\label{fig:CO dataset}
\end{figure}

This dataset was designed to test how CE algorithms perform when dealing with completeness issues. The actual difference between the two classes is the right side of the \emph{O}, which should be the fasts way for CNNs to differentiate the images. This main feature is part of the primitives, but patch extraction techniques may have issues detecting features that are important \emph{by omission}.

\newpage
\subsubsection{colorGB dataset}

The dataset \emph{colorGB} consists in detecting if a letter in the image is of color blue, or if all the letters are of color green. In this regard, the dataset contains four primitives. The first primitive $p_1$ can be either the character \emph{A} or \emph{B}, which is randomly selected and always appears in the images. The color of $p_1$ determines if the class is B blue, or G green. The second primitive $p_2$ is a random green character between \emph{C}, \emph{D} or blank (not appearing), yet, it is balanced between both classes. Finally, All images contain primitives $p_3$ and $p_4$ denoting a symbol \emph{$+$} and the background, which are filed with cotton and orange peel texture, respectively.

\begin{figure}[h]
\centering
\subfigure[Examples class B]{
\label{fig:colorGB dataset class B}
\includegraphics[width=0.11\textwidth]{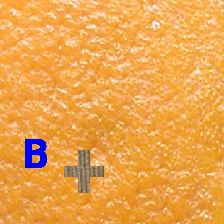}
\includegraphics[width=0.11\textwidth]{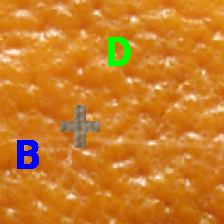}
}
\subfigure[Examples class G]{
\label{fig:colorGB dataset class G}
\includegraphics[width=0.11\textwidth]{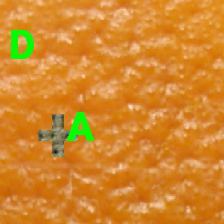}
\includegraphics[width=0.11\textwidth]{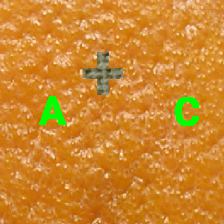}
}

\subfigure[Primitives $p_1$, $p_3$, and $p_4$ of class B.]{
\label{fig:colorGB dataset primitives class B}
\includegraphics[width=0.12\textwidth]{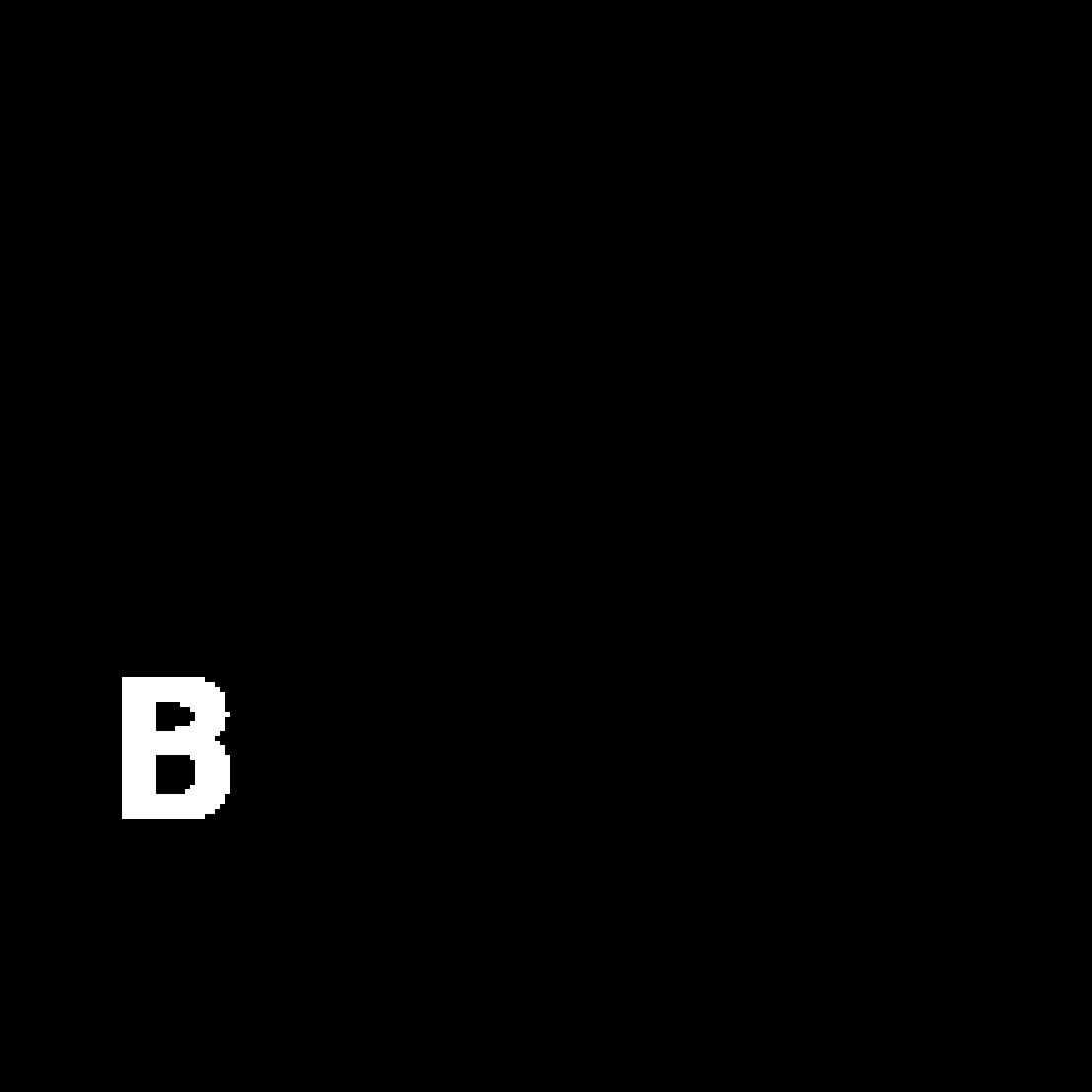}
\includegraphics[width=0.12\textwidth]{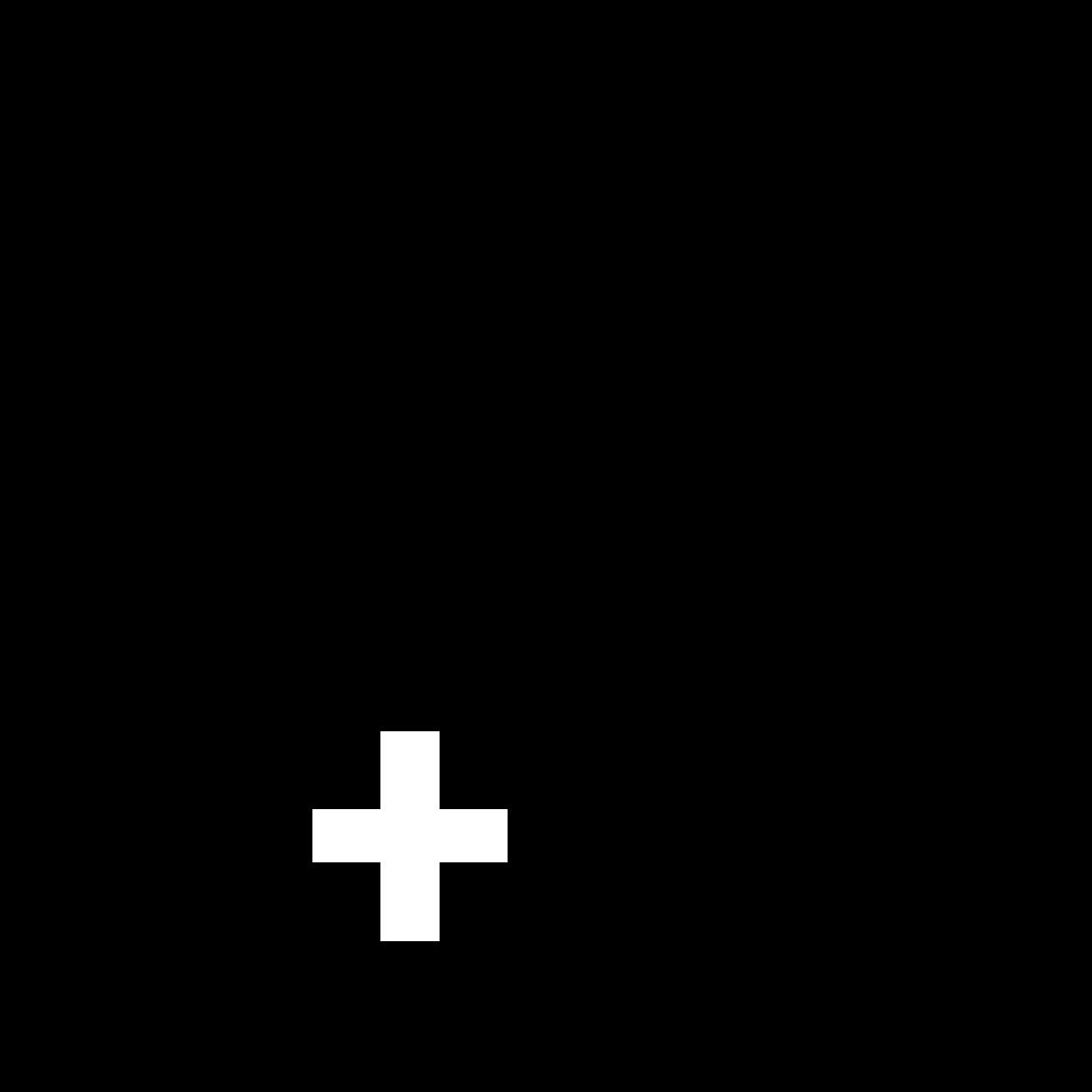}
\includegraphics[width=0.12\textwidth]{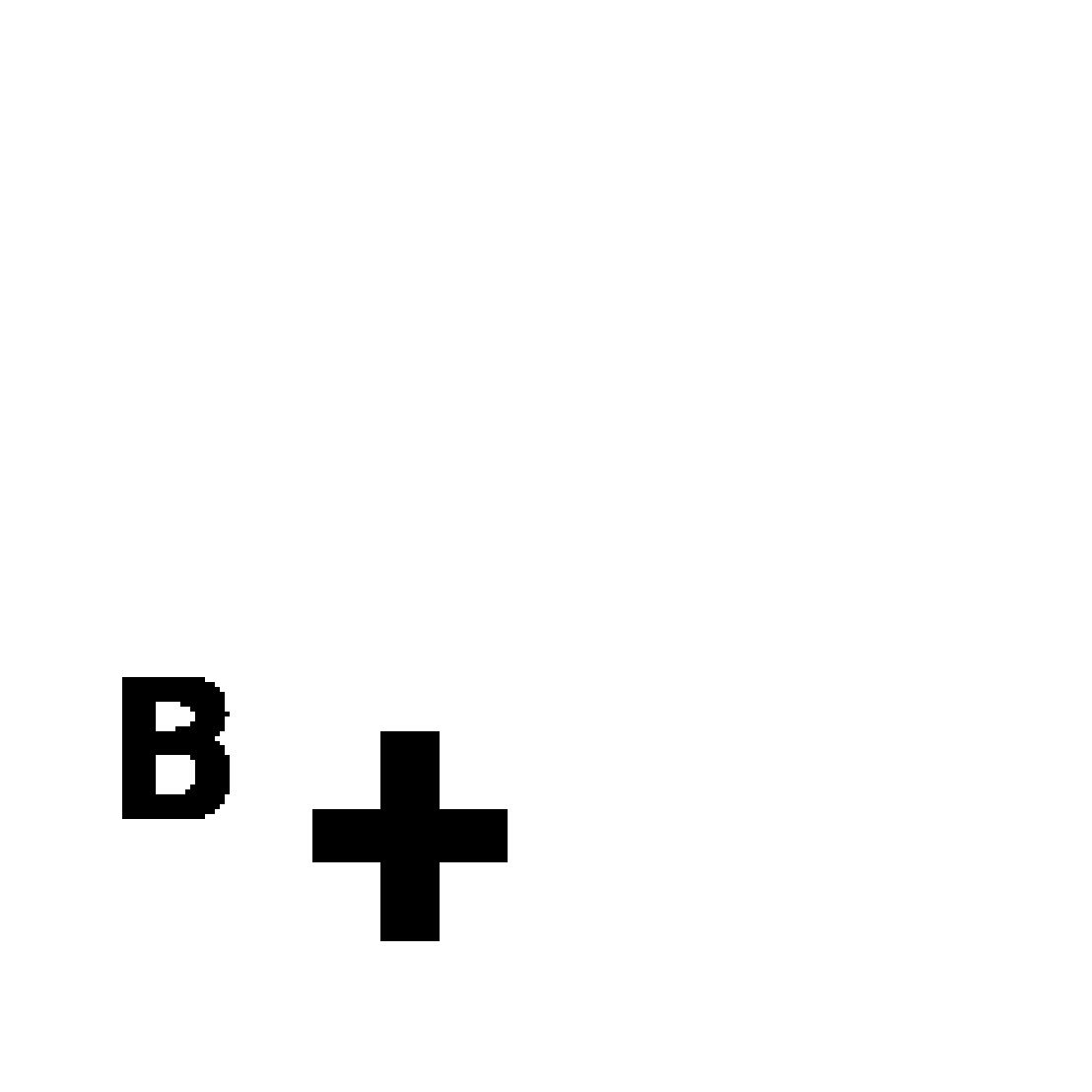}
}
\subfigure[Primitives $p_1$, $p_2$, $p_3$, and $p_4$ of class G.]{
\label{fig:colorGB dataset primitives class G}
\includegraphics[width=0.12\textwidth]{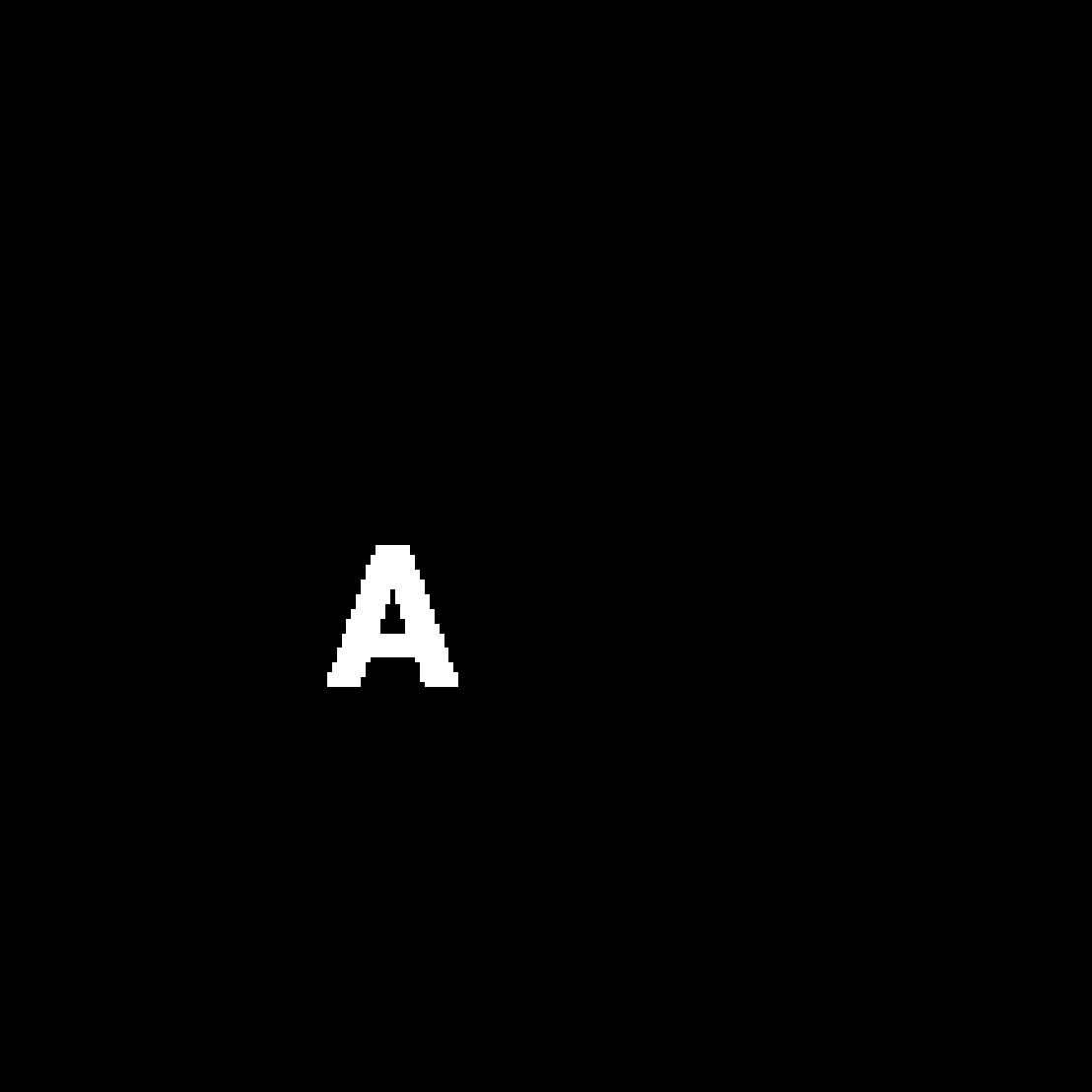}
\includegraphics[width=0.12\textwidth]{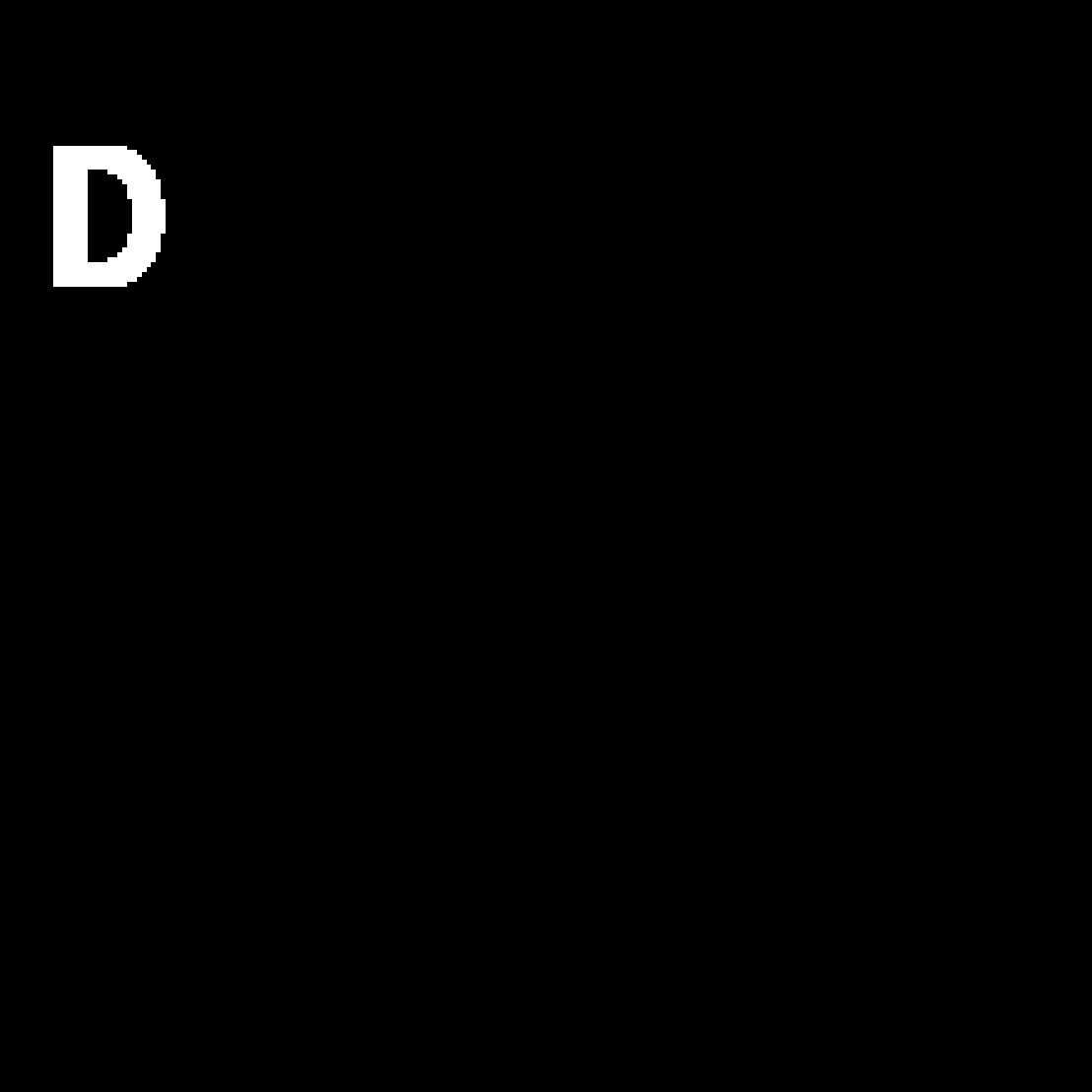}
\includegraphics[width=0.12\textwidth]{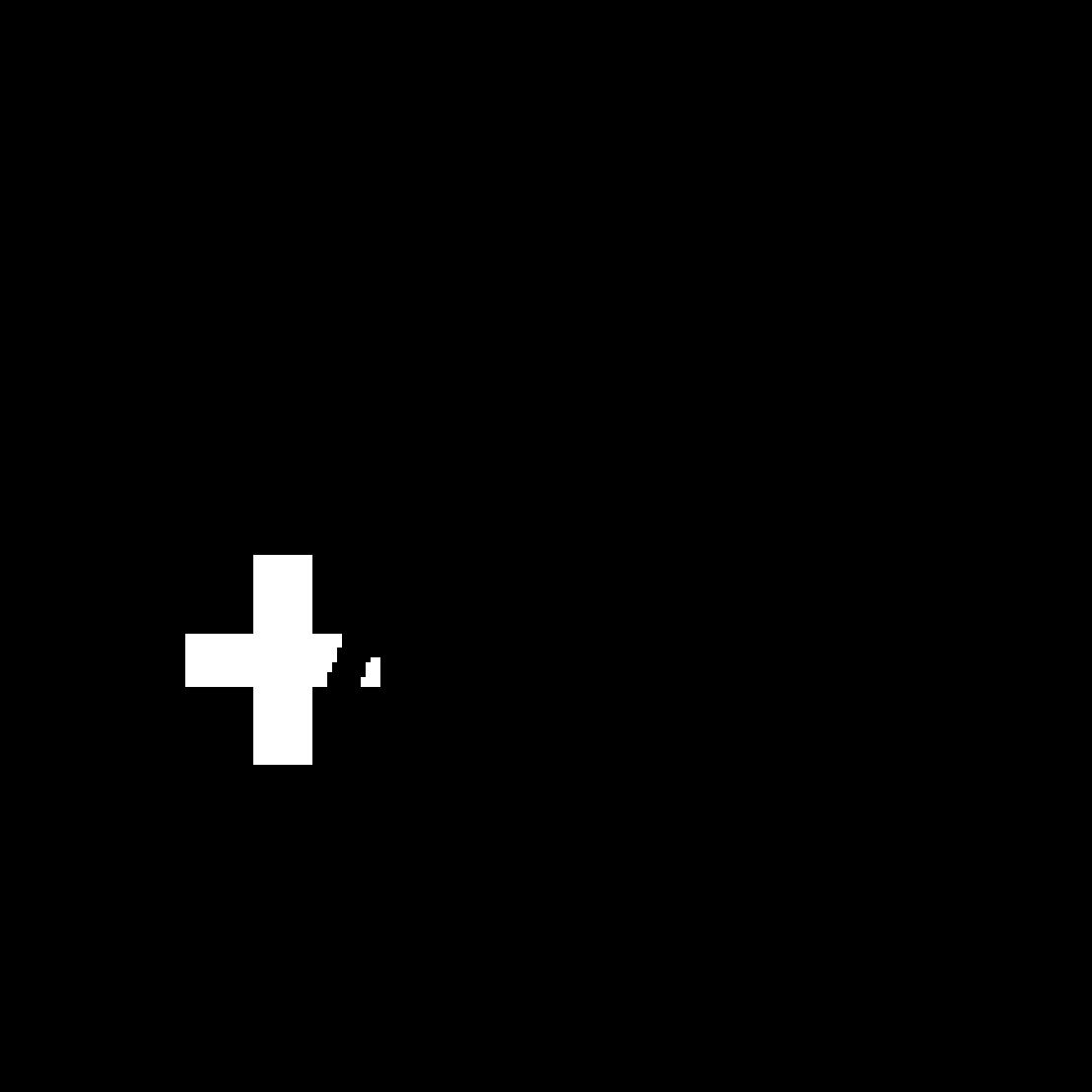}
\includegraphics[width=0.12\textwidth]{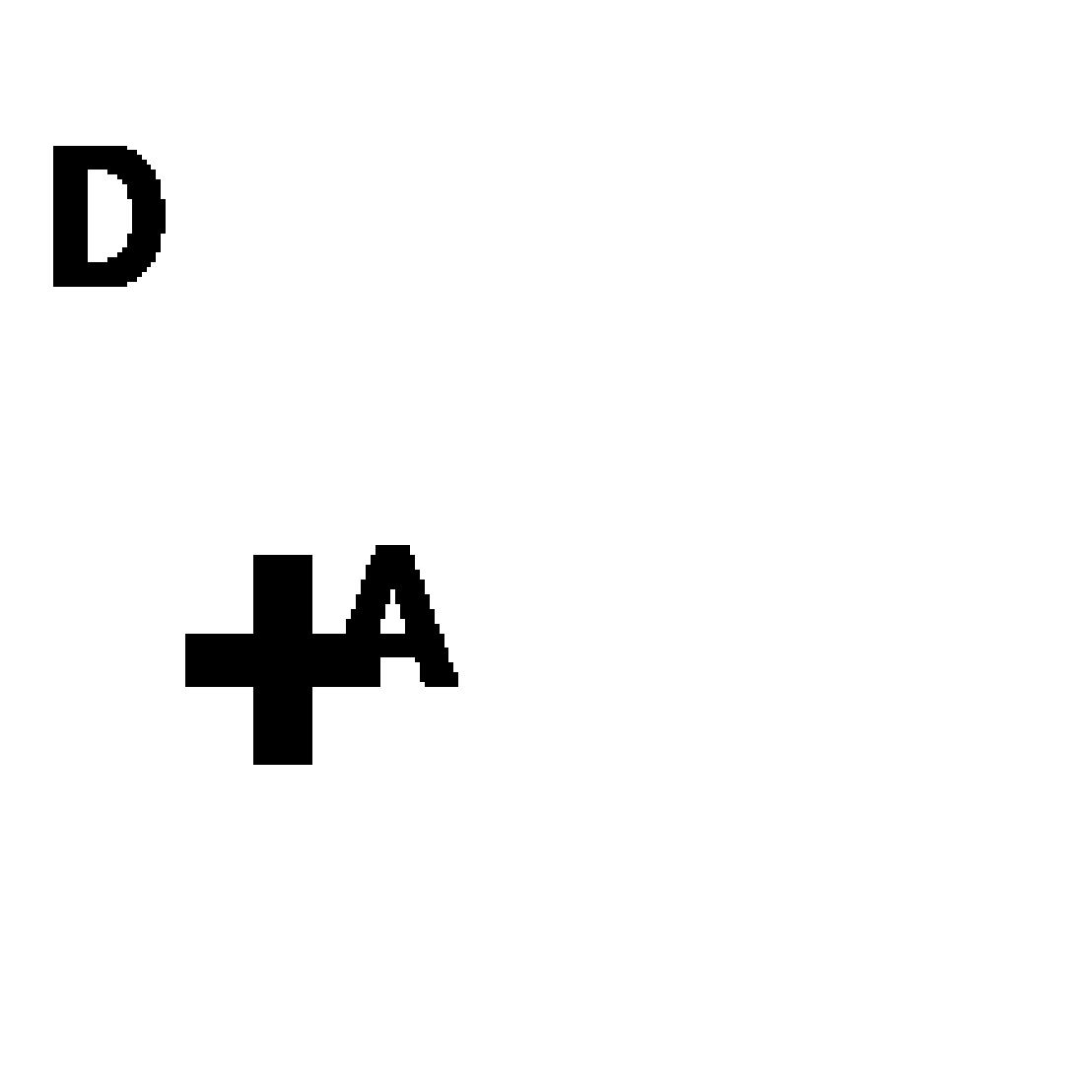}
}
\caption{Dataset \emph{colorGB}, where classes B (Figure \ref{fig:colorGB dataset class B}) and G (Figure \ref{fig:colorGB dataset class G}) are defined by the appearance of blue characters. Figure \ref{fig:colorGB dataset primitives class B} shows the primitives of class B, where $p_1$ is filled blue. Figure \ref{fig:colorGB dataset primitives class G} shows the primitives of class G, where $p_1$ always appears in green.}
\label{fig:colorGB dataset}
\end{figure}

This dataset is forcing a clear color different rather than a form difference between the classes. In theory, a model should converge towards blue and green characters (\emph{A},\emph{B}), possibly forcing a shortcut towards the blue color.

\newpage
\subsubsection{isA dataset}

The \emph{isA} dataset consists in the classification of whether the main primitive of an image is an \emph{A}. Its main primitive is the character \emph{A} ($p_1$) in blue, which appear in all the images of the class isA. The second primitive $p_2$, consists in one letter from B to H, also in blue, happening only in class notA. The third primitive $p_3$ refers to the background filled in gray.

\begin{figure}[h]
\centering
\subfigure[Examples class isA]{
\label{fig:isA dataset class isA}
\includegraphics[width=0.11\textwidth]{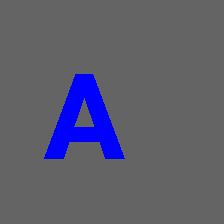}
\includegraphics[width=0.11\textwidth]{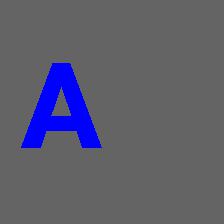}
}
\subfigure[Examples class notA]{
\label{fig:isA dataset class notA}
\includegraphics[width=0.11\textwidth]{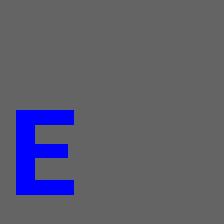}
\includegraphics[width=0.11\textwidth]{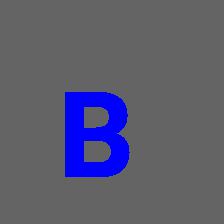}
}
\centering
\subfigure[Primitives $p_1$ and $p_3$ from class isA.]{
\label{fig:isA dataset primitives class isA}
\includegraphics[height=0.12\textwidth]{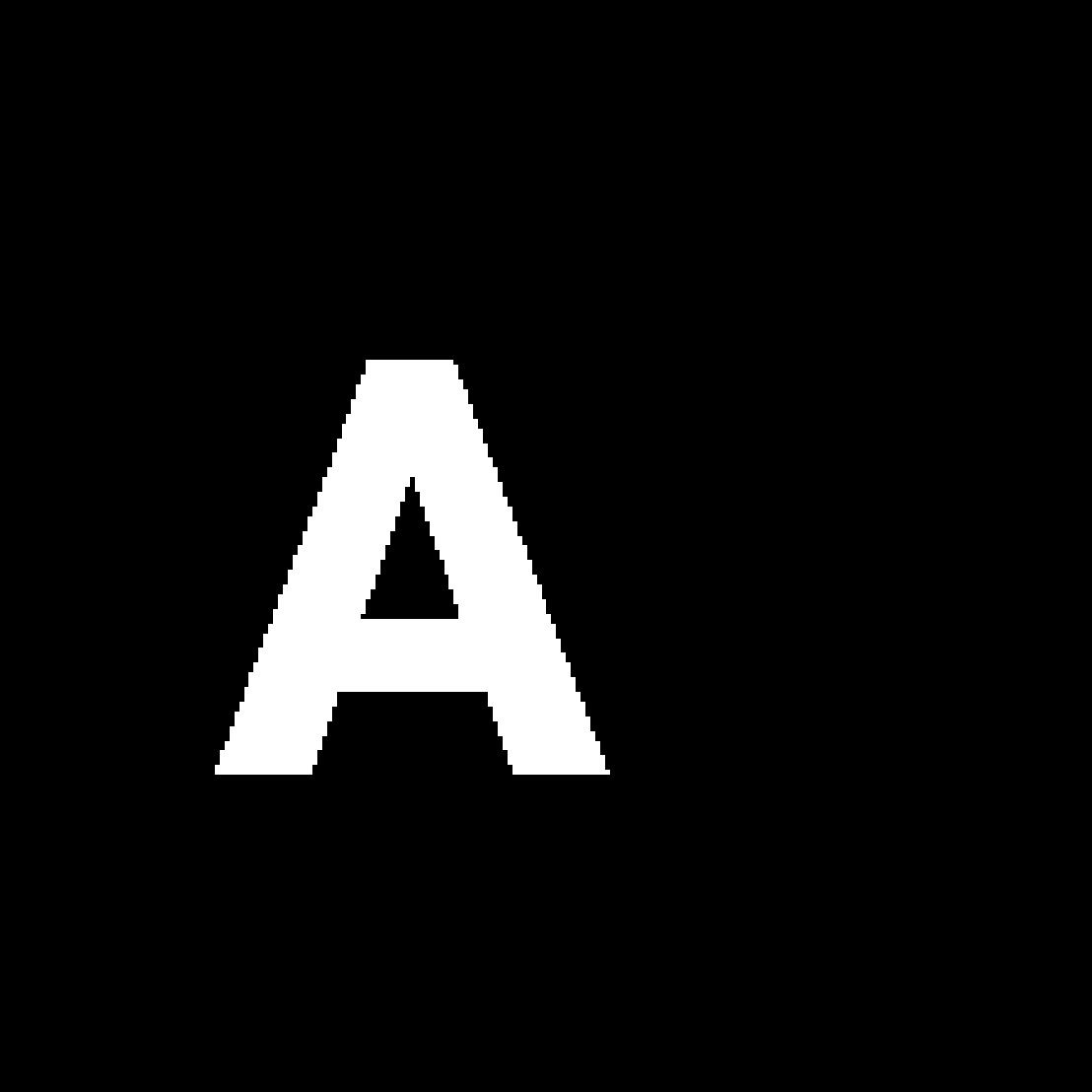}
\includegraphics[height=0.12\textwidth]{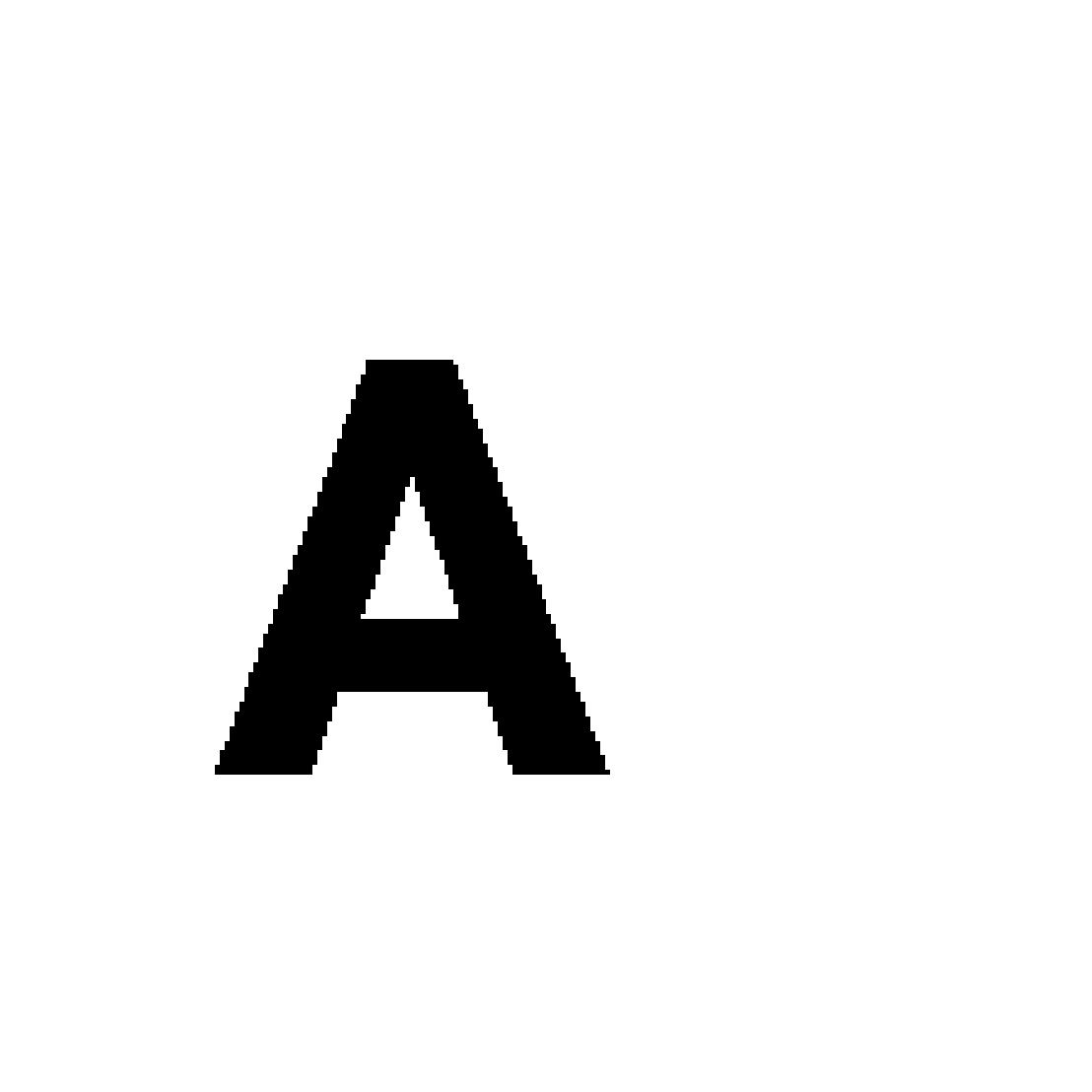}
}
\subfigure[Primitives $p_2$ and $p_3$ from class notA.]{
\label{fig:isA dataset primitives class notA}
\includegraphics[height=0.12\textwidth]{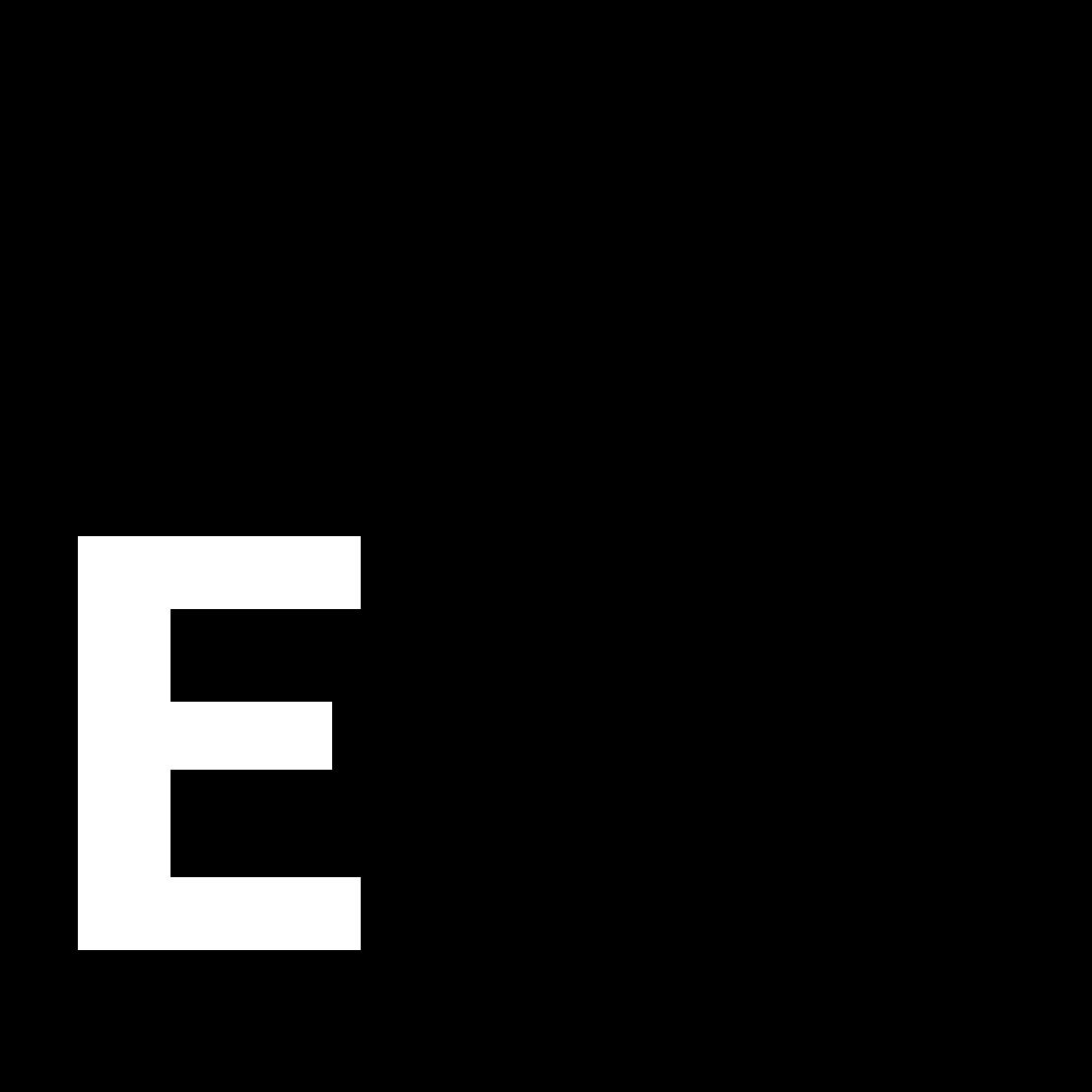}
\includegraphics[height=0.12\textwidth]{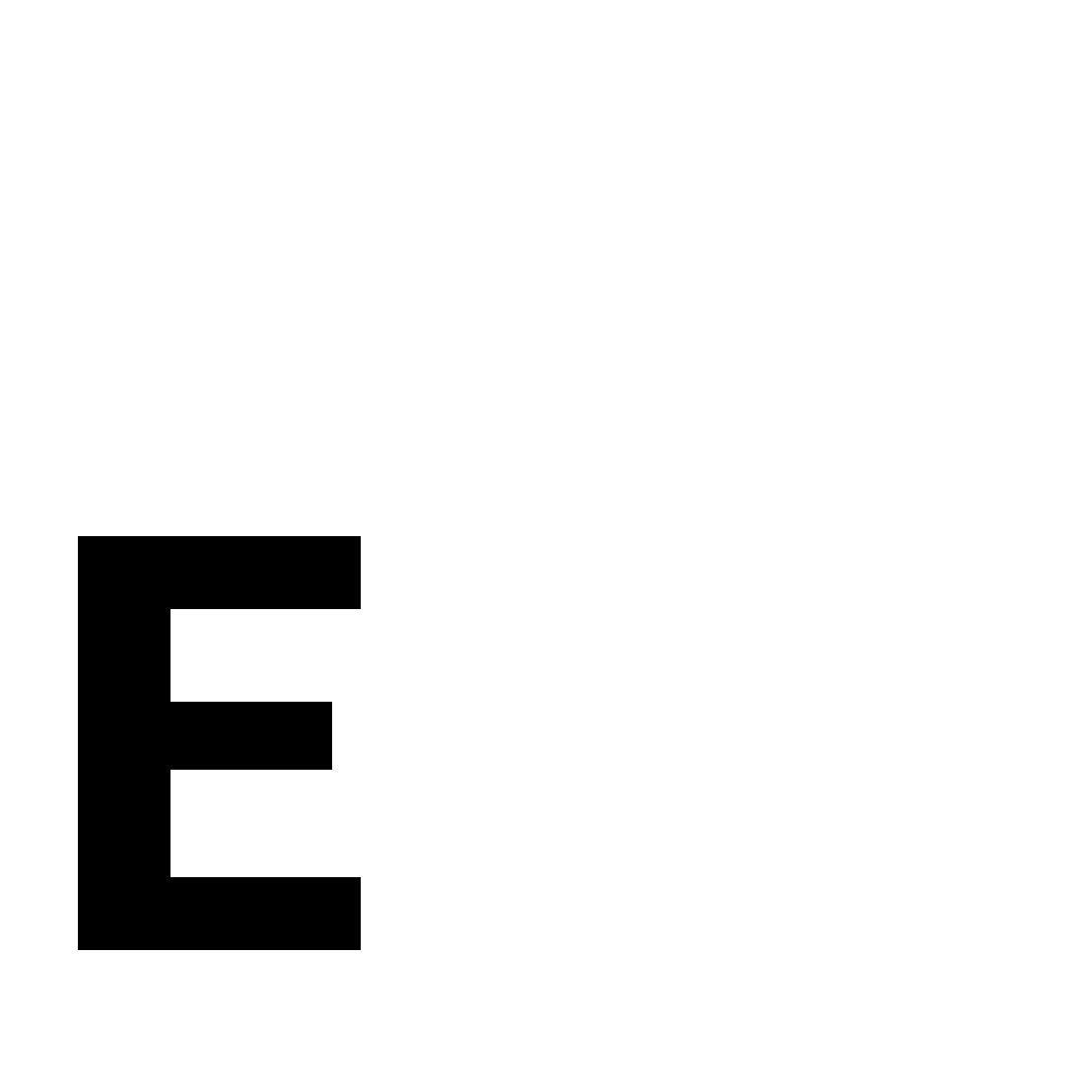}
}\caption{Dataset isA, examples of class isA and class notA are shown in Figures \ref{fig:CO dataset class C} and \ref{fig:CO dataset class O}, respectively. Figure \ref{fig:CO dataset primitives class C} shows the primitives $p_1$ and $p_3$ composing class isA. Figure \ref{fig:CO dataset class O} shows primitives $p_2$ and $p_3$ composing class notA.}
\label{fig:isA dataset}
\end{figure}

The complexity unbalance of this dataset aims to test the performance of CE algorithms in cases where shortcut learning is to be expected. Thus, it is to be expected that from the extracted concepts, one or more will be related to a region of the letter \emph{A}. This will then be measured via the spatial association of the concepts with the primitive $p_1$.

\clearpage
\section{Computational cost and bottlenecks}
\label{apd:Computational cost}

Concept extraction algorithms provide global explanations in human understandable terms to improve the interpretability of neural networks.
These algorithms can provide valuable insights, yet, their computational cost is significant. Most specifically, the presented methods, require the evaluation of the analyzed model a significant amount of times, as well as to execute clustering and regression techniques over large amounts of data. The computational cost of executing each one of these algorithms, not only depends on the CE method, but also on the model that is being analyzed. In this section we provide a rough approximation of the computational costs of each algorithm, in terms of their main operations. For each algorithm, we analyze the main operations on their phases of (1) identifying concepts, (2) importance scoring of concepts, and (3) usage to localize concepts, as of the authors implementation.

\subsection{ECLAD}

The first phase of ECLAD can be defined as the (1) Identification of concepts. This phase consists on the computation of LADs and execution of minibatch K-means. First, every batch of $n_i$ images  (from the dataset of $N_e$ images) is evaluated on the CNN, and the activation maps from the set $L$ of $N_l$ layers are extracted. After extracting the activation maps, they are resized and concatenated, to obtain the descriptor $d_{x_i}$ of each image, and latter flattening the descriptors of all images in the batch to form a vector of $h \ w \ n_i$ LADs, where each LAD has $n_{\mathrm{LAD}}$ dimensions (where $n_{\mathrm{LAD}}$ is the sum of units/neurons for all selected layers $L$). Finally, this vector is used to perform a step of the minibatch K-means. The time complexity of this phase is proportional to:

\begin{equation}
\label{eq:ECLAD cost 1}
\begin{split}
O({\mathrm{ECLAD}}_{(1)}) =& \frac{N_e}{n_i} ( n_i O(f_{\mathrm{CNN}}) + n_i N_l O(f_{\mathrm{resize}}) \\
& + O(\mathrm{mbkmean}(h \  w \  n_i, n_{\mathrm{LAD}}, k)))
\end{split}
\end{equation}

Where, $N_e$ denotes the size of the dataset; $n_i$ is the number of images composing each batch; $N_l$ is the number of layers on the set $L$; $h$ and $w$ denote the height and width of each input image; $n_{\mathrm{LAD}}$ is the sum of units in all layers of $L$; $O(f_{\mathrm{CNN}})$ represents the complexity of evaluating an image on the selected CNN; $O(f_{\mathrm{resize}})$ represents the complexity of resizing an activation map to size $(h, w)$; and $O(\mathrm{mbkmean}(h \ w \ n_i, n_{\mathrm{LAD}}, k))$ represents the complexity of executing one step of minibatch K-means, for $h \ w \ n_i$ points of $n_{\mathrm{LAD}}$ dimensions, to compute $k$ clusters.

Once the concepts have been identified, the phase (2) importance scoring of concepts starts. This phase requires the computation of $s_{x_i}^{k}$ for each image and each class, before aggregating the results to obtain $\mathrm{CS}_{c_j}^k$ and $\mathrm{RI}_{c_j}$. For each computation of $s_{x_i}^{k}$, the computation of $d_{x_i}$ and $g_{x_i}$ are required. Similarly, for the correct aggregation of $\mathrm{CS}_{c_j}^k$, each LAD must be associated to the centroids extracted through minibatch K-means. The time complexity of this phase is proportional to:

\begin{equation}
\label{eq:ECLAD cost 2}
\begin{split}
O({\mathrm{ECLAD}}_{(2)}) =& N_k N_e ( O(f_{\mathrm{CNN}}) + N_l O(f_{\mathrm{resize}}) \\
& + O(f_{\mathrm{\nabla CNN}}) + 2 N_l O(f_{\mathrm{resize}}) \\
& + O(f_{\mathrm{association}}) + O(f_{\mathrm{s}})) + O(f_{\mathrm{aggregation}})
\end{split}
\end{equation}

Where $N_k$ denotes the number of classes of the dataset; $O(f_{\mathrm{\nabla CNN}})$ represents the complexity of computing the gradient of the CNN; $O(f_{\mathrm{association}})$ represents the complexity of associating the LADs of an image to the centroids of the extracted concepts; $O(f_{\mathrm{s}})$ is the complexity of computing $s_{x_i}^{k}$; and $O(f_{\mathrm{aggregation}})$ represents the complexity of computing $\mathrm{CS}_{c_j}^k$ and $\mathrm{RI}_{c_j}$, based on the set of all $s_{x_i}^{k}$ and their associated concepts.

The rough cost of executing ECLAD is:

\begin{equation}
\label{eq:ECLAD cost all}
\begin{split}
O(\mathrm{ECLAD}) =& \\
& (1+N_k) N_e O(f_{\mathrm{CNN}}) \\
& + N_k N_e O(f_{\mathrm{\nabla CNN}}) \\
& + \frac{N_e}{n_i} O(\mathrm{mbkmean}(h \  w \  n_i, n_{\mathrm{LAD}}, k)) \\
& + (1 + 2 N_k) Ne N_l O(f_{\mathrm{resize}}) \\
& + N_k N_e O(f_{\mathrm{association}}) + N_k N_e O(f_{\mathrm{s}}) + O(f_{\mathrm{aggregation}})
\end{split}
\end{equation}

To evaluate new images, ECLAD's (3) usage to localize concepts, consists of three steps. First, the image is evaluated on the analyzed CNN, and the activation maps of the set of layers $L$ are extracted. Then, these activation maps are resized and aggregated. Finally, each LAD of the resulting descriptor $d_{x_i}$ is associated to the centroids of each concept. The time complexity of this phase is proportional to:

\begin{equation}
\label{eq:ECLAD cost 3}
\begin{split}
O({\mathrm{ECLAD}}_{(3)}) = O(f_{\mathrm{CNN}}) + N_l O(f_{\mathrm{resize}}) + O(f_{\mathrm{aggregation}})
\end{split}
\end{equation}

From these operations, the bottlenecks are $O(f_{\mathrm{CNN}})$, $O(\mathrm{mbkmean}(h \ w \ n_i, n_{\mathrm{LAD}}, k)$, and $O(f_{\mathrm{s}})$. Where $O(f_{\mathrm{CNN}})$ was performed on a GPU, and requires not only the memory to execute the CNN, but also to extract $d_{x_i}$ of dimensions $h \times w \times n_{\mathrm{LAD}}$. $O(\mathrm{mbkmean}(h \ w \ n_i, n_{\mathrm{LAD}}, k)$ was executed on the CPU, and could be speed up by using a GPU implementation of K-means, yet, it would also imply higher GPU requirements. Finally, $O(f_{\mathrm{s}})$ was performed on GPU, and required the multiplication of two matrices $d_{x_i}$ and $g_{x_i}$, which in case of being computed in minibatches, are of dimension $h \ w \ n_i \times n_{\mathrm{LAD}}$ each. This operation can arise practical problems with limited GPU resources.

\subsection{ACE}

For a fair comparison, we will discuss the analysis of a complete dataset, and for simplification, we will assume balanced classes. The (1) identification of concepts using ACE, consists on three steps. First, each image is segmented $n_s$ times using SLIC, to obtain a set of $n_p$ patches. Then, each patch is resized, padded, and evaluated on the CNN, to obtain the activation map of the selected layer. Finally, for each class, the set of $\frac{N_e}{N_k} n_p$ vectors, of $h_l \ w_l \ n_{\mathrm{layer}}$ dimensions are used to extract $k$ clusters using K-means (where $n_{\mathrm{layer}}$ denotes the number of dimensions of the selected layer). The time complexity of this phase is proportional to:

\begin{equation}
\label{eq:ACE cost 1}
\begin{split}
O({\mathrm{ACE}}_{(1)}) =& N_k ( \frac{N_e}{N_k} (n_s O_{\mathrm{SLIC}} + n_p O(f_{\mathrm{CNN}})) \\
& \quad + O_{kmean}(\frac{N_e}{N_k}, h_l \ w_l \ n_{\mathrm{layer}}, k))
\end{split}
\end{equation}

Where $N_e$ is the size of the dataset; $N_k$ is the number of classes of the dataset; $k$ is the number of concepts to extract; $n_s$ denotes the number of SLIC segmentations to perform (e.g., the ACE method segments an image to obtain [15,50,80] patches, in this case $n_s=3$); $n_p$ denotes the total number of patches extracted after filtering the SLIC segments (e.g., for the default parameters of ACE, $15+50+80>n_p>15$); $h_l$ and $w_l$ denote the height and width of the activation maps of the selected layer; $n_{\mathrm{layer}}$ denotes the number of units of the selected layer; $O_{\mathrm{SLIC}}$ denotes the complexity of executing the SLIC segmentation algorithm; $O(f_{\mathrm{CNN}})$ represents the complexity of evaluating an image on the selected CNN; $O_{kmean}(\frac{N_e}{N_k}, h_l \ w_l \ n_{\mathrm{layer}}, k))$ represent the complexity of executing the K-means algorithm for $\frac{N_e}{N_k}$ datapoints, of $h_l \ w_l \ n_{\mathrm{layer}}$ dimensions, and $k$ clusters.

After the extraction of concepts, the (2) importance scoring of concepts is performed using TCAV, for each concept of each class. The process of obtaining the CAV and TCAV score of a concept consists on four steps. First, a random set of images are sampled from the dataset to serve as a random concept. Similarly, a subset of $n_b$ images from the concept and random concept are sampled. Second, both subsets of images are evaluated on the CNN, to obtain the flattened activation maps of the selected TCAV layer. Third, a linear classifier is trained to differentiate both subsets of vectors (we obtain a CAV from this classifier). This linear classification is performed over a dataset of $2 n_b$ vectors of $h_l \ w_l \ n_{\mathrm{layer}}$ dimensions. Fourth, we compute the TCAV score for said CAV, by evaluating every image of the associated class in the model, obtaining the directional derivative on the corresponding layer and counting how many of these directional derivatives point on the same direction as the CAV, this proportion is the TCAV score. This process is repeated $n_{\mathrm{tcav}}$ times for each concept and an associated concept of random images (serving as a random concept for control). The resulting sets of TCAV scores and CAVs of the concept and random concept are then compared using a t-test, to obtain a p-value stating the statistical significance of the concept. The time complexity of this phase is proportional to:

\begin{equation}
\label{eq:ACE cost 2}
\begin{split}
O({\mathrm{ACE}}_{(2)}) =& N_k K ( \\
& \quad n_{\mathrm{tcav}} (O(f_{\mathrm{sample}}) \\
& \quad\quad + 2 n_b O(f_{\mathrm{CNN}}) \\
& \quad\quad + O(f_{\mathrm{lin-class}}) \\
& \quad\quad + \frac{N_e}{N_k} (O(f_{\mathrm{CNN}}) \\
& \quad\quad\quad + O(f_{\mathrm{\nabla CNN}}) \\
& \quad\quad\quad + O(f_{\mathrm{proj}})) \\
& ) + O(f_{\mathrm{t-test}}))
\end{split}
\end{equation}

Where $N_e$ is the size of the dataset; $N_k$ is the number of classes of the dataset; $n_{\mathrm{tcav}}$ refers to the number of subsample computations to TCAV scores for each concept, to later compute the statistical significance of the concept; $n_b$ is the size of each image subset to compute each TCAV score; $O(f_{\mathrm{sample}})$ represents the complexity of sampling $n_b$ images from a concept and random concept; $O(f_{\mathrm{CNN}})$ represents the complexity of executing the CNN and extracting the selected activation map; $O(f_{\mathrm{\nabla CNN}})$ represents the complexity of computing the gradient of the CNN with respect to the selected layer; $O(f_{\mathrm{lin-class}})$ represents the complexity of fitting a linear a stochastic gradient descent classifier to $2 n_b$ vectors of $h_l \ w_l \ n_{\mathrm{layer}}$ dimensions; $O(f_{\mathrm{proj}})$ represents the complexity of projecting the gradient of the network towards a CAV; 

The rough cost of executing ACE is:

\begin{equation}
\label{eq:ACE cost all}
\begin{split}
O(\mathrm{ACE}) = &     
N_e n_s O_{\mathrm{SLIC}} \\
& + ( N_e n_p + 2 N_k K n_{\mathrm{tcav}} n_b + K n_{\mathrm{tcav}} N_e) O(f_{\mathrm{CNN}}) \\ 
& + K n_{\mathrm{tcav}} N_e O(f_{\mathrm{\nabla CNN}}) \\
& + N_k O_{kmean}(\frac{N_e}{N_k}, h_l \  w_l \  n_{\mathrm{layer}}, k) \\
& + N_k K n_{\mathrm{tcav}} O(f_{\mathrm{lin-class}}) \\
& + N_k K n_{\mathrm{tcav}} O(f_{\mathrm{sample}}) \\
& + K n_{\mathrm{tcav}} N_e O(f_{\mathrm{proj}}) \\
& + N_k K O(f_{\mathrm{t-test}})
\end{split}
\end{equation}

To improve the efficiency of ACE, we evaluated the extracted patches, random images, and the images of every class a single time, and sampled the tensors when performing the TCAV computations. This implementation detail significantly reduced the time complexity to:

\begin{equation}
\label{eq:ACE cost all efficient}
\begin{split}
O(\mathrm{ACE}) =& N_e n_s O_{\mathrm{SLIC}} \\
& + ( N_e n_p + 2 N_k K n_b + N_e) O(f_{\mathrm{CNN}}) \\ 
& + N_e O(f_{\mathrm{\nabla CNN}}) \\
& + N_k O_{kmean}(\frac{N_e}{N_k}, h_l \  w_l \  n_{\mathrm{layer}}, k) \\
& + N_k K n_{\mathrm{tcav}} O(f_{\mathrm{lin-class}}) \\
& + N_k K n_{\mathrm{tcav}} O(f_{\mathrm{sample}}) \\
& + K n_{\mathrm{tcav}} N_e O(f_{\mathrm{proj}}) \\
& + N_k K O(f_{\mathrm{t-test}})
\end{split}
\end{equation}

To evaluate new images, ACE's (3) usage to localize concepts, consists of three steps. First, the image is segmented using SLIC $n_s$ times. Second, each patch is evaluated on the analyzed CNN, and the activation maps of the selected layer is extracted. Then, these flattened activation maps are compared with each CAV of the class concepts. Finally, the masks of each path are aggregated to obtain the localization result. The time complexity of this phase is proportional to:

\begin{equation}
\label{eq:ACE cost 3}
\begin{split}
O({\mathrm{ACE}}_{(3)}) =& n_s O_{\mathrm{SLIC}} \\
& + n_p O(f_{\mathrm{CNN}}) \\
& + n_p O(f_{comparison}) \\
& + O(f_{\mathrm{aggregation}})
\end{split}
\end{equation}

From these operations, $O(f_{\mathrm{CNN}})$, $O_{\mathrm{SLIC}}$, and $O_{kmean}(\frac{N_e}{N_k}, h_l \ w_l \ n_{\mathrm{layer}}, k)$ were the bottlenecks. $O(f_{\mathrm{CNN}})$ was performed on GPU, and also required the memory for the extraction of the selected layer. Both $O_{\mathrm{SLIC}}$ and $O_{kmean}(\frac{N_e}{N_k}, h_l \  w_l \  n_{\mathrm{layer}}, k)$ were performed on CPU, which significantly increase the time requirements of ACE.

\subsection{ConceptShap}

The phase of (1) identification of concepts with ConceptShap is performed by including an extra set of layers at a defined point of a CNN and training said layer for $n_{et}$ epochs. Each evaluation of the CNN is performed until a selected layer. Then, a linear projection of the resulting activation map is performed towards a lower dimensional space (of $k$ dimensions). Afterwards, an extra pair of layers $g$ are introduced to rescale the obtained tensor and obtain an activation map of the original size. Then, the rest of the CNN is evaluated as originally intended. In this process, the lower dimensional space is introduced as a concept space, and loss is added to it. The identification of concept is then performed by freezing the CNN weights and training the new layer (for $n_{et}$ epochs), to optimize the performance of the model as well as the extra losses introduced on the concept space. The time complexity of this phase is proportional to:

\begin{equation}
\label{eq:ConceptShap cost 1}
\begin{split}
O({\mathrm{ConceptShap}}_{(1)}) =& N_{et} n_{et} (O(f_{\mathrm{CNN}}) + O(f_{\mathrm{\nabla CNN}}) \\
& + O(f_{\mathrm{optim-step-CNN}}))
\end{split}
\end{equation}

Where $N_{et}$ refers to the training subset of the dataset $N_e$; $n_{et}$ refers to the number of epochs for training the added layers; $O(f_{\mathrm{CNN}})$ refers to the complexity of the evaluation of the CNN, including the evaluation of the new layers; $O(f_{\mathrm{\nabla CNN}})$ and $O(f_{\mathrm{optim-step-CNN}})$ refer to the complexity of computing the gradient of the CNN and performing an optimization step over the new layers.

After training the new projections and obtaining a concept space, each component of said space is scored, and their importance is computed based on Shapely values. Said Shapely values are obtained based on a Monte Carlo approximation. For each one of the $N_{\mathrm{MC}}$ samples of this approximation, the contribution is computed as the difference in completeness score between not ablating the concepts, and ablating them. The completeness score requires a complete evaluation of the validation set. In addition, when ablating the concepts, a retraining of the layers $g$ is performed for $n_{ev}$ epochs (to compute the completeness score). The time complexity of this phase is proportional to:

\begin{equation}
\label{eq:ConceptShap cost 2}
\begin{split}
O({\mathrm{ConceptShap}}_{(2)}) =& 
N_{\mathrm{MC}} ( N_{ev} n_{ev} (O(f_{\mathrm{CNN}}) + O(f_{\mathrm{\nabla CNN}}) \\
& \quad + O(f_{\mathrm{optim-step-CNN}})) + 2 N_{ev} O(f_{\mathrm{CNN}}))
\end{split}
\end{equation}

Where $N_{ev}$ refers to the validation subset of the $N_e$ dataset; $n_{ev}$ refers to the number of epochs to retrain the new layers at each computation of the completeness score; $N_{\mathrm{MC}}$ refers to the number of samples to use when computing the Monte Carlo approximate of the shapely values of the concepts; $O(f_{\mathrm{CNN}})$, $O(f_{\mathrm{\nabla CNN}})$,  $O(f_{\mathrm{optim-step-CNN}})$ refer to the computational complexity of evaluating the CNN, computing its gradient and performing an optimization step over the parameters of the new layers, respectively.

The rough cost of executing ConceptSHAP is:
\begin{equation}
\label{eq:ConceptShap cost all}
\begin{split}
O(\mathrm{ConceptShap}) =&  \\
& ( N_{et} n_{et} + N_{\mathrm{MC}} N_{ev} n_{ev} + 2 N_{\mathrm{MC}} N_{ev}) O(f_{\mathrm{CNN}}) \\
& + (N_{et} n_{et} + N_{\mathrm{MC}} N_{ev} n_{ev}) O(f_{\mathrm{\nabla CNN}}) \\
& + (N_{et} n_{et} + N_{\mathrm{MC}} N_{ev} n_{ev}) O(f_{\mathrm{optim-step-CNN}})
\end{split}
\end{equation}

To evaluate new images, ConceptShap's (3) usage to localize concepts, consists of three steps. First, the image is evaluated on the analyzed CNN, including the newly added layers for linear projection and resizing. Then, the activation map of the concept space is extracted, and based on a threshold, each dimension of said tensor is used as the mask of said concept. Finally, the extracted masks are resized to the original size of the image, and can be used to localize each concept. The time complexity of this phase is proportional to:

\begin{equation}
\label{eq:ConceptShap cost 3}
\begin{split}
O({\mathrm{ConceptShap}}_{(3)}) = O(f_{\mathrm{CNN}}) + O(f_{\mathrm{aggregation}}) + O(f_{\mathrm{threshold}})
\end{split}
\end{equation}

From these operations, the Monte Carlo approximation of the shapely values of each concept are resource intensive. Specially, since the computation of the contribution for each sample requires the retraining of the layers $g$. Yet, this operation is performed on GPU, which speeds it up significantly.

\subsection{Comparison}

The nature of the three algorithms differs significantly, and thus, they scale differently to specific parameters and operations.
As an example, ECLAD requires the evaluation of a CNN $(1+N_k) N_e$ times, and $\frac{N_e}{n_i}$ executions of minibatch K-means. ECLAD will perform efficiently (with respect to ACE and ConceptShap) for dataset with few classes (e.g. $N_k<20$). There is a tradeoff between speed and GPU requirements based on the minibatch size $n_i$, where it increases the required GPU memory (by a factor of $n_i \times n_{\mathrm{LAD}} \times \frac{h \ w}{h_l \ w_l}$, with respect to ACE and ConceptShap), yet, it speeds up the computations of $O(f_{\mathrm{CNN}})$, and $O(f_{\mathrm{s}})$.
In contrast ACE requires $(1+n_p) N_e + 2 N_k K n_b$ evaluations of the CNN, $N_e n_s$ executions of SLIC, and $N_k$ executions of K-means. This means that the number of executions of the CNN scales better to the number of classes. To increase the scalability of the method, the K-means of each class can be performed by minibatches (analog to ECLAD). Yet, the computational cost of executing $N_e n_s$ times SLIC, and $N_k K n_{\mathrm{tcav}}$ linear classifiers is significant, which makes it slower than ECLAD and ConceptShap for most cases.
ConceptShap evaluates the analyzed CNN roughly $n_{et} N_e + (N_{\mathrm{MC}}-1) N_{ev} n_{et} + 2 N_{\mathrm{MC}} N_{et}$ (if we consider $N_{et}=N_{ev}$), which means it doesn't require more resources, regardless of the number of classes. In contrast, it scales poorly with the number of concepts $k$, as it directly influences the number of samples $N_{\mathrm{MC}}$ required for the convergence of the Monte Carlo approximation of the shapely values of the concepts.

As a broad summary, ECLAD provides more granular explanations, scaling well to large datasets and number of concepts. It scales linearly for the number of classes $N_k$, which makes it preferable when dealing with a small number of classes (e.g. $N_k<20$).
ConceptShap scales well for a large number of classes, yet it scales poorly for the number of extracted concepts. As a caution, ConceptShap and Shapely values in general have issues when dealing with correlated concepts, which can be problematic when detecting spurious correlations and their importance for a CNN.
Finally, ACE scales better than ECLAD with respect to the number of classes, yet, it has a significant computational cost of executing SLIC and SDG linear classification. In this regard, ACE can be parallelized and executed per class, being a better fit for large datasets with a large number of classes (e.g. $N_k~1000$).
In our settings, ECLAD took the least time to execute, followed by ConceptShap, and finally ACE.

\clearpage
\section{Ablation study}
\label{apd:Ablation study}

In contrast to other concept extraction methods (e.g. ACE, ConceptShap), ECLAD proposes the upscaling and aggregation of of activation maps at different levels of a neural network. On a pixel level, these local aggregated descriptors are denoted as LADs, and are used as a basis for extracting similarly encoded areas through a clustering algorithm. In this ablation study, we explore four key components of our method. First, we explore the need for aggregating information of different layers. Second, we investigate the impact of the number of aggregated layers. Third, we examine the difference of using different numbers of clusters. Finally, we assess the impact of using multiple upscaling methods. 

Each study was performed using two models architectures (ResNet-18 \citep{DBLP:journals/corr/HeZRS15} and DenseNet-121 \citep{DBLP:journals/corr/HuangLW16a}) trained over the \textit{ABplus} and \textit{leather} datasets. From each architecture we selected eight equally distributed layers, named $l_1$ to $l_8$, from which subsets were used on each run. The plots shown below result from executing ECLAD with different sets of parameters over a DenseNet-121, trained on the \textit{ABplus} datasets, which are representative of both architectures and datasets.

\subsection{Aggregating activation maps}
\label{apd:Ablation aggregating activation maps}

\begin{figure*}[h]
\centering
\subfigure[$L=\{l_1\}$]{
\label{fig:scatter ECLAD dense 1L1}
\includegraphics[width=0.41\textwidth]{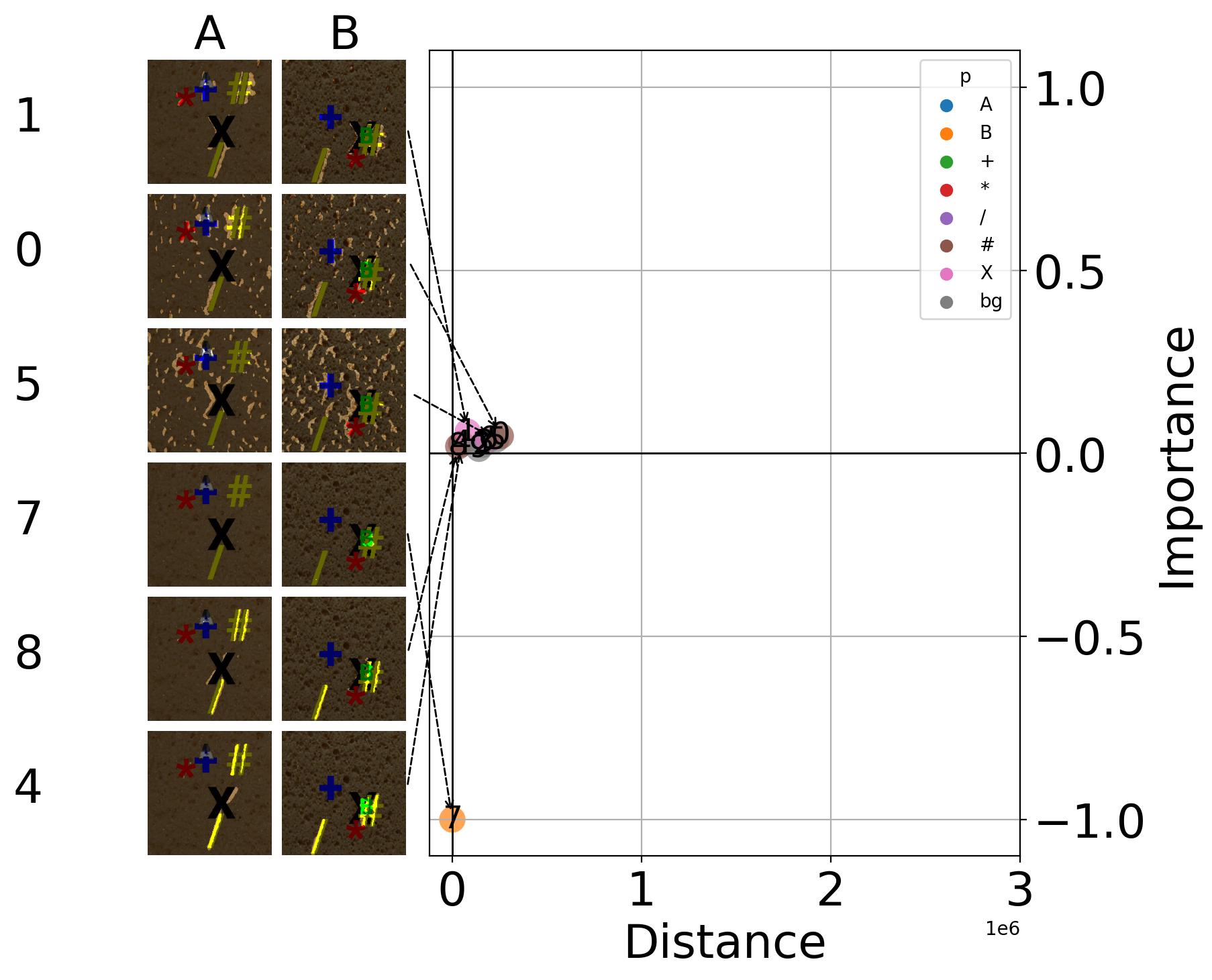}
}
\subfigure[$L=\{l_3\}$]{
\label{fig:scatter ECLAD dense 1L3}
\includegraphics[width=0.41\textwidth]{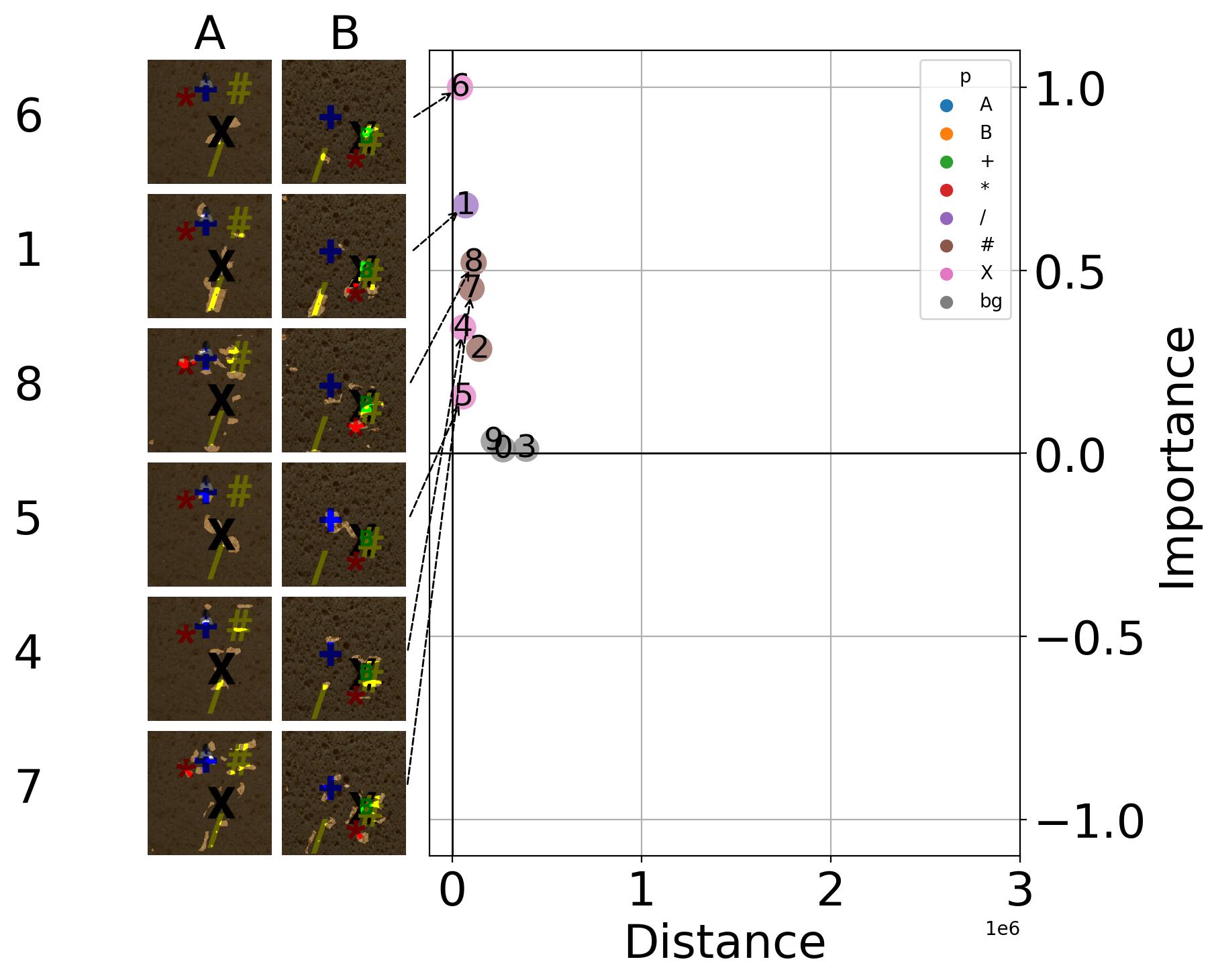}
}

\subfigure[$L=\{l_5\}$]{
\label{fig:scatter ECLAD dense 1L5}
\includegraphics[width=0.41\textwidth]{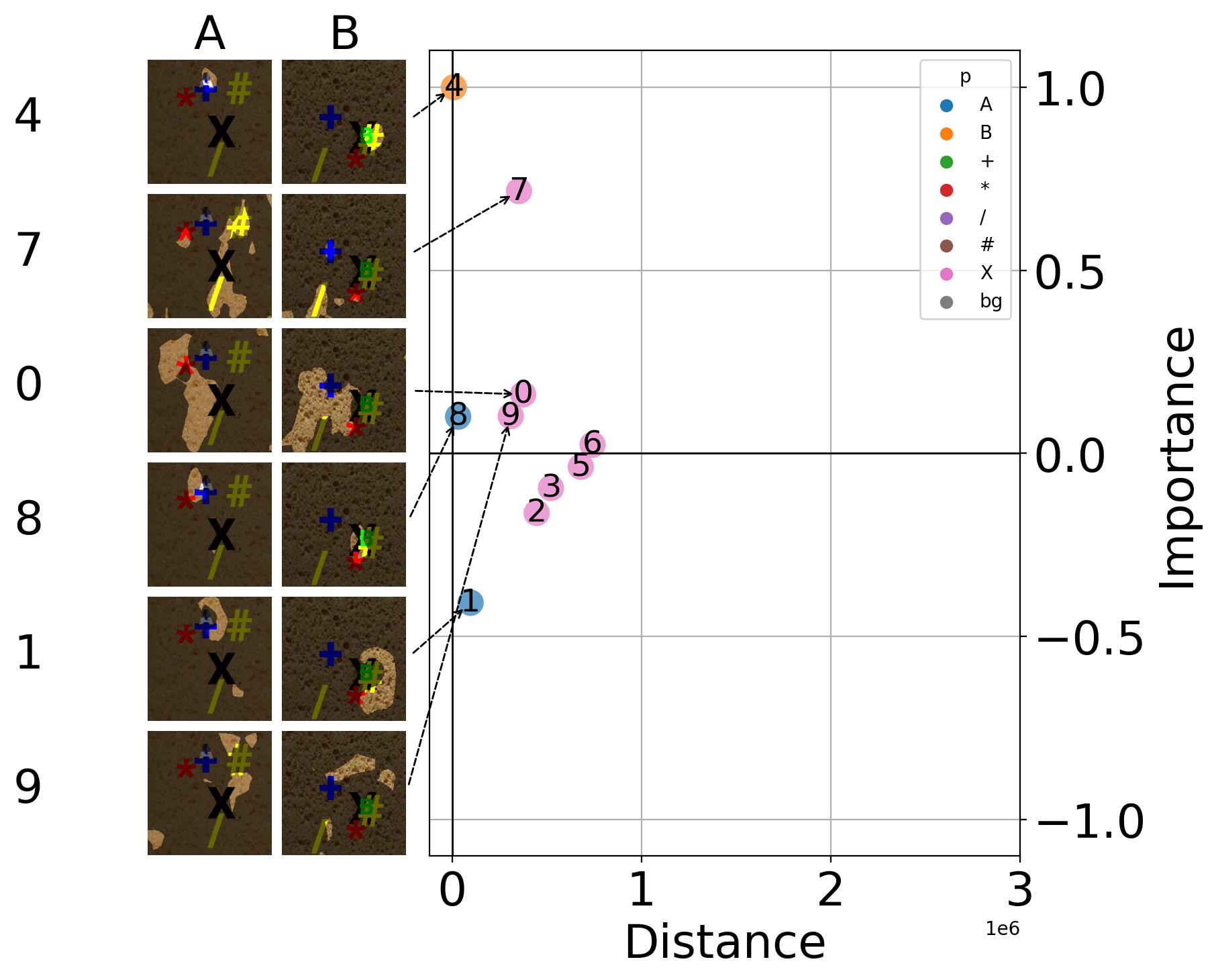}
}
\subfigure[$L=\{l_6\}$]{
\label{fig:scatter ECLAD dense 1L6}
\includegraphics[width=0.41\textwidth]{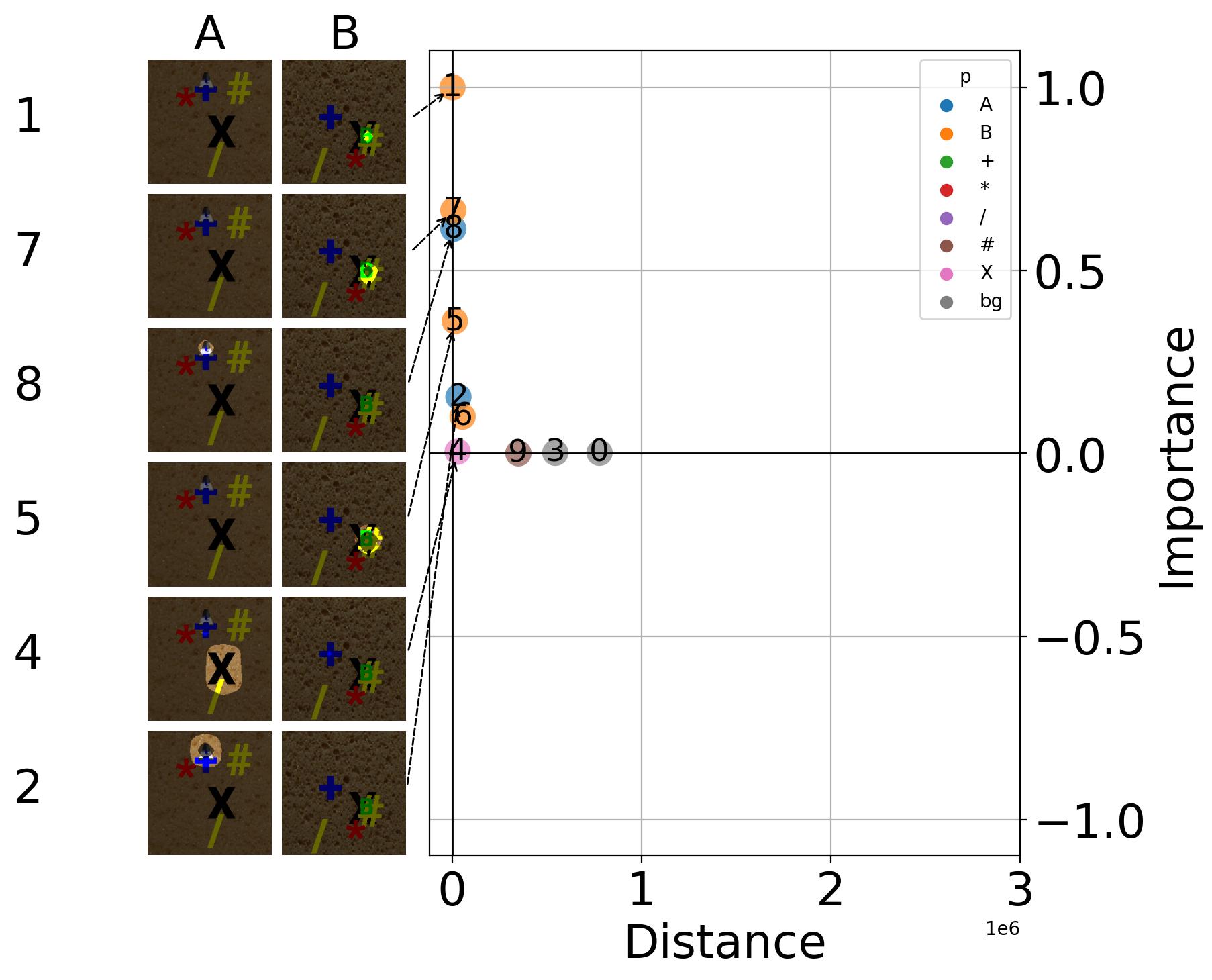}
}

\subfigure[$L=\{l_8\}$]{
\label{fig:scatter ECLAD dense 1L8}
\includegraphics[width=0.41\textwidth]{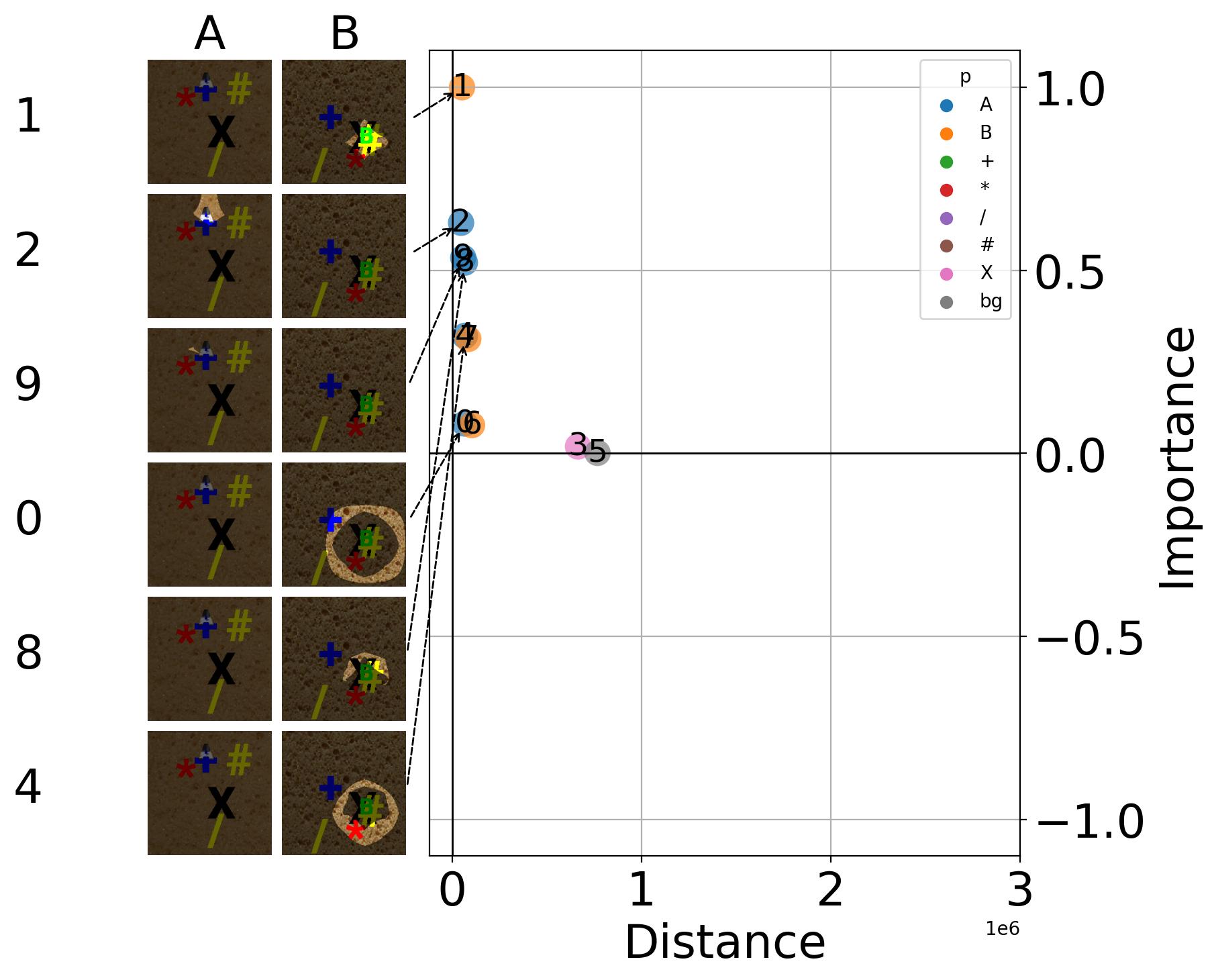}
}
\subfigure[$L=\{l_2, l_4, l_6, l_8\}$]{
\label{fig:scatter ECLAD dense 4L}
\includegraphics[width=0.41\textwidth]{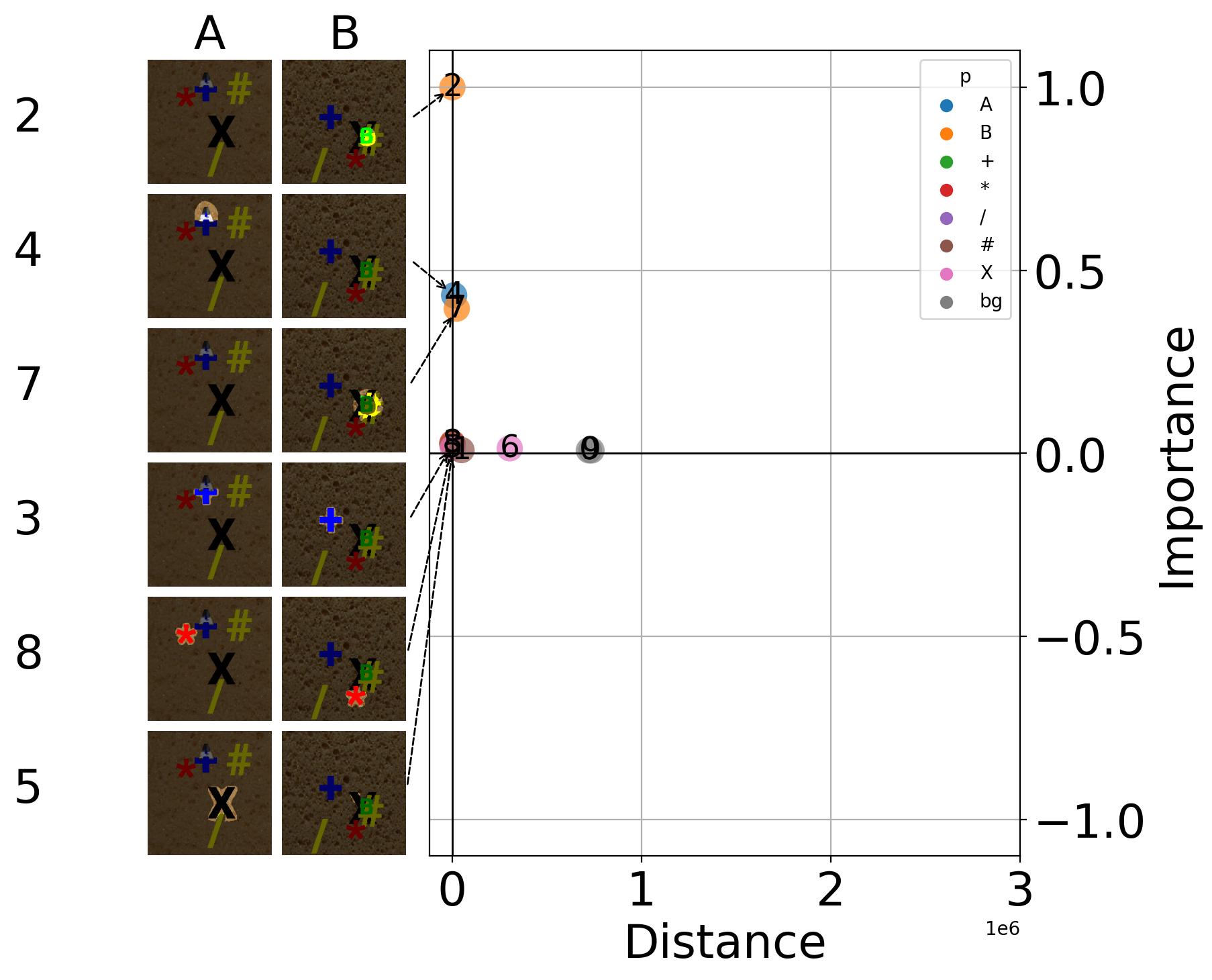}
}

\caption{
Concepts extracted from a DenseNet-121 trained in the ABplus dataset. Subfigures \ref{fig:scatter ECLAD dense 1L1} to \ref{fig:scatter ECLAD dense 1L8} contain the results of single layer executions, and subfigures \ref{fig:scatter ECLAD dense 4L} contains the results of aggregating four layers.
Concepts extracted from low layers lack abstract meaning and are related to texture, edges or color (e.g. $c_0$, $c_1$, $c_7$ in subfigure \ref{fig:scatter ECLAD dense 1L1}), while concepts extracted from higher layers disregard mid level concepts (e.g. + character on the images). The results from aggregating layers (subfigure \ref{fig:scatter ECLAD dense 4L}), contain abstract concepts relevant to the classification task (e.g. $c_2$, $c_4$), without loosing specificity nor disregarding mid level concepts (e.g. $c_8$, $c_5$, $c_3$).
}
\label{fig:scatter results sigle layers}
\end{figure*}

We compare the execution of ECLAD over single layers (across different depths of the network), with the standard execution using four layers equally distributed across the depth of the network. Similar to the figures shown in the result section, we provide scatter plots of association distance and importance of the extracted concepts for each run, as seen in Figure \ref{fig:scatter results sigle layers}.

\textbf{Combining layers from multiple depths allows the extraction of mid and high level concepts without the complexity of fine-tuning the selected layer}.
The results of performing CE over single low level layers generates concepts lacking from abstract meaning such as lateral edges found in $l1$ ($c_4$, $c_5$ in subfigure \ref{fig:scatter ECLAD dense 1L1}), or multiple entangled features such as yellow, green and black edges found in $l3$ ($c_6$, $c_1$ in subfigure \ref{fig:scatter ECLAD dense 1L3}).
When using a single high level layer, the generated concepts disregard mid level features such as the \textbf{*} or \textbf{+} characters in the images which are not found in $l_6$ nor $l_8$.
In addition, the latent representations of the features dilates, generating significant halo effects in $l_6$ and $l_8$ ($c_0$, $c_8$, $c_9$ in subfigure \ref{fig:scatter ECLAD dense 1L8}).
This makes the choice of a single layer, non trivial, as higher level layers miss existing concepts, and low level layers lack abstraction and disentanglement.
In contrast, the aggregation of equally distributed layers extracted high level features characteristics of the task, such as A and B ($c_2$, $c_4$ in subfigure \ref{fig:scatter ECLAD dense 4L}), as well as other existing high level concepts differentiated by the model, such as the \textbf{*} or \textbf{+} characters ($c_8$, $c_3$ in subfigure \ref{fig:scatter ECLAD dense 4L}). By aggregating multiple layers, the resulting concepts are more defined (mitigating the halo effect of higher layers), and the selection of layers for the analysis becomes less critical.

\subsection{Number of aggregated layers}
\label{apd:Ablation number of aggregated layers}

\begin{figure*}[h]
\centering
\subfigure[$L=\{l_1, l_7\}$]{
\label{fig:scatter ECLAD dense number 2L}
\includegraphics[width=0.41\textwidth]{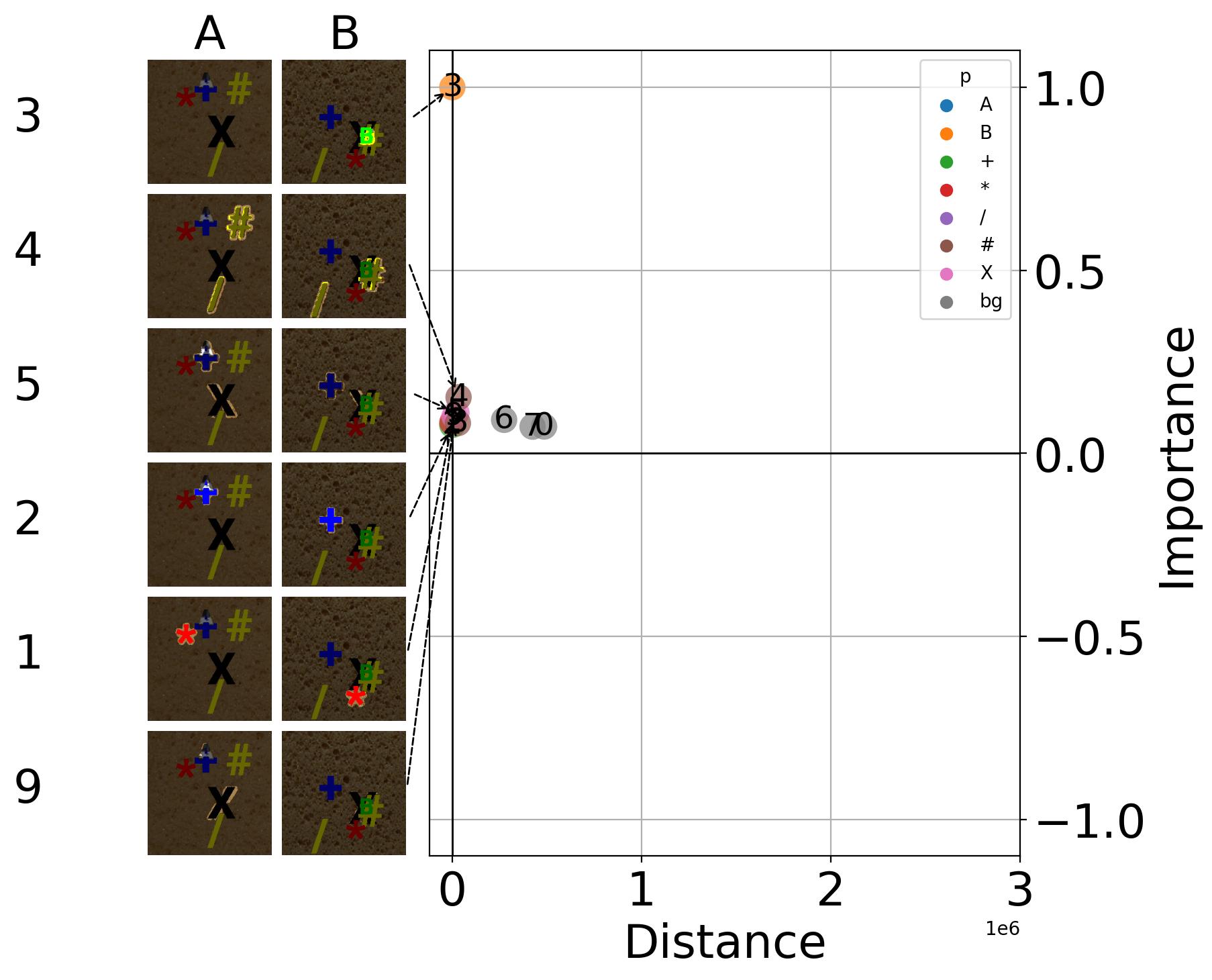}
}
\subfigure[$L=\{l_1, l_4, l_7\}$]{
\label{fig:scatter ECLAD dense number 3L}
\includegraphics[width=0.41\textwidth]{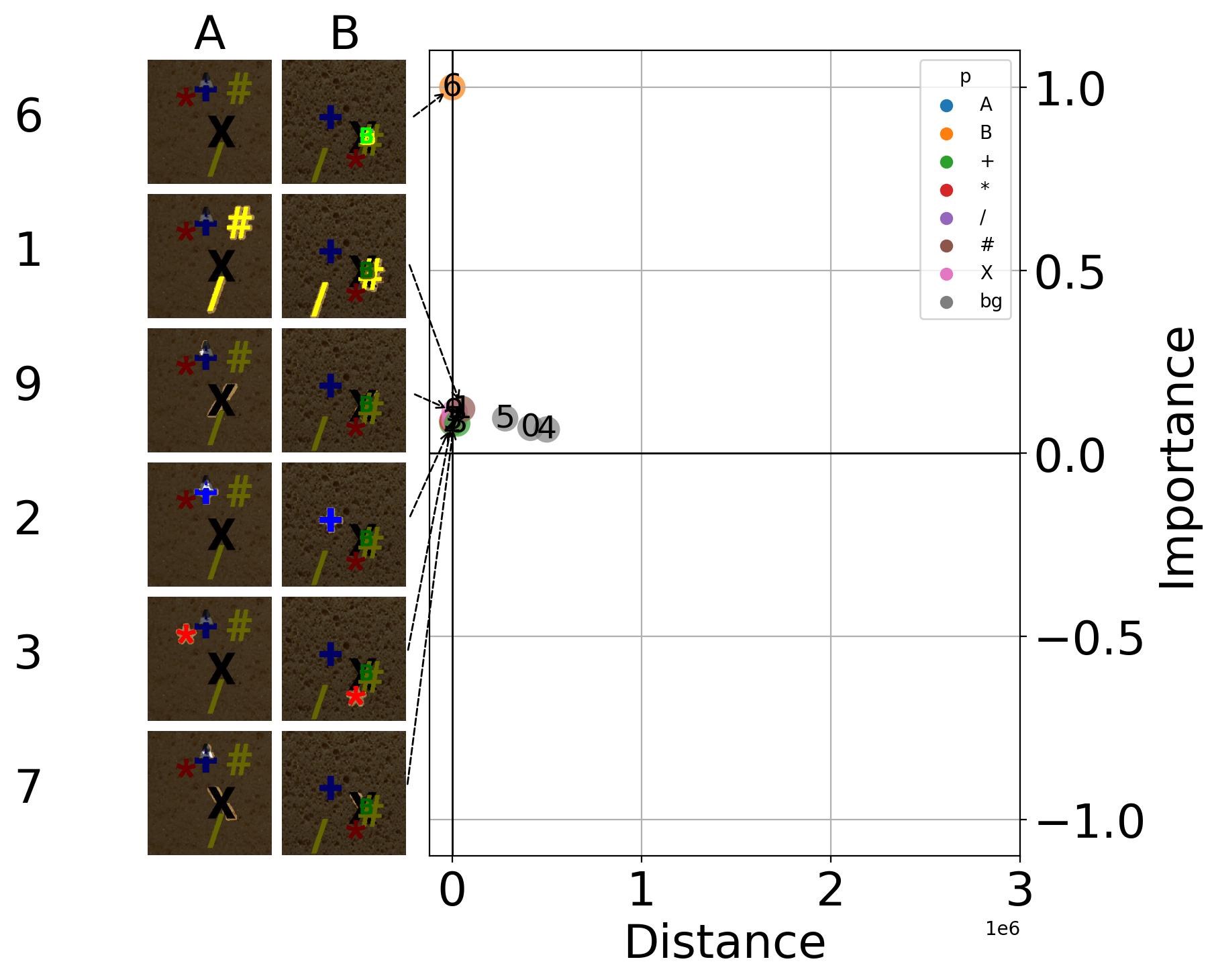}
}

\subfigure[$L=\{l_1, l_3, l_5, l_7\}$]{
\label{fig:scatter ECLAD dense number 4L}
\includegraphics[width=0.41\textwidth]{eclad_ABplus_densenet121_plateau_0_n10_mbKMeans_bilinear_4L}
}
\subfigure[$L=\{l_0, l_1, l_3, l_4, l_6, l_7\}$]{
\label{fig:scatter ECLAD dense number 6L}
\includegraphics[width=0.41\textwidth]{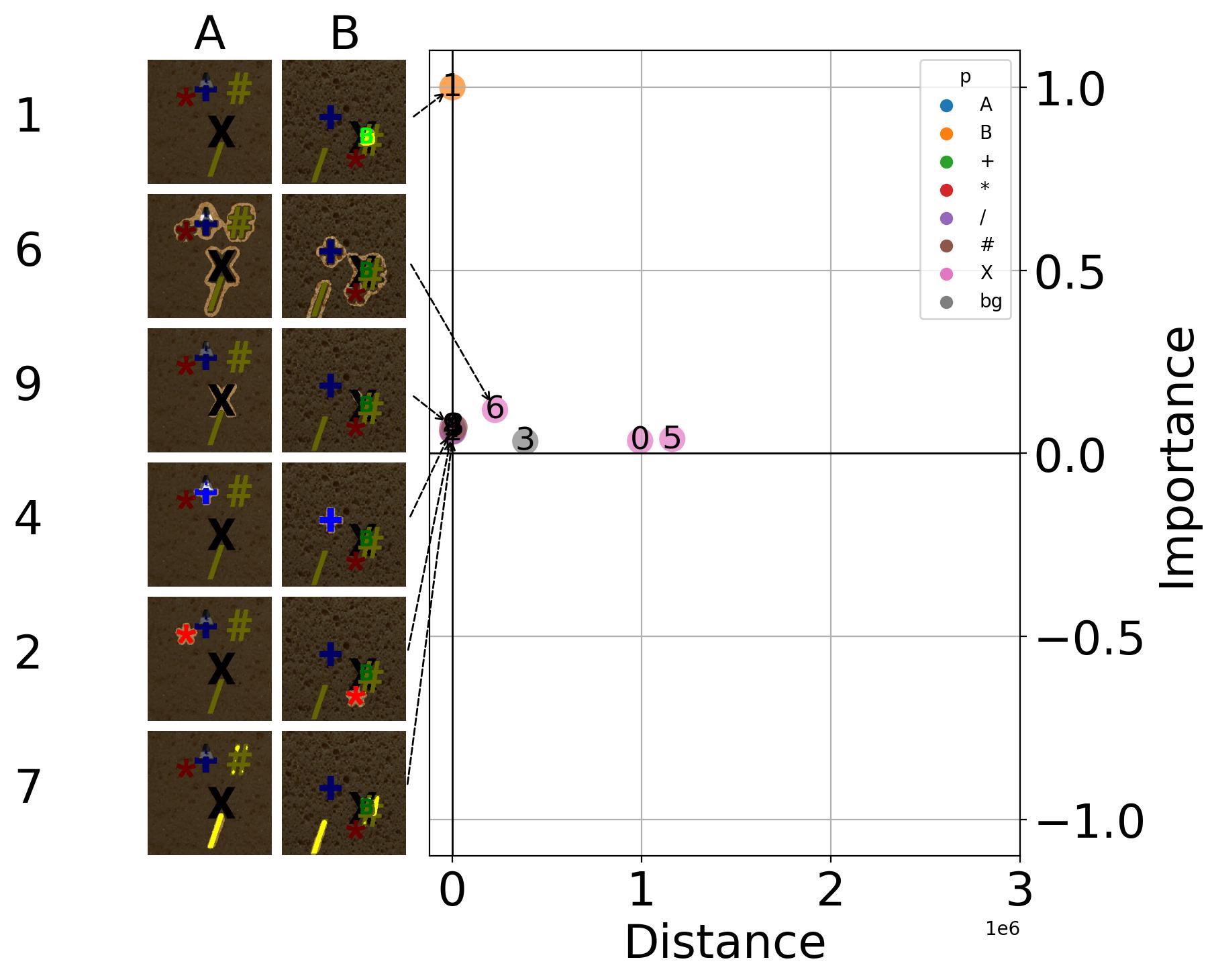}
}

\subfigure[$L=\{l_0, l_1, l_2, l_3, l_4, l_5, l_6, l_7\}$]{
\label{fig:scatter ECLAD dense number 8L}
\includegraphics[width=0.41\textwidth]{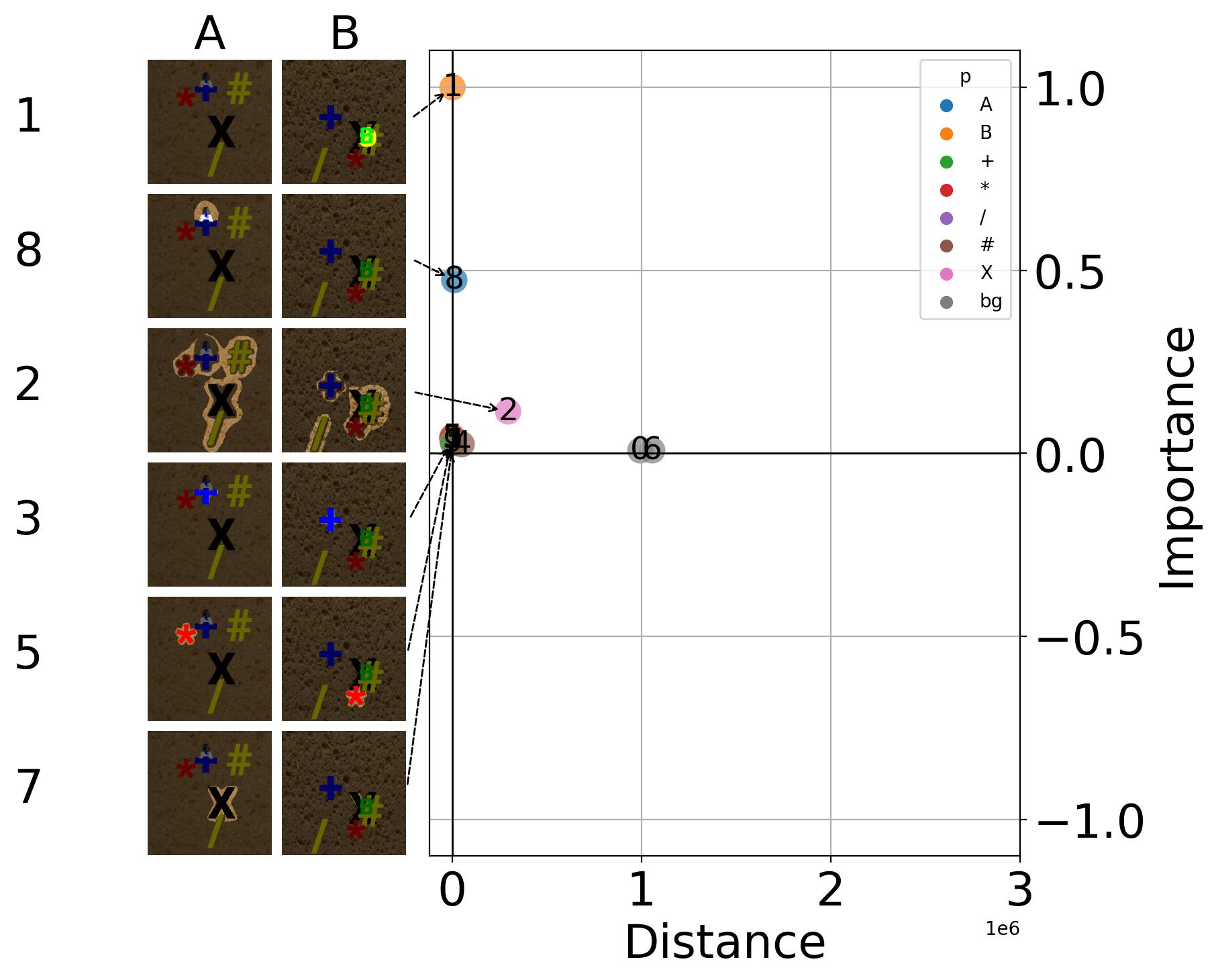}
}
\caption{
Concepts extracted from a DenseNet-121 trained in the ABplus dataset. Subfigures \ref{fig:scatter ECLAD dense number 2L} to \ref{fig:scatter ECLAD dense number 8L} contain results of executing ECLAD with 2 to 8 layers equally distributed through the depth of the model.
For two and three layers, the resulting concepts also include low level features such as edges ($c_9$ in subfigure \ref{fig:scatter ECLAD dense number 2L}, and $c_7$ in subfigure \ref{fig:scatter ECLAD dense number 3L}), and entangled concepts such as the characters \textbf{X} and \textbf{A} ($c_5$ in subfigure \ref{fig:scatter ECLAD dense number 2L}, and $c_9$ in subfigure \ref{fig:scatter ECLAD dense number 3L}).
runs with four to eighth layers provide a better extraction of disentangled concepts such as the characters \textbf{X}, \textbf{*} and \textbf{B} (e.g. $c_2$ and $c_5$ in subfigure \ref{fig:scatter ECLAD dense number 4L}, and $c_1$ and $c_7$ in subfigure \ref{fig:scatter ECLAD dense number 8L}). In addition, the halo effect of important concepts such as the character \textbf{B} progressively diminishes with the number of layers.
}
\label{fig:scatter results number of layers}
\end{figure*}

We compare the execution of ECLAD selecting different number of equally distributed layers (along the depth of the model). As discussed before, using a single layer for concept extraction can be problematic given the dilation of concepts through the CNNs, as well as the disappearance of mid level concepts through the network. In this section we compare the impact of using two or more layers for the concept extraction, the resulting scatter plots are shown in Figure \ref{fig:scatter results number of layers}.

\textbf{Including more layers mitigates the halo effect of important concepts and allows the inclusion of mid level concepts}.
The results of performing CE with two or three layers generates concepts including the most important ones, but also entangled representations of other concepts, such as edges ($c_9$ in subfigure \ref{fig:scatter ECLAD dense number 2L}, and $c_7$ in subfigure \ref{fig:scatter ECLAD dense number 3L}), and entangled concepts such as the characters \textbf{X} and \textbf{A} ($c_5$ in subfigure \ref{fig:scatter ECLAD dense number 2L}, and $c_9$ in subfigure \ref{fig:scatter ECLAD dense number 3L}).
A possible explanation can suggest that the selected layers did not have enough information for clearly separating the different concepts (e.g. \textbf{X} and \textbf{A}), as their representations are differentiated in middle layers.
ECLAD runs with more layers (e.g. 4 and 8) extract disentangled concepts as the possibility of including relevant layers increase. This can be observed in the characters \textbf{X}, \textbf{*} and \textbf{B} (e.g. $c_2$ and $c_5$ in subfigure \ref{fig:scatter ECLAD dense number 4L}, and $c_1$ and $c_7$ in subfigure \ref{fig:scatter ECLAD dense number 8L}).
Similarly, possible issues of halo effect in important concepts diminish with an increasing number of layers, which can be seen for concepts related to the character \textbf{B} with a halo in subfigure \ref{fig:scatter ECLAD dense number 4L}, which disappears with the subsequent inclusion of more layers. It must be mentioned that the computational cost increases with each new layer selected for the analysis, thus, our choice of four layers is a balance between computational cost and good performance.

\subsection{Number of clusters}
\label{apd:Ablation number of clusters}

\begin{figure*}[h]
\centering
\subfigure[$n_\mathrm{c} = 5$]{
\label{fig:scatter ECLAD dense n=5}
\includegraphics[width=0.41\textwidth]{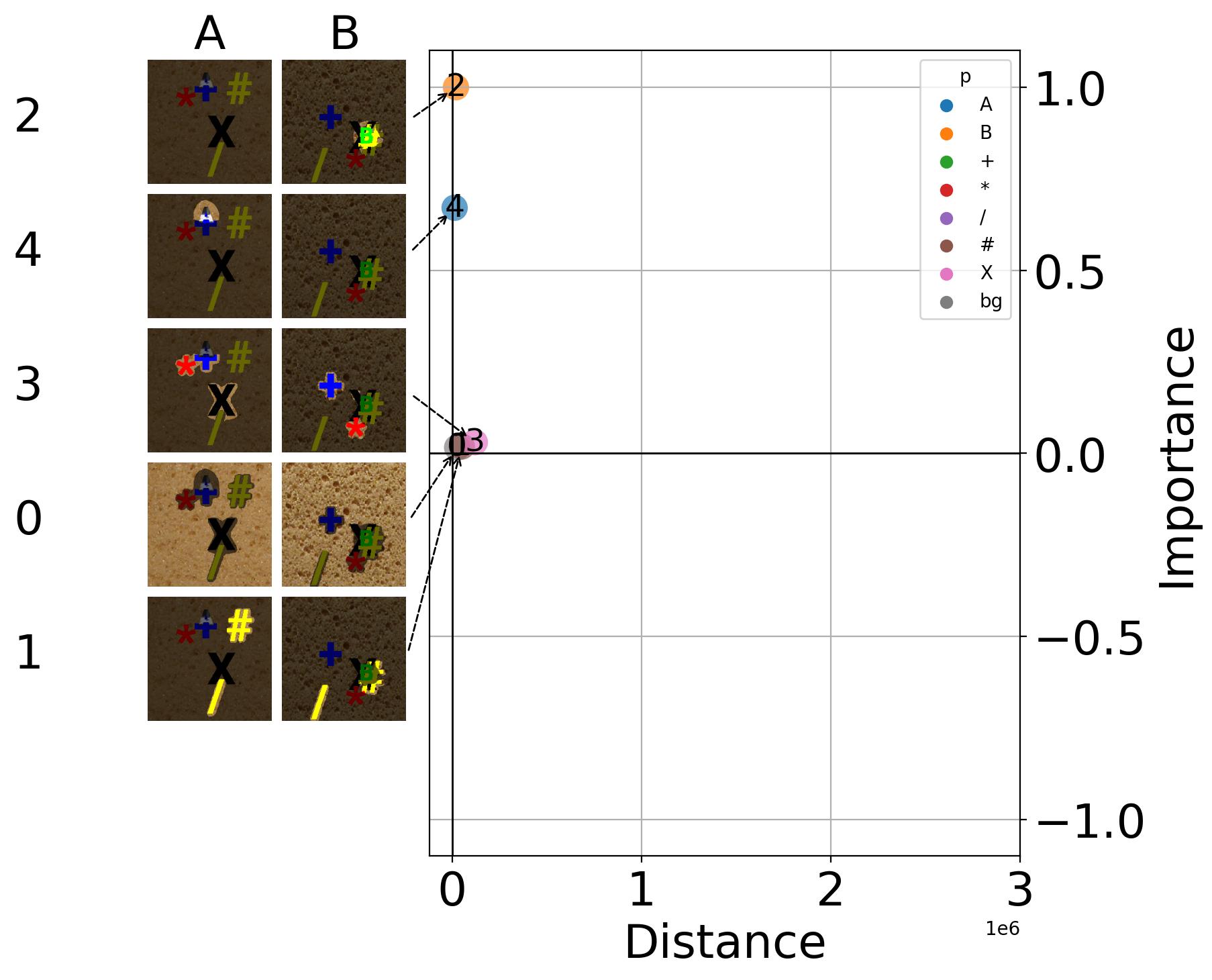}
}
\subfigure[$n_\mathrm{c} = 10$]{
\label{fig:scatter ECLAD dense n=10}
\includegraphics[width=0.41\textwidth]{eclad_ABplus_densenet121_plateau_0_n10_mbKMeans_bilinear_4L}
}

\subfigure[$n_\mathrm{c} = 20$]{
\label{fig:scatter ECLAD dense n=20}
\includegraphics[width=0.41\textwidth]{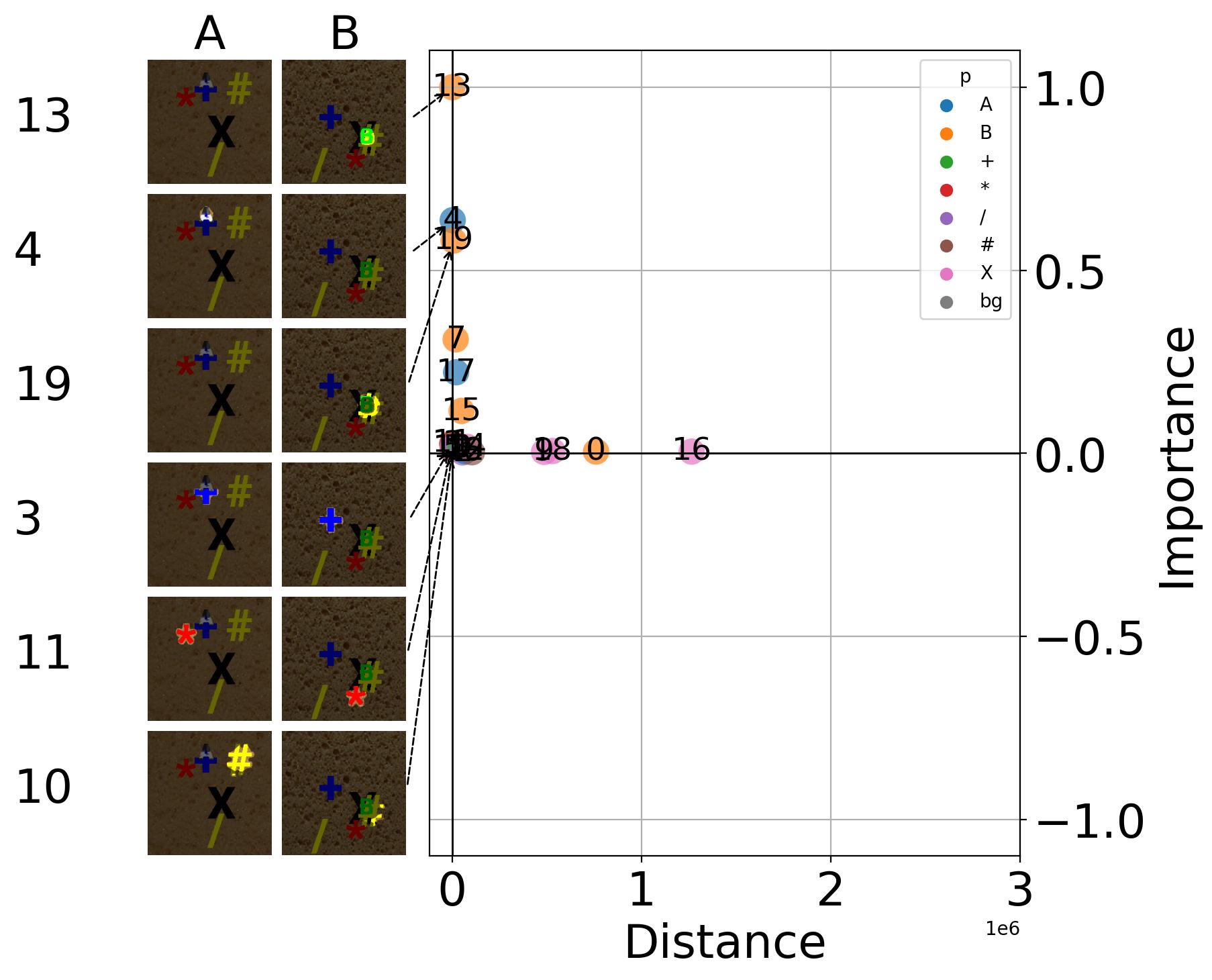}
}
\subfigure[$n_\mathrm{c} = 50$]{
\label{fig:scatter ECLAD dense n=50}
\includegraphics[width=0.41\textwidth]{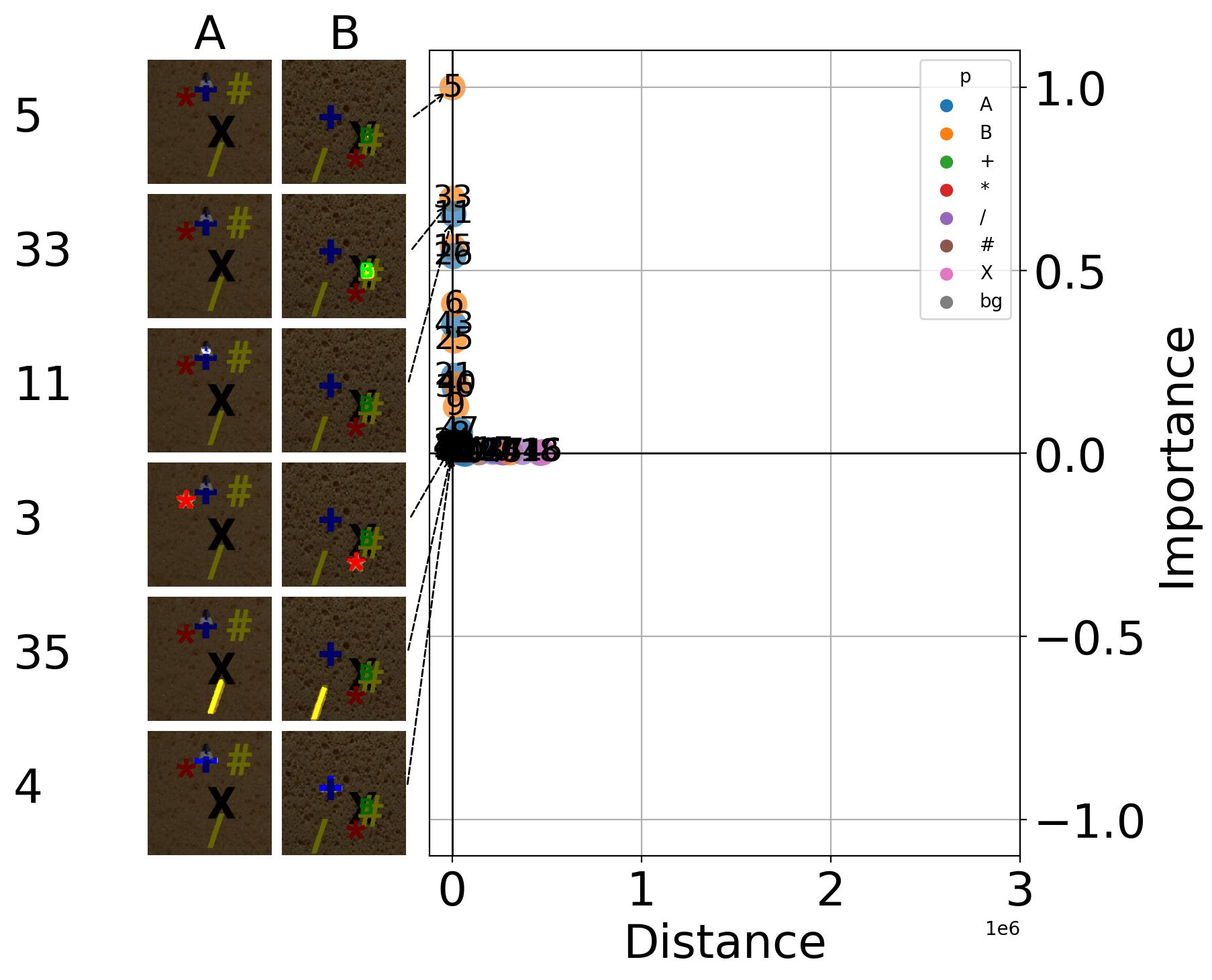}
}

\caption{
Concepts extracted from a DenseNet-121 trained in the ABplus dataset. Subfigures \ref{fig:scatter ECLAD dense n=5} to \ref{fig:scatter ECLAD dense n=50} contain results of executing ECLAD with 5 to 50 extracted concepts.
With $n_c$ of five, the important concepts are extracted correctly (character \textbf{B} and \textbf{A}, $c_2$ and $c_4$ in subfigure \ref{fig:scatter ECLAD dense n=5}), and unimportant concepts are presented together ($c_3$ in subfigure \ref{fig:scatter ECLAD dense n=5}).
With an increasing number of clusters, the different features start to disentangle, e.g. characters \textbf{X}, \textbf{*}, and \textbf{+}, in subfigure \ref{fig:scatter ECLAD dense n=10}.
Yet, for larger number of concepts such as 20 or 50, the original features such as the character \textbf{B}, start to be sliced into multiple concepts (e.g. $c_5$ and $c_{33}$ in subfigure \ref{fig:scatter ECLAD dense n=50}).
}
\label{fig:scatter results number of clusters}
\end{figure*}

The number of concepts to extract $n_c$ is an important parameter for ECLAD, as it determines the number of clusters to mine using minibatch k-means over subsets of LADs. In this section we compare executions of ECLAD with different numbers of k-means clusters $n_c$. The results of four runs with $n_c$ of 5, 10, 20 and 50 are shown in Figure \ref{fig:scatter results number of clusters}.

\textbf{A low number of concepts will group unimportant features, and a high number of concepts will slice important features. Nonetheless, the extraction and scoring of important features is consistent}.
For low number of clusters, such as 5, unimportant features are grouped together in a single concept. An example can be seen in concept $c_3$ from subfigure \ref{fig:scatter ECLAD dense n=5}, where the characters \textbf{X}, \textbf{+} and \textbf{*} were grouped on a single concept.
These features are then split when the number of clusters is increased, as seen for $n_c=10$ in subfigure \ref{fig:scatter ECLAD dense n=10}, with concepts $c_3$, $c_8$, and $c_5$.
For larger number of clusters such as 20 or 50, features are sliced into multiple concepts. An example can be seen in concepts $c_5$ and $c_33$ of subplot \ref{fig:scatter ECLAD dense n=50} which represent the center and surrounding of character \textbf{B}, respectively.
Nonetheless, the slices of important concepts are still being scored with the highest $\mathrm{RI}_{c_j}$, consistently across different numbers of concepts.

\subsection{Upscaling methods}
\label{apd:Ablation upscaling methods}

\begin{figure}[h]
\centering
\subfigure[$f_U = \mathrm{nearest \; interpolation}$.]{
\label{fig:scatter ECLAD nearest}
\includegraphics[width=0.41\textwidth]{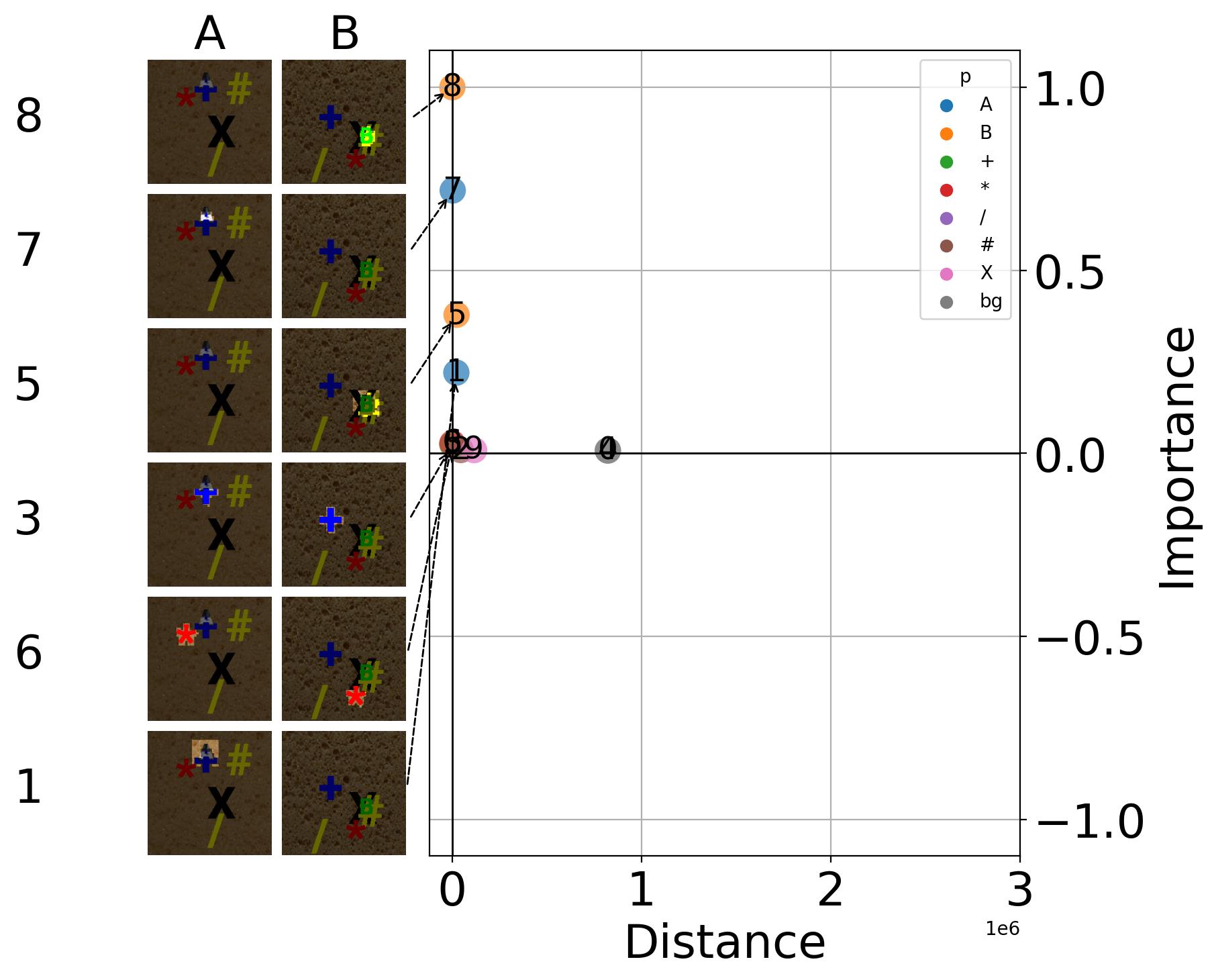}
}
\subfigure[$f_U = \mathrm{bilinear \; interpolation}$.]{
\label{fig:scatter ECLAD bilinear}
\includegraphics[width=0.41\textwidth]{eclad_ABplus_densenet121_plateau_0_n10_mbKMeans_bilinear_4L}
}
\subfigure[$f_U = \mathrm{bicubic \; interpolation}$.]{
\label{fig:scatter ECLAD bicubic}
\includegraphics[width=0.41\textwidth]{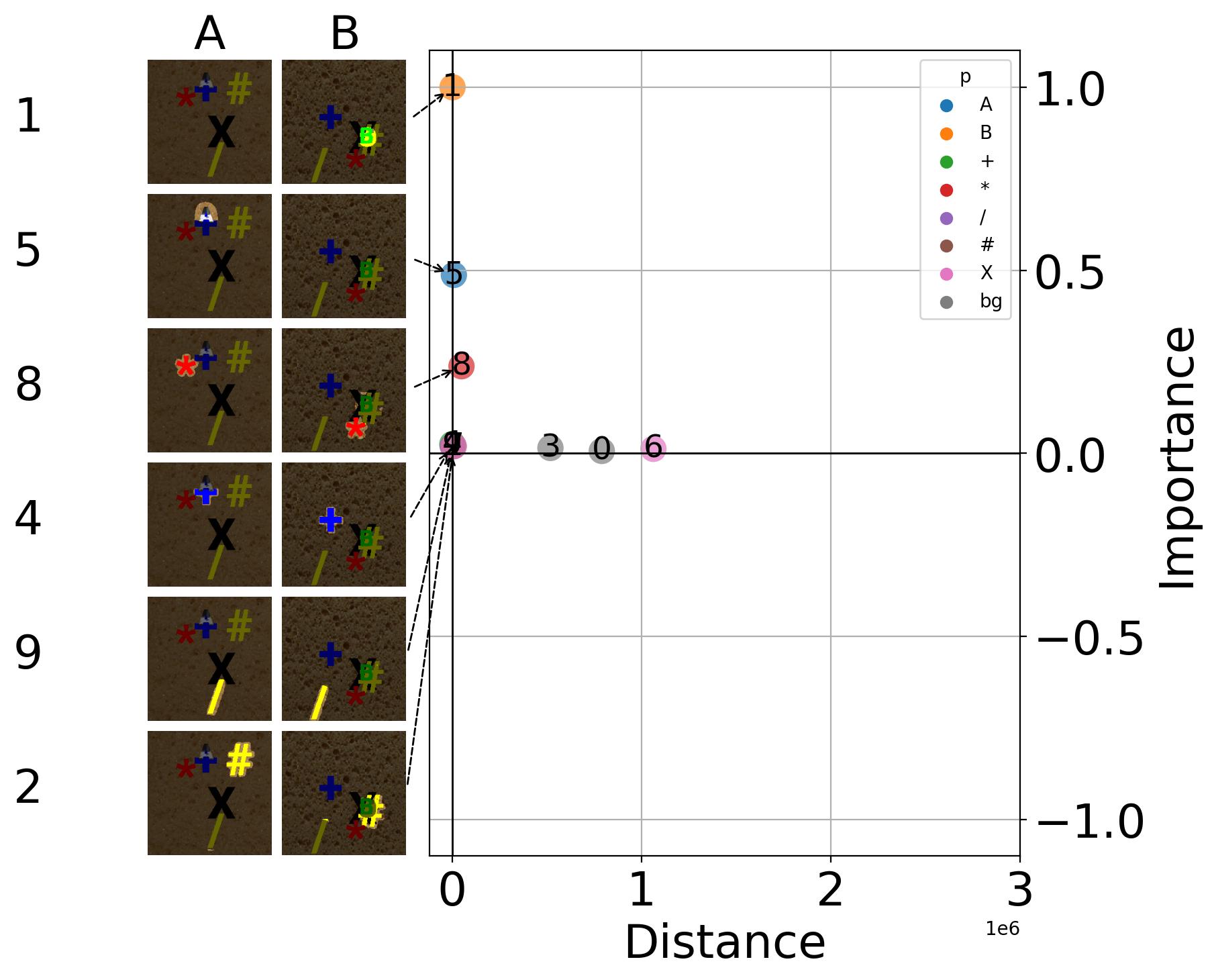}
}

\caption{
Concepts extracted from a DenseNet-121 trained in the ABplus dataset. Subfigures \ref{fig:scatter ECLAD nearest}, \ref{fig:scatter ECLAD bilinear}, and \ref{fig:scatter ECLAD bicubic} present the results for executing ECLAD with nearest interpolation, bilinear interpolation and bicubic interpolation respectively.
The resulting concepts of the three runs contain the main features of the dataset, characters \textbf{B} and \textbf{A}, which are extracted in concepts $c_8$, $c_7$ in subfigure \ref{fig:scatter ECLAD nearest}, $c_2$ and $c_4$ in subfigure \ref{fig:scatter ECLAD bilinear}, and $c_1$ and $c_5$ in subfigure \ref{fig:scatter ECLAD bicubic}.
}
\label{fig:scatter ECLAD upsampling dense}
\end{figure}

A key step of ECLAD is the upsampling of activation maps to obtain the image descriptors and LADs. The upsampling functions can have a significant impact when resizing small activation maps from high level layers. Thus we explore three alternatives in the runs below, in Figure \ref{fig:scatter ECLAD upsampling dense} we present ECLAD results of three runs using nearest interpolation, bilinear interpolation, and bicubic interpolation.

\begin{figure}[h]
\centering
\subfigure[$f_U = \mathrm{nearest \; interpolation}$.]{
\label{fig:scatter ECLAD nearest resnet}
\includegraphics[width=0.41\textwidth]{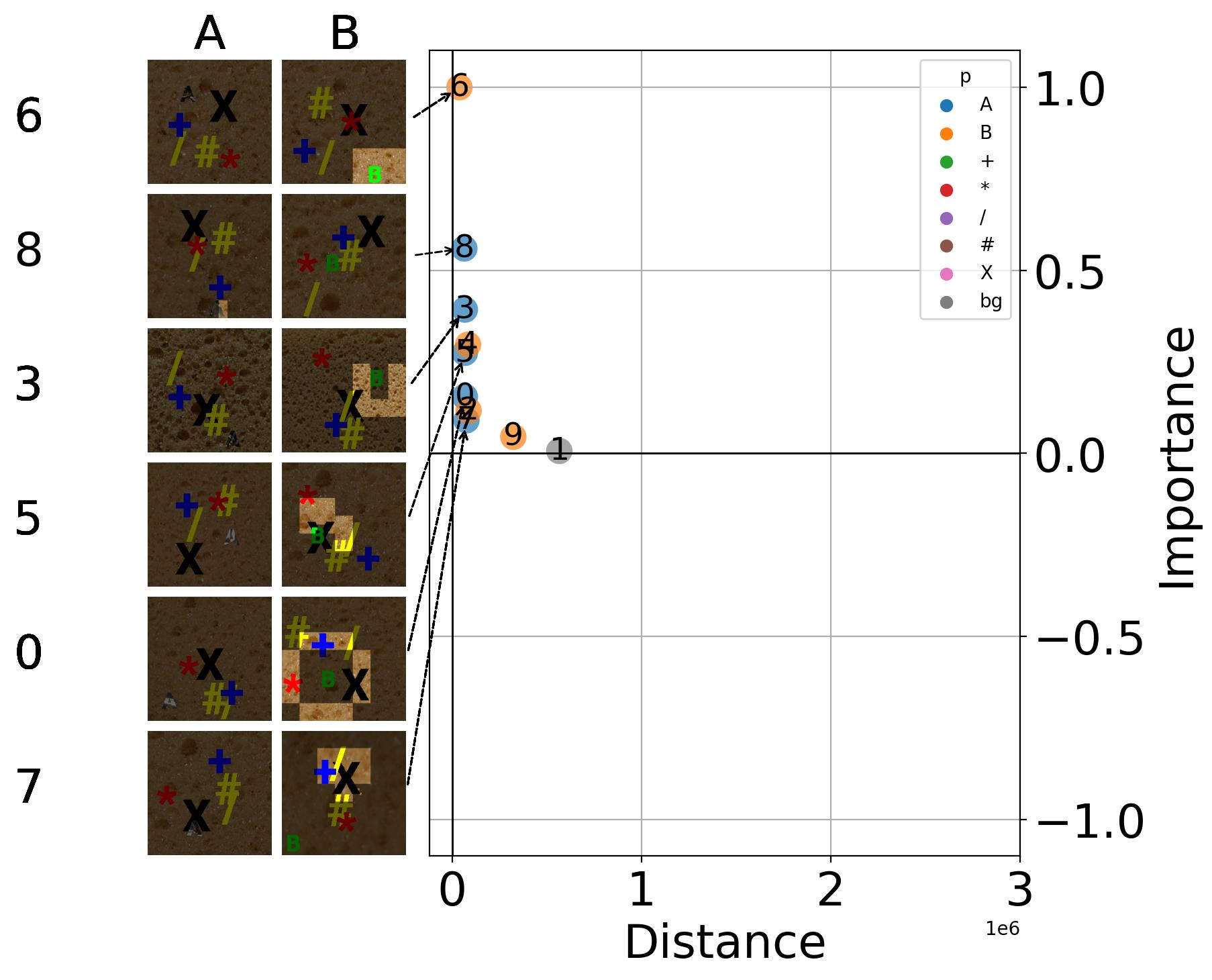}
}
\subfigure[$f_U = \mathrm{bilinear \; interpolation}$.]{
\label{fig:scatter ECLAD bilinear resnet}
\includegraphics[width=0.41\textwidth]{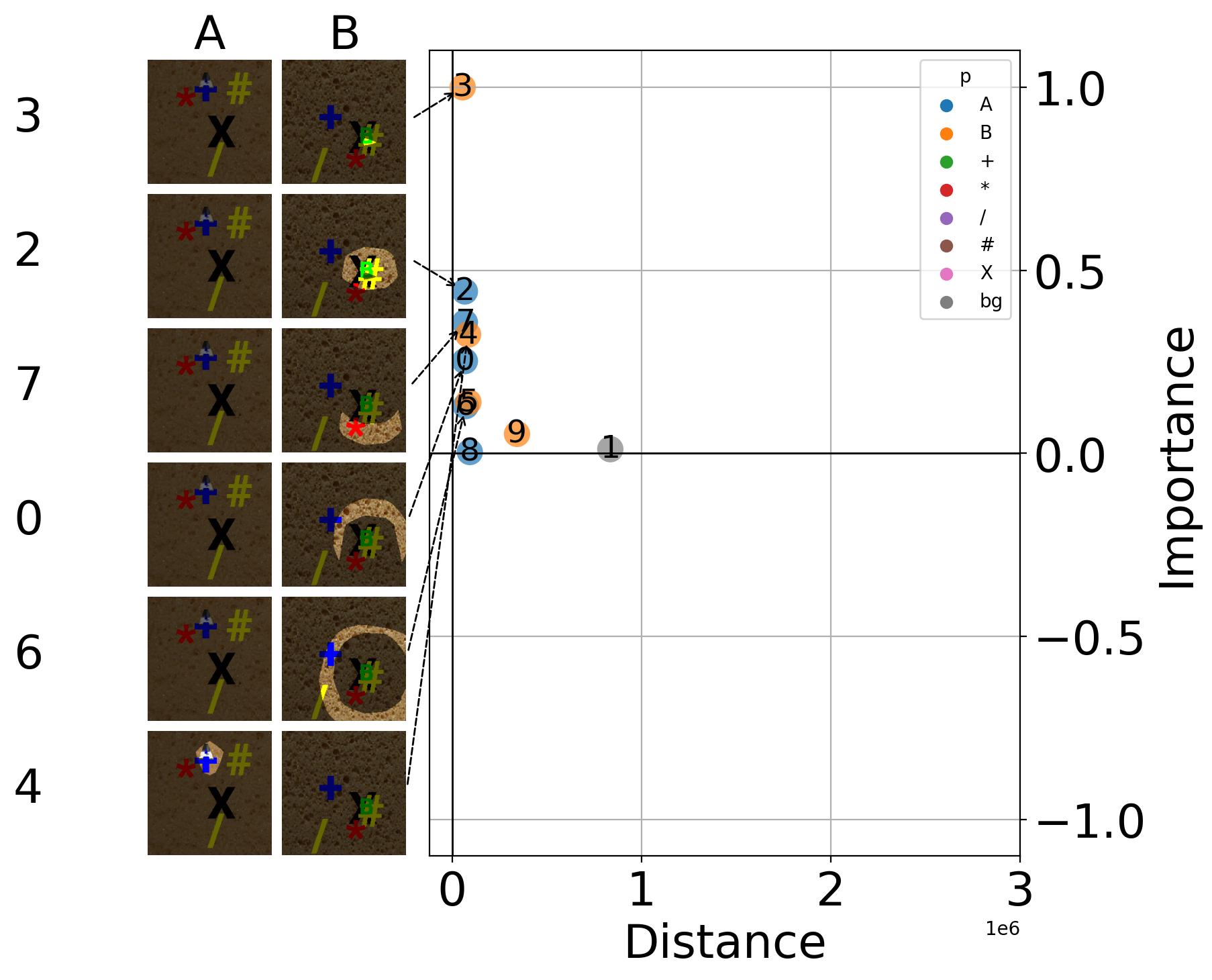}
}
\subfigure[$f_U = \mathrm{bicubic \; interpolation}$.]{
\label{fig:scatter ECLAD bicubic resnet}
\includegraphics[width=0.41\textwidth]{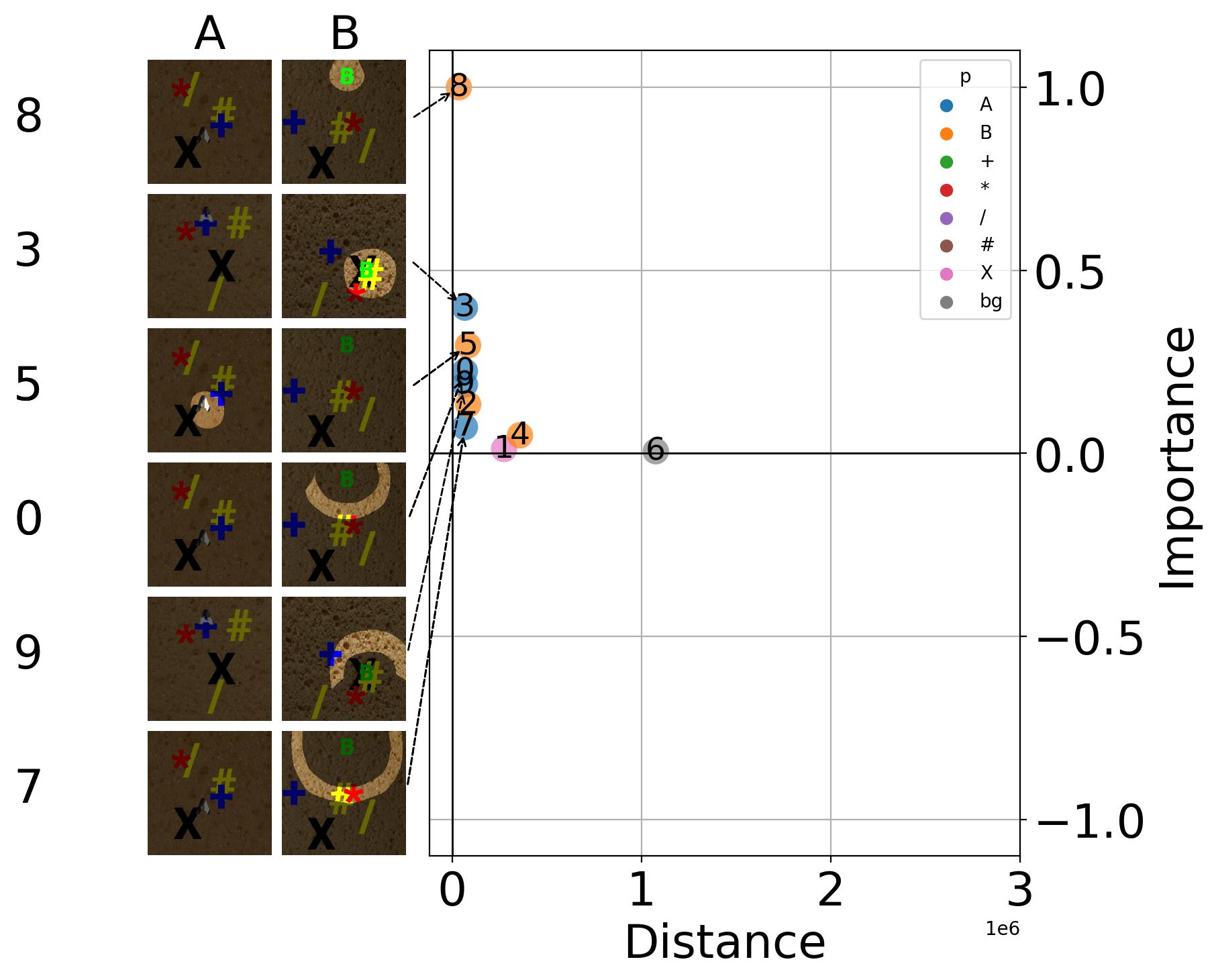}
}

\caption{
Concepts extracted from a Resnet-121 trained in the ABplus dataset. Subfigures \ref{fig:scatter ECLAD nearest resnet}, \ref{fig:scatter ECLAD bilinear resnet}, and \ref{fig:scatter ECLAD bicubic resnet} present the results for executing ECLAD with nearest interpolation, bilinear interpolation and bicubic interpolation respectively.
The resulting concepts of the three runs contain the main features of the dataset, characters \textbf{B} and \textbf{A}, which are extracted in concepts $c_6$, $c_8$ in subfigure \ref{fig:scatter ECLAD nearest resnet}, $c_3$ and $c_2$ in subfigure \ref{fig:scatter ECLAD bilinear resnet}, and $c_8$ and $c_5$ in subfigure \ref{fig:scatter ECLAD bicubic resnet}.
}
\label{fig:scatter ECLAD upsampling resnet}
\end{figure}

\textbf{Using coarse interpolation methods ($f_U$) will impact the boundaries of the extracted concepts, but not the concepts themselves}.
For the three methods, similar concepts where extracted, an example is the important character \textbf{B}, which is extracted as concepts $c_8$, $c_2$, $c_1$ for the nearest, bilinear, and bicubic interpolation runs respectively.
A similar example is the unimportant character \textbf{+}  which is extracted as concepts $c_3$, $c_3$, $c_4$ for the nearest, bilinear, and bicubic interpolation runs respectively.
Aside from the rough boundaries of the concepts, no perceivable effect was observed on the end result of the different runs. A similar behavior was observed when analysing the results for a ResNet-18 also trained on the ABplus dataset, as shown in Figure \ref{fig:scatter ECLAD upsampling resnet}

\clearpage

\section{Distance metric}
\label{apd:Distance metric}

A key contribution of the current manuscript is the proposal of an distance metric $\mathrm{DST}_{p_o,c_j}$ measuring the \textit{spatial association} between the masks of concepts and primitives. This metric takes into account overlapping and spatial closeness to mitigate the effect of associating off-centered and surrounding concept.
The case of off-centered concepts can arise when the representation of a feature shifts through the filters of CNN. This phenomenon can arise in relation with the depth of a CNN unless the activation maps of multiple depths are constrained.
The case of surrounding concepts can arise when a network recognizes the shape of a feature as important, and not the area of the feature itself. Thus, the the edges surrounding the shape may be recognized, and further propagated towards the exterior of an object.
We seek a metric capable of relating dataset primitives and concepts even in these exceptional cases. In this section we compare the proposed $\mathrm{DST}_{p_o,c_j}$ association distance, with the Jaccard score used in object detection, the normalized mutual information score, and the adjusted rand score used in clustering.

\begin{figure}[htb]
\centering
\includegraphics[width=0.41\textwidth]{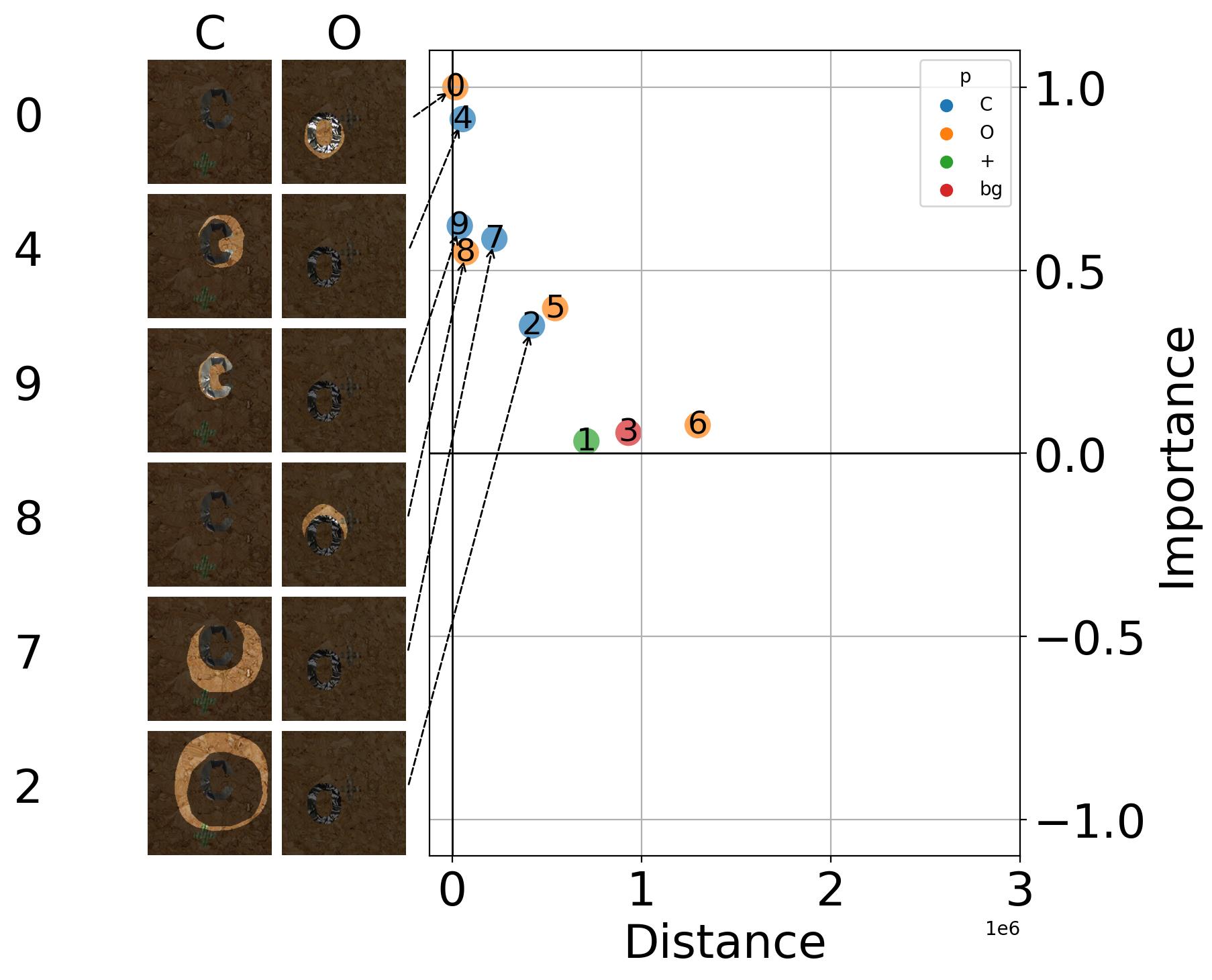}
\caption{
Concepts extracted from a ResNet-18 model trained over the CO synthetic dataset. Concept $c_4$ related to the missing right part of the character \textbf{C}. The concept is non-overlapping, yet, spatially related to character \textbf{C} on all images.
}
\label{fig:scatter ECLAD CO difference}
\end{figure}

As an example, the CO synthetic dataset consists in classifying images with a character \textbf{C} or a character \textbf{O} in them. Given the shape of both characters, the difference can be described as an extra right section for the character \textbf{O}, or a missing right section of the \textbf{C}. The two approaches for detecting both classes where seen in the experimentation process. For the current metric comparison, an example of overlapping related concepts is shown in Figure \ref{fig:scatter ECLAD CO difference}.

\begin{figure*}[htb]
\centering
\subfigure[Metric comparison for offsets of the character \textbf{A}.]{
\label{fig:overlap A}
\includegraphics[width=0.90\textwidth]{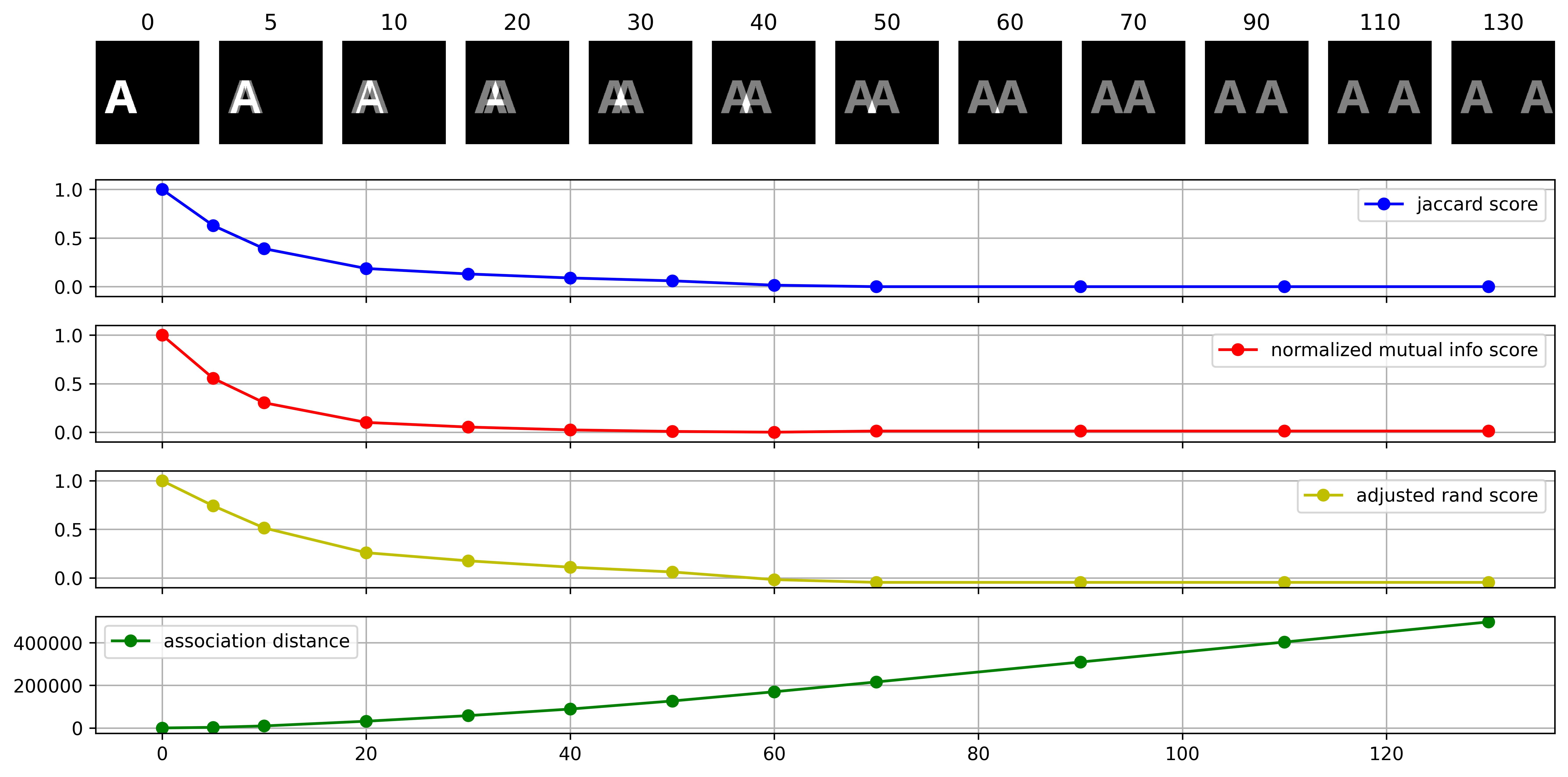}
}

\subfigure[Metric comparison for offsets of the character \textbf{O}.]{
\label{fig:overlap O}
\includegraphics[width=0.90\textwidth]{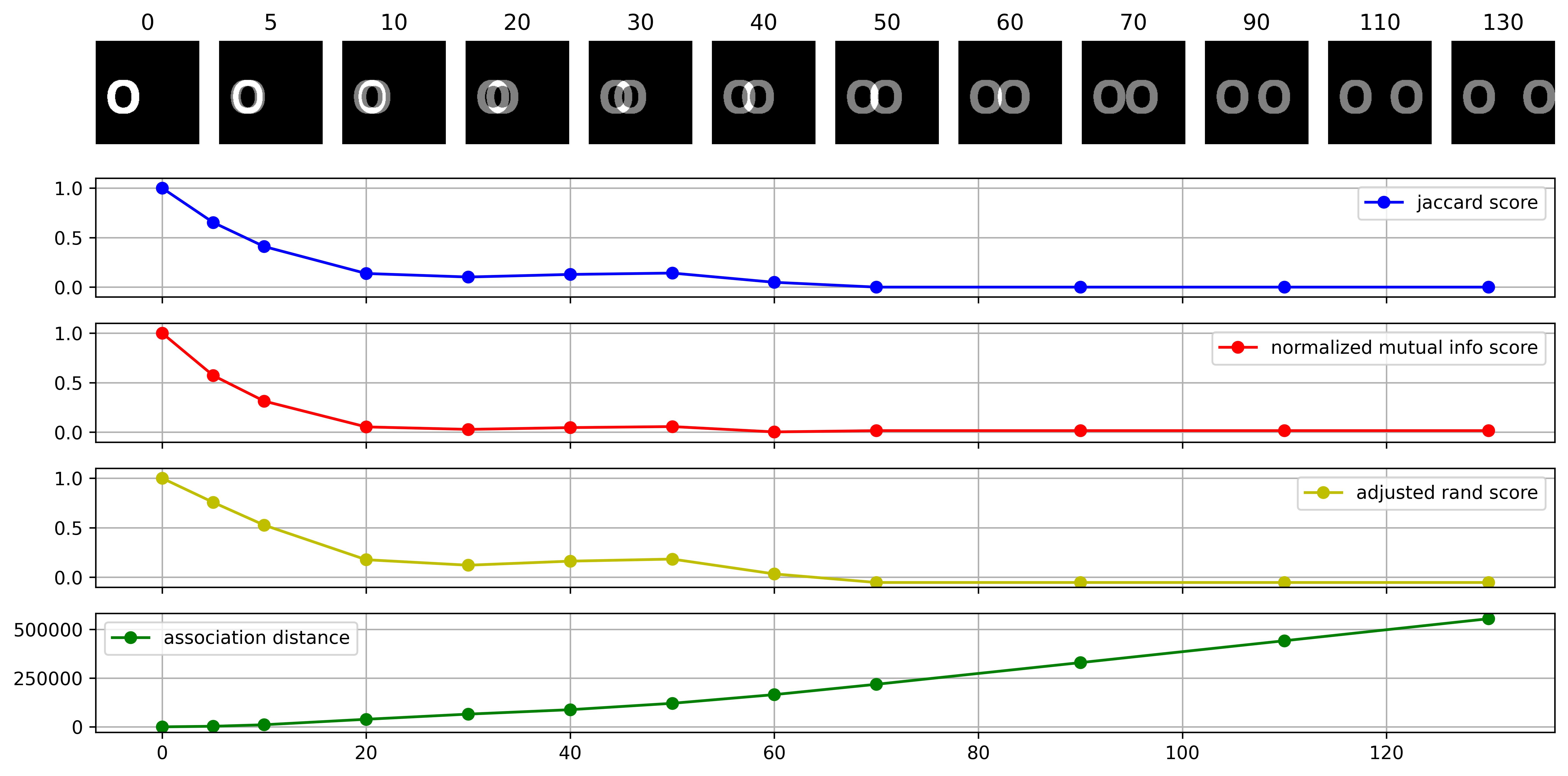}
}

\caption{
Evolution of metrics at various degrees of overlapping and offset for the character A in subfigure \ref{fig:overlap A}, and for the character O in subfigure \ref{fig:overlap O}.
In both cases, the jaccard score, normalized mutual information score, and adjusted rand score are proportional to the degree of overlapping, yet, do not show any difference for further offsets of the masks.
In comparison, the association distance captures the differences between the masks not only the overlapping cases (d<70), but also non overlapping shifts. 
}
\label{fig:metrics overlap}
\end{figure*}

On the first experimental setup we compare two masks emulating a primitive and a concept, with the same general form as the offset between the characters increase. In the figure \ref{fig:metrics overlap} we present the experimental results of comparing two masks of the character \textbf{A}, and two masks of the character \textbf{O}, at various offsets. White areas represent overlapped sections and gray areas represent non overlapping regions of both masks.

\textbf{The proposed association distance $\mathrm{DST}_{p_o,c_j}$ can express offsets between primitive and concept masks, with or without overlapping}.
Figure \ref{fig:overlap A}, shows as the three alternative metrics decrease monotonically with a bigger offset between masks, yet, the difference can only be measured while there is a degree of overlapping. When the two masks cease to overlap, the alternative metrics do not express the offset anymore.
In addition, depending on the geometry of the compared masks, the degree of overlapping may increase as the two masks shift from each other. This can be observed at 40 pixels offset on the subfigure \ref{fig:overlap O}, where the vertical section of the character \textbf{O} increases the overlapping. As a consequence, the Jaccard score, normalized mutual information score, and adjusted rand score, stop behaving monotonically with respect to the shift between the masks, which is undesired.


In other cases, a concept may represent the surroundings of a feature, or the missing counterpart of a form. In these cases, it is desired that an association metric is able to measure the degree of separation between the feature and the surrounding concept.

\begin{figure*}[htb]
\centering
\subfigure[Metric comparison for features surrounding the character \textbf{A}.]{
\label{fig:surround A}
\includegraphics[width=0.90\textwidth]{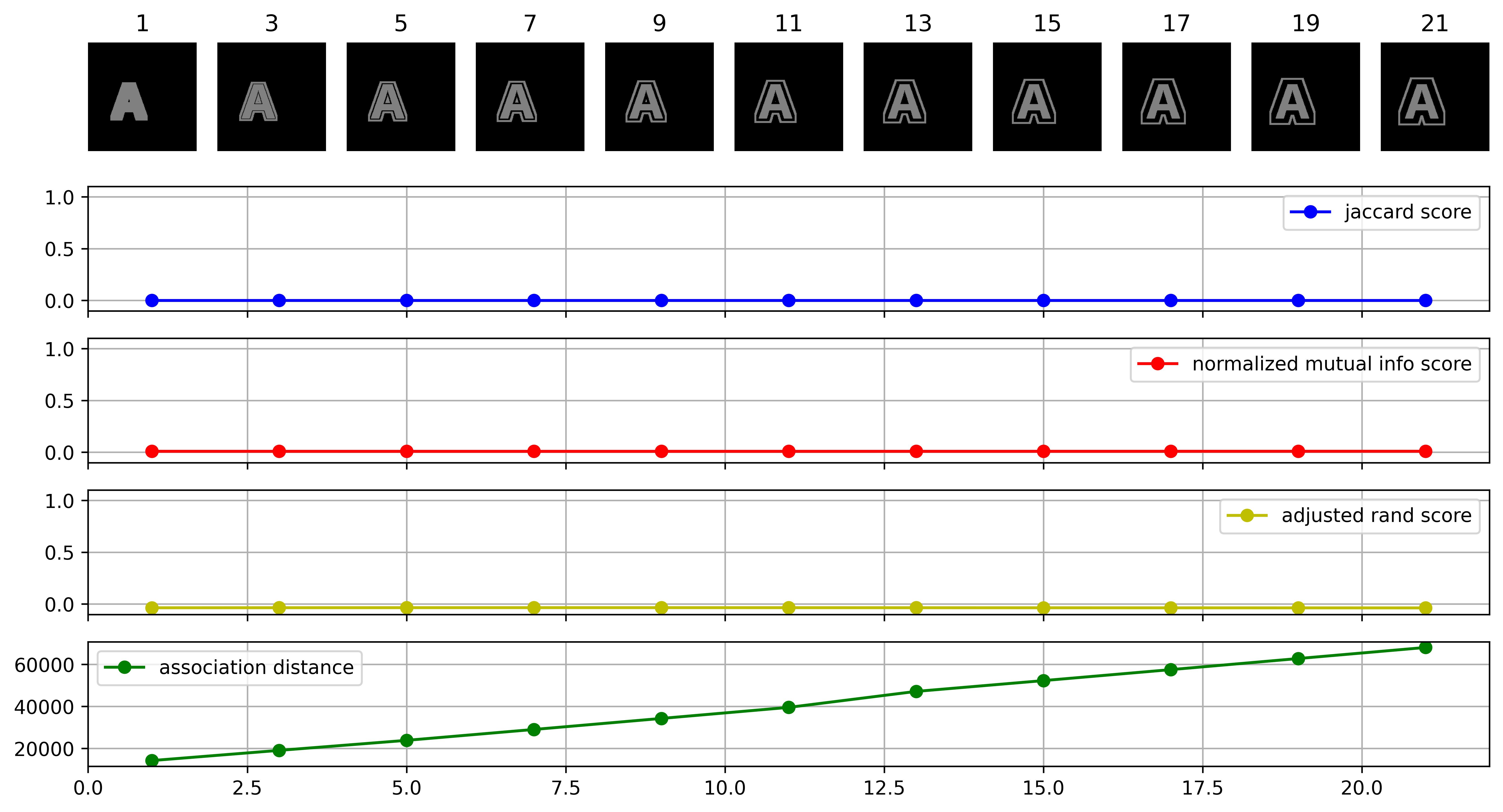}
}

\subfigure[Metric comparison for features surrounding the character \textbf{O}.]{
\label{fig:surround O}
\includegraphics[width=0.90\textwidth]{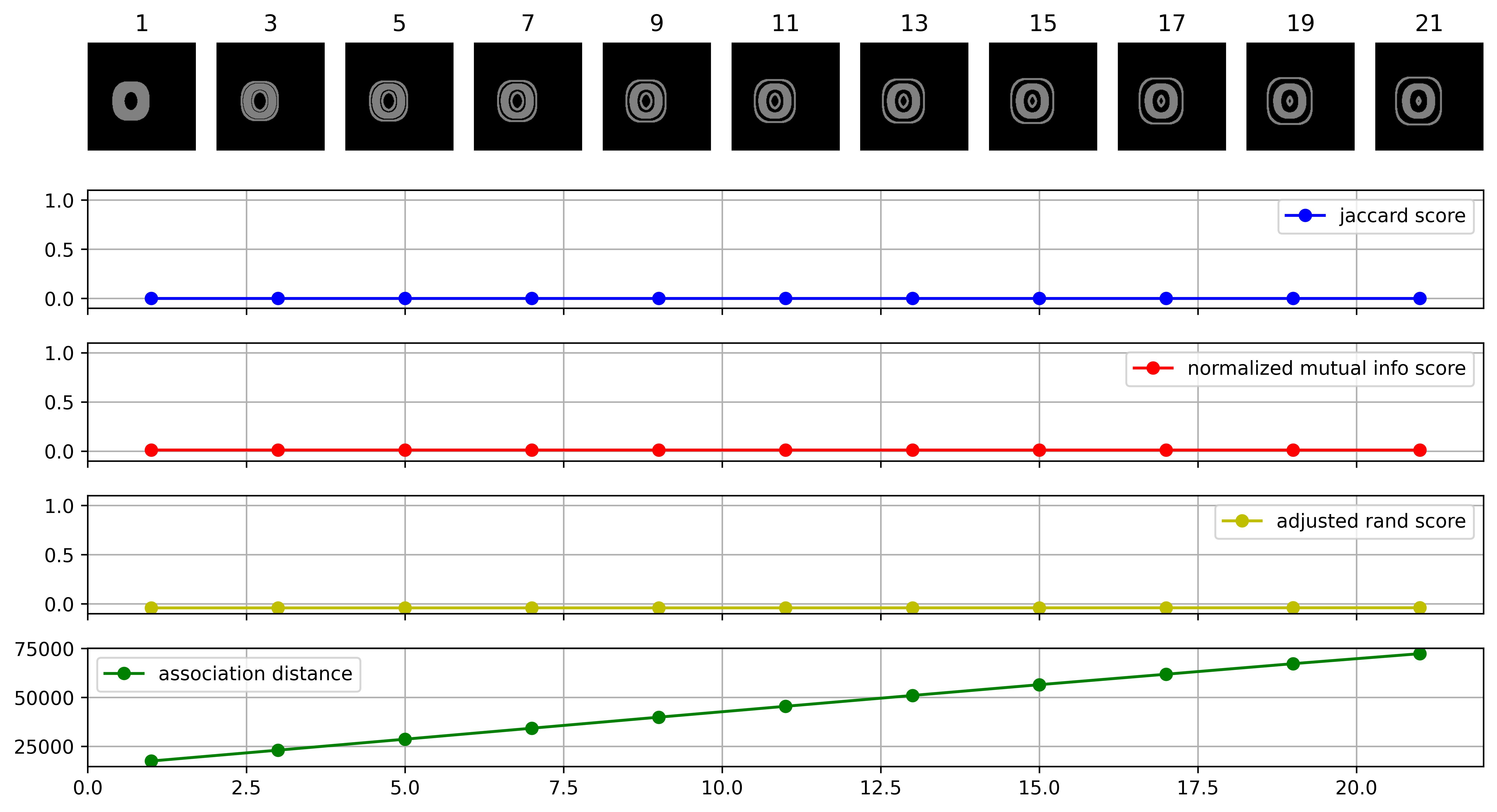}
}

\caption{
Evolution of metrics as a relate surrounding features distances itself from a shape. subfigure \ref{fig:surround A} contains the comparison of the character \textbf{A} and a concept surrounding it, initially including it's center, and subsequently only surrounding its exterior. Similarly, subfigure \ref{fig:surround O} contains the comparison between the character \textbf{O} and a surrounding concept.
In both cases, the association distance behaves monotonically as the primitive and the concepts distance increases, yet, other metrics cannot capture this behavior.
}
\label{fig:metrics surround}
\end{figure*}

\textbf{The proposed association distance $\mathrm{DST}_{p_o,c_j}$ can measure concepts surrounding primitives and express the degree of separation between both masks.}. Other metrics only allow the comparison of overlapping regions, which can be problematic.

%







\end{document}

%% file: math_commands.tex
\usepackage{amsmath,amsfonts,bm}

\def\eqref#1{equation~\ref{#1}}

\def\1{\bm{1}}

\DeclareMathAlphabet{\mathsfit}{\encodingdefault}{\sfdefault}{m}{sl}
\SetMathAlphabet{\mathsfit}{bold}{\encodingdefault}{\sfdefault}{bx}{n}

